\documentclass[twoside,11pt]{article}

\usepackage[preprint]{jmlr2e}
\usepackage{color}
\usepackage{amsmath}
\usepackage{mathtools}
\usepackage{booktabs,makecell}
\usepackage{multirow}
\usepackage{slashbox}
\usepackage{subcaption}
\usepackage{tikz}
\usepackage{tcolorbox}
\usetikzlibrary{shapes,decorations,arrows,calc,arrows.meta,fit,positioning}
\tikzset{
    -Latex,auto,node distance =1 cm and 1 cm,semithick,
    state/.style ={circle, draw, inner sep=0pt, minimum size = 18pt},
    point/.style = {circle, draw, inner sep=0.04cm,fill,node contents={}},
    bidirected/.style={Latex-Latex,dashed},
    el/.style = {inner sep=2pt, align=left, sloped},
    block/.style = {draw, rectangle, minimum height=3em, minimum width=3em},
}

\newcommand{\distri}{\mathcal{P}}
\newcommand{\distritar}{\widetilde{\mathcal{P}}}
\newcommand{\distrinoise}{\mathcal{E}}

\newcommand{\risk}{R}

\newcommand{\envs}{M}

\newcommand{\loss}{l}
\newcommand{\const}{c}

\newcommand{\invIB}{H}

\newcommand{\tagk}[1]{{\text{\scriptsize $(#1)$}}}

\newcommand{\tagpar}{\text{P}}
\newcommand{\tagdes}{\text{D}}
\newcommand{\tagint}{\mathcal{I}}

\newcommand{\lamMatch}{\lambda_{\text{match}}}
\newcommand{\lamCIP}{\lambda_{\text{CIP}}}

\newcommand{\noise}{\varepsilon}
\newcommand{\semB}{\mathbf{B}}
\newcommand{\semBX}{\mathbf{B}}
\newcommand{\sembv}{{b}}
\newcommand{\sembh}{{\omega}}
\newcommand{\interA}{a}
\newcommand{\intervtype}{g}
\newcommand{\betaDom}{\Theta}
\newcommand{\dimscau}{\ensuremath{r}}

\newcommand{\Xv}[1]{X^\tagk{#1}}
\newcommand{\Yv}[1]{Y^\tagk{#1}}
\newcommand{\distriv}[1]{\distri^\tagk{#1}}
\newcommand{\interAv}[1]{\interA^\tagk{#1}}
\newcommand{\noisev}[1]{\noise^\tagk{#1}}
\newcommand{\Xtar}{\widetilde{X}}
\newcommand{\Ytar}{\widetilde{Y}}
\newcommand{\interAtar}{\widetilde{\interA}}
\newcommand{\noisetar}{\widetilde{\noise}}
\newcommand{\noisecovX}{\Sigma}
\newcommand{\noisecovY}{\sigma}

\newcommand{\newYtar}[1]{\widetilde{\Upsilon}}

\newcommand{\tagcausal}{\text{Causal}}
\newcommand{\betacausal}{\beta_\tagcausal}
\newcommand{\tagolstarget}{\text{OLSTar}}
\newcommand{\betaolstarget}{\beta_\tagolstarget}
\newcommand{\tagolssource}{\text{OLSSrc}}
\newcommand{\betaolssourceone}[1]{\beta_{\tagolssource}^\tagk{#1}}
\newcommand{\tagolspool}{\text{SrcPool}}
\newcommand{\betaolspool}{\beta_\tagolspool}
\newcommand{\tagdipmeanmatchlin}{\text{DIP}}
\newcommand{\betadipmeanmatchlin}[1]{\beta_{\tagdipmeanmatchlin}^\tagk{#1}}
\newcommand{\tagdipmeanmix}{\text{DIP}\diamondsuit}
\newcommand{\betadipmeanmix}[1]{\beta_{\tagdipmeanmix}^\tagk{#1}}
\newcommand{\tagdipstdmatchlin}{\text{DIP-std}}

\newcommand{\tagdipstdpmatchlin}{\text{DIP-std$+$}}

\newcommand{\tagdipMMDmatchlin}{\text{DIP-MMD}}

\newcommand{\tagdiporaclemeanmatchlin}{\text{DIPOracle}}
\newcommand{\betadiporaclemeanmatchlin}[1]{\beta_{\tagdiporaclemeanmatchlin}^\tagk{#1}}
\newcommand{\tagdipmeanmatchabs}{\text{DIPAbs}}
\newcommand{\betadipmeanmatchabs}[1]{\beta_{\tagdipmeanmatchabs}^\tagk{#1}}
\newcommand{\tagdipweigh}{\text{DIPweigh}}

\newcommand{\tagcipmean}{\text{CIP}}
\newcommand{\betacipmean}{\beta_{\tagcipmean}}
\newcommand{\tagriimean}{\text{RII}}
\newcommand{\betariimean}{\beta_\tagriimean}
\newcommand{\tagcirmeanmatch}{\text{CIRM}}
\newcommand{\bcirmeanmatch}{\vartheta_{\tagcirmeanmatch}}
\newcommand{\betacirmeanmatch}[1]{{\beta_{\tagcirmeanmatch}^\tagk{#1}}}
\newcommand{\tagcirmweigh}{\text{CIRMweigh}}

\newcommand{\tagcirmi}{\text{CIRMI}}
\newcommand{\bcirmi}{\vartheta_{\tagcirmi}}
\newcommand{\betacirmi}[1]{\beta_{\tagcirmi}^\tagk{#1}}

\newcommand{\tagcipmeanmix}{\text{CIP}\diamondsuit}
\newcommand{\betacipmeanmix}{\beta_{\tagcipmeanmix}}
\newcommand{\tagcirmeanmatchmix}{\text{CIRM}\diamondsuit}
\newcommand{\bcirmeanmatchmix}{\vartheta_{\tagcirmeanmatchmix}}
\newcommand{\betacirmeanmatchmix}[1]{\beta_{\tagcirmeanmatchmix}^\tagk{#1}}

\newcommand{\tagriirmi}{\text{RIIRMI}}
\newcommand{\briirmi}{\vartheta_{\tagriirmi}}
\newcommand{\betariirmi}[1]{\beta_{\tagriirmi}^\tagk{#1}}

\newcommand{\obstar}{\tilde{\obs}}

\newcommand{\RKHS}{\mathcal{H}}

\newcommand{\disMMD}{\mathcal{D}_{\text{MMD}, \RKHS}}

\newcommand{\Qdip}{Q_\text{\tiny DIP}}
\newcommand{\Gdip}{G_\text{\tiny DIP}}
\newcommand{\Qdipv}[1]{{Q_\text{\tiny DIP}^\tagk{#1}}}
\newcommand{\Gdipv}[1]{{G_\text{\tiny DIP}^\tagk{#1}}}

\newcommand{\Gdipmixv}[1]{{G_\text{\tiny $\tagdipmeanmix$}^\tagk{#1}}}
\newcommand{\QdipY}{Q_2}
\newcommand{\GdipY}{G_2}
\newcommand{\Qcip}{Q_\text{\tiny CIP}}
\newcommand{\Gcip}{G_\text{\tiny CIP}}
\newcommand{\Pcip}{P_\text{\tiny CIP}}
\newcommand{\Qcirm}{\Qdip}

\newcommand{\Qcirmv}[1]{\Qdipv{#1}}
\newcommand{\Gcirmv}[1]{\Gdipv{#1}}
\newcommand{\Prii}{P_4}
\newcommand{\vrii}{v}
\newcommand{\Grii}{G_4}
\newcommand{\Qrii}{Q_4}
\newcommand{\Qriirmi}{Q_5}
\newcommand{\Griirmi}{G_5}

\newtheorem{assumption}{Assumption}

\long\def\comment#1{}
\definecolor{battleshipgrey}{rgb}{0.52, 0.52, 0.51}
\definecolor{darkgray}{rgb}{0.66, 0.66, 0.66}
\definecolor{darkgreen}{rgb}{0.0, 0.2, 0.13}
\definecolor{darkspringgreen}{rgb}{0.09, 0.45, 0.27}
\definecolor{dukeblue}{rgb}{0.0, 0.0, 0.61}
\definecolor{olivedrab7}{rgb}{0.24, 0.2, 0.12}
\definecolor{darkblue}{rgb}{0.0, 0.0, 0.55}
\definecolor{darkscarlet}{rgb}{0.34, 0.01, 0.1}
\definecolor{candyapplered}{rgb}{1.0, 0.03, 0.0}
\definecolor{ao(english)}{rgb}{0.0, 0.5, 0.0}
\definecolor{applegreen}{rgb}{0.55, 0.71, 0.0}
\definecolor{neworange}{RGB}{221,120,63}
\definecolor{newblue}{RGB}{62,76,209}

\newcommand{\olcomment}[1]{}

\newcommand{\defn}{:=}

\newcommand{\obs}{\ensuremath{n}}
\newcommand{\dims}{\ensuremath{d}}
\newcommand{\real}{\ensuremath{\mathbb{R}}}

\newcommand{\Ind}{\ensuremath{\mathbb{I}}}

\DeclareMathOperator*{\argmin}{arg\,min}
\newcommand{\Exs}{\ensuremath{{\mathbb{E}}}}
\newcommand{\Prob}{\ensuremath{{\mathbb{P}}}}

\newcommand{\Normal}{\ensuremath{\mathcal{N}}}

\DeclareMathOperator{\Var}{Var}

\DeclarePairedDelimiterX{\infdivx}[2]{(}{)}{%
  #1\;\delimsize\|\;#2%
}

\newcommand{\brackets}[1]{\left[ #1 \right]}
\newcommand{\parenth}[1]{\left( #1 \right)}

\newcommand{\braces}[1]{\left\{ #1 \right \}}
\newcommand{\abss}[1]{\left| #1 \right |}

\newcommand{\tp}{^\top}

\newcommand{\bmat}[1]{\begin{bmatrix} #1 \end{bmatrix}}

\newcommand{\vecnorm}[2]{\left\| #1\right\|_{#2}}

\usepackage{lastpage}
\jmlrheading{22}{2021}{1-\pageref{LastPage}}{10/20; Revised
11/21}{11/21}{20-1227}{Yuansi Chen and Peter B\"uhlmann}
\ShortHeadings{Domain adaptation under structural causal models}{Chen and B\"uhlmann}
\firstpageno{1}

\begin{document}

\title{Domain adaptation under structural causal models}

\author{\name Yuansi Chen \email yuansi.chen@stat.math.ethz.ch
       % \addr Seminar for Statistics\\
       % ETH Z\"urich\\
       % Z\"urich, Switzerland
       \AND
       \name  Peter B\"uhlmann \email buhlmann@stat.math.ethz.ch \\
       \addr Seminar for Statistics\\
       ETH Z\"urich\\
       Z\"urich, Switzerland}

\editor{Isabelle Guyon}

\maketitle

\begin{abstract}%   <- trailing '%' for backward compatibility of .sty file
  Domain adaptation (DA) arises as an important problem in statistical machine learning when the source data used to train a model is different from the target data used to test the model. Recent advances in DA have mainly been application-driven and have largely relied on the idea of a common subspace for source and target data. To understand the empirical successes and failures of DA methods, we propose a theoretical framework via structural causal models that enables analysis and comparison of the prediction performance of DA methods. This framework also allows us to itemize the assumptions needed for the DA methods to have a low target error. Additionally, with insights from our theory, we propose a new DA method called CIRM that outperforms existing DA methods when both the covariates and label distributions are perturbed in the target data. We complement the theoretical analysis with extensive simulations to show the necessity of the devised assumptions. Reproducible synthetic and real data experiments are also provided to illustrate the strengths and weaknesses of DA methods when parts of the assumptions in our theory are violated.
\end{abstract}


\section{Introduction} % (fold)
\label{sec:introduction}

Domain adaptation (DA) is a statistical machine learning problem in which one aims at learning a model from a labeled source dataset and expecting it to perform well on an unlabeled target dataset drawn from a different but related data distribution. Domain adaptation is considered a sub-field of transfer learning and also a sub-field of semi-supervised learning~\citep{pan2009survey}. The possibility of DA is inspired by the human ability to apply knowledge acquired on previous tasks to unseen tasks with minimal or no supervision. For example, it is common to believe that humans who learned driving in sunny days would adapt their skills to drive reasonably well in a rainy day without additional training. However, the scenario where the source and target data distribution is different (e.g. sunny vs. rainy) is difficult to handle for many machine learning systems. This is mainly because the classical statistical learning theory mostly focuses on statistical learning methods and guarantees when the training and test data are generated from the same distribution.

While existing DA theory is limited, more and more application scenarios have emerged where DA is needed and useful. DA is desired especially when obtaining unlabeled data is cheap while labeling data is difficult. The difficulties of labeling data typically arise due to the required human expertise or the large amount of human labor. For example, to annotate the ImageNet Large Scale Visual Recognition Challenge (ILSVRC) dataset~\citep{russakovsky2015imagenet} with more than 1.2 million labeled images, it costs an average worker on the Amazon Mechanical Turk (\href{https://www.mturk.com/}{www.mturk.com}) about 0.5 seconds per image~\citep{fei2010imagenet}. Thus, annotating more than 1.2 million images requires more than 150 human hours. The actual time needed for labeling is much longer, due to the additional time spent on label verification and on labeling the images which are not used in the final challenge. Now if one is only interested in a similar but different task such as classifying objects in oil paintings, one cannot expect the ILSVRC dataset which contains pictures of natural objects to be representative. As a consequence, one needs to collect new data. It is relatively easy to collect digital images for the new task, but it is costly to label them if human experts have to be involved. Similar situations where the labeling is costly emerge in many other fields such as part-of-speech tagging~\citep{ratnaparkhi1996maximum, ben2007analysis}, web-page classification~\citep{blum1998combining}, etc.

\begin{figure}[ht]

  \begin{minipage}{0.30\textwidth}
  \begin{tcolorbox}[width=\textwidth, colframe=gray!80, colback=newblue!40, boxsep=0mm, arc=3mm]
    \begin{center}
      {\small (i) causal prediction} \vspace{0.3cm}\\
      \begin{tikzpicture}
        % x node set with absolute coordinates
        \node[state] (x1) at (0,0) {$X_1$};
        \node[state] (x2) at (2,0) {$X_2$};
        \node (blackhammer) at (-0.8,0) {\includegraphics[angle = 0, scale = 0.04]{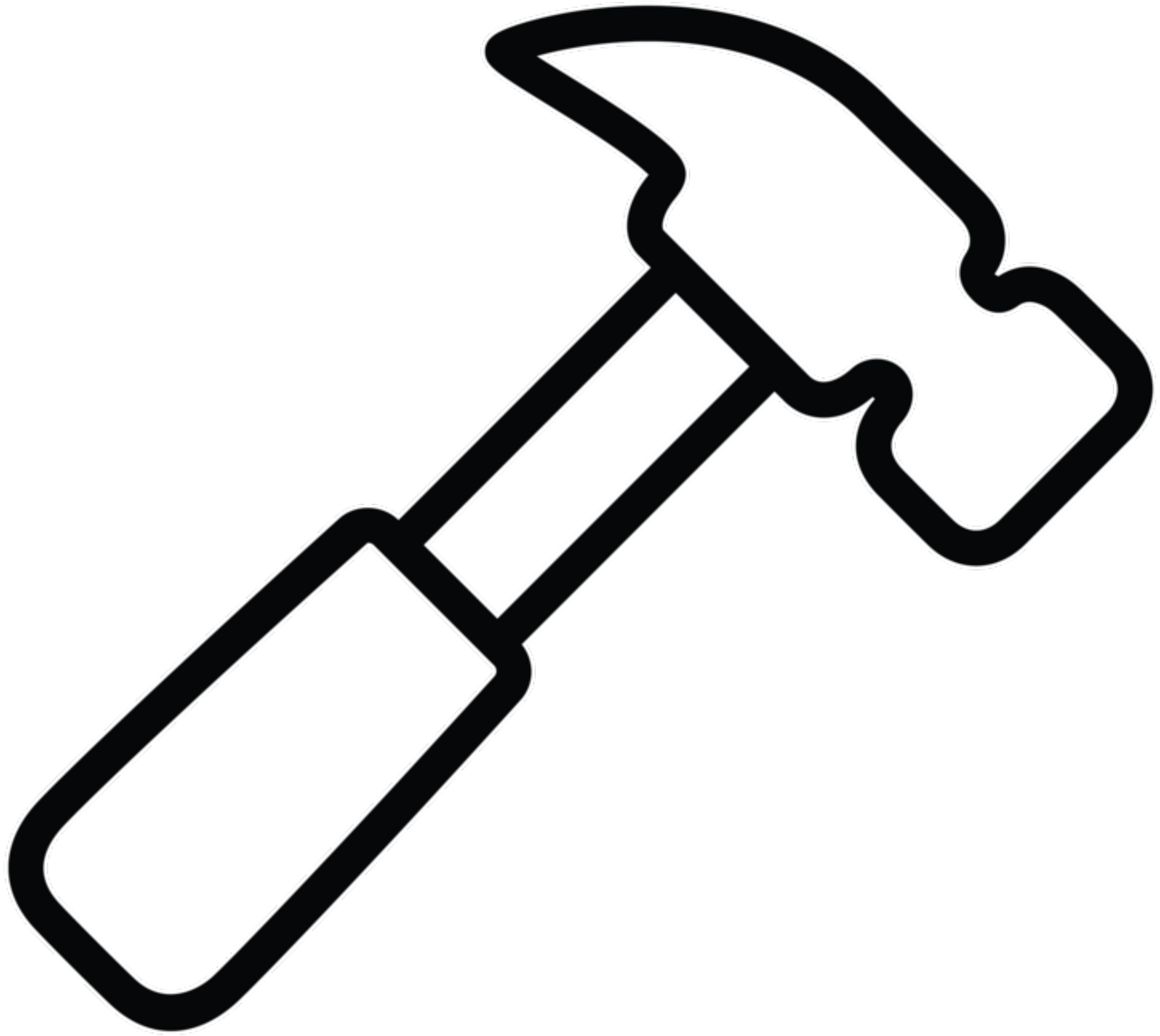}};

        % y node set relative to x.
        % Locations can be:
        % right,left,above,below,
        % above left,below right, etc
        \node[state] (y) at (1, -1.7) {$Y$};
        \node (whitehammer) at (0.2, -1.7) {\includegraphics[angle = 0, scale = 0.04]{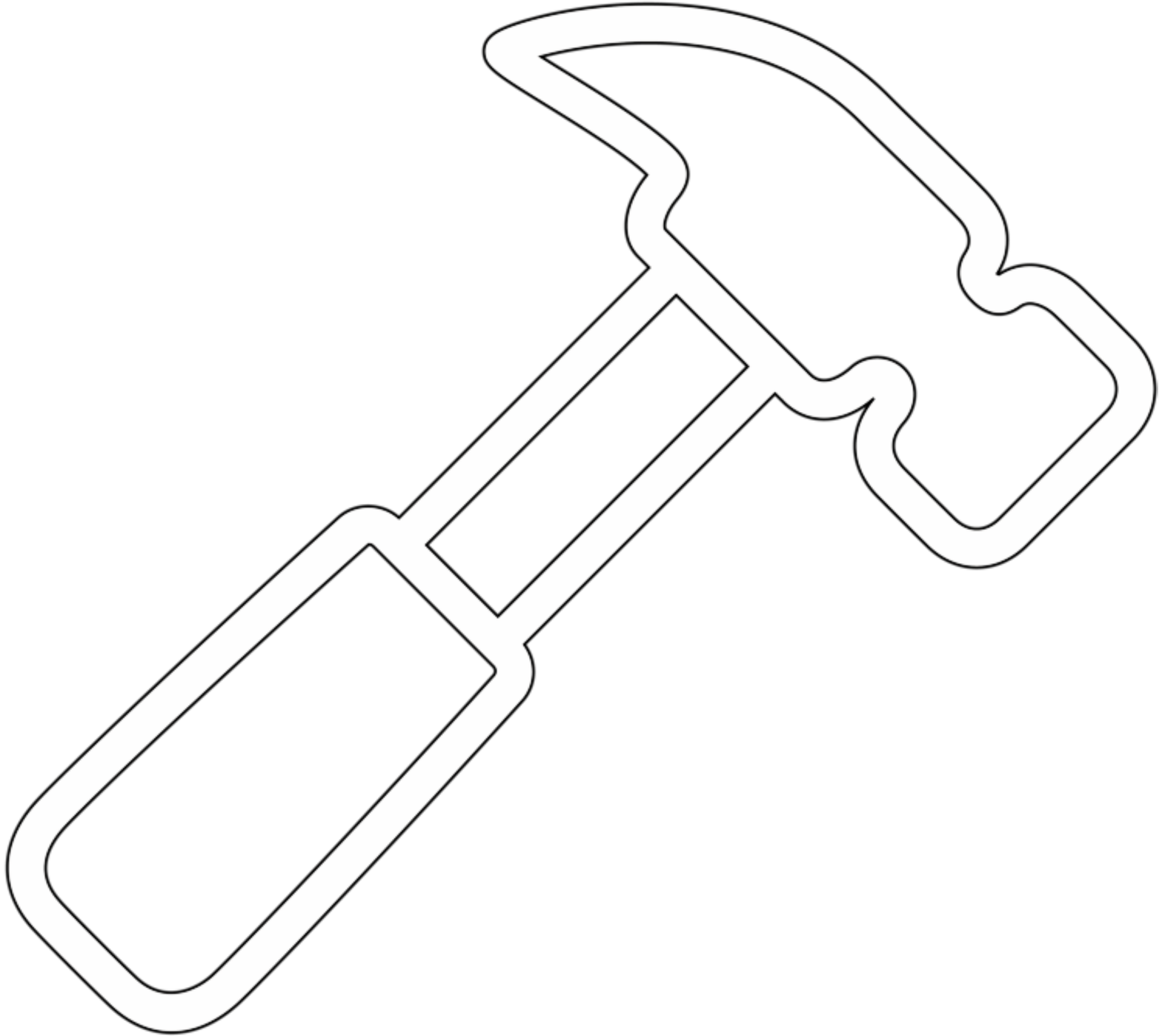}};

        % Directed edges
        \path (x1) edge (y);
        \path (x2) edge (y);
      \end{tikzpicture}
    \end{center}
  \end{tcolorbox}
  \end{minipage}
  \hfill
  \begin{minipage}{0.30\textwidth}
  \begin{tcolorbox}[width=\textwidth, colframe=gray!80, colback=neworange!40, boxsep=0mm, arc=3mm]
    \begin{center}
      {\small (ii) anticausal} \vspace{0.3cm}\\
      \begin{tikzpicture}
        % x node set with absolute coordinates
        \node[state] (x1) at (0,0) {$X_1$};
        \node[state] (x2) at (2,0) {$X_2$};
        \node (blackhammer) at (-0.8,0) {\includegraphics[angle = 0, scale = 0.04]{figures/hammer_black.pdf}};

        % y node set relative to x.
        % Locations can be:
        % right,left,above,below,
        % above left,below right, etc
        \node[state] (y) at (1, -1.7) {$Y$};
        \node (whitehammer) at (0.2,-1.7) {\includegraphics[angle = 0, scale = 0.04]{figures/hammer_white.pdf}};

        % Directed edges
        \path (y) edge (x1);
        \path (y) edge (x2);
      \end{tikzpicture}
    \end{center}
  \end{tcolorbox}
  \end{minipage}
  \hfill
  \begin{minipage}{0.30\textwidth}
  \begin{tcolorbox}[width=\textwidth, colframe=gray!80, colback=gray!40, boxsep=0mm, arc=3mm]
    \begin{center}
      {\small (iii) mixed} \vspace{0.3cm}\\
      \begin{tikzpicture}
        % x node set with absolute coordinates
        \node[state] (x1) at (0,0) {$X_1$};
        \node[state] (x2) at (2,0) {$X_2$};
        \node (blackhammer) at (-0.8,0) {\includegraphics[angle = 0, scale = 0.04]{figures/hammer_black.pdf}};

        % y node set relative to x.
        % Locations can be:
        % right,left,above,below,
        % above left,below right, etc
        \node[state] (y) at (1, -1.7) {$Y$};
        \node (whitehammer) at (0.2,-1.7) {\includegraphics[angle = 0, scale = 0.04]{figures/hammer_white.pdf}};

        % Directed edges
        \path (x1) edge (y);
        \path (y) edge (x2);
      \end{tikzpicture}
    \end{center}
  \end{tcolorbox}
  \end{minipage}
  \hfill

\caption{(i) causal prediction setting where the label $Y$ is the descendant of the covariates $X$. The hammers indicate the location of data perturbation. (2) anti-causal prediction setting where all the covariates $X$ are descendants of the label $Y$. (3) mixed-causal-anticausal setting where some covariates are descendants of the label $Y$ and some covariates are ancestors of $Y$. Would the causal direction of the data generation affect DA performance? Would changing the location of the perturbation (black hammer vs. white hammer) cause some DA methods to fail? }
\label{fig:three_settings}
\end{figure}

Despite the broad need of DA methods in practice, a priori it is impossible to provide a generic solution to DA. In fact, the DA problem is ill-posed if assumptions on the relationship between the source and target datasets are absent. One cannot learn a model with good performance if the target dataset can be arbitrary. As it is often difficult to specify the relationship between the source and target datasets, a large body of existing DA work are driven by applications. This line of work often focuses on developing new DA methods to solve very specific data problems. Despite the empirical success on these specific problems, it is not clear why DA succeeds or how applicable the proposed DA methods are to new related data problems. The growing development of domain adaptation calls forth a theoretical framework to analyze the existing methods and to guide the design of new procedures. More specifically, can one formulate the assumptions needed for a DA method to have a low target error? Can these assumptions be itemized, so that once specified the performance of different DA methods can be compared? More specifically, does the causal direction of the data generation or the type and location of data perturbation (Figure~\ref{fig:three_settings}) affect the choice of the best DA method?

To answer the above questions, this work develops a theoretical framework via structural causal models that enables the comparison of various existing DA methods. Since there are no clear winning DA methods in general, the performance of DA methods has to be analyzed and compared with precise assumptions on the underlying data structure. Through analysis on simple models, we aim to give insights on when and why one DA method outperform others.

\paragraph{Our contributions:} Our contributions are three-fold. First, we develop a theoretical framework via structural causal models (SCM) to analyze and compare the prediction performance of existing DA methods, such as domain invariant projection (DIP)~\citep{pan2010domain,baktashmotlagh2013unsupervised} and conditional invariance penalty (CIP) or conditional transferable component~\citep{gong2016domain,heinze2017conditional}, under precise assumptions relating source and target data. In particular, we show that under linear SCM the popular DA method DIP is guaranteed to have a low target error when the prediction problem is anticausal without label distribution perturbation. However, DIP fails to outperform the estimator trained solely on the source data when there are perturbations on the label distribution or when the prediction problem is either causal or mixed-causal-anticausal. Second, based on our theory, we introduce a new DA method called CIRM and develop its variants which can have better prediction performance than DIP in mixed-causal-anticausal DA scenarios and those with label distribution perturbation. Third, we illustrate via extensive simulations and real data experiments that our theoretical DA framework enables a better understanding of the success and failure of DA methods even in cases where presumably not all assumptions in our theory are satisfied. Our theory and experiments make it clear that knowing the relevant information about the data generation process such as the causal direction or the existence of label distribution perturbation is the key to the success of domain adaptation.

The rest of the paper is organized as follows. In Section~\ref{sec:related_work} we review previous ways to mathematically formulate the DA problem and summarize the existing DA methods. Section~\ref{sec:preliminary_and_problem_setup} contains background on structural causal models, our problem setup, a formal introduction of the DA methods to study and three simple motivating examples. In Section~\ref{sec:domain_adaptation_guarantees} we analyze and compare the performance of three DA methods (DIP, CIP and CIRM) under our theoretical framework. Based on whether the DA problem is causal, anticausal or mixed and whether there is label distribution perturbation, we identify scenarios where these DA methods are guaranteed to have low target errors. Section~\ref{sec:numerical_experiments} contains numerical experiments on synthetic and real datasets to illustrate the validity of our theoretical results and the implication of our theory for practical scenarios.

% section introduction (end)

\section{Related work} % (fold)
\label{sec:related_work}

Domain adaptation is a sub-field of the broader research area of transfer learning. More specifically, domain adaptation is also named transductive transfer learning~\citep{pan2009survey,redko2020survey}. In this paper, we concentrate on the DA problem without diving into the broader field of transfer learning. Consequently, many previous work on transfer learning are omitted for the sake of space. We direct the interested readers to the survey paper by~\cite{pan2009survey} and references therein for a literature review on transfer learning. Focusing on the DA problem, we first review existing theoretical frameworks of DA and then provide an overview of DA methods and algorithms.

\paragraph{Theoretical DA frameworks:}
The DA problem is ill-posed if one does not specify any assumptions on the relationship between the source and target data distribution. For this reason, depending on how this relationship is specified, many ways to formulate the DA problem have been introduced.

\cite{ben2007analysis} were the first to provide a DA prediction performance bound via Vapnik-Chervonenkis (VC) theory for classifiers from a generic hypothesis class. Since this bound is obtained without making explicit assumptions on the relationship between the source and target data distribution, it involves a divergence term that characterizes the closeness of the source and target distribution. A follow-up from \cite{ben2010impossibility} further formally proves the necessity of assuming similarity between source and target distribution to ensure learnability. Ben-David et al.'s work laid the foundation of many further studies that attempt to bound the target and source prediction performance difference via divergence measures~\citep{mansour2009domain,cortes2011domain,cortes2014domain,cortes2015adaptation,hoffman2018algorithms}. See also the survey paper by~\cite{redko2020survey} for a complete review of VC-theory-type DA prediction performance bounds.

One natural way to explicitly relate the source and target data distribution is to assume that both of them are generated via the same well-specified generative model and to treat the missing target label problem as a missing data problem. Given a well-specified probabilistic generative model, the problem of imputing the missing data is well-studied and it is commonly solved via the expectation maximization (EM) algorithm~\cite[see][]{mclachlan2007algorithm}. This idea of casting a DA problem to a missing data problem has been introduced in~\cite{amini2003semi} and~\cite{nigam2006semi}.

It is also possible to loosely relate the source and target data distribution by assuming that they differ only by a small amount. How to specify this ``small amount'' depends on applications. If one assumes that the source data distribution is contaminated so that it is $\epsilon$ total variation distance away from the target data distribution, then the problem goes back to the classical robust statistics literature~\citep{huber1964robust,yuan2019visual}. More recently, Wasserstein distance or $f$-divergence have been considered to describe the difference between source and target data distribution. These work and contributions are referred to as distributional robust learning~\citep{sinha2017certifying,duchi2018learning,gao2017wasserstein}. A closely related line of work directly assumes that target data points can be interpreted as source data points contaminated with arbitrary small additive noise quantified via norm constraints ($\ell_1$ or $\ell_\infty$). This direction is called adversarial machine learning~\citep{goodfellow2018making,raghunathan2018semidefinite}.

Another way to make the DA problem tractable is to assume that the conditional distribution $Y \mid X$ is invariant across source and target data, where $Y$ and $X$ denote the response (label) and covariates, respectively. The only difference between source and target distributions comes from the change in the distribution of the covariates $X$. This type of assumption is called the \textit{covariate shift} assumption~\citep{quionero2009dataset,sugiyama2012machine,storkey2009training}. Alternatively, assuming the other conditional distribution $X \mid Y$ distribution to be invariant is also plausible in certain applications. This approach has a similar name called the \textit{label shift} assumption~\citep{lipton2018detecting,azizzadenesheli2019regularized,garg2020unified}.

Finally on the causality side, it has been pointed out by~\cite{pearl2014external} that full specification of a structural causal model (SCM) allows to study transportability of learning methods on the relationship between variables in the structural causal model. We refer to this approach as full-SCM transfer learning. The full-SCM transfer learning approach is very powerful in describing many data generation models. However, the main drawback of this framework is that the full specification of the structural causal model might be difficult to learn in many applications with limited data. On the other hand, the pioneering work by~\cite{scholkopf2012causal} reveals that distinguishing between the causal and anticausal prediction may already be useful to facilitate the selection of DA and semi-supervised learning methods. Figuring out the right amount of causal information needed to carry out DA is one of our main motivations in this paper.

\paragraph{Previous DA methods:}
While it is in general helpful to have theoretical DA frameworks to relate the source and target data, theory is not essential for the development of new DA methods. A large number of DA methods and algorithms were introduced with the focus of addressing DA for specific datasets. Here we highlight several popular ones.

Self-training~\citep{amini2003semi} is one of the earliest DA methods which is originated in the semi-supervised learning literature~\citep[see the book by][]{chapelle2009semi}. The self-training algorithm begins with an estimator trained on the source data, and gradually labels a part of unlabeled target data and then updates the estimator with appropriate regularization after combining newly labeled target data. It has been shown to have good empirical performance on several computer vision domain adaptation tasks with small labeled source datasets~\citep{xie2019self,carmon2019unlabeled}. A theoretical analysis of the performance of self-training under a gradual shift assumption on Gaussian mixture data was recently provided by~\cite{kumar2020understanding}.

An important line of DA methods relate the source and target data by assuming the existence of a common subspace. \cite{pan2010domain} first came up with the idea of projecting the source and target data onto a reproducing kernel Hilbert space to preserve common properties and applied the idea to text classification datasets. The existence of an intermediate subspace that relates source and target data was made precise in~\cite{gopalan2011domain} for visual objection recognition datasets. This method was further developed and analyzed for sentiment analysis and web-page classification~\citep{blitzer2011domain,gong2012geodesic,muandet2013domain}. \cite{baktashmotlagh2013unsupervised} simplified the idea of enforcing a common subspace to adding a regularization term based on the maximum mean discrepancy (MMD)~\citep{gretton2012kernel}. Their method is named domain invariant projection (DIP) because the regularization term enforces a projection of the source and target data on the subspace to be invariant in distribution. Recently, with the development of deep neural networks and the introduction of generative adversarial nets based distributional distance measures, the common subspace approach was further extended to allow for neural network implementations~\citep{ganin2016domain,peng2019moment}.

The empirical success of DIP like DA methods and their neural network variants such as in~\cite{ganin2016domain} has sparked a wide discussion on the general validity of these methods. \cite{zhao2019learning} constructed a simple counterexample showing that domain invariant projection is not sufficient to guarantee successful domain adaptation. Furthermore, \cite{zhao2019learning} provided a lower bound of the target error when the label distribution is perturbed. \cite{johansson2019support} illustrated via a simple example the danger of the blind use of distributional distance such as MMD for domain invariant penalization. \cite{li2019target} and~\cite{combes2020domain} discussed the failure of DIP in the presence of target label perturbation and proposed label shift correction using conditional invariant features. However, it is not clear whether their proposed algorithms have any guarantees for estimating the conditional invariant features or achieving low target error. The mixed messages about the success and failure of DIP like DA methods motivate us to set up rigorous target error comparisons of DIP and other DA methods.

Another line of DA methods that is worth mentioning is the one that only makes use of source data. \cite{gong2016domain} introduced conditional transferable components which consist of features that are have invariant distributions given the label. The search of conditional transferable components is achieved via a penalty that matches the conditional distribution for any label across source environments.  A related idea was proposed by~\cite{heinze2017conditional}. They extract the conditionally invariant (or core) components (CICs) across data points that share the same identifier but have different style features. The invariance is enforced via adding a conditional variance penalty to the training loss. Enforcing the conditional invariance allows them to learn models that are robust across perturbed computer vision datasets. Later, other concepts of invariance beyond conditional invariance across source environments were developed, such as invariant risk minimization~\citep{arjovsky2019invariant}. It should be noted that, though with a different focus, the idea of using the heterogeneity across multiple source datasets to learn invariant or robust models has also appeared in the causal inference literature~\citep{,peters2016causal,meinshausen2018causality,rothenhausler2018anchor}. Based on these ideas, \cite{rojas2018invariant} and \cite{magliacane2017domain} provided guarantees for domain adaptation under the assumption that the conditional distribution of the label given some subset of covariates is invariant.

There are many other interesting DA methods that are less related to our work. For the sake of space, we direct the interested readers to the book by~\cite{chapelle2009semi} on semi-supervised learning and other surveys~\citep{zhu2005semi,wilson2018survey,wang2018deep} for additional references.

% section related_work (end)

\section{Preliminaries and problem setup} % (fold)
\label{sec:preliminary_and_problem_setup}

In this section, we first provide a brief summary of structural causal models, which are essential components of our theoretical framework. Then we formalize our domain adaptation problem setup and introduce the DA methods we study. Finally, we provide three simple motivating examples to illustrate the need of a DA theory.
\subsection{Background on structural causal models} % (fold)
\label{sub:structural_causal_models}
Structural causal models (SCMs)~\cite[see][]{pearl2000causality} are introduced to describe causal relationships between variables in a system. A SCM integrates the structural equation models (SEMs) used in economics and social sciences, the potential-outcome framework of \cite{neyman1923applications} and \cite{rubin1974estimating}, and the graphical models developed for probabilistic reasoning and causal analysis. A SCM can be seen as a set of generative equations that describe not only the data generation process of the observational data, but also that of the intervention data. We refer the readers to Chapter 7 of the book by~\cite{pearl2000causality} for a detailed description of SCM for causal inference. In the context of DA, SCMs can be used to describe both the data generation process of the source and target domains (or environments). The SCMs are specified via a set of structural equations with a corresponding causal graph to describe the relationship between variables.

While SCMs are very powerful tools to describe data generation processes in interventional environments, fully specifying a SCM for a DA problem has two main drawbacks in practice: first, defining the functional forms that relate variables in a SCM can be difficult for data involving many variables; second, even if the functional forms are specified, learning all the functions from data may result in a more complicated statistical learning task than the original DA problem. Focusing on solving DA problems, the way we address the two main drawbacks differentiates our work from the full-SCM transfer learning approach by \cite{pearl2014external}.

To address the first drawback and to ease our theoretical analysis, we adopt a simplification that replaces all the functional models in the SCM with linear models as in previous work~\citep{peters2016causal,rothenhausler2018anchor,rothenhausler2019causal}. We call this simplified SCM a linear SCM. The simplification allows us to develop rigorous DA theory and to study more complicated DA problems as extensions of the linear case. Regarding the second drawback, we focus on DA methods that can be applied without relying on the full specification of the SCM. Only when we analyze the performance of these DA methods, we bring in SCMs to set forth the assumptions needed for the DA methods to have low target errors.

% subsection structural_causal_models (end)

\subsection{Domain adaptation problem setup} % (fold)
\label{sub:domain_adaptation_problem_setup}
In this subsection, we set up the domain adaptation problem with $\envs$ ($\envs \geq 1$) labeled source environments and one unlabeled target environment. Although SCMs are general enough to also handle classification problems, we focus our theory on regression to simplify the presentation. Later in Section~\ref{sec:numerical_experiments}, we show the adaptation of our results to the classification setting through numerical experiments.

For the $m$-th source environment ($m \in \braces{1, 2, \cdots, \envs}$), we observe $\obs_m$ i.i.d. samples $S^\tagk{m} = \parenth{(x_1^\tagk{m}, y_1^\tagk{m}), \cdots, (x^\tagk{m}_{\obs_m}, y^\tagk{m}_{\obs_m})}$ drawn from the source data distribution $\distri^{\tagk{m}}$, with $(x_k^\tagk{m}, y_k^\tagk{m}) \in \real^{\dims + 1}$ for each $k \in \braces{1, 2, \cdots, \obs_m}$. The $m$-th dataset is also called the $m$-th source environment. Furthermore, there are $\obstar$ i.i.d. samples $\tilde{S} = \parenth{(\tilde{x}_1, \tilde{y}_1), \cdots, (\tilde{x}_{\obstar}, \tilde{y}_{\obstar})}$ from the target distribution $\distritar$, but we only observe the covariates $\tilde{S}_X = \parenth{\tilde{x}_1, \cdots, \tilde{x}_{\obstar}}$ from $\distritar_X$. Here $\distritar_X$ is used to denote the marginal distribution of $\distritar$ on $X$. The goal of the DA problem is to estimate a function $f_\beta: \real^\dims \mapsto \real$, mapping covariates to the label, parametrized by $\beta \in \betaDom$ so that the \textit{target population risk} is ``small''. Here the parameter space $\betaDom$ is a subset of a finite-dimensional space. The performance metric \textit{target population risk} for an estimator $f$ is defined as
\begin{align}
  \label{eq:target_pop_risk}
  \tilde{\risk}(f) = \Exs_{(X, Y) \sim \distritar} \brackets{\loss(f(X), Y)},
\end{align}
where $l$ is a loss function and it is set to the squared loss function $x \mapsto x^2$ if not specified otherwise. Similarly, we can define the $m$-th \textit{source population risk} as
\begin{align}
  \label{eq:source_pop_risk}
  \risk^\tagk{m}(f) = \Exs_{(X, Y) \sim \distriv{m}} \brackets{\loss(f(X), Y)}.
\end{align}

In addition to the risk on an absolute scale, one can quantify the target population risk achieved by an arbitrary estimator on a relative scale by comparing it with the oracle target population risk $\tilde{\risk}(f_{\beta_\text{oracle}})$, where $\beta_\text{oracle}$ is defined as
\begin{align*}
  \beta_\text{oracle} \in \argmin_{\beta \in \betaDom} \Exs_{(X, Y) \sim \distritar} \brackets{\loss(f_\beta(X), Y)}.
\end{align*}

If we don't assume any relationship between the source distribution $\distri^\tagk{m}$ and the target distribution $\distritar$, the target population risk of an estimator learned from the source and unlabeled target data can be arbitrarily larger than the oracle target population risk. Assumptions on the relationship between source and target distribution are needed to make the DA problem tractable. In this work, we consider the DA setting where source and target data are both generated through similar linear SCMs with additional structural assumptions on the interventions.

\paragraph{Domain adaptation under linear SCM with noise interventions: } For $m \in \braces{1, 2, \cdots, \envs}$, the data distribution $\distri^\tagk{m}$ of the $m$-th source environment is specified by the following data generation equations on $\parenth{\Xv{m}, \Yv{m}}$ from $\distri^\tagk{m}$,
\begin{align}
  \label{eq:lin_SEM_noise_intervention_src}
  \bmat{\Xv{m} \\ \Yv{m}} = \bmat{\semBX & \sembv \\ \sembh\tp & 0} \bmat{\Xv{m} \\ \Yv{m}} + \intervtype(\interAv{m}, \noisev{m}),
\end{align}
and the target data distribution $\distritar$ is specified via the same equation except for the noise distribution,
\begin{align}
  \label{eq:lin_SEM_noise_intervention_tar}
  \bmat{\Xtar \\ \Ytar} = \bmat{\semBX & \sembv \\ \sembh\tp & 0} \bmat{\Xtar \\ \Ytar} + \intervtype(\interAtar, \noisetar).
\end{align}
Here $\semBX \in \real^{\dims \times \dims}$ is an unknown constant matrix with zero diagonal such that $\Ind_\dims - {\semBX}$ is invertible, $\sembv \in \real^\dims$ and $\sembh \in \real^\dims$ are unknown constant vectors; $\noisev{m}$ and $\noisetar$ are $\dims + 1$ dimensional random vectors drawn from the same noise distribution $\distrinoise$; $g$ is a fixed function to model the change (or intervention) across source and target environments; $\interAv{m} \in \real^{\dims+1}$ is an unknown (random or non-random) intervention that changes from one environment to another. Note that the assumption on the invertibility of $\Ind_\dims - {\semBX}$ ensures the uniqueness of the data generation given a draw of the noise $\noisev{m}$. This assumption is in general weaker than requiring the corresponding causal graph to be directed acyclic~\citep{rothenhausler2019causal}.

The only difference between the source and target data distribution is due to the difference in intervention $\interAv{m}$ and $\interAtar$. According to the SCM, the way we specify the difference between source and target distribution is through the term $g(\interAtar, \noisetar)$. This type of intervention is often called noise intervention or soft intervention~\citep{eberhardt2007interventions, peters2016causal}. As a concrete example, if the mean shift noise intervention is considered, then we specify the function $g: \real^{\dims+1} \times \real^{\dims+1} \rightarrow \real^{\dims+1}$ as $(\interA, \noise) \mapsto \interA + \noise$,
and define the intervention term with a deterministic vector in $\real^{\dims+1}$. As another example, if the variance shift noise intervention is considered, then we specify $g$ as $(\interA, \noise) \mapsto \interA \odot \noise$,
where $\odot$ is the element-wise product. We focus our theoretical results on the mean shift noise intervention. Other types of noise interventions are discussed in numerical experiments. The linear SCM with noise interventions clearly does not cover all kinds of perturbations to the data. For example, our theory does not apply when the matrix $\semBX$ is no longer invariant across source and target environments. We show via numerical experiments that assuming that the perturbations are due to noise interventions is plausible in many settings.

% subsection domain_adaptation_problem_setup (end)

\subsection{Oracle and baseline DA methods} % (fold)
\label{sub:baseline_da_methods}
As we aim to compare existing and new DA methods rigorously under our theoretical framework, it is useful to start with several estimators to establish the basis of comparison.

First, we introduce two oracle estimators. They are defined using the unobserved information such as target labels or SCM parameters.
\begin{itemize}
  \item \textbf{OLSTar:} the population ordinary least squares (OLS) estimator on the target data.
  \begin{align}
    \label{eq:estimator_pop_olstarget}
    f_{\tagolstarget}(x) &\defn x \tp\betaolstarget + \beta_{\tagolstarget, 0} \notag \\
    \betaolstarget, \beta_{\tagolstarget, 0}  &\defn \argmin_{\beta, \beta_0} \Exs_{(X, Y) \sim \distritar}\parenth{Y - X\tp\beta - \beta_0}^2.
  \end{align}
  This is the oracle target population estimator when we restrict the function class to be linear. Hence the target risk of OLSTar defines the lowest target risk that any linear DA estimator can achieve.
  \item \textbf{Causal:} the population causal estimator via the linear SCM
  \begin{align}
    \label{eq:estimator_pop_causal}
    f_{\tagcausal}(x) &\defn x \tp\betacausal \notag \\
    \betacausal &\defn \sembh,
  \end{align}
  where $\sembh$ appeared in the last row of the SCM matrix in Equation~\eqref{eq:lin_SEM_noise_intervention_src}.
  Note that this formulation of the causal estimator assumes that there is no intervention on $Y$ and the intercept is also zero. The Causal estimator is closely related to distributional robust estimators. That is, the Causal estimator is the robust estimator which achieves the minimum worse-case risk when the perturbations on the covariates are allowed to be arbitrary~\citep{buhlmann2020invariance}. However, in our DA setting where observed target covariates provide additional information, it is no longer clear whether Causal still achieves the lowest target risk.
\end{itemize}
Second, we introduce two population estimators that only use the source data.
\begin{itemize}
  \item \textbf{OLSSrc$^\tagk{m}$:} the population OLS estimator on the single $m$-th source environment.
  \begin{align}
    \label{eq:estimator_pop_olssource1}
    f_{\tagolssource}^\tagk{m}(x) &\defn x \tp\betaolssourceone{m} + \beta_{\tagolssource, 0}^\tagk{m} \notag \\
    \betaolssourceone{m}, \beta_{\tagolssource, 0}^\tagk{m} &\defn \argmin_{\beta, \beta_0} \Exs_{(X, Y) \sim \distriv{m}}\parenth{Y - X\tp\beta - \beta_0}^2.
  \end{align}
  \item \textbf{SrcPool:} the population OLS estimator by pooling all source data together.
  \begin{align}
    \label{eq:estimator_pop_olspool}
    f_{\tagolspool}(x) &\defn x \tp\betaolspool + \beta_{\tagolspool, 0} \notag \\
    \betaolspool, \beta_{\tagolspool, 0} &\defn \argmin_{\beta, \beta_0} \Exs_{(X, Y) \sim \distri^{\text{allsrc}}}\parenth{Y - X\tp\beta - \beta_0}^2,
  \end{align}
  where $\distri^{\text{allsrc}}$ is the uniform mixture over $\envs$ source distributions $\braces{\distriv{m}}_{m = \braces{1, \cdots, \envs}}$. We omitted the SrcPool formulation with weighted mixtures because it is not the main focus of our study.
\end{itemize}
These two estimators are natural estimators in the classical statistical learning setting when the source and target data share the same distribution. A DA method that has larger target risk than SrcPool is clearly unfavorable. One goal throughout our paper is to understand under which conditions DA methods are guaranteed to outperform OLSSrc$^\tagk{m}$ and SrcPool.

% subsection baseline_da_methods (end)

\subsection{Advanced DA methods} % (fold)
\label{sub:advanced_da_methods}
In this subsection, we introduce three advanced DA methods. The first two have been introduced previously and the third one is our new DA method.

First, we consider the DA method called \textit{domain invariant projection} (DIP). DIP is a subspace-based DA method which aims at learning a common intermediate subspace that relates the source and target domains. The specific form of DIP we study follows from~\cite{baktashmotlagh2013unsupervised}. Specifically, the population DIP estimator involves the following optimization problem
\begin{align}
  \label{eq:dip_general_form}
  f_{\text{DIP}}(x) &\defn u_{\text{DIP}}\circ v_{\text{DIP}} (x) \notag \\
  u_{\text{DIP}}, v_{\text{DIP}} &\defn \argmin_{u \in \mathcal{U}, v \in \mathcal{V}} \Exs_{(X, Y) \sim \distriv{1}, \tilde{X} \sim \distritar} \loss{\parenth{u \circ v(X), Y} + \lambda \cdot \mathcal{D}(v(X), v(\tilde{X}))},
\end{align}
where $\mathcal{D}(\cdot, \cdot)$ measures the distance between two distribution, $\mathcal{U}$ and $\mathcal{V}$ are function classes that are specified through expert knowledge of the problem and $\lambda$ is a positive regularization parameter. DIP, in its simple form, only uses a single source environment. Hence, without loss of generality, we used the first source data environment.

In~\cite{baktashmotlagh2013unsupervised}, maximum mean discrepancy (MMD) is used as the distributional distance measure and both $\mathcal{U}$ and $\mathcal{V}$ are set to be linear mappings. This DIP idea of matching the mappings of source and target data is extended later in many DA papers with various choices of distance and function classes~\cite[see e.g.][]{ghifary2016scatter,li2018domain}. A noteworthy line of follow-up work consist of replacing the distribution distance and function classes in DIP with neural networks. For example, \cite{ganin2016domain} introduced domain-adversarial neural network (DANN) which uses generative adversarial nets (GAN) in place of MMD to measure distributional distance and makes both $\mathcal{U}$ and $\mathcal{V}$ to be neural networks.

Analyzing the most generic form of DIP is out of the scope of this study. Instead, we start with a simple DIP formulation.
\begin{itemize}
  \item \textbf{DIP$^\tagk{m}$-mean:} the population DIP estimator where mean squared difference is used as distributional distance, $\mathcal{V}$ is linear and $\mathcal{U}$ is the singleton of the identity mapping and $\lambda$ is chosen to be $\infty$. DIP, in its simple form, only uses the data from one source environment and the target covariates. This form of DIP is defined as
  \begin{align}
    \label{eq:estimator_pop_dipmeanmatchlin}
    f_{\tagdipmeanmatchlin}^\tagk{m}(x) &\defn x \tp\betadipmeanmatchlin{m} + \beta_{\tagdipmeanmatchlin, 0}^\tagk{m} \notag \\
    \betadipmeanmatchlin{m}, \beta_{\tagdipmeanmatchlin, 0}^\tagk{m} &\defn \argmin_{\beta, \beta_0}\  \Exs_{(X, Y) \sim \distriv{m}}\parenth{Y - X\tp\beta - \beta_0}^2 \notag \\
    &\text{s.t.\ } \Exs_{X \sim \distriv{m}_X} \brackets{X \tp\beta} = \Exs_{X \sim \distritar_X}\brackets{X \tp\beta}.
  \end{align}
  For simplicity, we use the shorthand notation DIP$^\tagk{m}$ to refer to DIP$^\tagk{m}$-mean. The constraint in Equation~\eqref{eq:estimator_pop_dipmeanmatchlin} is called the DIP matching penalty.
\end{itemize}

Second, we introduce the \textit{conditional invariant penalty} (CIP) estimator. Unlike DIP which projects the source and target covariates to the same subspace, CIP directly uses the label information in multiple source environments to look for the conditionally invariant components. CIP only makes use of source data.

\begin{itemize}
  \item \textbf{CIP-mean:} the population conditional invariance penalty (CIP) estimator where the conditional mean is matched across multiple source environments.
  \begin{align}
    \label{eq:estimator_pop_cipmean}
    f_{\tagcipmean}(x) &\defn x \tp\betacipmean + \beta_{\tagcipmean, 0} \notag \\
    \betacipmean, \beta_{\tagcipmean, 0} &\defn \argmin_{\beta, \beta_0}\ \frac{1}{\envs} \sum_{m=1}^\envs \Exs_{(X, Y) \sim \distriv{m}}\parenth{Y - X\tp\beta - \beta_0}^2 \notag \\
    \text{s.t.\ } \Exs_{(X, Y) \sim \distriv{m}} &\brackets{X \tp\beta \mid Y} = \Exs_{(X, Y) \sim \distriv{1}} \brackets{X \tp\beta \mid Y} a.s., \forall m \in \braces{2, \cdots, \envs},
  \end{align}
  where the equality between the conditional expectation is in the sense of almost sure equality of random variables. Note that CIP naturally requires $\envs \geq 2$, because when $\envs=1$ the CIP constraint becomes vacuous.
  For simplicity, we use the shorthand notation CIP to refer to CIP-mean.
\end{itemize}
CIP puts more regression weights on the conditionally invariant components from multiple source datasets via the conditional invariance constraint in Equation~\eqref{eq:estimator_pop_cipmean}.
The idea of conditional invariance penalty in the context of anticausal learning has appeared in multiple papers with slightly different settings. \cite{gong2016domain} introduced conditional transferable components which are in fact conditionally invariant features. However, unlike the formulation above, \cite{gong2016domain} propose to learn the conditional transferable components with only one source environment which requires assumptions that are difficult to check. On the other hand, the algorithm from \cite{heinze2017conditional} learns the conditionally invariant features from a single source dataset if multiple observations of data points that share the same identifier are present. They use their conditional variance penalty to enforce their algorithm to learn the conditionally invariant features in their specific datasets with identifiers. The same idea can be applied if we replace identifiers with source environments.

Third, we introduce our new DA estimator \textit{conditional invariant residual matching} (CIRM).

\begin{itemize}
  \item \textbf{CIRM$^\tagk{m}$-mean:} the population conditional invariant residual matching estimator that uses all source environments to compute CIP and the $m$-th source environment to perform risk minimization.
  \begin{align}
    \label{eq:estimator_pop_cirmeanmatch}
    f_{\tagcirmeanmatch}^\tagk{m}(x) &\defn x \tp\betacirmeanmatch{m} + \beta_{\tagcirmeanmatch, 0}^\tagk{m} \notag \\
    \betacirmeanmatch{m}, \beta_{\tagcirmeanmatch, 0}^\tagk{m} &\defn \argmin_{\beta, \beta_0}\  \Exs_{(X, Y) \sim \distriv{m}}\parenth{Y - X\tp\beta - \beta_0}^2 \notag \\
    \text{s.t.\ } \Exs_{X \sim \distriv{m}_X} &\brackets{\beta \tp \parenth{ X - \parenth{X \tp\betacipmean} \bcirmeanmatch }} = \Exs_{X \sim \distritar_X} \brackets{\beta \tp \parenth{ X - \parenth{X \tp\betacipmean} \bcirmeanmatch }},
\end{align}
  where
  \begin{align}
    \label{eq:estimator_pop_cirmeanmatch_b_choice}
    \bcirmeanmatch \defn \frac{\Exs_{(X, Y)\sim \distri^{\text{allsrc}} }\brackets{X \cdot (Y - \Exs[Y])}}{ \Exs_{(X, Y)\sim \distri^{\text{allsrc}} } \brackets{\parenth{X\tp \betacipmean - \Exs[X\tp \betacipmean] } \cdot \parenth{Y-\Exs[Y]}}},
  \end{align}
  with $\distri^{\text{allsrc}}$ denoting the uniform mixture of all source distributions.
  For simplicity, we use the shorthand notation CIRM$^\tagk{m}$ to refer to CIRM$^\tagk{m}$-mean.
\end{itemize}
At first glance, the CIRM estimator is a combination of DIP and CIP. CIRM first uses CIP to compute a linear combination of conditionally invariant components to serve as a proxy of the label $Y$. To tackle the label distribution perturbation, CIRM then performs the DIP-type matching on the residual obtained by regressing $Y$ on its proxy. We can show that the joint distribution of the residual together with the covariates do not suffer from the label distribution perturbation. CIRM is designed to be applied in DA scenarios with label distribution perturbation. Note that the idea of using domain-invariant features to tackle label distribution shift was proposed previously in~\cite{li2019target} and in~\cite{combes2020domain}, while it was not clear how to formulate the exact algorithm so that its success and failure conditions can be analyzed. The intuition behind the CIRM construction becomes clearer after we state Theorem~\ref{thm:mutiple_source_anticausal_mean_shift}.

The comparison of DA methods in the following sections is centered around DIP, CIP, CIRM and their variants. To keep track of these variants, we provide a summary of all DA methods appeared in this paper in Table~\ref{tab:summary_pop_da_methods} of Appendix~\ref{sub:population_DA_methods}.

% subsection advanced_da_methods (end)

\subsection{Simple motivating examples} % (fold)
\label{sub:simple_motivating_examples}
Before we dive into theoretical comparisons of the DA methods, we go through three simple examples to illustrate the assumptions needed for DA methods to have low target risks. The simple examples have data generated via low-dimensional SCMs so that the DA estimators can be easily computed and understood.

\subsubsection{Example 1: causal prediction} % (fold)
\begin{figure}[ht]
  \centering
  \begin{minipage}{0.30\textwidth}
  \begin{tcolorbox}[width=\textwidth, colframe=gray!80, colback=neworange!40, boxsep=0mm, arc=3mm]
  \begin{tikzpicture}
    % x node set with absolute coordinates
    \node[state] (x1) at (0,0) {$X_1$};
    \node[state] (x2) at (0,1) {$X_2$};
    \node[state] (x3) at (0,2) {$X_3$};
    \node[regular polygon,regular polygon sides=4,draw,fill=pink] (i) at (0,3.5) {$A$};

    % y node set relative to x.
    % Locations can be:
    % right,left,above,below,
    % above left,below right, etc
    \node[state] (y) [right =of x1] {$Y$};

    % Directed edges
    \path (x2) edge (y);
    \path (x1) edge (y);

    \path (i) edge[dashed, bend right=60] (x1);
    \path (i) edge[dashed, bend right=60] (x2);
    \path (i) edge[dashed, bend right=60] (x3);
  \end{tikzpicture}
  \end{tcolorbox}
  \end{minipage}
  \begin{minipage}{0.66\textwidth}
  \begin{equation*}
    \begin{aligned}[c]
      \Xv{1}_1 &= \noisev{1}_{X_1} + 1 \\
      \Xv{1}_2 &= \noisev{1}_{X_2} + 1 \\
      \Xv{1}_3 &= \noisev{1}_{X_3} + 1 \\
      \Yv{1} &= \Xv{1}_1 + \Xv{1}_2 + \noisev{1}_Y,
    \end{aligned}
    \hspace{1cm}
    \begin{aligned}[c]
      \Xtar_1 &= \noisetar_{X_1} - 1 \\
      \Xtar_2 &= \noisetar_{X_2} - 1 \\
      \Xtar_3 &= \noisetar_{X_3} + 1 \\
      \Ytar &= \Xtar_1 + \Xtar_2 + \noisetar_Y,
    \end{aligned}
  \end{equation*}
  \end{minipage}
  \caption{The causal diagram and structural equations for the source and target environments in Example 1: causal prediction.}
  \label{fig:causal_diagram_predict_effect_from_cause}
\end{figure}
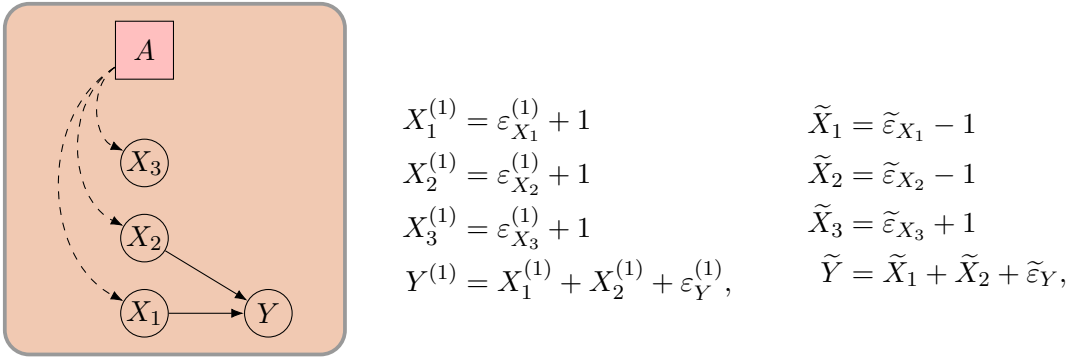

\label{ssub:example_1_causal_prediction}
Example 1 has one source environment and one target environment. The data in the source and target environment are generated independently according to the following SCMs, with the causal diagram on the left and the structural equations on the right of Figure~\ref{fig:causal_diagram_predict_effect_from_cause}.

Here the noise variables follow independent Gaussian distributions with mean zero and variance $0.1$ for $X$ and variance $0.2$ for $Y$, namely, $
  \noisev{1}_{X_1}, \noisev{1}_{X_2}, \noisev{1}_{X_3}, \noisetar_{X_1}, \noisetar_{X_2}, \noisetar_{X_3} \sim \Normal(0, 0.1)$, $\noisev{1}_Y, \noisetar_Y \sim \Normal(0, 0.2)$.
The type of intervention is mean shift noise intervention. That is, the function $g$ in Equation~\eqref{eq:lin_SEM_noise_intervention_src} is taken to be $g: (\interAv{1}, \noisev{1}) \mapsto \interAv{1} + \noisev{1}$. The intervention is $\interAv{1} = \bmat{1 & 1 & 1 & 0} \tp$ for the source environment and it is $\interAtar = \bmat{-1 & -1 & 1 & 0} \tp$ for the target environment. This example is called causal prediction, because the covariates $X$ are parents of the label $Y$. In other words, we are predicting the effect from the causes as illustrated in the causal diagram in Figure~\ref{fig:causal_diagram_predict_effect_from_cause}.

Given the source and target distribution, the population DA estimators in the previous subsection can be computed explicitly. We obtain
\begin{align*}
  \bmat{\betaolstarget \\ \beta_{\tagolstarget, 0}} &= \bmat{1 & 1 & 0 & 0} \tp,
  &\bmat{\betacausal \\ \beta_{\tagcausal, 0}} &= \bmat{1 & 1 & 0 & 0} \tp, \\
  \bmat{\betaolssourceone{1}\\ \beta_{\tagolssource, 0}^\tagk{1}} &= \bmat{1 & 1 & 0 & 0} \tp,
  &\bmat{\betadipmeanmatchlin{1} \\ \beta_{\tagdipmeanmatchlin, 0}^\tagk{1}} &= \bmat{\frac{1}{3} & \frac{1}{3} & -\frac{2}{3} & -2} \tp.
\end{align*}
Note that OLSSrc$^\tagk{1}$, Causal and OLSTar share the same estimate $\beta$, while DIP$^\tagk{1}$ does not.
The corresponding population source and target risks are summarized in the first two rows of Table~\ref{tab:population_risks_motivating_examples}. DIP$^\tagk{1}$ has a larger target population risk than OLSSrc$^\tagk{1}$. In this example of causal prediction, using the additional target covariate information via DIP is making the target prediction performance worse. This is because DIP matching penalty in Equation~\eqref{eq:estimator_pop_dipmeanmatchlin} which is not satisfied by the oracle estimator OLSTar, makes DIP too restrictive.
% subsubsection example_1_predicting_effect_from_cause (end)
\begin{table}[ht]
    \centering
    % \begin{adjustwidth}{-.2in}{-.3in}
    % {\renewcommand{\arraystretch}{.5}
    \begin{tabular}{c|ccccc}
        \toprule
        % \thead{\bf \# Grad. evals $\to$ \\
        % \bf Algorithm $\downarrow$}
        \backslashbox{\thead{\bf Risk}}{\thead{\bf Methods}} & \thead{\bf OLSTar \\ (oracle)} & \thead{\bf Causal} & \thead{\bf OLSSrc$^\tagk{1}$} & \thead{\bf DIP$^\tagk{1}$} & \thead{\bf DIPAbs$^\tagk{1}$} \\ [4mm] \midrule
        \thead{Ex 1, source risk $\risk^\tagk{1}$} & 0.200 & 0.200 & 0.200 & 0.333 & -  \\ [2mm]
        \thead{Ex 1, target risk $\tilde{\risk}$} & {\bf 0.200} & 0.200 & 0.200 & 16.333 & - \\ [2mm] \midrule
        \thead{Ex 2, source risk $\risk^\tagk{1}$} & 2.600 & 0.200 & 0.040 & 0.086 & 0.044 \\ [2mm]
        \thead{Ex 2, target risk $\tilde{\risk}$} & {\bf 0.040} & 0.200 & 2.600  & 0.086 & 0.667 \\ [2mm] \midrule
        \thead{Ex 3, source risk $\risk^\tagk{1}$} & 0.200 & 0.200 & 0.040 & 0.066 & -  \\ [2mm]
        \thead{Ex 3, target risk $\tilde{\risk}$} & {\bf 0.040} & 1.200 & 0.200 & 4.066 & -\\
         [2mm]
    \bottomrule
    \end{tabular}
    % }
    % \end{adjustwidth}
    \caption{Population source and target risks in two motivating examples (the lower the better). The oracle target risk is highlighted in bold. In Example 1 of causal prediction, DIP$^\tagk{1}$ performs worse than Causal and OLSSrc$^\tagk{1}$ in terms of target population risk. In Example 2 of anticausal prediction, DIP$^\tagk{1}$ performs better than Causal and OLSSrc$^\tagk{1}$, but it is still not as good as the oracle estimator. DIPAbs$^\tagk{1}$ is better in terms of population source risk can have worse target population risk than Causal. In Example 3 of anticausal prediction with intervention on $Y$, DIP$^\tagk{1}$ performs worse than OLSSrc$^\tagk{1}$ in terms of target population risk. }
    \label{tab:population_risks_motivating_examples}
\end{table}

% subsubsection example_1_causal_prediction (end)

\subsubsection{Example 2: anticausal prediction} % (fold)
\label{ssub:example_2_anti_causal_prediction}
Example 2 has one source environment and one target environment. The data in the source and target environment are generated independently according to the following SCMs, with the causal diagram on the left and the structural equations on the right of Figure~\ref{fig:causal_diagram_predict_cause_from_effect}.
\begin{figure}[ht]
  \centering
  \begin{minipage}{0.30\textwidth}
  \begin{tcolorbox}[width=\textwidth, colframe=gray!80, colback=newblue!40, boxsep=0mm, arc=3mm]
  \begin{tikzpicture}
    % x node set with absolute coordinates
    \node[state] (x1) at (0,0) {$X_1$};
    \node[state] (x2) at (0,1) {$X_2$};
    \node[state] (x3) at (0,2) {$X_3$};
    \node[regular polygon,regular polygon sides=4,draw,fill=pink] (i) at (0,3.5) {$A$};

    % y node set relative to x.
    % Locations can be:
    % right,left,above,below,
    % above left,below right, etc
    \node[state] (y) [right =of x1] {$Y$};

    % Directed edge
    % \path (y) edge (x3);
    \path (y) edge (x2);
    \path (y) edge (x1);

    \path (i) edge[dashed, bend right=60] (x1);
    \path (i) edge[dashed, bend right=60] (x2);
    \path (i) edge[dashed, bend right=60] (x3);
  \end{tikzpicture}
  \end{tcolorbox}
\end{minipage}
\begin{minipage}{0.66\textwidth}
\begin{equation*}
  \begin{aligned}[c]
    \Xv{1}_1 &= \Yv{1} + \noisev{1}_{X_1} + 1 \\
    \Xv{1}_2 &= \Yv{1} + \noisev{1}_{X_2} + 1\\
    \Xv{1}_3 &= \noisev{1}_{X_3} + 1\\
    \Yv{1} &= \noisev{1}_Y,
  \end{aligned}
  \hspace{2cm}
  \begin{aligned}[c]
    \Xtar_1 &= \Ytar + \noisetar_{X_1} - 1 \\
    \Xtar_2 &= \Ytar + \noisetar_{X_2} - 1 \\
    \Xtar_3 &= \noisetar_{X_3} - 1 \\
    \Ytar &= \noisetar_Y,
  \end{aligned}
\end{equation*}
\end{minipage}
  \caption{The causal diagram and structural equations for the source and target environments in Example 2: anticausal prediction}
  \label{fig:causal_diagram_predict_cause_from_effect}
\end{figure}

Here the noise variables follow independent Gaussian distributions with mean zero and variance $0.1$ for $X$ and variance $0.2$ for $Y$, namely, $
  \noisev{1}_{X_1}, \noisetar_{X_1}, \noisev{1}_{X_2}, \noisetar_{X_2}, \noisev{1}_{X_3}, \noisetar_{X_3} \sim \Normal(0, 0.1)$, $\noisev{1}_Y, \noisetar_Y \sim \Normal(0, 0.2).$
The type of intervention $g$ is mean shift noise intervention, which is the same as in Example 1. The intervention is $\interAv{1} = \bmat{1 & 1 & 1 & 0} \tp$ for the source environment and it is $\interAtar = \bmat{-1 & -1 & -1 & 0} \tp$ for the target environment. Compared to Example 1, the main difference is that the causal direction between the covariates $X$ and the label $Y$ has changed. This example is called anticausal prediction, because the covariates $X$ are descendants of the label $Y$. We are predicting the cause from the effects as illustrated in the causal diagram in Figure~\ref{fig:causal_diagram_predict_cause_from_effect}.

In addition to the DA estimators used in the previous example, we introduce one more variant of DIP, \textit{DIPAbs}. It is a made-up estimator to show that arbitrary choice of the function classes $\mathcal{U}, \mathcal{V}$ in DIP formulation~\eqref{eq:dip_general_form} without consideration on the data generation process is in general a bad idea.
\begin{itemize}
  \item \textbf{DIPAbs$^\tagk{m}$-mean:} the population DIP estimator where mean squared difference is used as distributional distance, $\mathcal{V}$ is element-wise absolute value followed linear mapping and $\mathcal{U}$ is singleton of identity mapping and regularization parameter $\lambda$ is chosen to be $\infty$. For the $m$-th source environment, it is defined as
  \begin{align}
    \label{eq:estimator_pop_dipmeanmatchabs}
    f_{\tagdipmeanmatchabs}^\tagk{m}(x) &\defn \abss{x} \tp\betadipmeanmatchabs{m} + \beta_{\tagdipmeanmatchabs, 0}^\tagk{m} \notag \\
    \betadipmeanmatchabs{m}, \beta_{\tagdipmeanmatchabs, 0}^\tagk{m} &\defn \argmin_{\beta, \beta_0}\  \Exs_{(X, Y) \sim \distriv{m}}\parenth{Y - \abss{X}\tp\beta - \beta_0}^2 \notag\\
    & \text{s.t.\ } \Exs_{X \sim \distriv{m}_X} \brackets{\abss{X} \tp\beta} = \Exs_{X \sim \distritar_X}\brackets{\abss{X} \tp\beta}.
  \end{align}
\end{itemize}
The population estimators can be computed explicitly
\begin{align*}
  \bmat{\betaolstarget \\ \beta_{\tagolstarget, 0}} &= \bmat{\frac{2}{5} & \frac{2}{5} & 0 & \frac{4}{5}} \tp,
  &\bmat{\betacausal \\ \beta_{\tagcausal, 0}} &= \bmat{0 & 0 & 0 & 0} \tp, \\
  \bmat{\betaolssourceone{1} \\ \beta_{\tagolssource, 0}^\tagk{1}} &= \bmat{\frac{2}{5} & \frac{2}{5} & 0 & -\frac{4}{5}} \tp,
  &\bmat{\betadipmeanmatchlin{1} \\ \beta_{\tagdipmeanmatchlin, 0}^\tagk{1}} &= \bmat{\frac{2}{7} & \frac{2}{7} & -\frac{4}{7} & 0} \tp, \\
  \bmat{\betadipmeanmatchabs{1} \\ \beta_{\tagdipmeanmatchabs, 0}^\tagk{1}} &= \bmat{-\frac{2}{5} & -\frac{2}{5} & 0 & \frac{4}{5}} \tp.
\end{align*}
For the other four estimators, none of the estimated $\beta$ perfectly agrees with that of the oracle OLSTar.
The corresponding population source and target risks are summarized in the third and fourth rows of Table~\ref{tab:population_risks_motivating_examples}. DIP$^\tagk{1}$ improves upon Causal and OLSSrc$^\tagk{1}$ on the target population risk. However its target population risk is not as small as that of the oracle estimator OLSTar. DIPAbs$^\tagk{1}$ also uses the target covariate information as DIP$^\tagk{1}$ does, but DIPAbs$^\tagk{1}$ performs worse than Causal or DIP$^\tagk{1}$ in terms of target population risk.

% subsubsection example_2_anti_causal_prediction (end)

\subsubsection{Example 3: anticausal prediction when Y is intervened on} % (fold)
\label{ssub:example_3_anti_causal_prediction_when_y_is_intervened_on}
Example 3 is also an anticausal prediction problem similar to Example 2, except that the label $Y$ is also intervened on. The data in the source and target environment are generated independently according to the following SCMs, with the causal diagram on the left and the structural equations on the right of Figure~\ref{fig:causal_diagram_predict_cause_from_effect_intervY}.
\begin{figure}[ht]
  \centering
  \begin{minipage}{0.30\textwidth}
  \begin{tcolorbox}[width=\textwidth, colframe=gray!80, colback=newblue!40, boxsep=0mm, arc=3mm]
  \begin{tikzpicture}
    % x node set with absolute coordinates
    \node[state] (x1) at (0,0) {$X_1$};
    \node[state] (x2) at (0,1) {$X_2$};
    \node[regular polygon,regular polygon sides=4,draw,fill=pink] (i) at (0,3) {$A$};

    % y node set relative to x.
    % Locations can be:
    % right,left,above,below,
    % above left,below right, etc
    \node[state] (y) [right =of x1] {$Y$};

    % Directed edge
    % \path (y) edge (x3);
    \path (y) edge (x2);
    \path (y) edge (x1);

    % \path (i) edge[bend right=60] (x1);
    \path (i) edge[dashed, bend left=30] (y);
    \path (i) edge[dashed, bend right=60] (x1);
    \path (i) edge[dashed, bend right=60] (x2);
  \end{tikzpicture}
  \end{tcolorbox}
  \end{minipage}
  \begin{minipage}{0.66\textwidth}
  \begin{equation*}
    \begin{aligned}[c]
        \Xv{1}_1 &= \Yv{1} + \noisev{1}_{X_1} + 1 \\
        \Xv{1}_2 &= -\Yv{1} + \noisev{1}_{X_2} + 1 \\
        \Yv{1} &= \noisev{1}_Y + 1,
    \end{aligned}
    \hspace{2cm}
    \begin{aligned}[c]
        \Xtar_1 &= \Ytar + \noisetar_{X_1} - 1 \\
        \Xtar_2 &= -\Ytar + \noisetar_{X_2} - 1 \\
        \Ytar &= \noisetar_Y - 1,
    \end{aligned}
  \end{equation*}
  \end{minipage}
  \caption{The causal diagram and structural equations for the source and target environments in Example 3: anticausal prediction when Y is intervened on}
  \label{fig:causal_diagram_predict_cause_from_effect_intervY}
\end{figure}

Here the noise variables follow independent Gaussian distributions with $
  \noise_{X_1}, \noisetar_{X_1} \sim \Normal(0, 0.1)$, $\noise_{X_2}, \noisetar_{X_2} \sim \Normal(0, 0.1)$, $\noise_Y, \noisetar_Y \sim \Normal(0, 0.2)$.
The type of intervention is still mean shift noise intervention. The intervention is $\interAv{1} = \bmat{1 & 1 & 1} \tp$ for the source environment and it is $\interAtar = \bmat{-1 & -1 & -1} \tp$ for the target environment. Compared to Example 2, the main difference is that the intervention on $Y$ is nonzero as illustrated in the causal diagram in Figure~\ref{fig:causal_diagram_predict_cause_from_effect_intervY}.

The population estimators can be computed explicitly
\begin{align*}
  \bmat{\betaolstarget \\ \beta_{\tagolstarget, 0}} &= \bmat{\frac{2}{5} & -\frac{2}{5} & -\frac{1}{5}} \tp,
  &\bmat{\betacausal \\ \beta_{\tagcausal, 0}} &= \bmat{0 & 0 & 0} \tp, \\
  \bmat{\betaolssourceone{1} \\ \beta_{\tagolssource, 0}^\tagk{1}} &= \bmat{\frac{2}{5} & -\frac{2}{5} & \frac{1}{5}} \tp,
  &\bmat{\betadipmeanmatchlin{1} \\ \beta_{\tagdipmeanmatchlin, 0}^\tagk{1}} &= \bmat{0 & -\frac{2}{3} & 1} \tp.
\end{align*}
Note that DIP$^\tagk{1}$ has zero weight on the first coordinate but non-zero weight on the second because $\Xv{1}_2$ and $\Xtar_2$ share the same distribution.
The corresponding population source and target risks are summarized in the last two rows of Table~\ref{tab:population_risks_motivating_examples}. In Example 3 when $Y$ is intervened on, DIP$^\tagk{1}$ is again worse than OLSSrc$^\tagk{1}$ on the target population risk.
% subsubsection example_3_anti_causal_prediction_when_y_is_intervened_on (end)

\subsubsection{Lessons from the simple motivating examples} % (fold)
\label{ssub:lessons_from_the_simple_motivating_examples}
The three simple motivating examples reveal three observations. First, even though DIP has a low target risk in the anticausal prediction setting (Example 2), DIP is not likely to outperform OLSSrc in the causal prediction setting (Example 1). The fact that the additional target covariate information is not useful in causal prediction problem was previously pointed out by~\cite{scholkopf2012causal}. Consequently, DIP-type methods~\citep{baktashmotlagh2013unsupervised,ganin2016domain} which make use of the additional target covariate information can make target performance worse in causal prediction problems. Second, not all DIP variants have low target risks in the anticausal prediction setting. DIPAbs$^\tagk{1}$, despite using the target covariate information as DIP$^\tagk{1}$ does, performs worse than Causal and DIP$^\tagk{1}$. In general, it is dangerous to treat DA methods as generic solutions that always work without consideration on the data generation process. We remark that DIPAbs$^\tagk{1}$ is a made-up method. But one could imagine a case where the first part $\mathcal{V}$ in DIP$^\tagk{1}$ is set to be a neural network that can approximate a large class of functions, then it is no longer clear the DIP matching penalty that matches the source and target covariate distributions always helps to improve the target population risk. This danger of learning blindly invariant representations was pointed out previously via a simple nonlinear data generation model by~\cite{zhao2019learning}. Third, when $Y$ is intervened on, DIP can perform worse than OLSSrc. Actually, none of the DA methods in Table~\ref{tab:population_risks_motivating_examples} can handle the intervention on $Y$ well. The weakness of DIP in the presence of label perturbation has been observed previously in~\cite{zhao2019learning,li2019target,combes2020domain}. Especially, \cite{combes2020domain} proposed the idea of using conditional invariant components for label shift correction. However, since their proposed method has no guarantees for estimating the conditional invariant components, it is not clear whether their methods have any guarantees for correcting the label shift.

The three examples are designed to demonstrate that the popular DA method DIP does not always outperform baseline methods such as OLSSrc or Causal. Besides, it also shows that the assumption of both source and data being generated from linear SCMs is not sufficient to guarantee DIP to have a low target risk. Additional assumptions on the data generation or on the intervention are needed. For example, knowing the causal direction of the data generation model or whether the label distribution is perturbed are all crucial information. Based on these examples, we ask the following questions:
\begin{enumerate}
  \item In addition to the linear SCM assumption, what other assumptions are needed for DIP to perform better than Causal or OLSSrc? If such assumptions exist, can one quantify the gap between the target risk achieved by DIP and the oracle target risk?
  \item If there is label distribution perturbation like in Example 3, are there DA methods that can outperform OLSSrc and have target risk guarantees?
  \item If the prediction direction is a mix of causal and anticausal, is domain adaptation still beneficial?
\end{enumerate}
In the sequel, we address these questions one by one. The first question is addressed in Section~\ref{sub:anti_causal_domain_adaptation_without_intervention_on_y}. The second question is dealt with in Section~\ref{sub:anti_causal_domain_adaptation_with_intervention_on_y} via the introduction of CIP and CIRM. The general solution to the third question remains open. Naive applications of DIP and CIRM are not optimal. We provide a partial answer to the third question in Section~\ref{sub:mixed_causal_anti_causal_domain_adaptation} and show that the mixed-causal anticausal DA problem can be reduced to the anticausal DA problem when the causal variables have been already identified.
% subsubsection lessons_from_the_simple_motivating_examples (end)

% subsection simple_motivating_examples (end)

% section preliminary_and_problem_setup (end)

\section{Domain adaptation with theoretical guarantees} % (fold)
\label{sec:domain_adaptation_guarantees}
In this section, to answer the three questions in the last section with rigor, we establish target risk guarantees for the DA estimators DIP, CIP and CIRM in three settings. Subsection~\ref{sub:anti_causal_domain_adaptation_without_intervention_on_y} focuses on the DIP performance in the anticausal DA setting without intervention on $Y$. Subsection~\ref{sub:anti_causal_domain_adaptation_with_intervention_on_y} demonstrates the difficulty of DIP in the anticausal DA setting with interventions on $Y$ and then proves the advantage of CIP and CIRM over DIP when multiple source environments are available. Subsection~\ref{sub:mixed_causal_anti_causal_domain_adaptation} shows how domain adaptation is still possible in the mixed causal anticausal DA setting.

\subsection{Anticausal domain adaptation without intervention on Y} % (fold)
\label{sub:anti_causal_domain_adaptation_without_intervention_on_y}

In this subsection, we study the anticausal domain adaptation with the additional assumption that the intervention on the covariates $X$ is a mean shift noise intervention and there is no intervention on the label $Y$. In the anticausal domain adaptation, all the covariates $X$ are descendants of the label $Y$ in the SCM. First, we derive the target risk bound for DIP under these assumptions with a single source environment. Then we show how to make use of more source environments to improve the performance of DIP. Finally, we discuss ways to relax the mean shift noise intervention assumption.

\subsubsection{Target risk guarantees of DIP with a single source environment} % (fold)
\label{ssub:target_risk_guarantees_of_dip}
Before we state the main theorem, we introduce another oracle estimator to simplify the theorem statement. \textit{DIPOracle} minimizes the mean squared error on target data while it uses the same matching penalty as DIP.
\begin{itemize}
  \item \textbf{DIPOracle$^\tagk{m}$-mean:} the population DIP estimator which uses target labels and the covariate distribution from the $m$-th source and target environment.
  \begin{align}
    \label{eq:estimator_pop_diporaclemeanmatchlin}
    f_{\tagdiporaclemeanmatchlin}^\tagk{m}(x) &\defn x \tp\betadiporaclemeanmatchlin{m} + \beta_{\tagdiporaclemeanmatchlin, 0}^\tagk{m} \notag \\
    \betadiporaclemeanmatchlin{m}, \beta_{\tagdiporaclemeanmatchlin, 0}^\tagk{m} &\defn \argmin_{\beta, \beta_0}\  \Exs_{(X, Y) \sim \distritar} \parenth{Y - X\tp\beta - \beta_0}^2 \notag \\
    &\text{s.t.\ } \Exs_{X \sim \distriv{m}_X} \brackets{X \tp\beta} = \Exs_{X \sim \distritar_X}\brackets{X \tp\beta}.
  \end{align}
\end{itemize}
Next, we state the data generating assumptions needed for DIP to have target risk guarantees. Since a single source environment is considered in this subsection, without loss of generality, we assume that the first source environment is used.
\begin{assumption}
  \label{ass:assumption_single_source_anti_causal}
  Each data point in the source environment is generated i.i.d. according to distribution $\distriv{1}$ specified by the following SCM
  \begin{align*}
    \bmat{\Xv{1} \\ \Yv{1}} = \bmat{\semBX & \sembv \\ \sembh\tp & 0} \bmat{\Xv{1} \\ \Yv{1}} + \bmat{\interAv{1}_X \\ \interAv{1}_Y} + \bmat{\noisev{1}_X \\ \noisev{1}_Y},
  \end{align*}
  each data point in the target environment is generated i.i.d. according to distribution $\distritar$ specified with the same SCM except for the mean shift intervention term
  \begin{align*}
    \bmat{\Xtar \\ \Ytar} = \bmat{\semBX & \sembv \\ \sembh\tp & 0} \bmat{\Xtar \\ \Ytar} + \bmat{\interAtar_X \\ \interAtar_Y} + \bmat{\noisetar_X \\ \noisetar_Y},
  \end{align*}
  where $\Ind_\dims - \semBX \in \real^{\dims \times \dims}$ is invertible, the prediction problem is anticausal i.e. $\sembh=0$, $\sembv \in \real^\dims$, the intervention terms $\bmat{\interAv{1}_X \\ \interAv{1}_Y}$ and $\bmat{\interAtar_X \\ \interAtar_Y}$ are vectors in $\real^{\dims+1}$, the noise terms $\bmat{\noisev{1}_X \\ \noisev{1}_Y}$ and $\bmat{\noisetar_X \\ \noisetar_Y}$ share the same distribution with
  \begin{align*}
    \Exs\brackets{\noisev{1}_X} = \Exs\brackets{\noisetar_X} = 0, \quad \Exs\brackets{\noisev{1}_X {\noisev{1}_X}\tp } = \Exs\brackets{\noisetar_X {\noisetar_X}\tp } =  \noisecovX \in \real^{\dims \times \dims}, \\
    \Exs\brackets{\noisev{1}_Y} = \Exs\brackets{\noisetar_Y} = 0, \quad \Exs\brackets{{\noisev{1}_Y}^2} = \Exs\brackets{{\noisetar_Y}^2} = \noisecovY^2 \in \real.
  \end{align*}
  Additionally, the noise terms on $X$ and $Y$ are uncorrelated
    $\Exs \brackets{\noisev{1}_Y \cdot \noisev{1}_X} = \Exs \brackets{\noisetar_Y \cdot \noisetar_X} = 0$.
\end{assumption}
Note that Assumption~\ref{ass:assumption_single_source_anti_causal} does not require $\noisecovX$ to be diagonal, meaning that the noise terms of $X$ can be correlated. Consequently, this assumption allows for unobserved confounders affecting the $X$ variables to exist. Under the assumptions above and that there is no intervention on $Y$, we obtain the following theorem on the target risk of DIP.
\begin{theorem}
  \label{thm:single_source_anticausal_mean_shift}
  Under the data generation Assumption~\ref{ass:assumption_single_source_anti_causal} and the assumption of no intervention on $Y$ i.e. $\interAv{1}_Y = \interAtar_Y = 0$,  the target population risks of OLSTar, OLSSrc$^\tagk{1}$ and DIP$^\tagk{1}$-mean satisfy
  \begin{align}
    \label{eq:single_source_pop_risk_olsoracle}
    \tilde{\risk}\parenth{f_{\tagolstarget}} &= \frac{\noisecovY^2}{1 + \noisecovY^2 \sembv\tp \noisecovX^{-1}\sembv}, \\
    \label{eq:single_source_pop_risk_olssrc}
    \tilde{\risk}\parenth{f_{\tagolssource}^\tagk{1}} &= \frac{\noisecovY^2}{1 + \noisecovY^2 \sembv\tp \noisecovX^{-1} \sembv } + \frac{\parenth{\noisecovY^2 \sembv \tp \noisecovX^{-1} \parenth{\interAv{1}_X - \interAtar_X}}^2}{\parenth{1 + \noisecovY^2 \sembv \tp \noisecovX^{-1} \sembv}^2}, \\
    \label{eq:single_source_pop_risk_dipmeanmatchlin}
    \tilde{\risk}\parenth{f_{\tagdipmeanmatchlin}^\tagk{1}} &= \tilde{\risk}\parenth{f_{\tagdiporaclemeanmatchlin}^\tagk{1}} = \frac{\noisecovY^2}{1 + \noisecovY^2 \sembv\tp \noisecovX^{-\frac{1}{2}} \Gdipv{1} \noisecovX^{-\frac{1}{2}} \sembv },
  \end{align}
  where $\Gdipv{1} = \noisecovX^{1/2} \Qdipv{1} \parenth{\Qdipv{1}  \tp \noisecovX \Qdipv{1}}^{-1} \Qdipv{1}\tp \noisecovX^{1/2}$ is a projection matrix with rank $\dims-1$; $\Qdipv{1} \in \real^{\dims \times \dims-1}$ is a matrix with columns formed by the vectors that complete the vector $u^\tagk{1}$ to an orthonormal basis. Here $u^\tagk{1} = \frac{\interAv{1}_X - \interAtar_X}{ \vecnorm{\interAv{1}_X - \interAtar_X}{2}}$
  when the source and target distribution is not identical i.e. $\interAv{1}_X \neq \interAtar_X$. When $\interAv{1}_X = \interAtar_X$, meaning that the source distribution equals the target distribution, then $u^\tagk{1} = 0$ and we have $\tilde{\risk}\parenth{f_{\tagdipmeanmatchlin}^\tagk{1}} =  \tilde{\risk}\parenth{f_{\tagolssource}^\tagk{1}}  = \tilde{\risk}\parenth{f_{\tagolstarget}}$.
\end{theorem}
The proof of Theorem~\ref{thm:single_source_anticausal_mean_shift} is provided in Appendix~\ref{sub:proof_of_theorem_1}. Comparing Equation~\eqref{eq:single_source_pop_risk_dipmeanmatchlin} with Equation~\eqref{eq:single_source_pop_risk_olsoracle}, the target population risk of DIP$^\tagk{1}$ is larger than that of OLSTar but it is lower than the target population risk of the Causal estimator (which equals to $\noisecovY^2$). The target population risk of OLSSrc$^\tagk{1}$ depends on the magnitude of the difference in intervention $\vecnorm{\interAv{1}_X - \interAtar_X}{2}$, while the target risk of DIP$^\tagk{1}$ is independent of that magnitude. Consequently when the difference in intervention becomes large, DIP$^\tagk{1}$ can outperform OLSSrc$^\tagk{1}$. Moreover, the first equality of Equation~\eqref{eq:single_source_pop_risk_dipmeanmatchlin} shows that DIP$^\tagk{1}$ achieves the same target population risk as DIPOracle$^\tagk{1}$. DIPOracle$^\tagk{1}$ differs from OLSTar by only one linear constraint. The connection between DIP$^\tagk{1}$ and DIPOracle$^\tagk{1}$ intuitively explains why DIP$^\tagk{1}$ target risk should be close to that of OLSTar.

In fact, with additional assumptions on how the interventions $\interAv{1}_X$ and $\interA_X$ are positioned and simplifications on the covariance matrix $\noisecovX$, we obtain the following corollary which clearly highlights the difference between the target population risk of DIP and the oracle target population risk of OSLTar.

\begin{corollary}
  \label{cor:single_source_anticausal_mean_shift}
  In addition to the assumptions in Theorem~\ref{thm:single_source_anticausal_mean_shift}, suppose $\noisecovX = \frac{\noisecovY^2}{\rho} \Ind_\dims$ with $\rho > 0$, then
  \begin{align}
    \label{eq:single_source_pop_risk_olsoracle_gaussian}
    \tilde{\risk}\parenth{f_{\tagolstarget}} &= \frac{\noisecovY^2}{1 + \rho \vecnorm{\sembv}{2}^2}, \\
    \label{eq:single_source_pop_risk_olssrc_gaussian}
    \tilde{\risk}\parenth{f_{\tagolssource}^\tagk{1}} &= \frac{\noisecovY^2}{1 + \rho \vecnorm{\sembv}{2}^2} + \frac{\parenth{{u^\tagk{1}} \tp \sembv}^2 \vecnorm{\interAv{1}_X - \interAtar_X}{2}^2}{\parenth{1 + \rho \vecnorm{\sembv}{2}^2 }^2}, \\
    \label{eq:single_source_pop_risk_dipmeanmatchlin_gaussian_eq}
    \tilde{\risk}\parenth{f_{\tagdipmeanmatchlin}^\tagk{1}} &= \frac{\noisecovY^2}{1 + \rho \vecnorm{\sembv}{2}^2 -  \rho\parenth{{u^\tagk{1}}\tp\sembv}^2},
  \end{align}
  where $u^\tagk{1} = \frac{\interAv{1}_X - \interAtar_X}{ \vecnorm{\interAv{1}_X - \interAtar_X}{2}}$ if $\interAv{1}_X \neq \interAtar_X$ and $u^\tagk{1} =0$ otherwise.

  Additionally, if $\interAv{1}_X - \interAtar_X$ is generated randomly from the Gaussian distribution $\Normal(0, \tau \Ind_\dims^2)$, then for a constant $t$ satisfying $0 < t \leq \frac{\dims}{2}$, with probability at least $1 - \exp(-t/16) - 2\exp(-t/4)$, we have
  \begin{align}
    \label{eq:single_source_pop_risk_dipmeanmatchlin_gaussian_ineq}
    \tilde{\risk}\parenth{f_{\tagolssource}^\tagk{1}} &\leq \frac{\noisecovY^2}{1 + \rho \vecnorm{\sembv}{2}^2} + \frac{\tau^2 t \vecnorm{\sembv}{2}^2}{2\parenth{1 + \rho \vecnorm{\sembv}{2}^2}^2}, \notag \\
    \tilde{\risk}\parenth{f_{\tagdipmeanmatchlin}^\tagk{1}} &\leq \frac{\noisecovY^2}{1 + \rho \parenth{1 - \frac{t}{\dims}} \vecnorm{\sembv}{2}^2}.
  \end{align}
\end{corollary}
The proof of Corollary~\ref{cor:single_source_anticausal_mean_shift} is provided in Appendix~\ref{sub:proof_of_cor_1}. Under the conditions of Corollary~\ref{cor:single_source_anticausal_mean_shift}, Equation~\eqref{eq:single_source_pop_risk_dipmeanmatchlin_gaussian_eq} shows that the gap between the target population risks of DIP$^\tagk{1}$ and OLSTar is smaller when the direction of the difference in intervention $u^\tagk{1}$ becomes less aligned with the vector $\sembv$. When the intervention is generated randomly from a Gaussian distribution, the bound~\eqref{eq:single_source_pop_risk_dipmeanmatchlin_gaussian_ineq} shows that the gap between the target population risks of DIP$^\tagk{1}$ and OLSTar is small with high probability. The gap only comes from an order $\frac{1}{\dims}$ term in the denominator of the target risk. So when the dimension $\dims$ is large, this difference in target population risk between DIP$^\tagk{1}$ and OLSTar becomes negligible. As for OLSSrc, it is now clear from Corollary~\ref{cor:single_source_anticausal_mean_shift} that the target risk of OLSSrc$^\tagk{1}$ depends on the magnitude of the difference in intervention $\tau$. For brevity, we only provided the target risk upper bound of OLSSrc$^\tagk{1}$. The lower bound should hold similarly up to constant factors because the risks are derived with equality in Equation~\eqref{eq:single_source_pop_risk_olssrc_gaussian}-\eqref{eq:single_source_pop_risk_dipmeanmatchlin_gaussian_eq} and tight Gaussian concentration bounds are used to obtain the upper bounds. As the magnitude of the difference in intervention $\tau$ increases, OLSSrc$^\tagk{1}$ will eventually have a larger target risk than DIP$^\tagk{1}$.

The intuition behind the success of DIP$^\tagk{1}$ over OLSSrc$^\tagk{1}$ is that the DIP matching penalty in Equation~\eqref{eq:estimator_pop_dipmeanmatchlin} allows the DIP$^\tagk{1}$ estimator to equalize the source and target covariate intervention by projecting it to a common space. As a result, given the DIP matching penalty, computing least squares on the source data is the same as computing least squares on the target data. This observation also explains why the target risk of DIP$^\tagk{1}$ matches that of DIPOracle$^\tagk{1}$. Having this intuition in mind, it is not hard to imagine extending the guarantees of DIP$^\tagk{1}$ to the generic form of DIP in Equation~\eqref{eq:dip_general_form} under appropriate assumptions. Specifically, under Assumption~\ref{ass:assumption_single_source_anti_causal}, as long as the function class $\mathcal{V}$ is chosen such that the DIP matching penalty $\mathcal{D}(v(X), v(\tilde{X}))$ ensures the conditionals $v(X) \mid Y$ and $v(\tilde{X}) \mid Y$ to have the same distribution, then no matter how $\mathcal{U}$ is chosen we always match the target risks of DIP and DIPOracle. Moreover, we know that target risk of DIPOracle is close to the oracle target risk over the function class $\braces{ u \circ v \mid  u \in \mathcal{U}, v \in \mathcal{V} }$. So the target risk of DIP in this extended setting is also close to the oracle one.

% subsubsection target_risk_guarantees_of_dip (end)

\subsubsection{Benefit of more source environments for DIP} % (fold)
\label{ssub:Benefit_of_more_source_environments_for_dip}
We show how to make use of more independent source environments to improve the performance of DIP. Here we consider $\envs$ ($\envs \geq 2$) source environments, where for each $m\in \braces{1, \ldots, \envs}$ the $m$-th source environment is generated independently according to the source environment in Assumption~\ref{ass:assumption_single_source_anti_causal} except for the unknown interventions $\interAv{m}_X$ and we still have $\interAv{m}_Y = 0$. First, we state a corollary revealing that it is possible to pick the best source environment for DIP based on the best source risk. Second, we discuss the reason behind the performance improvement after picking the best source environment.

First, based on the proof of Theorem~\ref{thm:single_source_anticausal_mean_shift}, we derive the following corollary on the source population risk.
\begin{corollary}
  \label{cor:single_source_anticausal_mean_shift_source_pop_risk}
  Under the data generation Assumption~\ref{ass:assumption_single_source_anti_causal} and the assumption of no intervention on $Y$ i.e. $\interAv{1}_Y = \interAtar_Y = 0$, the source population risk of DIP$^\tagk{1}$-mean as defined in Equation~\eqref{eq:source_pop_risk} satisfies
  \begin{align}
    \label{eq:single_source_source_pop_risk_dipmeanmatchlin}
    \risk^\tagk{1}\parenth{f_{\tagdipmeanmatchlin}^\tagk{1}} = \tilde{\risk}\parenth{f_{\tagdipmeanmatchlin}^\tagk{1}}.
  \end{align}
\end{corollary}
The proof of Corollary~\ref{cor:single_source_anticausal_mean_shift_source_pop_risk} is provided in Appendix~\ref{sub:proof_of_cor_2}. According to Corollary~\ref{cor:single_source_anticausal_mean_shift_source_pop_risk}, one can read off the target population risk directly from the source population risk. Since the choice of source environment 1 is arbitrary, the same result holds for any of the $\envs$ source environments. We have $\risk^\tagk{m}\parenth{f_{\tagdipmeanmatchlin}^\tagk{m}} = \tilde{\risk}\parenth{f_{\tagdipmeanmatchlin}^\tagk{m}}$ for any $m \in \braces{1, \ldots, \envs}$. Consequently, having more source environments allows one to apply DIP between each source environment and the target environment one by one, and then pick the estimator $f_{\tagdipmeanmatchlin}^\tagk{m}$ with the lowest source population risk to reduce the target population risk.

Second, we reason about the type of source environments that reduces the target population risk the most. According to Equation~\eqref{eq:single_source_pop_risk_olsoracle} and~\eqref{eq:single_source_pop_risk_dipmeanmatchlin} in Theorem~\ref{thm:single_source_anticausal_mean_shift}, the projection matrix $\Gdipv{1}$ is the term that makes the target population risks of DIP$^\tagk{1}$ and DIP$^\tagk{m}$ different ($m \neq 1$). If the vector $\noisecovX^{-1/2} \sembv$ is in the span of the projection matrix $\Gdipv{1}$, then DIP$^\tagk{1}$ achieves the oracle target population risk. Otherwise, the target population risk of DIP$^\tagk{1}$ depends on the norm of the component of $\noisecovX^{-1/2}\sembv$ outside the span of the projection matrix $\Gdipv{1}$.

The above intuition becomes clearer under the additional assumptions of Corollary~\ref{cor:single_source_anticausal_mean_shift}. There we obtain a simple form for the projection matrix $\Gdipv{1} = \Ind_\dims - u^\tagk{1} {u^\tagk{1}}\tp$,
where $u^\tagk{1} = \frac{\interAv{1}_X - \interAtar_X}{\vecnorm{\interAv{1}_X - \interAtar_X}{2}}$ assuming $\interAv{1}_X \neq \interAtar_X$.
Under the same assumptions, the denominator in the DIP target risk in Equation~\eqref{eq:single_source_pop_risk_dipmeanmatchlin} becomes $1 + \vecnorm{\sembv}{2}^2 - \parenth{{u^\tagk{1}}\tp \sembv}^2$.
If $\interAv{1}_X - \interAtar_X$ is orthogonal to $\sembv$ then DIP$^\tagk{1}$ achieves the oracle target population risk. Otherwise, DIP$^\tagk{1}$ has larger target population risk than OLSTar. With more source environments, it is more likely to find a source environment $m$ such that $\interAv{m}_X - \interAtar_X$ is closer to be orthogonal to $\sembv$.

Based on the above intuition, we propose the following DIP variant that provides a weighting choice to take advantage of multiple source environments.
\begin{itemize}
  \item \textbf{DIPweigh-mean:} the population DIP estimator that makes use of multiple source environments based on the source risks.
  \begin{align}
    \label{eq:estimator_pop_dipweigh}
    f_{\tagdipweigh}(x) &\defn \frac{1}{\sum_{m=1}^{\envs} e^{-\eta \cdot s_m}}\sum_{m=1}^{\envs} e^{-\eta \cdot s_m}\parenth{x \tp\betadipmeanmatchlin{m} + \beta_{\tagdipmeanmatchlin, 0}^\tagk{m}} \notag \\
    s_m &\defn \risk^\tagk{m}\parenth{f_{\tagdipmeanmatchlin}^\tagk{m}}.
  \end{align}
  $\eta > 0$ is a constant. Choosing $\eta$ to be $\infty$ is equivalent to choosing the source estimator with the lowest source risk. DIPweigh weights all the source predictions based on the source risk of each source environment.
\end{itemize}
Here we introduce the weighted version of DIP rather than directly selecting the source estimator with the lowest source risk. The main intuition is that in finite sample, averaging the predictions from several source environments with low source risks can take advantage of a larger sample size. For example, in the presence of a less related source environment with very large sample size, one might still include it if the reduction in variance outweighs the increase in bias.
% subsubsection Benefit_of_more_source_environments_for_dip (end)

\subsubsection{Failure scenarios of DIP} % (fold)
\label{ssub:failure_scenarios_of_dip}
The target population risk guarantees of DIP in Theorem~\ref{thm:single_source_anticausal_mean_shift} rely on the anticausal data generation Assumption~\ref{ass:assumption_single_source_anti_causal} and the assumption of no intervention on $Y$. We already showed in Example 1 that DIP cannot outperform OLSSrc in the causal prediction setting. Even in the anticausal prediction setting, we identify two scenarios where DIP$^\tagk{1}$-mean might have large target risk: when the DIP matching penalty is not well-suited for the underlying type of the intervention that generates the data and when there is intervention on $Y$.

Consider the following data generation model where the source environment is generated from the following SCM with interventions on the variance
\begin{align*}
  \bmat{\Xv{1} \\ \Yv{1}} = \bmat{\semBX & \sembv \\ \sembh\tp & 0} \bmat{\Xv{1} \\ \Yv{1}} + \bmat{ \interAv{1}_X \odot \noisev{1}_X \\ \noisev{1}_Y},
\end{align*}
and the target environment is generated from the same SCM except that the intervention on $X$ takes the form $\interAtar_X \odot \noisetar_X$.
Here $\odot$ denotes the $\dims$-dimensional element-wise multiplication. This intervention is called variance shift noise intervention. The intervention term $\interAv{1}_X$ and $\interAtar_X$ are fixed vectors in $\real^\dims$, and the noise terms are kept the same as in Assumption~\ref{ass:assumption_single_source_anti_causal}. Under the data generation model in this subsection, the matching penalty in DIP$^\tagk{1}$-mean~\eqref{eq:estimator_pop_dipmeanmatchlin}
is always satisfied because both the left and right hand sides are zero. Consequently, DIP$^\tagk{1}$-mean becomes the same estimator as OLSSrc. Matching the mean between source and target distribution in this case is no longer a good idea, because the intervention is on the variance rather than the mean.

To adapt to the new type of intervention, one can consider \textit{DIP-std} which puts the matching penalty on the standard deviations, \textit{DIP-std+} which puts the matching penalty on the means, standard deviations and 25\% quantiles, or \textit{DIP-MMD} which uses more generic distributional matching via MMD. The exact formulations consist of replacing the DIP mean matching penalty in Equation~\eqref{eq:estimator_pop_dipmeanmatchlin} with the appropriate matching penalties and they are formally introduced in Appendix~\ref{sub:population_DA_methods}.

If the assumptions are set up such that the intervention type agrees with the matching penalty in DIP, analyzing the theoretical performance of DIP-std, DIP-std+ or DIP-MMD under linear SCMs works similarly as analyzing that of DIP-mean. Hence we leave out the theoretical guarantees of DIP-std, DIP-std+ or DIP-MMD for brevity. The empirical performance of DIP-std+ and DIP-MMD is shown through simulations and real data experiments in Section~\ref{sec:numerical_experiments}.

The second failure scenario of DIP is when there is intervention on $Y$. This failure scenario of DIP is already noticeable from Example 3 in Section~\ref{ssub:example_3_anti_causal_prediction_when_y_is_intervened_on}. Here we provide a corollary that quantifies the additional target population risk made by DIP if there is intervention on $Y$.

\begin{corollary}
  \label{cor:single_source_anticausal_mean_shift_intervention_on_y}
  Under the data generation Assumption~\ref{ass:assumption_single_source_anti_causal}, the target population risk of OLSTar and DIP$^\tagk{1}$-mean satisfy
  \begin{align}
    \label{eq:single_source_pop_risk_dipmeanmatchlin_intery}
    \tilde{\risk}\parenth{f_{\tagolstarget}} &= \frac{\noisecovY^2}{1 + \noisecovY^2 \sembv\tp \noisecovX^{-1}\sembv} \notag \\
    \tilde{\risk}\parenth{f_{\tagdipmeanmatchlin}^\tagk{1}} &= \frac{\noisecovY^2}{1 + \noisecovY^2 \sembv\tp \noisecovX^{-\frac{1}{2}} \GdipY \noisecovX^{-\frac{1}{2}} \sembv } + \parenth{\interAv{1}_Y - \interAtar_Y}^2,
  \end{align}
  where $\GdipY = \noisecovX^{1/2} \QdipY \parenth{\QdipY  \tp \noisecovX \QdipY}^{-1} \QdipY\tp \noisecovX^{1/2}$ is a projection matrix with rank $\dims-1$. $\QdipY \in \real^{\dims \times \dims-1}$ is a matrix with columns formed by the vectors that complete the vector $\frac{\interAv{1}_X + \interAv{1}_Y \sembv - \interAtar_X - \interAtar_Y \sembv}{ \vecnorm{\interAv{1}_X + \interAv{1}_Y \sembv - \interAtar_X - \interAtar_Y \sembv}{2}}$ to an orthonormal basis.
\end{corollary}
The proof of Corollary~\ref{cor:single_source_anticausal_mean_shift_intervention_on_y} is provided in Appendix~\ref{sub:proof_of_cor_2}. According to Equation~\eqref{eq:single_source_pop_risk_dipmeanmatchlin_intery} in Corollary~\ref{cor:single_source_anticausal_mean_shift_intervention_on_y}, the target population risk has an extra term that depends on the difference between target and source $Y$ intervention. The target population risk of DIP is increases as the difference between target and source $Y$ intervention increases. This corollary highlights the result that DIP has a large target risk when the intervention on $Y$ is large.

% subsubsection failure_scenarios_of_dip (end)

% subsection anti_causal_domain_adaptation_without_intervention_on_y (end)

\subsection{Anticausal domain adaptation with intervention on Y} % (fold)
\label{sub:anti_causal_domain_adaptation_with_intervention_on_y}
% \olcomment{
% \paragraph{Subsection outline:}
% \begin{enumerate}
%   \item Theorem 2: risk bound for CIP and CIRM for anticausal Y-intervention mean-shift case
%   % \item The residual independent constraint: RII, RIIRM, CIRMri
%   \item Discussion on number of source environments needed
% \end{enumerate}
% }

In this subsection, we study the anticausal domain adaptation where there is intervention on the label $Y$. As illustrated in Example 3 in Section~\ref{ssub:example_3_anti_causal_prediction_when_y_is_intervened_on} and in the last section, the intervention on $Y$ can cause DIP to have a large target population risk. In fact, allowing arbitrary intervention on both $X$ and $Y$ can also lead to unidentifiable cases where two data generation models result in the same target covariate distribution in the anticausal domain adaptation. If it were the case, any DA estimator based solely on the target covariate distribution will have large target risk. The following example illustrates one concrete unidentifiable case.

\paragraph{A simple unidentifiable example: }  The target data is generated form the following SCM
\begin{align*}
  \Ytar &= \noisetar_Y + \interAtar_Y \\
  \Xtar &= 2 \Ytar + \noisetar_X + \interAtar_X,
\end{align*}
where $\Xtar$ is $1$-dimensional random variable, $\noisetar_Y \sim \Normal(0, 1)$ and $\noisetar_X \sim \Normal(0, 0.1)$. Assume we observe the target covariate distribution $\Xtar \sim \Normal(3, 1.1)$. Without observing the target label $\Ytar$ distribution, it is impossible to tell whether the intervention is $(\interAtar_X, \interAtar_Y) = (0, 1.5)$, $(1, 1)$ or $(3, 0)$. This is because all three interventions result in the same target covariate distribution. However, the conditional distribution $\Ytar \mid \Xtar$ are different. Consequently, any estimator of the conditional mean $\Exs\brackets{\Ytar \mid \Xtar}$ without access to the target label distribution can not get it correct.

To make the anticausal domain adaptation problem with intervention on $Y$ tractable, additional assumptions on the structure of the interventions is needed. Here we adopt one type of assumptions introduced by~\cite{gong2016domain,heinze2017conditional}. This line of work assumes the existence of conditionally invariant components (CICs) across all source and target environments. That is, there exists an unknown transformation $\mathcal{T}$ of the covariates such that the conditional distribution $\mathcal{T}(X) \mid Y$ is invariant across source and target environments.

In the following, we first prove target population risk guarantees for CIP and CIRM under the conditionally invariant components assumption. We show that the CIP and CIRM target risks are much less dependent on the $Y$ intervention than the DIP target risk. Finally we discuss extensions and variants of CIP and CIRM.

\subsubsection{Target risk guarantees of CIP and CIRM with multiple source environments and conditionally invariant components} % (fold)
\label{ssub:target_risk_guarantees_of_cip_and_cirm_with_multiple_source_environments_and_cics}
We start by stating the data generation assumptions for CIP and CIRM to have target risk guarantees, namely a SCM and the existence of conditionally invariant components.

\begin{assumption}
  \label{ass:assumption_multiple_source_anti_causal}
  There are $\envs$ ($\envs \geq 2$) source environments. Each data point in the $m$-th source environment is generated i.i.d. according to distribution $\distriv{m}$ specified by the following SCM
  \begin{align*}
    \bmat{\Xv{m} \\ \Yv{m}} = \bmat{\semBX & \sembv \\ \sembh\tp & 0} \bmat{\Xv{m} \\ \Yv{m}} + \bmat{\interAv{m}_X \\ \interAv{m}_Y} + \bmat{\noisev{m}_X \\ \noisev{m}_Y}.
  \end{align*}
  Each data point in the target environment is generated i.i.d. according to distribution $\distritar$ specified with the same SCM except for the mean shift intervention term
  \begin{align*}
    \bmat{\Xtar \\ \Ytar} = \bmat{\semBX & \sembv \\ \sembh\tp & 0} \bmat{\Xtar \\ \Ytar} + \bmat{\interAtar_X \\ \interAtar_Y} + \bmat{\noisetar_X \\ \noisetar_Y},
  \end{align*}
  where $\Ind_\dims - \semBX$ is invertible, the prediction problem is anticausal $\sembh=0$, the intervention term $\bmat{\interAv{1}_X \\ \interAv{1}_Y}$ and $\bmat{\interAtar_X \\ \interAtar_Y}$ are vectors in $\real^{\dims+1}$, the noise term $\bmat{\noisev{m}_X \\ \noisev{m}_Y}$ and $\bmat{\noisetar_X \\ \noisetar_Y}$ share the same distribution with
  \begin{align*}
    \Exs\brackets{\noisev{m}_X} = \Exs\brackets{\noisetar_X} = 0, \quad \Exs\brackets{\noisev{m}_X {\noisev{m}_X}\tp } = \Exs\brackets{\noisetar_X {\noisetar_X}\tp } =  \noisecovX, \\
    \Exs\brackets{\noisev{m}_Y} = \Exs\brackets{\noisetar_Y} = 0, \quad \Exs\brackets{{\noisev{m}_Y}^2} = \Exs\brackets{{\noisetar_Y}^2} = \noisecovY^2.
  \end{align*}
  In addition, the noise terms on $X$ and $Y$ are uncorrelated \mbox{$
    \Exs \brackets{\noisev{m}_Y \cdot \noisev{m}_X} = \Exs \brackets{\noisetar_Y \cdot \noisetar_X} = 0$}.
  The interventions do not span the whole space (to ensure the existence of conditionally invariant components)
  \begin{align}
    \label{eq:assumption_multiple_source_existence_CIC}
    \text{dim}\parenth{\text{span}\parenth{\interAv{2}_X - \interAv{1}_X, \ldots, \interAv{\envs}_X - \interAv{1}_X} } =  p \leq \dims - 1,
  \end{align}
  and the target $X$ intervention is in the span of source $X$ interventions
  \begin{align}
    \label{eq:assumption_multiple_source_target_interv_in_span}
    \interAtar_X - \interAv{1}_X \in \text{span}\parenth{\interAv{2}_X-\interAv{1}_X, \ldots, \interAv{\envs}_X-\interAv{1}_X}.
  \end{align}
\end{assumption}

Compared to Assumption~\ref{ass:assumption_single_source_anti_causal}, Assumption~\ref{ass:assumption_multiple_source_anti_causal} involves multiple source environments instead of one. Also, the way the source and target environments are generated up to Equation~\eqref{eq:assumption_multiple_source_existence_CIC} is the same as in Assumption~\ref{ass:assumption_single_source_anti_causal}. Equation~\eqref{eq:assumption_multiple_source_existence_CIC} assumes that the $X$ interventions do not span the whole covariate space $\real^\dims$. It ensures the existence of a invariant linear projection of the covariates given the label $Y$. More precisely, this assumption allows us to find a vector $v$ which is orthogonal to the $\text{span}\parenth{\parenth{\Ind_\dims - \semBX}^{-1}\parenth{\interAv{2}_X-\interAv{1}_X}, \ldots, \parenth{\Ind_\dims - \semBX}^{-1}\parenth{\interAv{\envs}_X-\interAv{1}_X}}$ and the projection of the covariates $X\tp v$ has the same intervention term across source environments.  This particular linear projection of the covariates becomes one conditionally invariant component. To make sure the same conditionally invariant component is also valid in the target environment, Equation~\eqref{eq:assumption_multiple_source_target_interv_in_span} requires that the target $X$ intervention is in the span of source $X$ interventions.

Given the data generation assumption above, we have the following theorem on the target risk of CIP and CIRM.

\begin{theorem}
  \label{thm:mutiple_source_anticausal_mean_shift}
  Under the data generation Assumption~\ref{ass:assumption_multiple_source_anti_causal}, the population target risks of CIP-mean and CIRM$^\tagk{m}$-mean for any $m \in \braces{1, \ldots, \envs}$ satisfy
  \begin{align}
    \label{eq:multiple_source_pop_risk_olstar}
    \tilde{\risk}\parenth{f_{\tagolstarget}} &= \frac{\noisecovY^2}{1 + \noisecovY^2 \sembv\tp \noisecovX^{-1} \sembv }, \\
    \label{eq:multiple_source_pop_risk_cipmean}
    \tilde{\risk}\parenth{f_{\tagcipmean}} &= \frac{\noisecovY^2 + \Delta_Y }{1 + \parenth{\noisecovY^2 + \Delta_Y} \sembv\tp \noisecovX^{-\frac{1}{2}} \Gcip \noisecovX^{-\frac{1}{2}} \sembv } + \frac{ \parenth{\interAtar_Y - \bar{\interA}_Y}^2- \Delta_Y }{\parenth{1 + \parenth{\noisecovY^2 + \Delta_Y} \sembv\tp \noisecovX^{-\frac{1}{2}} \Gcip \noisecovX^{-\frac{1}{2}}\sembv }^2}, \\
    \label{eq:multiple_source_pop_risk_cirm}
    \tilde{\risk}\parenth{f_{\tagcirmeanmatch}^\tagk{m}} &= \frac{\noisecovY^2}{1 + \noisecovY^2 \sembv\tp \noisecovX^{-\frac{1}{2}} \Gcirmv{m} \noisecovX^{-\frac{1}{2}} \sembv } + \frac{\parenth{\interAv{m}_Y - \interAtar_Y}^2}{\parenth{1 + \noisecovY^2 \sembv\tp \noisecovX^{-\frac{1}{2}} \Gcirmv{m} \noisecovX^{-\frac{1}{2}} \sembv}^2},
  \end{align}
  where $\bar{\interA}_Y = \frac{1}{\envs} \sum_{j=1}^\envs \interAv{j}_Y$, $\Delta_Y = \frac{1}{\envs} \sum_{j=1}^\envs \parenth{\interAv{j}_Y - \bar{\interA}_Y}^2$.
  \mbox{$\Gcip = \noisecovX^{1/2} \Qcip \parenth{\Qcip  \tp \noisecovX \Qcip}^{-1} \Qcip\tp \noisecovX^{1/2}$} is a projection matrix with rank $\dims-p$ where $\Qcip \in \real^{\dims \times (\dims-p)}$ is a matrix with columns formed by an orthonormal basis of the orthogonal complement of $\text{span}(\interAv{2}_X - \interAv{1}_X, \ldots, \interAv{\envs}_X - \interAv{1}_X)$. $\Gcirmv{m}$ is a projection matrix defined in the same way as $\Gdipv{1}$ in Theorem~\ref{thm:single_source_anticausal_mean_shift}.
\end{theorem}

The proof of Theorem~\ref{thm:mutiple_source_anticausal_mean_shift} is provided in Appendix~\ref{sub:proof_of_theorem_2}. Comparing the target population risk of CIRM$^\tagk{m}$ in Equation~\eqref{eq:multiple_source_pop_risk_cirm} with that of DIP$^\tagk{m}$ in Equation~\eqref{eq:single_source_pop_risk_dipmeanmatchlin_intery}, CIRM$^\tagk{m}$ reduces the dependency on the difference in $Y$ intervention $\interAv{m}_Y - \interAtar_Y$. This is because we always have $1 + \noisecovY^2 \sembv\tp \noisecovX^{-\frac{1}{2}} \Gcirmv{m} \noisecovX^{-\frac{1}{2}} \sembv > 1$.
The target population risk of CIRM$^\tagk{m}$ when there is intervention on $Y$ in Equation~\eqref{eq:multiple_source_pop_risk_cirm} has an additional term depending on $\interAv{m}_Y - \interAtar_Y$ when compared to that of DIP$^\tagk{m}$ when there is no intervention on $Y$ in Equation~\eqref{eq:single_source_pop_risk_dipmeanmatchlin}. The additional term becomes close to zero when $\noisecovY^2 \sembv\tp \noisecovX^{-\frac{1}{2}} \Gcirmv{m} \noisecovX^{-\frac{1}{2}} \sembv$ is large.

In fact, without additional assumptions to Assumption~\ref{ass:assumption_multiple_source_anti_causal}, the dependence on $\interAv{m}_Y - \interAtar_Y$ or similar terms is unavoidable for any DA estimator. Because the Assumption~\ref{ass:assumption_multiple_source_anti_causal} does not prevent $\sembv$ from being in the $\text{span}\parenth{\interAv{2}_X - \interAv{1}_X, \ldots, \interAv{\envs}_X - \interAv{1}_X}$, it can still result in a slightly unidentifiable example similar to the one presented at the beginning of Section~\ref{sub:anti_causal_domain_adaptation_with_intervention_on_y}. See Appendix~\ref{sec:cirm_variants} for ways to get rid of the $\interAv{m}_Y - \interAtar_Y$ dependence at the cost of small additional target risk.

In the same spirit of Corollary~\ref{cor:single_source_anticausal_mean_shift}, we present a corollary that puts additional assumptions on the positions of the interventions $\interAv{m}_X$ and $\interAtar_X$ to make the results in Theorem~\ref{thm:mutiple_source_anticausal_mean_shift} easier to understand.

\begin{corollary}
  \label{cor:multiple_source_anticausal_mean_shift}
  In addition to Assumption~\ref{ass:assumption_multiple_source_anti_causal}, suppose $\noisecovX = \frac{\noisecovY^2}{\rho} \Ind_\dims$ with $\rho > 0$, then
  \begin{align}
    \label{eq:multiple_source_pop_risk_olsoracle_gaussian}
    \tilde{\risk}\parenth{f_{\tagolstarget}} &= \frac{\noisecovY^2}{1 + \rho \vecnorm{\sembv}{2}^2}, \\
    \label{eq:multiple_source_pop_risk_cip_gaussian}
    \tilde{\risk}\parenth{f_{\tagcipmean}} &= \frac{\noisecovY^2}{1 + \rho \vecnorm{\sembv}{2}^2 - \rho \parenth{\Pcip\tp \sembv}^2} + \frac{\parenth{\interAtar_Y - \bar{\interA}_Y}^2 - \Delta_Y}{\brackets{1 + \rho \vecnorm{\sembv}{2}^2 - \rho \parenth{\Pcip\tp \sembv}^2}^2}, \\
    \label{eq:multiple_source_pop_risk_cirm_gaussian}
    \tilde{\risk}\parenth{f_{\tagcirmeanmatch\tagk{m}}} &= \frac{\noisecovY^2}{1 + \rho \vecnorm{\sembv}{2}^2 - \rho \parenth{{u^\tagk{m}}\tp \sembv}^2} + \frac{\parenth{\interAtar_Y - \interAv{m}_Y}^2}{\brackets{1 + \rho \vecnorm{\sembv}{2}^2 - \rho \parenth{{u^\tagk{m}}\tp \sembv}^2}^2},
  \end{align}
  where $\bar{\interA}_Y = \frac{1}{\envs} \sum_{m=1}^\envs \interAv{m}_Y$, $\Delta_Y = \parenth{\interAtar_Y - \bar{\interA}_Y}^2 - \frac{1}{\envs}\sum_{m=1}^\envs \parenth{\interAv{m}_Y - \bar{\interA}_Y}^2$, $\Pcip = \Qcip^{\perp} \in \real^{\dims \times p}$ is the matrix with columns formed by the orthonormal basis of $\text{span}\parenth{\interAv{2}_X-\interAv{1}_X, \ldots, \interAv{\envs}_X - \interAv{1}_X}$ and $u^\tagk{m} = \frac{\interAv{m}_X - \interAtar_X}{\vecnorm{\interAv{m}_X - \interAtar_X}{2}}$. Additionally, if $\interAv{2}_X - \interAv{1}_X, \ldots, \interAv{\envs}_X - \interAv{1}_X, \xi$ are generated independently from Gaussian distribution $\Normal\parenth{0, \tau^2 \Ind_\dims}$ and $\interAtar_X - \interAv{1}_X = \Pcip \xi$, $\dims \geq 6\envs$, a constant $0 < t \leq \dims/2$, then with probability at least $\parenth{1 - 2 \exp(-\dims/32) - 2\exp(-\envs/32)} \cdot \parenth{1 - \exp(-t/16) - 2 \exp(-t/4)}$, we have $\text{dim}\parenth{\text{span}\parenth{\interAv{2}_X-\interAv{1}_X, \ldots, \interAv{\envs}_X - \interAv{1}_X}} = \envs - 1$ and additionally
  \begin{align}
    \label{eq:multiple_source_pop_risk_cirm_gaussian_ineq}
    \tilde{\risk}\parenth{f_{\tagcipmean}} &\leq \frac{\noisecovY^2}{1 + \rho \parenth{1 - \frac{3(\envs-1)}{\dims}} \vecnorm{\sembv}{2}^2} + \frac{\parenth{\interAtar_Y - \bar{\interA}_Y}^2 - \Delta_Y}{\brackets{1 + \rho \parenth{1 - \frac{3(\envs-1)}{\dims}} \vecnorm{\sembv}{2}^2}}, \\
    \tilde{\risk}\parenth{f_{\tagcipmean}} &\geq \frac{\noisecovY^2}{1 + \rho \parenth{1 - \frac{(\envs-1)}{3\dims}} \vecnorm{\sembv}{2}^2} + \frac{\parenth{\interAtar_Y - \bar{\interA}_Y}^2 - \Delta_Y}{\brackets{1 + \rho \parenth{1 - \frac{(\envs-1)}{3\dims}} \vecnorm{\sembv}{2}^2}}, \\
    \tilde{\risk}\parenth{f_{\tagcirmeanmatch\tagk{m}}} &\leq \frac{\noisecovY^2}{1 + \rho \parenth{1 - \frac{t}{\dims}}\vecnorm{\sembv}{2}^2} + \frac{\parenth{\interAtar_Y - \interAv{m}_Y}^2}{\brackets{1 + \rho \parenth{1 - \frac{t}{\dims}}\vecnorm{\sembv}{2}^2}}.
  \end{align}
  % \begin{align}
  %   \label{eq:multiple_source_pop_risk_cip_gaussian_projection_ineq}
  %   \frac{\envs-1}{3\dims} \vecnorm{\sembv}{2}^2 \leq \parenth{\Pcip \tp \sembv}^2 \leq \frac{3(\envs-1)}{\dims} \vecnorm{\sembv}{2}^2, \text{ and }
  % \end{align}
  % \begin{align}
  %   \label{eq:multiple_source_pop_risk_cirm_gaussian_projection_ineq}
  %   \parenth{{u^\tagk{m}} \tp \sembv}^2 \leq \frac{t}{\dims} \vecnorm{\sembv}{2}^2.
  % \end{align}
\end{corollary}
The proof of Corollary~\ref{cor:multiple_source_anticausal_mean_shift} is provided in Appendix~\ref{sub:proof_of_corollary_cor:multiple_source_anticausal_mean_shift}. Under the assumptions of Corollary~\ref{cor:multiple_source_anticausal_mean_shift}, it is clear that for $\envs$ and $\dims$ sufficiently large, with high probability, CIRM$^\tagk{m}$ has smaller target population risk than CIP when all $Y$ interventions are equal $\interAv{1}_Y= \ldots = \interAv{\envs}_Y = \interAtar_Y$. This is because the last term in Equation~\eqref{eq:multiple_source_pop_risk_cip_gaussian} become zero can be ignored and $\frac{\envs-1}{3\dims} \geq \frac{t}{\dims}$ when $\envs \geq 3t + 1$. Intuitively, CIP only uses the conditionally invariant components (roughly $\dims - p$ coordinates) to build the estimator, while CIRM$^\tagk{m}$ takes advantage of the other coordinates of $X$ (roughly $\dims -1$ coordinates).

The intuition behind CIRM becomes apparent after the theorem statement. CIRM$^\tagk{m}$ is indeed a combination of DIP$^\tagk{m}$ and CIP: it applies CIP in the first step to obtain the conditionally invariant component which serves as a proxy of the label $Y$ to correct for the intervention on $Y$, then it applies DIP$^\tagk{m}$ with the corrected target covariate distribution to improve the target prediction performance. The conditionally invariant components identified by CIP are useful to constitute a proxy of $Y$ but they alone do not predict $Y$ very well especially when there are only a few of them. DIP has good target risk guarantees when there is no $Y$ intervention. When there is $Y$ intervention, CIRM uses CIP to correct for the $Y$ distribution perturbation and to create a residual which does not suffer from intervention, then apply DIP on the residual.

Since CIRM can be seen as applying DIP on the label-distribution-corrected source and target data, it is natural to introduce a weighted version of CIRM called CIRMweigh similar to DIPweigh~\eqref{eq:estimator_pop_dipweigh}. A precise formulation of CIRMweigh can be found in Appendix~\ref{sub:population_DA_methods}.

Under the assumptions of Corollary~\ref{cor:multiple_source_anticausal_mean_shift}, we summarize the target risk upper bounds of all DA methods in this paper in Table~\ref{tab:summary_target}. For brevity, we only present the target population risk upper bounds under high probability when the interventions are generated i.i.d. Gaussian $\Normal\parenth{0, \tau^2 \Ind_\dims}$. The upper bounds are tight up to constant factors in the denominators because we first proved the exact target risks with equality and then applied Gaussian concentration to obtain the upper bounds. Consequently, the upper bounds serve the comparison purpose.

\begin{table}[ht]
    \centering
    % \begin{adjustwidth}{-.2in}{-.3in}
    % {\renewcommand{\arraystretch}{.8}
    \begin{tabular}{cllc}
        \toprule
         & \thead{ \bf Estimator}  & \thead{
         \bf Target population risk upper bound\\interventions under general position}
        \\ \midrule
        & \thead{ OLSTar \\(oracle)}
        & $\displaystyle\frac{\noisecovY^2}{1 + \rho\vecnorm{\sembv}{2}^2}$
        \\[5mm]
        \multirow{3}{*}{\makecell{No intervention on $Y$ \\ (Corollary~\ref{cor:single_source_anticausal_mean_shift})}}
        &\thead{ OLSSrc$^\tagk{1}$}
        & $\displaystyle{\frac{\noisecovY^2}{1 + \rho\vecnorm{\sembv}{2}^2} + \frac{c\cdot\tau^2\vecnorm{\sembv}{2}^2}{\brackets{1 + \rho \vecnorm{\sembv}{2}^2}^2}}$
        \\[5mm]
        &\thead{ DIP$^\tagk{1}$}
        & $\displaystyle\frac{\noisecovY^2}{1 + \rho(1 - \frac{c}{\dims})\vecnorm{\sembv}{2}^2}$
        \\[5mm]
        \hline
        &\thead{DIP$^\tagk{m}$}
        & $\displaystyle\frac{\noisecovY^2}{1 + \rho(1 - \frac{c}{\dims})\vecnorm{\sembv}{2}^2} + \parenth{\interAtar_Y - \interAv{m}_Y}^2$
        \\[5mm]
        \multirow{3}{*}{\makecell{Intervention on $Y$ \\
        with CICs \\ $\envs$ sources \\ (Corollary~\ref{cor:multiple_source_anticausal_mean_shift})}}
        &\thead{CIP}
        & $\displaystyle\frac{\noisecovY^2}{1 + \rho(1 - \frac{c(\envs-1)}{\dims})\vecnorm{\sembv}{2}^2} + \frac{\parenth{\interAtar_Y - \bar{\interA}_Y}^2- \Delta_Y}{\brackets{1 + \rho(1 - \frac{c(\envs-1)}{\dims})\vecnorm{\sembv}{2}^2}^2}$
        \\[5mm]
        &\thead{CIRM$^\tagk{m}$}
        & $\displaystyle\frac{\noisecovY^2}{1 + \rho(1 - \frac{c}{\dims})\vecnorm{\sembv}{2}^2} + \frac{\parenth{\interAtar_Y - \interAv{m}_Y}^2}{\brackets{1 + \rho(1 - \frac{c}{\dims})\vecnorm{\sembv}{2}^2}^2}$
        \\[2mm]
        \bottomrule
    \end{tabular}
    % }
    % \end{adjustwidth}
    \caption{Summary of target population risk for DA estimators for anticausal domain adaptation under the assumptions of Corollary~\ref{cor:single_source_anticausal_mean_shift} and Corollary~\ref{cor:multiple_source_anticausal_mean_shift}. Here $c > 0$ is a generic constant to improve readability, which can be different for different methods. For simplicity, we only present the target population risk upper bounds holding with high probability when the interventions are generated i.i.d. Gaussian $\Normal(0, \tau^2 \Ind_\dims)$ and when both $\envs$ and $\dims$ are large. Under the assumptions of Corollary~\ref{cor:single_source_anticausal_mean_shift}, the target risk of OLSSrc$^\tagk{1}$ depends on the magnitude of the difference in intervention $\tau$, while the other DA methods in the table do not: DIP$^\tagk{1}$ outperforms OLSSrc$^\tagk{1}$. Under the assumptions of Corollary~\ref{cor:multiple_source_anticausal_mean_shift}, when intervention on $Y$ becomes large, DIP$^\tagk{m}$ is worse than CIP or CIRM$^\tagk{m}$. CIRM$^\tagk{m}$ slightly outperforms CIP.}
    \label{tab:summary_target}
\end{table}

% subsubsection target_risk_guarantees_of_cip_and_cirm_with_multiple_source_environments_and_cics (end)

\subsubsection{On relaxing the assumptions needed for CIP and CIRM} % (fold)
\label{ssub:on_relaxing_the_assumptions_needed_for_cip_and_cirm}
We discuss two different relaxations of Assumption~\ref{ass:assumption_multiple_source_anti_causal}.

First, the two assumptions~\eqref{eq:assumption_multiple_source_existence_CIC} and~\eqref{eq:assumption_multiple_source_target_interv_in_span} in Assumption~\ref{ass:assumption_multiple_source_anti_causal} can be replaced by a single assumption, namely there exists a non-zero vector $v \in \real^\dims$ such that $v\tp \Xv{1}, \ldots, v \tp \Xv{\envs}$ and $v \tp \Xtar$ all have the same distribution. This is also equivalent to saying that the projection of the covariates on the direction $v$ is a conditionally invariant component. Stating the CIC assumption as we did in Assumption~\ref{ass:assumption_multiple_source_anti_causal} has the advantage of separating the assumptions on the source interventions from those on the target intervention.

When the dimension $\dims$ is larger than the number of source environments $\envs$, the assumption~\eqref{eq:assumption_multiple_source_existence_CIC} on the dimension of the span of the source $X$ interventions is always satisfied. In the case of $\envs < \dims$, the more source environments there are, the more likely that the target $X$ intervention is in the span of the source $X$ interventions.

Second, the mean shift noise intervention in Assumption~\ref{ass:assumption_multiple_source_anti_causal} can be relaxed to other types of noise interventions. For example, the noise intervention can be the standard deviation shift as we did for DIP in Section~\ref{ssub:failure_scenarios_of_dip}. We may consider CIP-std which puts the conditional invariance penalty on the standard deviations, CIP-std+ which puts the conditional invariance penalty on the means, standard deviations and $25\%$ quantiles or CIP-MMD which uses generic distributional matching to make sure the conditional distribution of $v \tp X$ is invariant across source environments. CIRM can be extended similarly. We do not provide theoretical guarantees of these extensions and we only show their empirical performance through simulation and data experiments in Section~\ref{sec:numerical_experiments}.

% subsubsection on_relaxing_the_assumptions_needed_for_cip_and_cirm (end)

% subsection anti_causal_domain_adaptation_with_intervention_on_y (end)

\subsection{Mixed-causal-anticausal domain adaptation} % (fold)
\label{sub:mixed_causal_anti_causal_domain_adaptation}

In this subsection, we consider the mixed-causal-anticausal DA setting where both causal and anticausal variables are present. In terms of the SCMs, neither the vector $\sembv$ nor $\sembh$ in Equation~\eqref{eq:lin_SEM_noise_intervention_src} is assumed to be zero. In terms of the graph associated with the SCM, some variables are descendants of the label $Y$ and some variables are ancestors of $Y$. As we have seen in Example 1 and Example 2 in Section~\ref{sub:simple_motivating_examples}, DA methods such as DIP perform worse than Causal or OLSSrc in the causal setting, while DIP can have smaller target population risk than Causal or OLSSrc in the anticausal setting. When both causal and anticausal variables are present, it is no longer clear whether more sophisticated DA methods would be better than Causal or OLSSrc. To gain some intuition, we first show a simple mixed-causal-anticausal DA example to illustrate how the standard DIP, CIP and CIRM can have worse target risk than Causal. Then we introduce new algorithms to tackle the mixed-causal-anticausal DA problem.

\subsubsection{A mixed-causal-anticausal DA example to illustrate the failure of the standard DIP, CIP and CIRM}
\begin{figure}[ht]
  \centering
  \begin{minipage}{0.30\textwidth}
  \begin{tcolorbox}[width=\textwidth, colframe=gray!80, colback=gray!40, boxsep=0mm, arc=3mm]
  \begin{tikzpicture}
    % x node set with absolute coordinates
    \node[state] (x2) at (0,0) {$X_1$};
    \node[state] (x3) at (3,1) {$X_2$};
    \node[state] (x4) at (3,0) {$X_3$};
    \node[regular polygon,regular polygon sides=4,draw,fill=pink] (i) at (0,3) {$A$};

    % y node set relative to x.
    % Locations can be:
    % right,left,above,below,
    % above left,below right, etc
    \node[state] (y) at (1.5,0) {$Y$};

    % Directed edge
    % \path (y) edge (x3);
    % \path (x1) edge (y);
    \path (x2) edge (y);
    \path (y) edge (x3);
    \path (y) edge (x4);

    % \path (i) edge[dashed, bend right=30] (x1);
    \path (i) edge[dashed, bend right=30] (x2);
    \path (i) edge[dashed, bend left=30] (x3);
    \path (i) edge[dashed, bend left=30] (y);
  \end{tikzpicture}
  \end{tcolorbox}
\end{minipage}
\begin{minipage}{0.66\textwidth}
\begin{equation*}
  \begin{aligned}[c]
    \Xv{m}_1 &= \noisev{m}_{X_1} + \interAv{m}_{X_1} \\
    \Xv{m}_2 &= \Yv{m} + \noisev{m}_{X_2} + \interAv{m}_{X_2} \\
    \Xv{m}_3 &= \Yv{m} + \noisev{m}_{X_3}\\
    \Yv{m} &= \Xv{m}_1 + 0 + \interAv{m}_Y
  \end{aligned}
  \hspace{0.5cm}
  \begin{aligned}[c]
    \Xtar_1 &= \noisetar_{X_1} + 0 \\
    \Xtar_2 &= \Ytar + \noisetar_{X_2} + 0 \\
    \Xtar_3 &= \Ytar + \noisetar_{X_3} \\
    \Ytar &= \Xtar_1 + 0 + 0\\
  \end{aligned}
\end{equation*}
\end{minipage}
  \caption{The causal diagram and structural equations for the $m$-th source ($m \in \braces{1, \ldots, \envs}$) and target environments in Example 4}
  \label{fig:causal_diagram_mixed_causal_anticausal}
\end{figure}
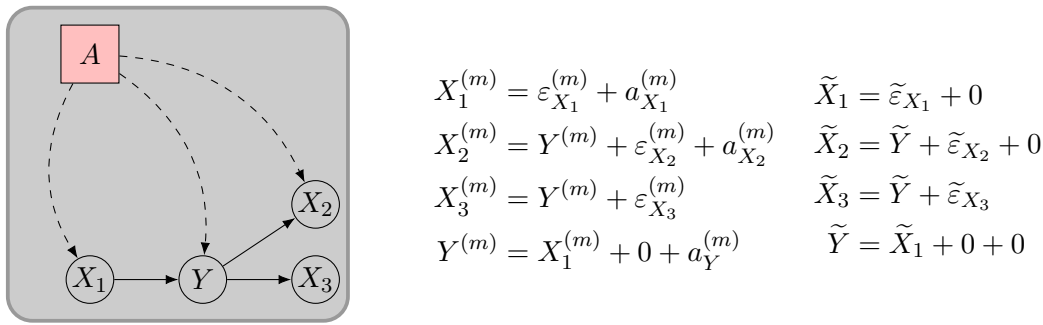

Here we provide a simple mixed-causal-anticausal DA problem to illustrate the difficulty of the mixed-causal-anticausal setting. Example 4 is a DA problem with a large number of source environments ($\envs \geq 4$) and one target environment. The data in the source and target environment are generated independently according to the following SCMs, with the causal diagram on the left and the structural equations on the right of Figure~\ref{fig:causal_diagram_mixed_causal_anticausal}.

Here the noise variables follow independent Gaussian distributions with $
  \noisev{m}_{X_i}, \noisetar_{X_i} \sim \Normal(0, 0.1)$, $i \in \braces{1, \ldots, 3}$. The noise variable of $Y$ is set to zero.
The type of intervention is mean shift noise intervention. The variable $X_3$ is never intervened on, so $X_3$ is conditionally invariant. For the first source environment, the intervention $\interAv{1} = \bmat{1 & 1 & 0 & 1}\tp$. For $m\geq 2$, $\interAv{m} = \bmat{\interAv{m}_{X_1} & \interAv{m}_{X_2} & 0 & \interAv{m}_Y} \tp$ where each coordinate is generated i.i.d. from $\Normal(0, 10)$ and we ensure that the $M-1$ interventions span a subspace of dimension 3. For the target environment, the intervention is $\bmat{0 & 0 & 0 & 0}\tp$. Compared to Example 3, the inclusion of the causal covariates $X_1$ and $X_2$ makes the DA problem more complicated.

The population estimators when $\envs$ tends to $\infty$ can be computed explicitly
\begin{align*}
  \bmat{\betaolstarget \\ \beta_{\tagolstarget, 0}} &= \bmat{1 & 0 & 0 & 0}\tp,
  &\bmat{\betacausal \\ \beta_{\tagcausal, 0}} &= \bmat{1 & 0 & 0 & 0 } \tp, \\
  \bmat{\betaolspool \\ \beta_{\tagolspool, 0}} &= \bmat{0 & 0 & 1 & 0} \tp,
  &\bmat{\betadipmeanmatchlin{1} \\ \beta_{\tagdipmeanmatchlin, 0}^\tagk{1}} &= \bmat{\frac{4}{3} & -\frac{1}{3} & -\frac{1}{6} & 2} \tp, \\
  \bmat{\betacipmean \\ \beta_{\tagcipmean, 0}} &= \bmat{0 & 0 & 1 & 0} \tp,
  &\bmat{\betacirmeanmatch{1} \\ \beta_{\tagcirmeanmatch, 0}^\tagk{1} } &= \bmat{1 & 0 & 0 & 1} \tp. \\
\end{align*}
Note that with a large number of source environments, CIP and SrcPool both identify the conditional invariant component $X_3$. The Causal estimator is the best because we made the noise variable on $Y$ to be $0$ and the intervention on $Y$ in the target to be 0. DIP$^\tagk{1}$ does not find the Causal estimator because the $Y$ intervention makes it difficult to align the covariate interventions between source and target. CIRM$^\tagk{1}$ does not find the Causal estimator because even if it can correct for the label intervention, matching on the causal covariate $X_1$ is not a good idea due to similar reason we have seen applying DIP for causal prediction in Example 1. The corresponding population source and target risks are summarize in Table~\ref{tab:population_risks_example_4}. We conclude that unlike in the anticausal DA setting, in the presence of both causal and anticausal covariates, CIRM which is based on the idea of domain invariant projection after label correction no longer outperforms SrcPool.

\begin{table}[ht]
    \centering
    % \begin{adjustwidth}{-.2in}{-.3in}
    % {\renewcommand{\arraystretch}{.5}
    \begin{tabular}{c|cccccc}
        \toprule
        % \thead{\bf \# Grad. evals $\to$ \\
        % \bf Algorithm $\downarrow$}
        \backslashbox{\thead{\bf Risk}}{\thead{\bf Methods}} & \thead{\bf OLSTar \\ (oracle)} & \thead{\bf Causal} & \thead{\bf SrcPool} & \thead{\bf DIP$^\tagk{1}$} & \thead{\bf CIP} & \thead{\bf CIRM$^\tagk{1}$} \\ [4mm] \midrule
        \thead{Ex 4, source risk $\risk^\tagk{1}$} & 1.000 & 1.000 & 0.100 & 0.000 & 0.100 & 0.000\\ [2mm]
        \thead{Ex 4, target risk $\tilde{\risk}$} & {\bf 0.000} & 0.000 & 0.100 & 4.000 & 0.100 & 1.000 \\
         [2mm]
    \bottomrule
    \end{tabular}
    % }
    % \end{adjustwidth}
    \caption{Population source and target risks in Example 4 (the lower the better). The oracle target risk is highlighted in bold. In Example 3 of mixed-causal-anticausal prediction with intervention on $Y$, DIP$^\tagk{1}$, CIP and CIRM$^\tagk{1}$ all perform worse than Causal in terms of target population risk. }
    \label{tab:population_risks_example_4}
\end{table}

One might argue that the superior performance of Causal over SrcPool is made up because we set the target label intervention $\interAtar_Y$ to be zero. The concern is valid in general. However, it is not difficult to observe that no matter how we set $\interAtar_Y$ there exist better estimators than SrcPool. In fact, if we know $X_1$ is the only causal variable, we can regress $Y, X_2, X3$ on $X_1$ and consider the prediction problem on the residuals. The resulting prediction problem is anticausal with intervention on $Y$. Consequently, it can be solved with DA methods discussed in Section~\ref{sub:anti_causal_domain_adaptation_with_intervention_on_y}. We formulate this idea precisely in the next subsection.

\subsubsection{New DA methods for mixed-causal-anticausal domain adaptation}

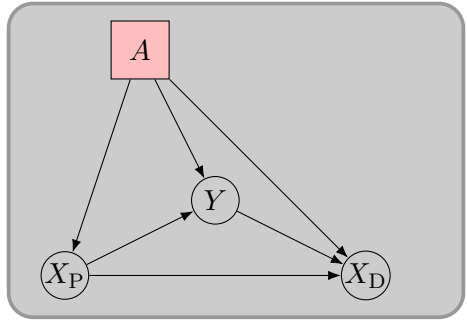
\begin{figure}[ht]
  \centering
  \begin{tcolorbox}[width=0.4\textwidth, colframe=gray!80, colback=gray!40, boxsep=0mm, arc=3mm]
  \begin{tikzpicture}
    % x node set with absolute coordinates
    \node[state] (x1) at (0,0) {$X_\tagpar$};
    \node[state] (x2) at (4,0) {$X_\tagdes$};
    \node[regular polygon,regular polygon sides=4,draw,fill=pink] (i) at (1,3) {$A$};

    % y node set relative to x.
    % Locations can be:
    % right,left,above,below,
    % above left,below right, etc
    \node[state] (y) at (2,1) {$Y$};

    % Directed edges
    \path (y) edge (x2);
    \path (x1) edge (y);
    \path (x1) edge (x2);

    \path (i) edge (x1);
    \path (i) edge (x2);
    \path (i) edge (y);
  \end{tikzpicture}
  \end{tcolorbox}
  \caption{The causal diagram for our mixed-causal-anticausal domain adaptation setup.}
  \label{fig:causal_diagram_mixed}
\end{figure}

Instead of studying the most general mixed-causal-anticausal domain adaptation, we first restrict ourselves to the setting where the ``rough causal structure around $Y$'' is known. That is, we know which covariates are the descendants of $Y$, denoted as $X_{\tagdes}$, and we also know which covariates are the parents of $Y$ or the parents of $X_{\tagdes}$ (denoted as $X_{\tagpar}$) as shown in Figure~\ref{fig:causal_diagram_mixed}. Assuming the ``rough causal structure'', we show that this mixed-causal-anticausal problem can be reduced to the familiar problem in the previous subsections. Assumption~\ref{ass:assumption_multiple_source_mixed} makes data generation requirements precise.

\begin{assumption}
  \label{ass:assumption_multiple_source_mixed}
  There are $\envs$ ($\envs \geq 1$) source environments. Each data point in the $m$-th source environment is generated i.i.d. according to distribution $\distriv{m}$ specified by the following SCM
  \begin{align*}
    \bmat{\Xv{m}_{\tagpar} \\ \Xv{m}_{\tagdes} \\ \Yv{m}} = \bmat{\semBX_\tagpar & 0  & 0 \\ \semBX_\text{D-P} & \semBX_\tagdes  & \sembv_\tagdes \\ \sembh_\tagpar \tp & 0 & 0} \bmat{\Xv{m}_{\tagpar} \\ \Xv{m}_{\tagdes} \\ \Yv{m}} + \bmat{\interAv{m}_{X, \tagpar} \\ \interAv{m}_{X, \tagdes} \\ \interAv{m}_Y} + \bmat{\noisev{m}_{X, \tagpar} \\ \noisev{m}_{X, \tagdes} \\ \noisev{m}_Y}.
  \end{align*}
  Each data point in the target environment is generated i.i.d. according to distribution $\distritar$ specified with the same SCM except for the mean shift intervention term
  \begin{align*}
    \bmat{\Xtar_{\tagpar} \\ \Xtar_{\tagdes} \\ \Ytar } = \bmat{\semBX_\tagpar & 0  & 0 \\ \semBX_\text{D-P} & \semBX_\tagdes  & \sembv_\tagdes \\ \sembh_\tagpar \tp & 0 & 0} \bmat{\Xtar_{\tagpar} \\ \Xtar_{\tagdes} \\ \Ytar} + \bmat{\interAtar_{X, \tagpar} \\ \interAtar_{X, \tagdes} \\ \interAtar_Y} + \bmat{\noisetar_{X, \tagpar} \\ \noisetar_{X, \tagdes} \\ \noisetar_Y},
  \end{align*}
  where $\semB_{\tagpar}\in \real^{\dimscau \times \dimscau}$, $\semBX_\tagdes \in \real^{(\dims-\dimscau) \times (\dims-\dimscau)}$ and $\semBX_\text{D-P} \in \real^{(\dims-\dimscau) \times \dimscau}$ are unknown constant matrices such that $\Ind_\dims - \bmat{\semBX_\tagpar & 0 \\ \semBX_\text{D-P} & \semBX_\tagdes}$ is invertible,
  $\sembh_\tagpar \in \real^{\dimscau}$ and $\sembv_\tagdes \in \real^{\dims - \dimscau}$ are unknown constant vectors, the intervention terms $\bmat{\interAv{m}_{X, \tagpar} \\ \interAv{m}_{X, \tagdes} \\ \interAv{m}_Y}$ and $\bmat{\interAtar_{X, \tagpar} \\ \interAtar_{X, \tagdes} \\ \interAtar_Y}$ are vectors in $\real^{\dims+1}$, the noise terms $\bmat{\noisev{m}_{X, \tagpar} \\ \noisev{m}_{X, \tagdes} \\ \noisev{m}_Y}$ and $\bmat{\noisetar_{X, \tagpar} \\ \noisetar_{X, \tagdes} \\ \noisetar_Y}$ share the same distribution with
  \begin{align*}
    \Exs\bmat{\noisev{m}_{X, \tagpar} \\ \noisev{m}_{X, \tagdes}} = \Exs\bmat{\noisetar_{X, \tagpar} \\ \noisetar_{X, \tagdes}} = 0, &\quad \Exs\brackets{\bmat{\noisev{m}_{X, \tagpar} \\ \noisev{m}_{X, \tagdes}} \bmat{\noisev{m}_{X, \tagpar} \\ \noisev{m}_{X, \tagdes}}\tp } = \Exs\brackets{\bmat{\noisetar_{X, \tagpar} \\ \noisetar_{X, \tagdes}} \bmat{\noisetar_{X, \tagpar} \\ \noisetar_{X, \tagdes}}\tp } =  \bmat{\noisecovX_\tagpar & 0 \\ 0 & \noisecovX_\tagdes}, \\
    \Exs\brackets{\noisev{1}_Y} = \Exs\brackets{\noisetar_Y} = 0, &\quad \Exs\brackets{{\noisev{1}_Y}^2} = \Exs\brackets{{\noisetar_Y}^2} = \noisecovY^2.
  \end{align*}
  Additionally, the noise terms on $X$ and $Y$ are uncorrelated, namely, $
    \Exs \brackets{\noisev{m}_Y \cdot \bmat{\noisev{m}_{X, \tagpar} \\ \noisev{m}_{X, \tagdes}}} = \Exs \brackets{\noisetar_Y \cdot \bmat{\noisetar_{X, \tagpar} \\ \noisetar_{X, \tagdes}}} = 0$.
\end{assumption}
Compared to the SCMs in Assumption~\ref{ass:assumption_single_source_anti_causal} and~\ref{ass:assumption_multiple_source_anti_causal}, Assumption~\ref{ass:assumption_multiple_source_mixed} introduces additional covariates $X_\tagpar$ which are the parents of $Y$ or $X_\tagdes$ and are located at the first $\dimscau$ coordinates. In the most general DA problem, this information of which covariates are parents may not be available. The information about causal relationships can be obtained for example from domain experts or from the causal discovery literature. The causal discovery problem is well studied and we refer the readers to \cite{glymour2019review} for a review of the causal discovery methods. We leave the most general mixed-causal-anticausal domain adaptation and the design of new DA methods to future work.

Based on Assumption~\ref{ass:assumption_multiple_source_mixed}, we can introduce the intermediate random variables
\begin{align}
  \label{eq:def_intermediate_X_Y}
  \Xv{m}_{\tagint} &\defn \Xv{m}_{\tagdes} - \parenth{\Ind_{\dims-\dimscau} - \semBX_\tagdes}^{-1}\semBX_\text{D-P} \Xv{m}_{\tagpar} - \parenth{\Ind_{\dims-\dimscau} - \semBX_\tagdes}^{-1}\sembv_\tagdes \sembh_\tagpar\tp \Xv{m}_{\tagpar} \notag \\
  \Yv{m}_{\tagint} &\defn \Yv{m} - \sembh_\tagpar\tp \Xv{m}_{\tagpar} \notag \\
  \Xtar_{\tagint} &\defn \Xtar_{\tagdes} - \parenth{\Ind_{\dims-\dimscau} - \semBX_\tagdes}^{-1}\semB_\text{D-P} \Xtar_{\tagpar} - \parenth{\Ind_{\dims-\dimscau} - \semBX_\tagdes}^{-1}\sembv_\tagdes \sembh_\tagpar\tp \Xtar_{\tagpar} \notag \\
  \Ytar_{\tagint} &\defn \Ytar - \sembh_\tagpar\tp \Xtar_{\tagpar}.
\end{align}
The intermediate random variable can be seen as the residuals after regressing on the parent variables $X_\tagpar$.
We observe that the intermediate random variables satisfy the following SCMs
\begin{align*}
  \bmat{\Xv{m}_{\tagint} \\ \Yv{m}_{\tagint}} &= \bmat{\semBX_\tagdes & \sembv_\tagdes \\ 0 & 0} \bmat{\Xv{m}_{\tagint} \\ \Yv{m}_{\tagint}} + \bmat{\interAv{m}_{X, \tagdes} \\ \interAv{m}_Y} + \bmat{\noisev{m}_{X, \tagdes} \\ \noisev{m}_Y} \\
  \bmat{\Xtar_{\tagint} \\ \Ytar_{\tagint}} &= \bmat{\semBX_\tagdes & \sembv_\tagdes \\ 0 & 0} \bmat{\Xtar_{\tagint} \\ \Ytar_{\tagint}} + \bmat{\interAtar_{X, \tagdes} \\ \interAtar_Y} + \bmat{\noisetar_{X, \tagdes} \\ \noisetar_Y}.
\end{align*}
Using the intermediate random variables $(\Xv{m}_{\tagint}, \Yv{m}_{\tagint})$, the mixed-causal-anticausal DA problem can be reduced to the anticausal DA problem. Intuitively, it works as follows. First, regress $Y$ and $X_\tagdes$ on the first $\dimscau$ covariates $X_\tagpar$. Second, create a transformed dataset with new covariates and new labels from the corresponding residuals like how we define the intermediate random variables. Third, apply the DA methods in the anticausal DA setting to the transformed dataset. Finally, bring back the original covariates and labels in the final estimator.

Based on the above intuition, we introduce the following estimators for the mixed-causal-anticausal DA setting.
\begin{itemize}
  \item \textbf{DIP$\diamondsuit^\tagk{m}$-mean:} the population domain invariant projection estimator for the mixed-causal-anticausal DA setting.
  \begin{align}
    \label{eq:estimator_pop_dipmeanmix}
    f_{\tagdipmeanmix}^\tagk{m}(x) &\defn x \tp \bmat{\gamma^\tagk{m} - \Gamma^\tagk{m} \betadipmeanmix{m} \\ \betadipmeanmix{m}} + \beta_{\tagdipmeanmix, 0}^\tagk{m} \notag \\
    \betadipmeanmix{m}, \beta_{\tagdipmeanmix, 0}^\tagk{m} &\defn \argmin_{\beta, \beta_0}\  \Exs_{(X_\tagpar, X_\tagdes, Y) \sim \distriv{m}}\parenth{Y_\tagint - X_\tagint\tp\beta - \beta_0}^2 \notag \\
    &\text{s.t.\ } \Exs_{\parenth{X_\tagpar, X_\tagdes} \sim \distriv{m}_X} \brackets{X_\tagint \tp\beta} = \Exs_{\parenth{X_\tagpar, X_\tagdes} \sim \distritar_X}\brackets{X_\tagint \tp\beta},
  \end{align}
  where $Y_\tagint \defn Y - X_\tagpar \tp \gamma^\tagk{m}$, $X_\tagint \defn X_\tagdes - {\Gamma^\tagk{m}} \tp X_\tagpar$, and the regression weights $\gamma^\tagk{m}$ and $\Gamma^\tagk{m}$ are defined as
  \begin{align}
    \label{eq:solve_gamma_dipmix}
    \gamma^\tagk{m}, \gamma_0^\tagk{m} &\defn \argmin_{\gamma, \gamma_0 \in \real^{\dimscau} \times \real} \Exs_{\parenth{X_\tagpar, X_\tagdes, Y} \sim \distriv{m}} \parenth{Y - X_\tagpar \tp \gamma - \gamma_0}^2, \\
    \label{eq:solve_Gamma_dipmix}
    \Gamma^\tagk{m}, \Gamma_0^\tagk{m} &\defn \argmin_{\Gamma, \Gamma_0 \in \real^{\dimscau \times (\dims-\dimscau)} \times \real^{\dims-\dimscau}} \Exs_{\parenth{X_\tagpar, X_\tagdes} \sim \distriv{m}_X} \vecnorm{X_\tagdes - \Gamma \tp X_\tagpar - \Gamma_0}{2}^2.
  \end{align}
\end{itemize}
\begin{corollary}
  \label{cor:single_source_mixed_mean_shift}
  Under the data generation Assumption~\ref{ass:assumption_multiple_source_mixed}, $\envs=1$ and the assumption of no intervention on $Y$ i.e. $\interAv{1}_Y = \interAtar_Y = 0$, the target population risks of OLSTar and DIP$\diamondsuit^\tagk{1}$-mean satisfy
  \begin{align}
    \label{eq:single_source_mixed_risk_olsoracle}
    \tilde{\risk}\parenth{f_{\tagolstarget}} &= \frac{\noisecovY^2}{1 + \noisecovY^2 \sembv_\tagdes \tp \noisecovX_\tagdes^{-1}\sembv_\tagdes}, \\
    \label{eq:single_source_mixed_risk_dipmeanmix}
    \tilde{\risk}\parenth{f_{\tagdipmeanmix}^\tagk{1}} &= \frac{\noisecovY^2}{1 + \noisecovY^2 \sembv_\tagdes\tp \noisecovX_\tagdes^{-\frac{1}{2}} \Gdipmixv{1} \noisecovX_\tagdes^{-\frac{1}{2}} \sembv_\tagdes },
  \end{align}
  where $\Gdipmixv{1}$ is a projection matrix with rank $\dims-1$.
\end{corollary}
The proof of Corollary~\ref{cor:single_source_mixed_mean_shift} is provided in Appendix~\ref{sub:proof_of_cor_6}. The target risk results in Corollary~\ref{cor:single_source_mixed_mean_shift} is very similar to those in Theorem~\ref{thm:single_source_anticausal_mean_shift}. This is because we reduce the mixed-causal-anticausal DA problem without intervention on $Y$ under Assumption~\ref{ass:assumption_multiple_source_mixed} to the anticausal DA problem without intervention on $Y$ under Assumption~\ref{ass:assumption_single_source_anti_causal}. According to Equation~\eqref{eq:single_source_mixed_risk_dipmeanmix}, the target population risk of DIP$\diamondsuit^\tagk{1}$ is lower than the target risk of Causal (which equals to $\noisecovY^2$).

Based on the intuition of DIP$\diamondsuit$, a similar strategy can be applied to extend CIP and CIRM to the mixed-causal-anticausal DA problems. The precise formulations of the extensions CIP$\diamondsuit$ and CIRM$\diamondsuit$ are introduced in Appendix~\ref{sub:population_DA_methods}.

Just as Corollary~\ref{cor:single_source_mixed_mean_shift} serves as the equivalent of Theorem~\ref{thm:single_source_anticausal_mean_shift} in the mixed-causal-anticausal DA setting, the equivalent of Theorem~\ref{thm:mutiple_source_anticausal_mean_shift} can be established for CIP$\diamondsuit$ and CIRM$\diamondsuit$. Additionally, we can also introduce the weighted version of CIRM$\diamondsuit$, called CIRM$\diamondsuit$weigh, following the discussion of CIRMweigh. The details are omitted.

% subsection mixed_causal_anti_causal_domain_adaptation (end)

% section anti_causal_domain_adaptation_guarantees (end)

\section{Numerical experiments} % (fold)
\label{sec:numerical_experiments}

In this section, we numerically compare DA methods in simulated, synthetic and real datasets. The experiments are used to validate our theoretical results in finite sample data, to illustrate DA failure modes when assumptions are violated and to provide ideas of adapting DA methods to scenarios where not all assumptions are satisfied. Section~\ref{sub:finite_sample_formulations_and_hyperparameter_choices} formulates the finite-sample DA estimators from the population DA ones to make sure our implementations are reproducible. Section~\ref{sub:linear_sem_simulations} contains simulation experiments from deterministic or randomly generated linear SCMs. Section~\ref{sub:mnist_experiments_with_synthetic_interventions} discusses the performance of DA estimators on the MNIST dataset with synthetic interventions. Finally, Section~\ref{sub:real_data_experiments_with_unknown_interventions} illustrates through a real data experiment that DA can be difficult in practice when not much domain knowledge about the data generating model is available.

Our code to reproduce all the numerical experiments is publicly available in the Github repository \href{https://github.com/yuachen/CausalDA}{https://github.com/yuachen/CausalDA}.

\subsection{Finite-sample formulations and hyperparameter choices} % (fold)
\label{sub:finite_sample_formulations_and_hyperparameter_choices}
Here we introduce the regularized formulations of the finite-sample DIP, CIP and CIRM. The DIP matching penalty in Equation~\eqref{eq:estimator_pop_dipmeanmatchlin} and the conditional invariance penalty in Equation~\eqref{eq:estimator_pop_cipmean} are enforced via regularization terms. The finite-sample versions of their variants can be formulated similarly after translating the constraints to regularization terms. For the sake of space, they are presented in Appendix~\ref{sub:finite_sample_formulation_of_DA_methods}.

\begin{itemize}
  \item \textbf{DIP$^\tagk{m}$-mean-finite:} the finite-sample formulation of the DIP$^\tagk{m}$-mean estimator~\eqref{eq:estimator_pop_dipmeanmatchlin}. The mean squared difference is used as distributional distance and is enforced via a regularization term,
  \begin{align}
    \label{eq:estimator_finite_dipmeanmatchlin}
    \hat{f}_{\tagdipmeanmatchlin}^\tagk{m}(x) &\defn x \tp \hat{\beta}_{\tagdipmeanmatchlin}^\tagk{m} + \hat{\beta}_{\tagdipmeanmatchlin, 0}^\tagk{m} \notag \\
    \hat{\beta}_{\tagdipmeanmatchlin}^\tagk{m}, \hat{\beta}_{\tagdipmeanmatchlin, 0}^\tagk{m} &\defn \argmin_{\beta, \beta_0}\  \frac{1}{\obs_m}\sum_{i=1}^{\obs_m} \parenth{y_i^\tagk{m} - {x_i^\tagk{m}}\tp\beta - \beta_0}^2 +  \notag \\
    &\hspace{3cm} \lamMatch  \parenth{\frac{1}{\obs_m} \sum_{i=1}^{\obs_m} {x_i^\tagk{m}}\tp\beta - \frac{1}{\obstar}\sum_{i=1}^{\obstar} \tilde{x}_i\tp\beta}^2,
  \end{align}
  where $\lamMatch$ is a positive regularization parameter that controls the match between the covariate mean of the source and target environment.
  \item \textbf{CIP-mean-finite:} the finite sample formulation of the CIP-mean estimator~\eqref{eq:estimator_pop_cipmean}. The conditional mean is matched across source environments and is enforced via a regularization term,
  \begin{align}
    \label{eq:estimator_finite_cipmean}
    \hat{f}_{\tagcipmean}(x) &\defn x \tp\hat{\beta}_\tagcipmean + \hat{\beta}_{\tagcipmean, 0}  \notag \\
    \hat{\beta}_{\tagcipmean}, \hat{\beta}_{\tagcipmean, 0} &\defn \argmin_{\beta, \beta_0}\ \frac{1}{\envs} \sum_{m=1}^\envs \frac{1}{\obs_m} \sum_{i=1}^{\obs_m} \parenth{y_i^\tagk{m} - {x_i^\tagk{m}}\tp\beta - \beta_0}^2 + \notag \\
    & \hspace{3cm}\frac{\lamCIP}{\envs^2} \sum_{m,k=1}^\envs \parenth{\frac{1}{\obs_m} \sum_{i=1}^{\obs_m} {z_{\text{cond}, i}^\tagk{m}}\tp \beta - \frac{1}{\obs_k} \sum_{i=1}^{\obs_k} {z_{\text{cond}, i}^\tagk{k}}  \tp \beta }^2,
  \end{align}
  where $z_{\text{cond}, i}^\tagk{k} = x_i^\tagk{k} -  \frac{y_i^\tagk{k} \bar{y}^\tagk{k} }{ \parenth{\frac{1}{\obs_k} \sum_{j=1}^{\obs_k} {y_j^\tagk{k}}^2}} {x_i^\tagk{k}}$, $\bar{y}^\tagk{k} = \frac{1}{\obs_k} \sum_{j=1}^{\obs_k} y_j^\tagk{k}$ and $\lamCIP$ is a positive regularization parameter that controls the strength of the conditional invariant penalty. In the finite-sample formulation, the conditional expectation in the constraint of population formulation~\eqref{eq:estimator_pop_cipmean} is computed via regressing $X\tp \beta$ on $Y$. As a result, $z_{\text{cond}, i}^\tagk{k}$'s are the residuals after regressing $x_i^\tagk{k}$ on $y_i^{\tagk{k}}$.
  \item \textbf{CIRM$^\tagk{m}$-mean-finite:} the finite-sample formulation of the CIRM$^\tagk{m}$-mean estimator~\eqref{eq:estimator_pop_cirmeanmatch}. The residual after removing conditionally invariant components is matched between source and target environments. The matching is enforced via a regularization term.
  \begin{align}
    \label{eq:estimator_finite_cirmeanmatch}
    \hat{f}_{\tagcirmeanmatch}^\tagk{m}(x) &\defn x \tp\hat{\beta}_{\tagcirmeanmatch}^\tagk{m} + \hat{\beta}_{\tagcirmeanmatch, 0}^\tagk{m} \notag \\
    \hat{\beta}_{\tagcirmeanmatch}^\tagk{m}, \hat{\beta}_{\tagcirmeanmatch, 0}^\tagk{m} &\defn \argmin_{\beta, \beta_0}\  \frac{1}{\obs_m} \sum_{i=1}^{\obs_m} \parenth{y_i^\tagk{m} - {x_i^\tagk{m}}\tp\beta - \beta_0}^2 \notag \\
    & \hspace{3cm} + \lamMatch \parenth{\frac{1}{\obs_m} \sum_{i=1}^{\obs_m} {z_{\text{res}, i}^\tagk{m}} \tp \beta - \frac{1}{\tilde{\obs}} \sum_{i=1}^{\tilde{\obs}} {\tilde{z}_{\text{res}, i}} \tp \beta }^2,
  \end{align}
  where $z_{\text{res}, i}^\tagk{m} = x_i^\tagk{m} - \parenth{{x_i^\tagk{m}}\tp \hat{\beta}_\tagcipmean} \hat{\vartheta}_{\tagcirmeanmatch}$ with $\hat{\vartheta}_{\tagcirmeanmatch}$ defined as
  \begin{align*}
    \hat{\vartheta}_{\tagcirmeanmatch} = \frac{\frac{1}{\envs} \sum_{m=1}^\envs \frac{1}{\obs_m} \sum_{i=1}^{\obs_m} \parenth{y_i^\tagk{m} - \bar{y}^\tagk{m}}x_i^\tagk{m} }{\frac{1}{\envs} \sum_{m=1}^\envs \frac{1}{\obs_m} \sum_{i=1}^{\obs_m} \parenth{y_i^\tagk{m} - \bar{y}^\tagk{m}}\parenth{{x_i^\tagk{m}} \tp \hat{\beta}_\tagcipmean - \frac{1}{\obs_m} \sum_{j=1}^{\obs_m} {x_j^\tagk{m}} \tp \hat{\beta}_\tagcipmean }},
  \end{align*}
  and $\lamMatch$ is a positive regularization parameter similar to the one define in DIP. Just like the population CIRM depends on the population CIP, the finite-sample CIRM also depends on the finite-sample CIP with the same regularization parameter $\lamCIP$.
\end{itemize}

% subsection finite_sample_formulations_and_hyperparameter_choices (end)

\paragraph{Regularization parameter choices: } Choosing the regularization parameters in finite-sample DA formulations such as $\lamMatch$ in DIP formulation~\eqref{eq:estimator_finite_dipmeanmatchlin} is a difficult subject. Because the DA setting here assumes no access to any target labels, one cannot get good estimates of the target performance easily. The classical model selection strategies based on a validation set are no longer applicable in domain adaptation.

% First, we consider the $\lamMatch$ choice in the DIP formulation~\eqref{eq:estimator_finite_dipmeanmatchlin}. Intuitively, if $\lamMatch$ is too small, then the regularization has no effect. DIP solution is close to the OLSSrc solution for small regularization $\lamMatch$. On the ther hand, if $\lamMatch$ is too large, then the regularization term in Equation~\eqref{eq:estimator_finite_dipmeanmatchlin} overfits the finite sample noise in the DIP penalty and cannot achieve a small source risk. Based on the intuition, we would like to choose $\lamMatch$ large enough but not too large to avoid overfitting the finite sample noise.

We make use of the fact that when DIP works as Theorem~\ref{thm:single_source_anticausal_mean_shift} predicts, the source population risk of DIP equals to the target population risk as shown in Corollary~\ref{cor:single_source_anticausal_mean_shift_source_pop_risk}. The source finite-sample risk is close to the target finite-sample risk up to finite sample errors. So we would like to choose $\lamMatch$ large enough so that the DIP matching penalty takes effect, but not too large so that the source finite risk remains reasonably small. We propose to choose the largest $\lamMatch$ so that the source finite-sample risk is less than two times of the source finite-sample risk when $\lamMatch$ is set to zero. The choice of ``two times'' is arbitrary here, the precise amount depends on the desired target finite-sample risk,  the sample size and the variance of the source finite risk estimate. The precise amount is specified for each DA dataset separately. The regularization parameter $\lamMatch$ in DIPweigh is chosen similarly.

Next, we consider the $\lamCIP$ choice in CIP formulation~\eqref{eq:estimator_finite_cipmean}. Since CIP only uses source data and never touches the target data, we leave a small part of each source data out for validation and choose $\lamCIP$ based on the validation source data. We choose $\lamCIP$ so that the average source risk across source environment is small.

Finally, the finite-sample CIRM formulation~\eqref{eq:estimator_finite_cirmeanmatch} requires the choice of both $\lamMatch$ and $\lamCIP$. The $\lamCIP$ in CIRM is the same as the $\lamCIP$ in CIP. With $\lamCIP$ fixed, we choose $\lamMatch$ in CIRM as we did for DIP.

\subsection{Linear SCM simulations} % (fold)
\label{sub:linear_sem_simulations}
In this section, we numerically compare DA estimators via simulations from linear SCMs. First, we consider seven simulations with data generated via linear SCMs with mean shift noise interventions. The first three simulations aim at illustrating the results in Theorem~\ref{thm:single_source_anticausal_mean_shift} and Theorem~\ref{thm:mutiple_source_anticausal_mean_shift}. The last four are to show the performance of DA estimators when at least one assumption is misspecified. Second, we consider two simulations where the interventions are variance shift noise interventions. Through these two simulations, we demonstrate the necessity of adapting the DIP matching penalties according to the type of interventions. The simulation settings are summarized in Table~\ref{tab:summary_simulations}.
\begin{table}[ht]
    \centering
    % \begin{adjustwidth}{-.2in}{-.3in}
    {\renewcommand{\arraystretch}{.4}
    \begin{tabular}{cccccccc}
        \toprule
         \thead{\bf Sim \\ \bf Num} & \thead{ \bf \# Src \\\bf envs} & \thead{
         \bf Causal \\ \bf Direction}  & \thead{ \bf Interv X \\ \bf type}
         & \thead{\bf Interv \\ \bf on Y?} & \thead{\bf Has \\\bf CIC? } & \thead{\bf Better \\ estimator(s)} & \thead{\bf Baseline \\ estimator(s) }
        \\ \midrule\\
        (i) & single & anticausal & mean shift & N & - & DIP$^\tagk{1}$ & OLSSrc$^\tagk{1}$
        \\[2mm]
        \hline\\
        (ii) & multiple & anticausal & mean shift & N & - & DIPweigh & \shortstack{\small OLSPool \\ \small DIP\tagk{1}}
        \\[2mm]
        \hline\\
        (iii) & multiple & anticausal & mean shift & Y & Y & CIRMweigh & \shortstack{\small OLSPool \\ \small DIPweigh}
        \\[2mm]
        \hline\\
        {\color{red} (iv)} & single & {\color{red} causal} & mean shift & N & - & - & OLSSrc$^\tagk{1}$
        \\[2mm]
        \hline\\
        {\color{red} (v)} & single & {\color{red} mixed} & mean shift & N & - & DIP$\diamondsuit^\tagk{1}$ & \shortstack{\small OLSSrc$^\tagk{1}$ \\ \small DIP$^\tagk{1}$}
        \\[2mm]
        \hline\\
        {\color{red} (vi)} & multiple & {anticausal} & mean shift & Y & {\color{red} N} & - & OLSPool
        \\[2mm]
        \hline\\
        {\color{red} (vii)} & multiple & {\color{red} mixed} & mean shift & Y & Y & CIRM$\diamondsuit$weigh & \shortstack{\small OLSPool \\ \small CIRMweigh}
        \\[2mm]
        \hline\\
        {\color{red} (viii)} & single & {anticausal} & {\color{red} var shift} & N & - & \shortstack{\small DIP-std+ \\ \small DIP-MMD} & \shortstack{\small OLSSrc$^\tagk{1}$ \\ \small DIP}
        \\[2mm]
        \hline\\
        {\color{red} (ix)} & multiple & {anticausal} & {\color{red} var shift} & Y & Y & \shortstack{\small CIRMweigh-std+ \\ \small CIRMweigh-MMD} & OLSPool
        \\[2mm]
        \bottomrule
    \end{tabular}
    }
    % \end{adjustwidth}
    \caption{List of linear SCM simulations. The first three simulations are done under the correct assumptions according to the main theorems. Starting from Simulation (iv), some assumptions are modified to show the performance of DA methods under misspecified assumptions (highlighted in red). The last two simulations are done with variance shift noise intervention. ``Y'' means yes, ``N'' means no, ``-'' means the information is irrelevant in that setting.}
    \label{tab:summary_simulations}
\end{table}

\subsubsection{Linear SCM with mean shift noise interventions} % (fold)
\label{ssub:linear_sem_with_mean_shifts}
We consider seven simulations with data generated via linear SCMs with mean shift noise interventions.
In all experiments in this subsection, we use large sample size $\obs=5000$ for each environment, $\obs=5000$ target test data for the evaluation of the target risk and we fix the regularization parameters $\lamMatch = 10.0$ and $\lamCIP = 1.0$ to focus the discussions on the choice of DA estimators.

\begin{figure}[ht]
  \begin{minipage}{0.75\textwidth}
    \centering
    \includegraphics[width=0.99\textwidth]{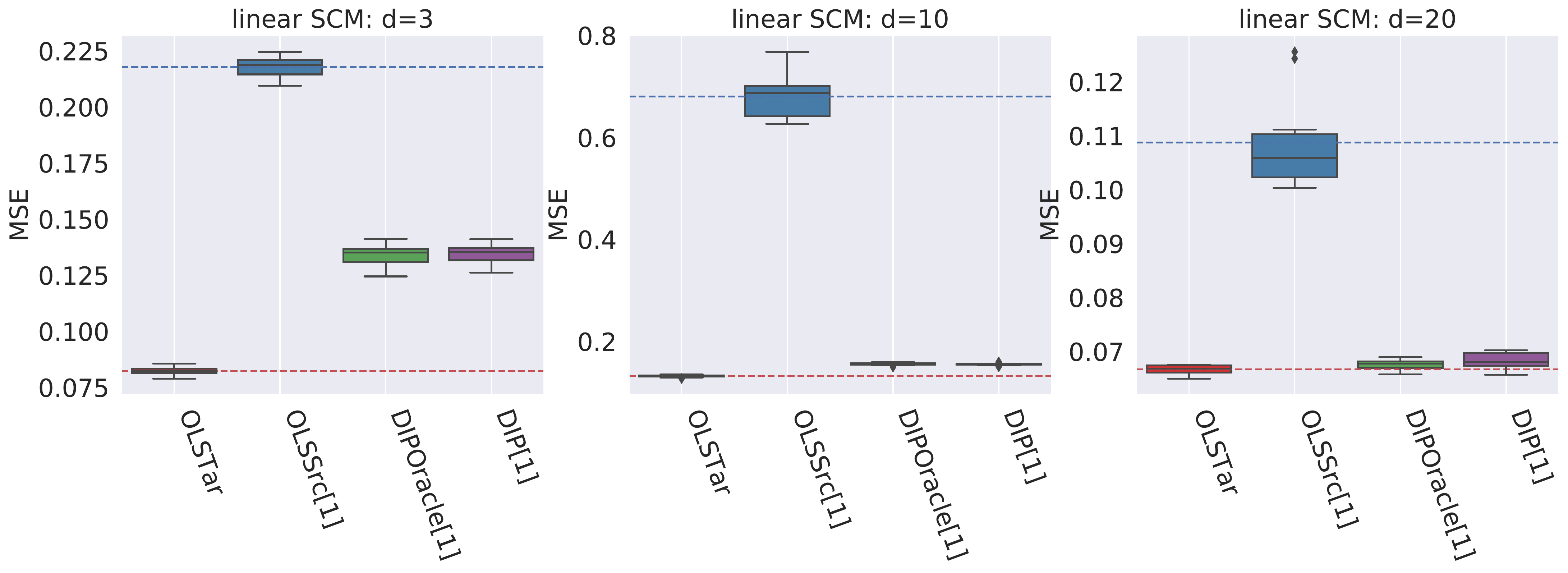}
  \end{minipage}\hfill
  \begin{minipage}{0.25\textwidth}
    \centering
    \includegraphics[width=0.99\textwidth]{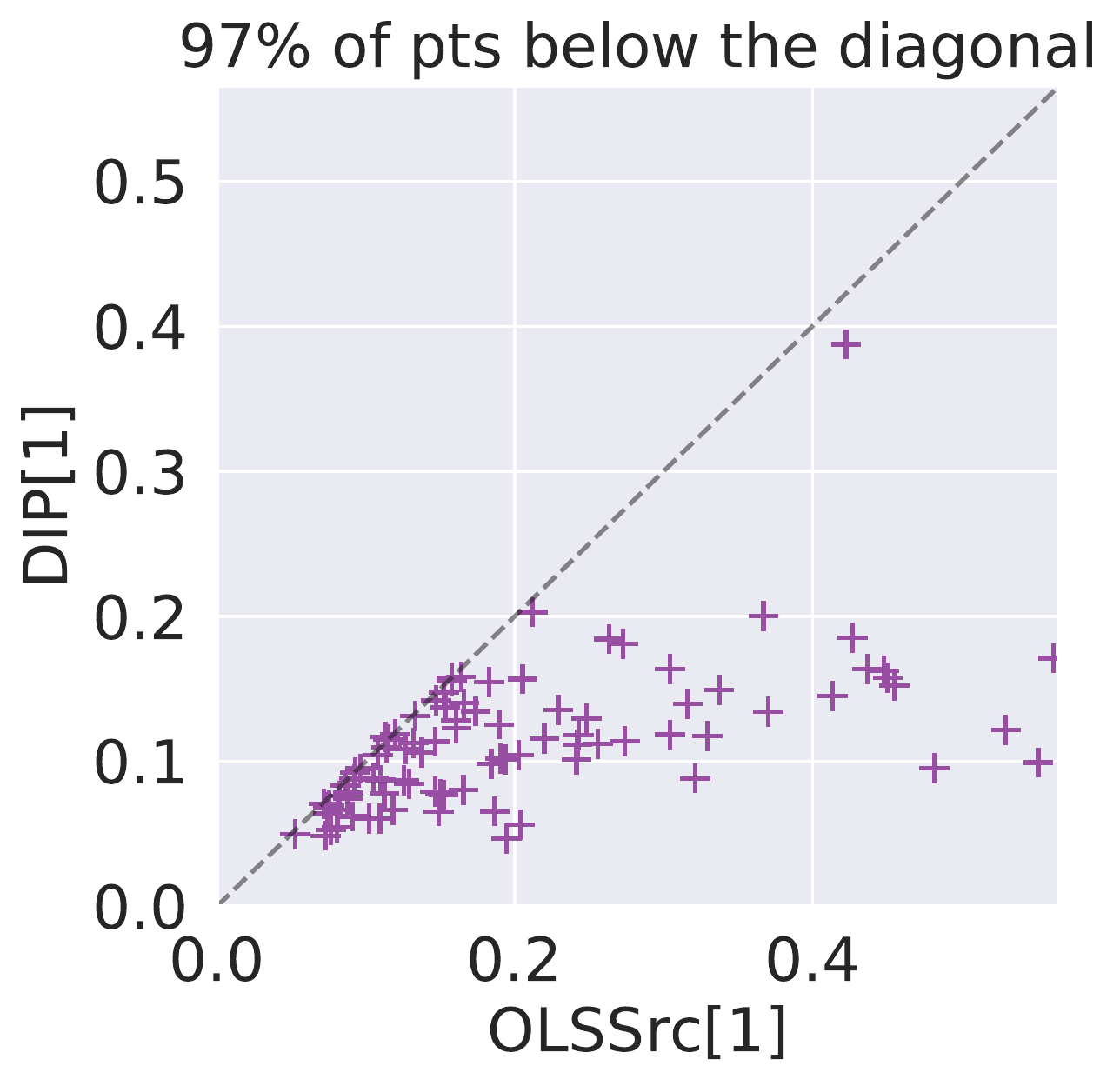}
  \end{minipage}\hfill
  \caption{The target risk comparison in simulation (i) single source anticausal DA without Y intervention (the lower the better). In all three datasets ($\dims = 3, 10, 20$) in the left three plots, DIP$^\tagk{1}$ has lower target risk than OLSSrc$^\tagk{1}$. The target risk of DIP$^\tagk{1}$ is close to that of DIPOracle$^\tagk{1}$. The last plot shows that in 98 out of 100 random coefficient $\semBX$ data generations ($\dims=10$), DIP$^\tagk{1}$ has lower target risk than OLSSrc$^\tagk{1}$. }
  \label{fig:sim_6_1_1}
\end{figure}

\paragraph{(i) Single source anticausal DA without Y intervention: } We consider three datasets of dimension $\dims=3, 10, 20$. In each dataset, there is one source environment and one target environment generated according to Assumption~\ref{ass:assumption_single_source_anti_causal}.
The matrix $\semB$ and the vector $\sembv$ is specified as follows
\begin{itemize}
  \item For $\dims=3$, the matrix $\semBX = 0 $, $\sembv = \brackets{ 1, -1, 3} \tp $ and $\sembh = 0$.
  \item For $\dims=10, 20$, the matrix $\semBX$ is upper triangular with diagonal zero and with each entry drawn i.i.d. from $\Normal(0, 0.25)$.  Each entry of $\sembv$ i.i.d. from $\Normal(0, 1)$. $\sembh = 0$.
\end{itemize}
The interventions $\interAv{1}_X$ and $\interAtar_X$ are generated i.i.d. from $\Normal(0, \Ind_\dims)$ and stay the same for all data points in one environment.
For each data point, $\noisev{1}$ or $\noise$ is generated i.i.d. from the standard Gaussian distribution $\Normal(0, \Ind_{\dims+1})$.
For each dataset, the matrix $\semBX$ and the vector $\sembv$ of the SCM are generated once; the noise and interventions are generated i.i.d. $10$ times with source and target sample size $\obs = \obstar = 5000$. The boxplot of the 10 target risks is reported for OLSTar, OLSSrc$^\tagk{1}$, DIP$^\tagk{1}$ and DIPOracle$^\tagk{1}$ in Figure~\ref{fig:sim_6_1_1}. The target risk of OLSTar is highlighted in red dashed horizontal line. The target risk of OLSSrc$^\tagk{1}$ is highlighted in blue dashed horizontal line. The first three plots in Figure~\ref{fig:sim_6_1_1} show that the target risk of DIP$^\tagk{1}$ is similar to that of DIPOracle$^\tagk{1}$ and it is very close to OLSTar as predicted by Theorem~\ref{thm:single_source_anticausal_mean_shift}.

The target risk performance of DA estimators depends on the how the matrix $\semB$ and the vector $\sembv$ are generated. To make the comparison less dependent on the randomness in the matrix $\semB$ and the vector $\sembv$, we complement the boxplot with a scatterplot that compares the target risks of DIP$^\tagk{1}$ and OLSSrc$^\tagk{1}$ over 100 random generations of the matrix $\semB$ for $\dims=10$. The last plot in Figure~\ref{fig:sim_6_1_1} shows that in 97 out of 100 random data generations, DIP$^\tagk{1}$ has lower target risk than OLSSrc$^\tagk{1}$.

\paragraph{(ii) Multiple source anticausal DA without Y intervention: } We consider three datasets of covariate dimension $\dims=3, 10, 20$. In each dataset, there is $\envs \geq 1$ source environments and one target environment generated according to Assumption~\ref{ass:assumption_single_source_anti_causal}. The matrix $\semB$, the vector $\sembv$, the interventions $\interA$ and noises $\noise$ are generated as in simulation (i). The only difference is the availability of multiple ($\envs \geq 1$) source environments.

\begin{figure}[ht]
    \begin{minipage}{0.75\textwidth}
    \centering
    \includegraphics[width=0.99\textwidth]{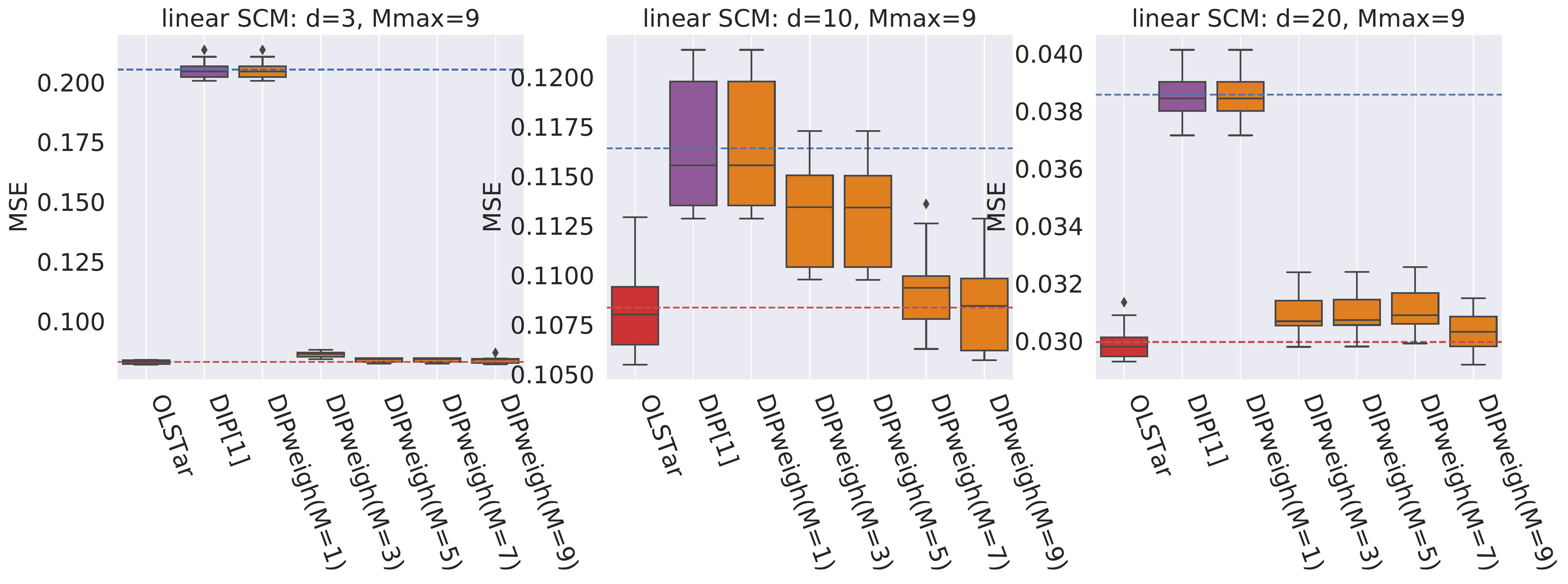}
  \end{minipage}\hfill
  \begin{minipage}{0.25\textwidth}
    \centering
    \includegraphics[width=0.99\textwidth]{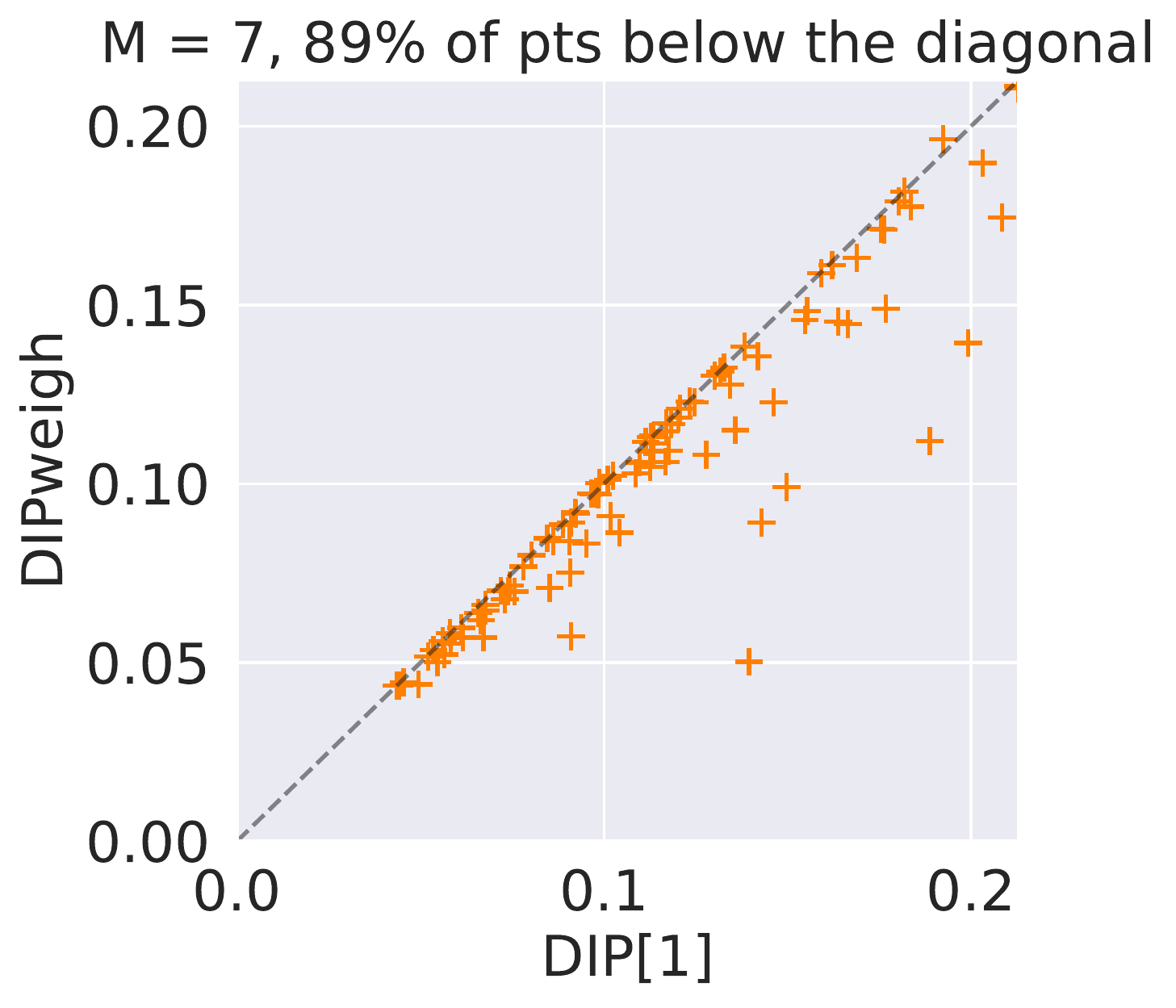}
  \end{minipage}\hfill
  \caption{The target risk comparison in simulation (ii) multiple source anticausal DA without Y intervention (the lower the better). The left three plots show that with more number of source environments, DIPweigh perform better. The last plot shows that in 89 out of 100 simulations, DIPweigh with $\envs=8$ has lower target risk than DIP$^\tagk{1}$. }
  \label{fig:sim_6_1_2}
\end{figure}

The boxplot of the 10 target risks are reported for OLSTar, DIP$^\tagk{1}$ and DIPweigh with increasing number of source environments in Figure~\ref{fig:sim_6_1_2}. The constant $\rho$ in DIPweigh formulation~\eqref{eq:estimator_pop_dipweigh} is fixed to be $1000$. To make the comparison less dependent on the randomness in the matrix $\semB$, we complement the boxplot with a scatterplot that compares the target risks of DIPweigh($\envs=8$) and DIP$^\tagk{1}$ over 100 random generations of the matrix $\semB$ for $\dims=10$. Figure~\ref{fig:sim_6_1_2} shows that in the anticausal DA setting without Y intervention, the more source environments the lower the target risk DIPweigh can achieve.

\paragraph{(iii) Multiple source anticausal DA with Y intervention and with CICs: } We fix the dimension $\dims=20$ and the number of source environments $\envs=14$. The source and target datasets are generated similarly to simulation (ii) with two exceptions: 1. there are interventions on $Y$ for the target environment: $\interAtar_Y$ follows $\Normal(0, 1)$ and it is generated once then it is fixed for the target environment. 2. there are conditionally invariant components (CICs): the interventions on $X$ only apply to first $10$ coordinates. That is, $\interAv{m}_X$ and $\interAtar_X$ have the last $10$ coordinates equal to zero. Note that we don't explicitly assume Equation~\eqref{eq:assumption_multiple_source_target_interv_in_span} in Assumption~\ref{ass:assumption_multiple_source_anti_causal}. Instead we expect that a large number of source environments will span the vector space with last $10$ coordinates zero and make the assumption~\eqref{eq:assumption_multiple_source_target_interv_in_span} satisfied.

\begin{figure}[ht]
  \begin{minipage}{0.25\textwidth}
    \centering
    \includegraphics[width=0.99\textwidth]{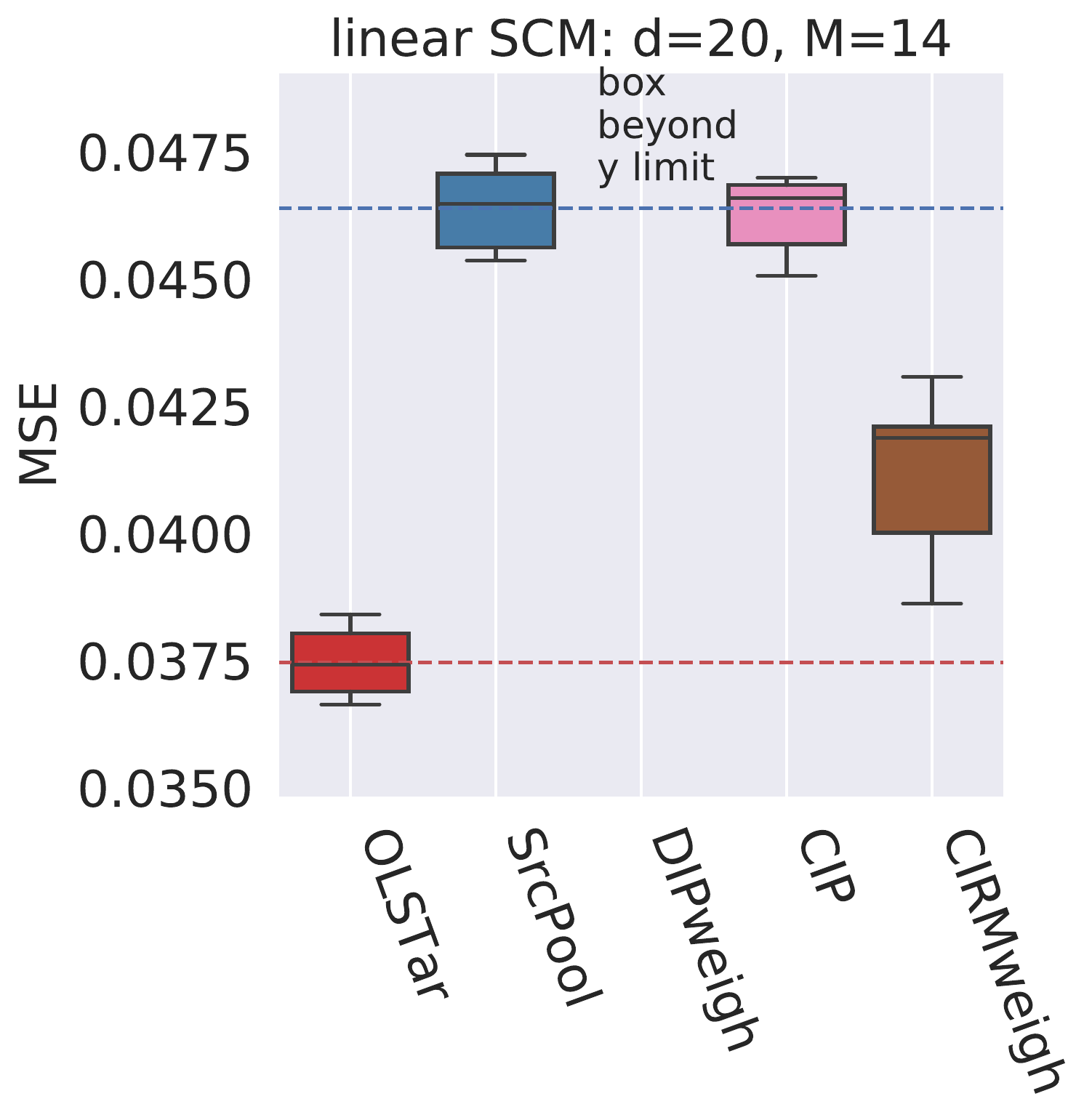}
  \end{minipage}\hfill
  \begin{minipage}{0.75\textwidth}
    \centering
    \includegraphics[width=0.99\textwidth]{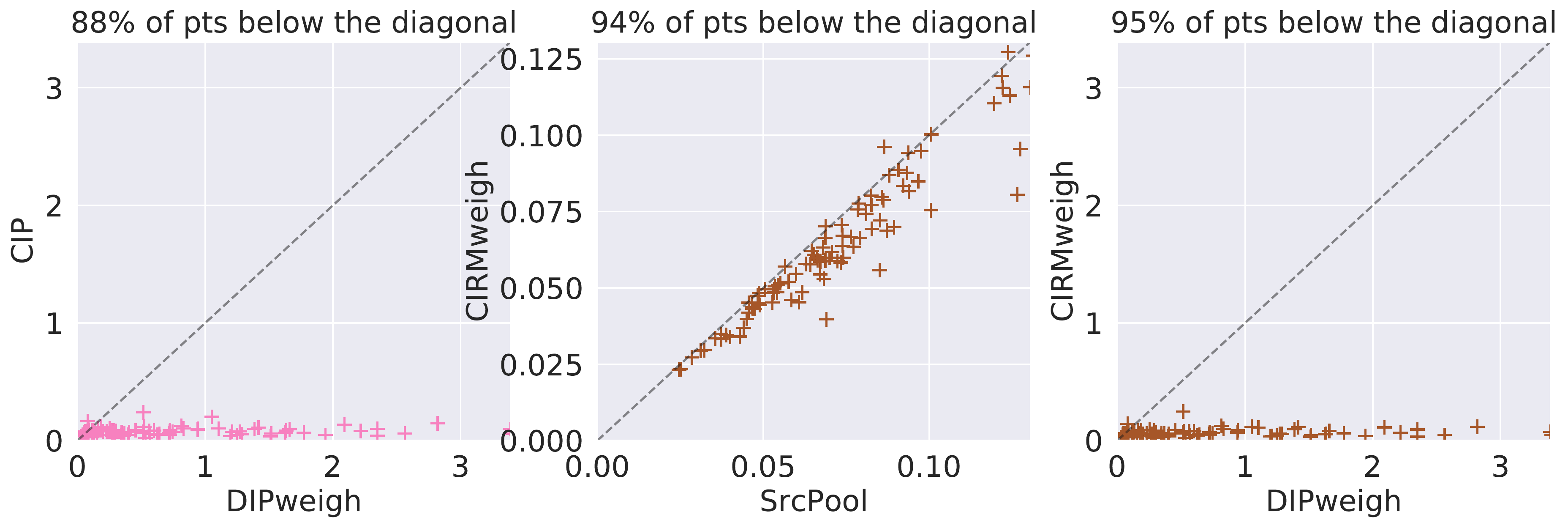}
  \end{minipage}\hfill
  \caption{Target risk comparison in simulation (iii) multiple source anticausal DA with Y intervention and with CICs (the lower the better). CIRMweigh has smaller target risk than SrcPool. In the presence of Y intervention, the target risk of DIPweigh can be much larger than that of SrcPool. The scatterplots are for 100 random coefficient $\semBX$ data generations. }
  \label{fig:sim_6_1_3}
\end{figure}

The left most plot in Figure~\ref{fig:sim_6_1_3} compares the boxplot of the target risks of OLSTar, SrcPool, DIPweigh, CIP and CIRMweigh with one time generation of $\semB$ and $10$ random generations of the interventions and the noises. CIRMweigh has lower target risk than SrcPool and DIPweigh. The right three plots in Figure~\ref{fig:sim_6_1_3} show the scatterplots comparing the pair-wise target risks in 100 random generations of $\semB$ and the interventions for the following pairs: CIP vs DIPweigh, CIRMweigh vis SrcPool. CIRMweigh vs DIPweigh. The number of points below the diagonal are reported in the titles.

Both CIP and CIRM have lower target risk than DIPweigh. DIPweigh loses target risk guarantees because of the intervention on $Y$. CIRMweigh also outperforms SrcPool. Note that due to finite sample errors, there are rare cases (about 5\%) where CIRM does not outperform SrcPool or DIPweigh.

\begin{figure}[ht]
  \begin{minipage}{0.75\textwidth}
    \centering
    \includegraphics[width=0.99\textwidth]{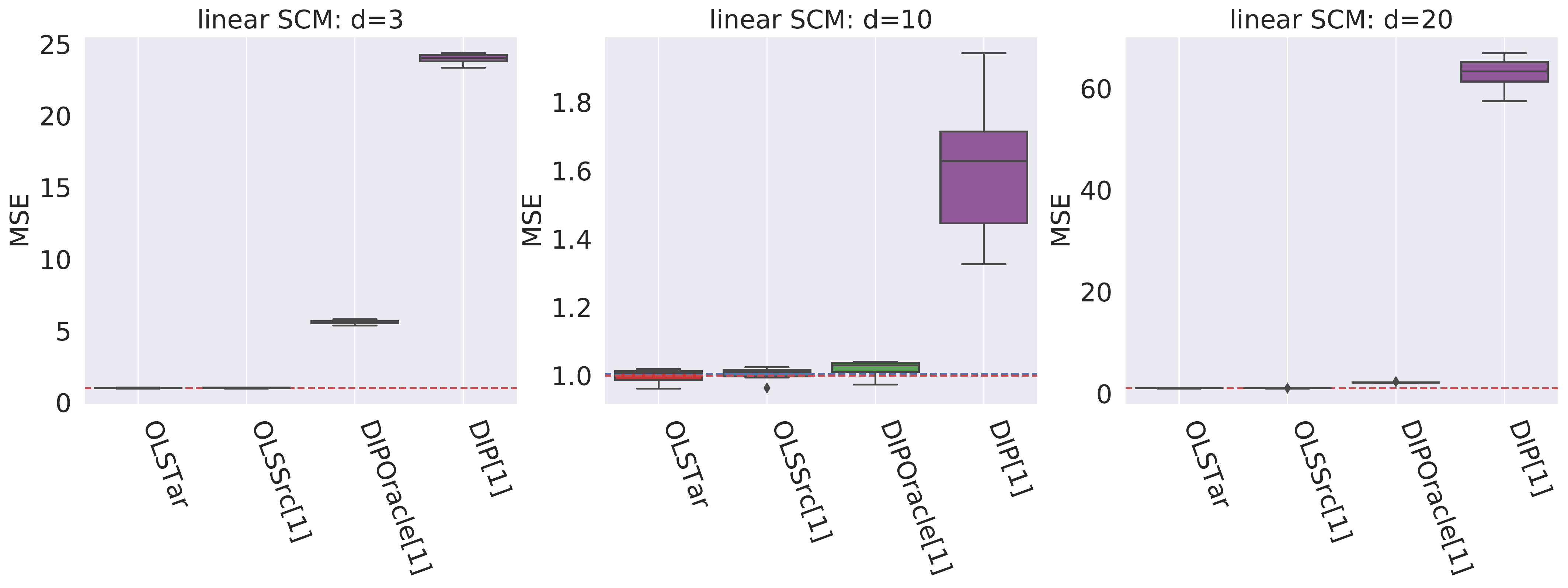}
  \end{minipage}\hfill
  \begin{minipage}{0.25\textwidth}
    \centering
    \includegraphics[width=0.99\textwidth]{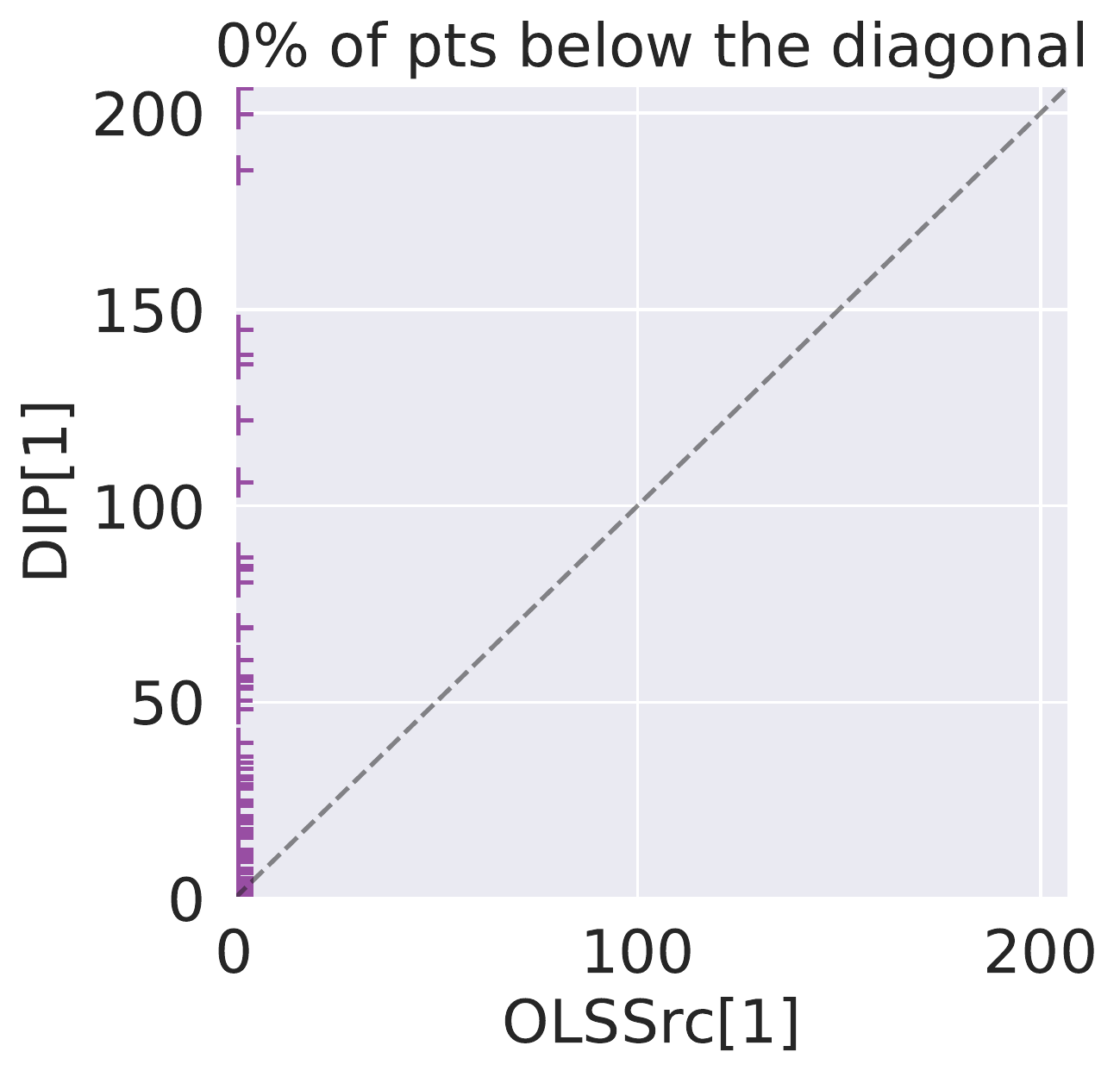}
  \end{minipage}\hfill
  \caption{Target risk comparison in simulation (iv) single source causal domain adaptation without Y intervention (the lower the better). DIP$^\tagk{1}$ always has larger target risk than OLSSrc$^\tagk{1}$ in the causal domain adaptation problem over 100 random coefficient B data generations ($\dims=10$). }
  \label{fig:sim_6_1_4}
\end{figure}

\paragraph{(iv) Single source {\color{red}causal} DA without Y intervention: } We consider three datasets of covariate dimension $\dims=3, 10, 20$ similar to simulation (i) except that the prediction direction is changed from anticausal to causal: $\sembh \neq 0$ and $\sembv = 0$. Figure~\ref{fig:sim_6_1_4} shows that DIP$^\tagk{1}$ can have larger target risk than OLSSrc$^\tagk{1}$. This simulation result makes it clear that DIP is not useful for causal DA in general. Example 3 in Section~\ref{ssub:example_3_anti_causal_prediction_when_y_is_intervened_on} is not just a pathological failure example of DIP.

\begin{figure}[ht]
  \centering
  \begin{minipage}{0.49\textwidth}
    \centering
    \includegraphics[width=0.99\textwidth]{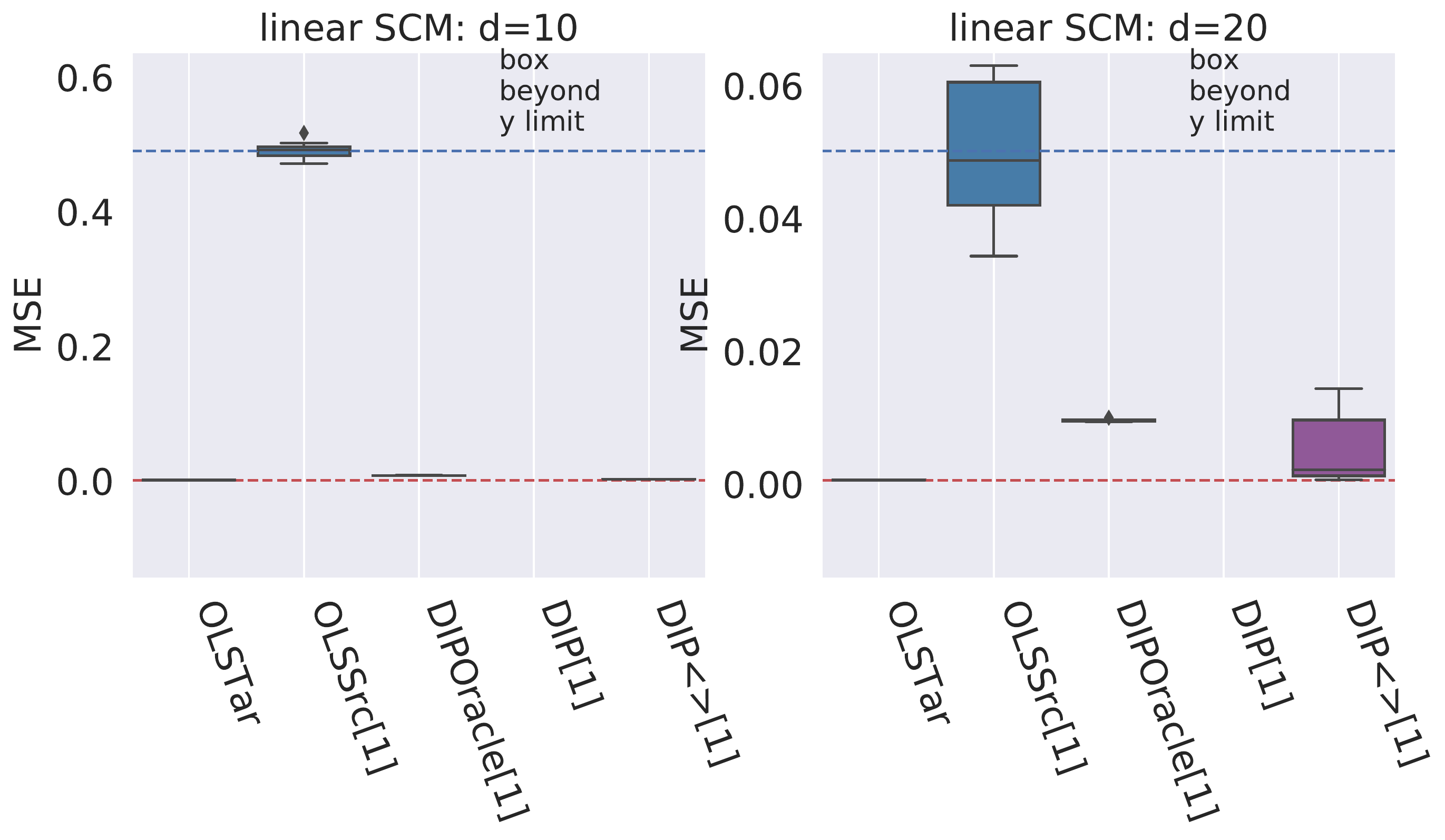}
  \end{minipage}
  \begin{minipage}{0.24\textwidth}
    \centering
    \includegraphics[width=0.99\textwidth]{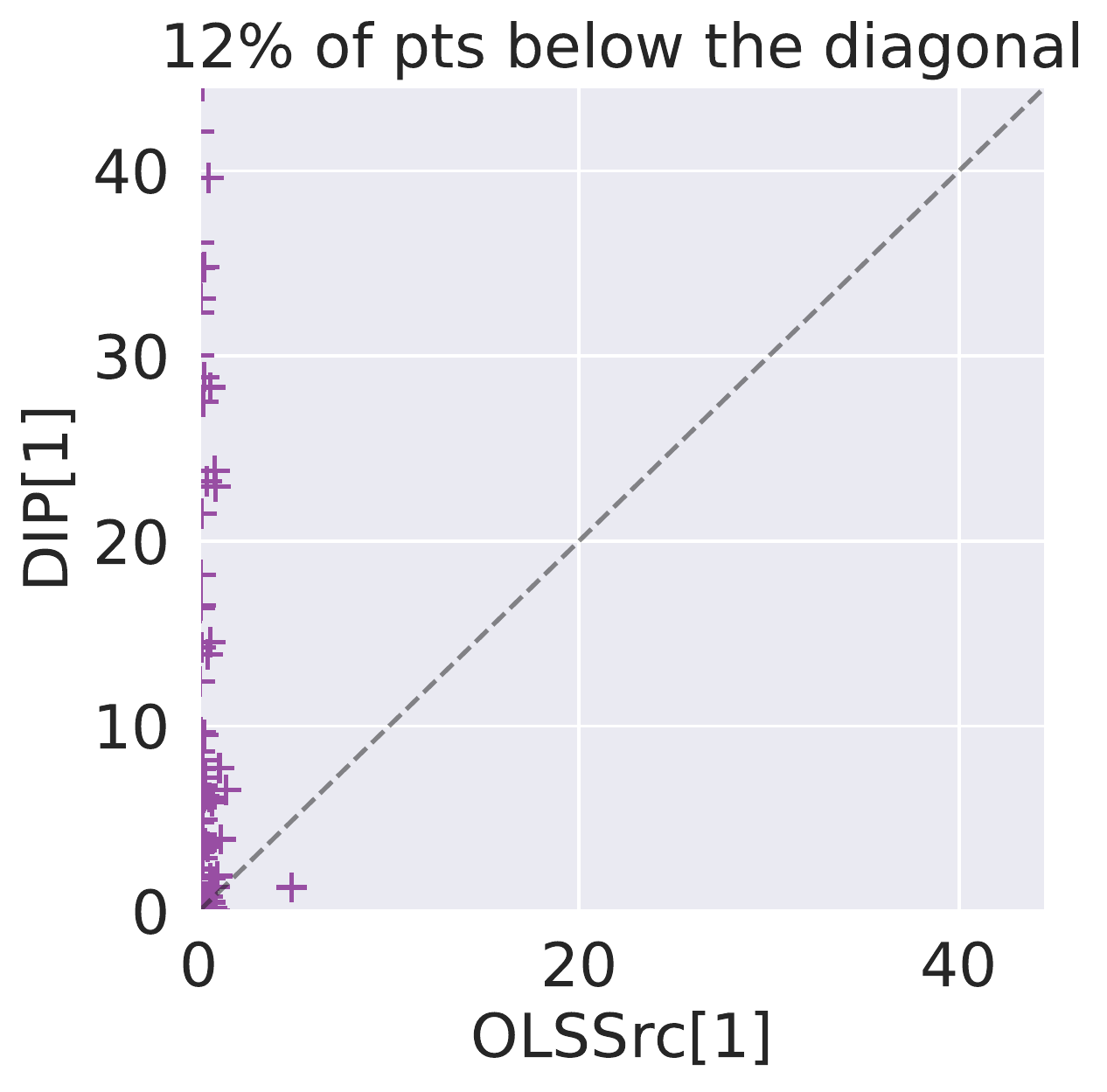}
  \end{minipage}
  \begin{minipage}{0.24\textwidth}
    \centering
    \includegraphics[width=0.99\textwidth]{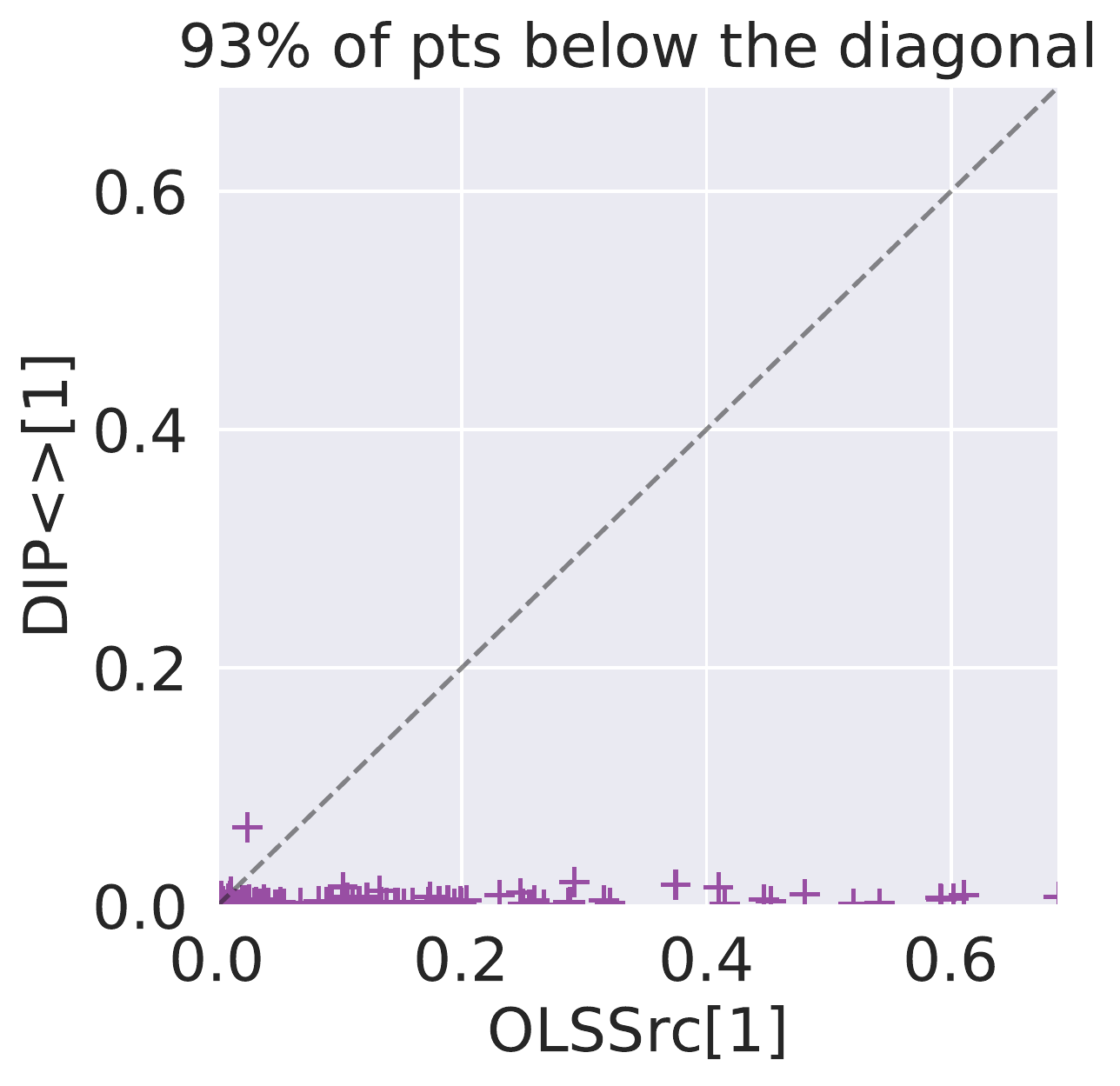}
  \end{minipage}\hfill
  \caption{Target risk comparison in simulation (v) single source mixed DA without Y intervention (the lower the better). DIP$^\tagk{1}$ has larger target risk than OLSSrc$^\tagk{1}$ in the mixed causal anticausal DA setting. DIP$\diamondsuit^\tagk{1}$ with ground-truth causal variables outperforms OLSSrc$^\tagk{1}$. The two scatterplots are over 100 random coefficient $\semBX$ data generations $(\dims=10)$.}
  \label{fig:sim_6_1_5}
\end{figure}

\paragraph{(v) Single source {\color{red} mixed} DA without Y intervention: } We consider two datasets of covariate dimension $\dims=10, 20$ similar to simulation (i) except that the prediction direction is changed from anticausal to mixed-causal-anticausal: The odd coordinates of $\sembv$ are set to zero and the even coordinates are nonzero. For $\sembh$, it is the contrary: the odd coordinates of $\sembh$ are nonzero and the even ones are zero. The matrix $\semB$ is generated according to Assumption~\ref{ass:assumption_multiple_source_mixed} with one block zero. The left two plots in Figure~\ref{fig:sim_6_1_5} show boxplots of target risks of OLSTar, OLSSrc$^\tagk{1}$, DIPOracle$^\tagk{1}$, DIP$^\tagk{1}$ and DIP$\diamondsuit^\tagk{1}$ for random generation of the matrix $\semB$. DIP$\diamondsuit^\tagk{1}$ uses the ground-truth knowledge of the nonzero coordinates of $\sembv$. In mixed causal-anticausal DA, DIP$^\tagk{1}$ has larger risk than OLSSrc$^\tagk{1}$. With the ground-truth knowledge of the causal variables, DIP$\diamondsuit^\tagk{1}$ still have lower risk than OLSSrc$^\tagk{1}$. The right two plots in Figure~\ref{fig:sim_6_1_5} confirms the result via scatterplot comparisons of OLSSrc$^\tagk{1}$, DIPOracle$^\tagk{1}$, DIP$^\tagk{1}$ and DIP$\diamondsuit^\tagk{1}$ after 100 runs.

\begin{figure}[ht]
  \includegraphics[width=0.99\textwidth]{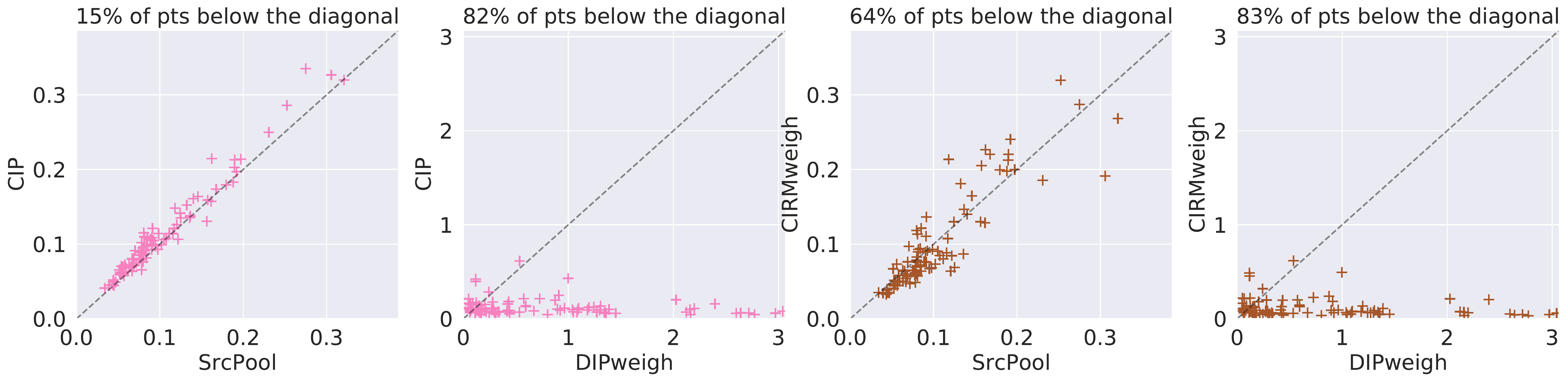}
  \caption{Target risk comparison in simulation (vi) multiple source anticausal DA with Y intervention and without CICs (the lower the better). Without CICs, CIRMweigh no longer outperforms SrcPool. }
  \label{fig:sim_6_1_6}
\end{figure}

\paragraph{(vi) Multiple source anticausal DA with Y intervention and {\color{red} without CICs}: } We consider a simulation ($\dims=20$, $\envs=14$) similar to simulation (iii) except that there are no conditionally invariant components (CICs). Figure~\ref{fig:sim_6_1_6} shows that without CICs, CIRMweigh no longer outperforms SrcPool. Only in 64 of 100 simulations CIRMweigh has lower target risk than SrcPool. Interestingly, CIRMweigh still has a large chance of having smaller target risk than DIPweigh. This may be due to the fact that even though there are no pure conditional invariant components, CIP may still be able to pick a combination of covariates that is relatively less sensitive to $X$ interventions.

\begin{figure}[ht]
  \includegraphics[width=0.99\textwidth]{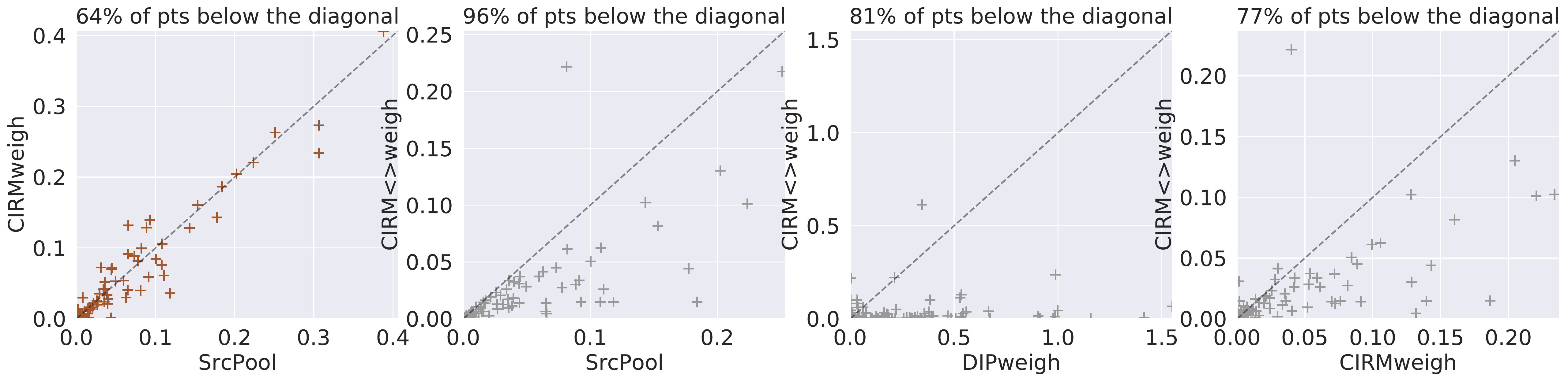}
  \caption{Target risk comparison in simulation (vii) multiple source mixed DA with Y intervention and with CICs (the lower the better). The scatterplots are over 100 random coefficient $\semBX$ data generations. CIRMweigh has lower target risk than SrcPool only in 64 of 100 simulations. However, knowing the indexes of the true causal covariates, the oracle estimator CIRM$\diamondsuit$weigh has lower target risk than SrcPool in 96 of 100 simulations. CIRM$\diamondsuit$weigh also outperforms DIPweigh and CIRMweigh. The scatterplots are over 100 random coefficient $\semBX$ data generations. }
  \label{fig:sim_6_1_7}
\end{figure}

\paragraph{(vii) Multiple source {\color{red} mixed} DA with Y intervention and with CICs: } We consider a simulation ($\dims=20$, $\envs=14$) similar to simulation (iii) except that the prediction direction is changed from anticausal to mixed causal anticausal. The odd coordinates of $\sembv$ is set to zero and the even ones are nonzero. The even coordinates of $\sembh$ is set to zero and the odd ones are nonzero. The first five even coordinates are set to be anticausal CICs. Additionally, we tuned down the variance of $\noisev{m}_Y$ to be 0.01 to make the causal part important for prediction. Figure~\ref{fig:sim_6_1_7} shows that in mixed causal and anticausal DA, CIRMweigh has lower target risk than SrcPool only in 64 of 100 simulations. However, with the true causal covariates, the oracle estimator CIRM$\diamondsuit$weigh has lower target risk than SrcPool in 96 of 100 simulations.

\subsubsection{Linear SCM with variance shift noise interventions} % (fold)
\label{ssub:linear_scm_with_other_interventions}
We consider two simulations with data generated via linear SCMs with variance shift noise interventions. Since DIP-std+ and DIP-MMD are involved, we have to adapt the regularization parameter choice strategy described at the beginning of Section~\ref{sec:numerical_experiments}. We apply all DIP variants with the  regularization parameter $\lamMatch$ ranging in the set $\braces{10^{-5}, 10^{-4}, \cdots, 10^4, 10^5}$. We choose the largest $\lamMatch$ such that the source risk is smaller than $\min(2r, r+0.01)$ as the final parameter. Here $r$ is the source risk of OLSSrc. Based on the average source risk, $\lamCIP$ is still chosen to be $1.0$.

Since DIP-std+, DIP-MMD, CIRMweigh-std+ and CIRMweigh-MMD do not have closed form solutions, we use Pytorch's gradient descent to optimize these methods. Specifically, Adam's optimizer is used with step-size $10^{-4}$ and iteration number 2000 epochs to ensure convergence.

\begin{figure}[ht]
  \begin{minipage}{0.25\textwidth}
    \centering
    \includegraphics[width=0.99\textwidth]{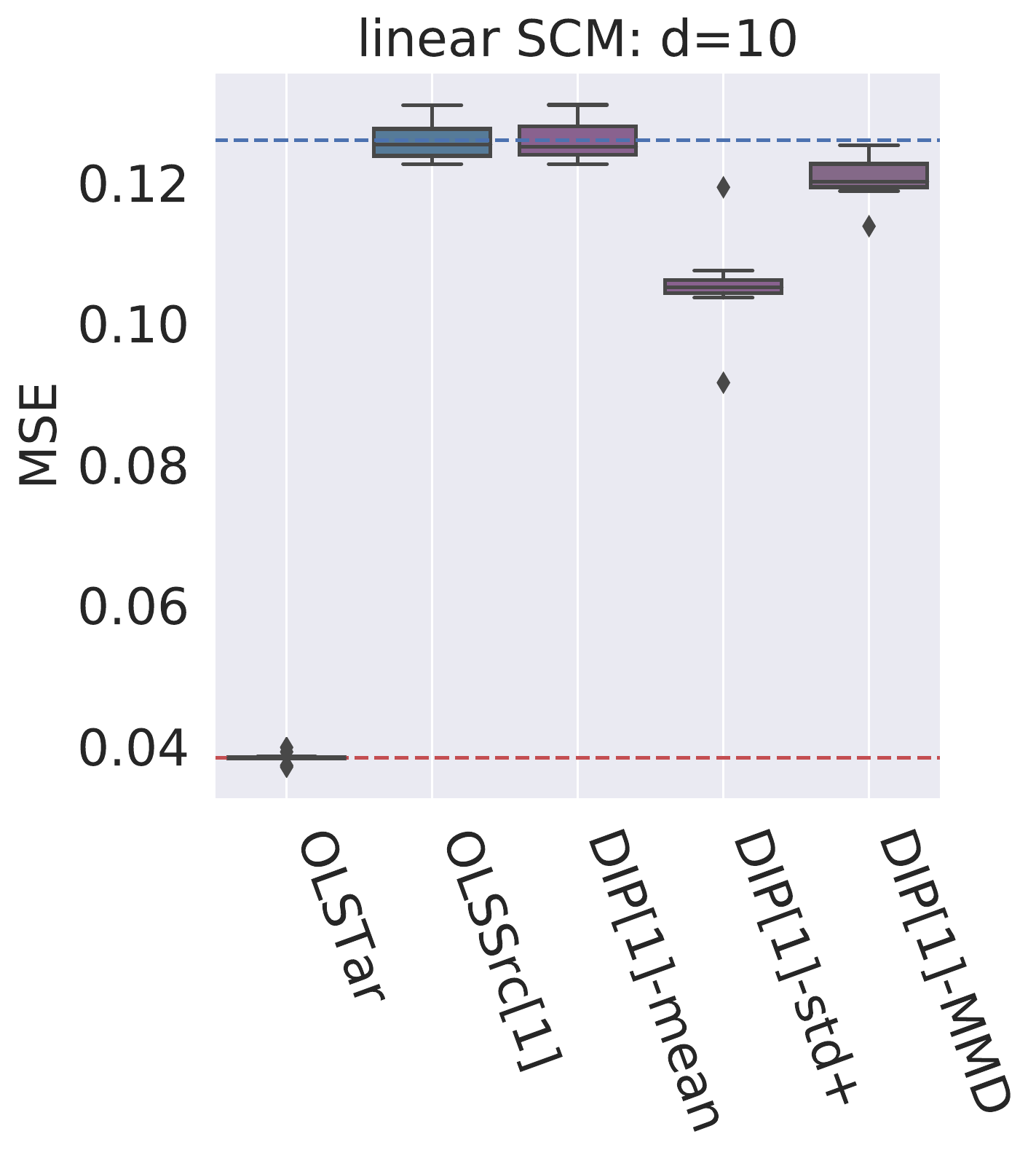}
  \end{minipage}\hfill
  \begin{minipage}{0.25\textwidth}
    \centering
    \includegraphics[width=0.99\textwidth]{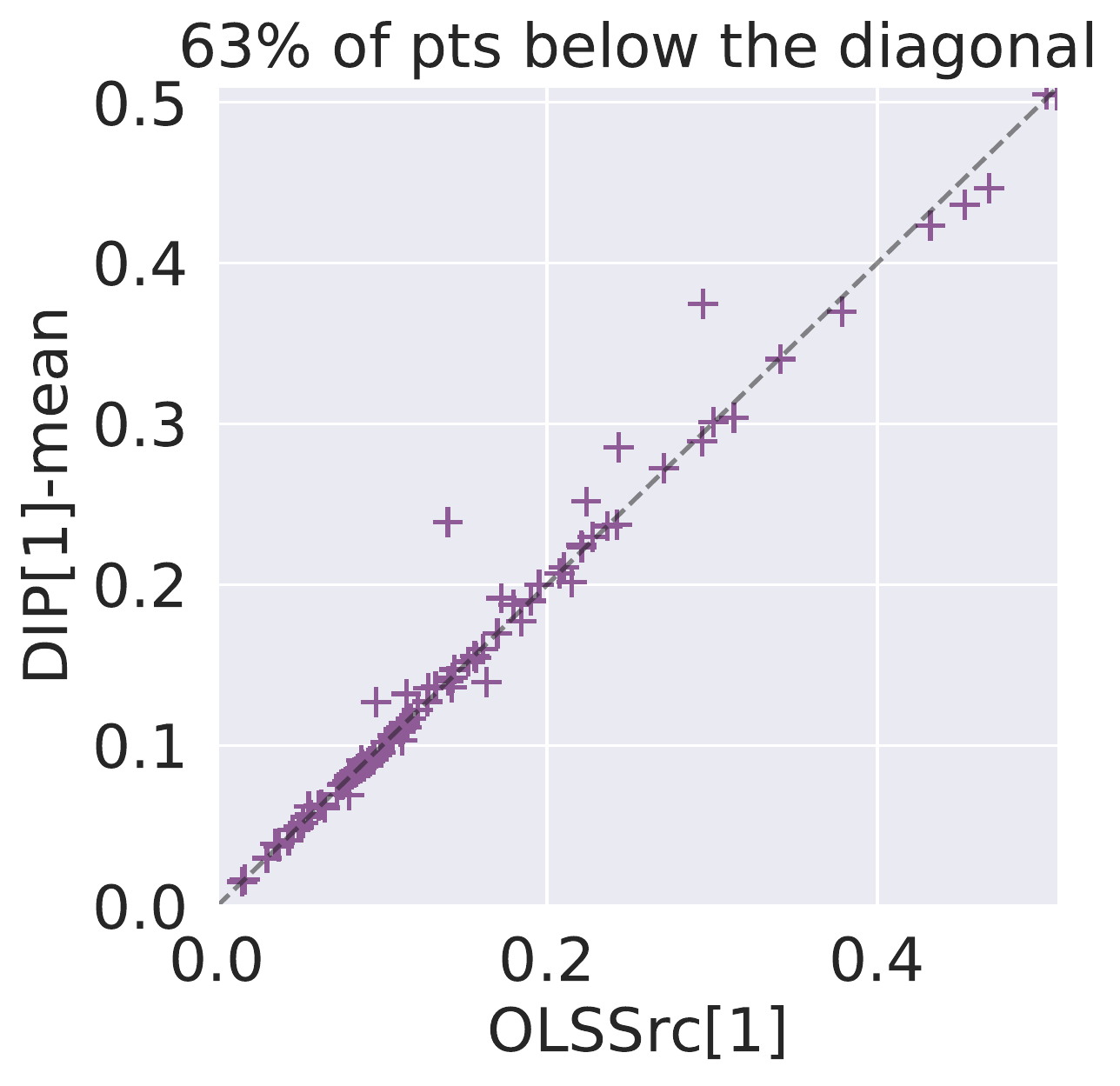}
  \end{minipage}\hfill
  \begin{minipage}{0.25\textwidth}
    \centering
    \includegraphics[width=0.99\textwidth]{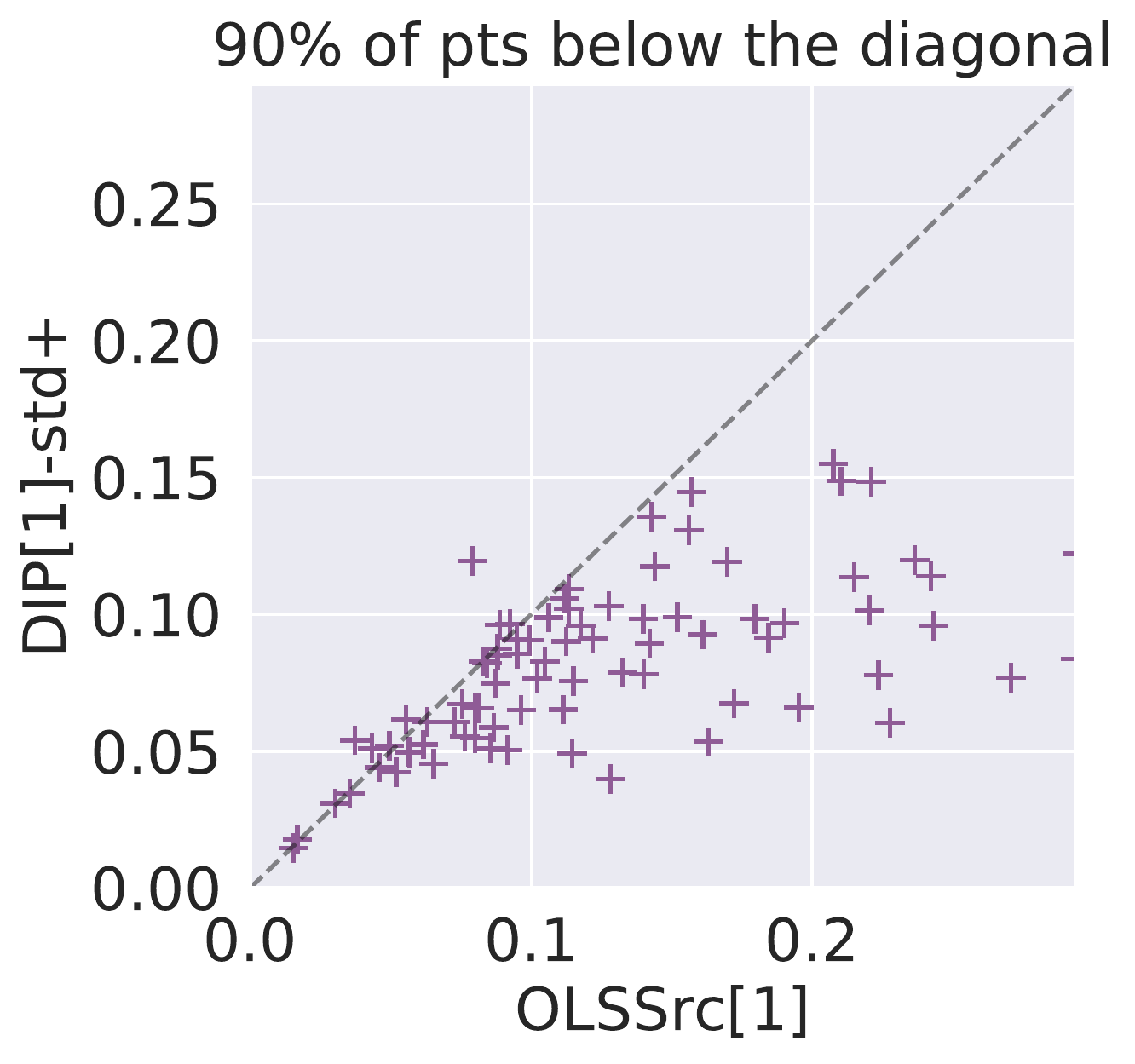}
  \end{minipage}\hfill
  \begin{minipage}{0.25\textwidth}
    \centering
    \includegraphics[width=0.99\textwidth]{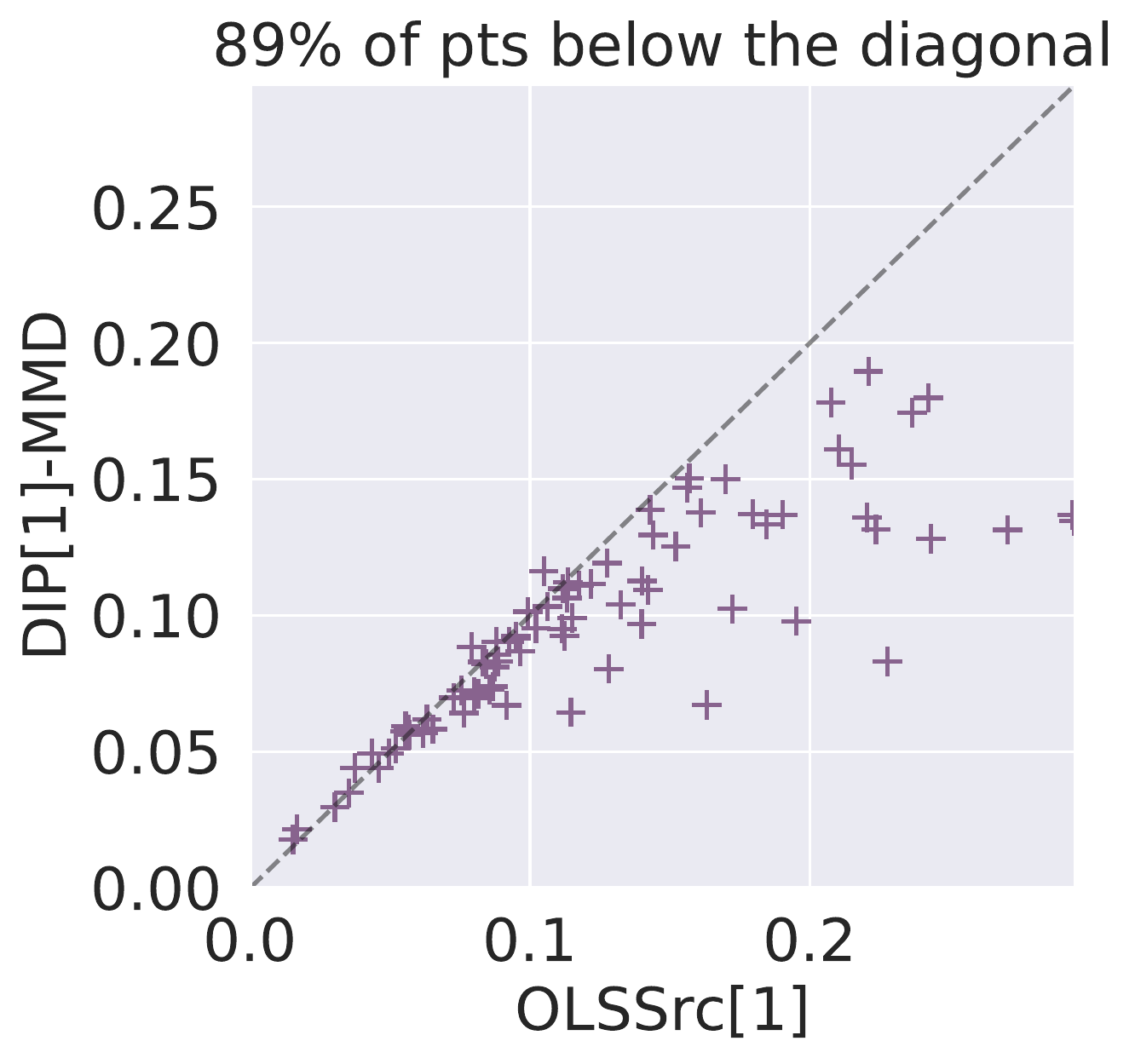}
  \end{minipage}\hfill
  \caption{Target risk comparison in simulation (viii) single source causal domain adaptation with variance intervention without Y intervention (the lower the better). Left: boxplots of the target risks. DIP-mean has the same target risk than Src[1]. DIP-std+ and DIP-MMD outperform Src[1]. Right: three scatterplots of 100 runs comparing DIP-mean, DIP-std+ and DIP-MMD with Src[1]. }
  \label{fig:sim_6_2_sv1}
\end{figure}

\paragraph{(viii) Single source anticausal DA without Y intervention + variance shift noise intervention:} We consider a simulation ($\dims=10$) similar to simulation (i) except that the type of intervention is changed from mean shift noise intervention to variance shift noise intervention. The intervention affects the variance of the noise $g(\interAv{1}_x, \noisev{1}_X) = \interAv{1}_X \odot \noisev{1}_X$.
The interventions $\interAv{1}_X$ and $\interAtar_X$ are still generated i.i.d. from $\Normal(0, \Ind_\dims)$ and stay the same for all data points in one environment. The left-most plot in Figure~\ref{fig:sim_6_2_sv1} shows the boxplot of the target risks of OLSTar, OLSSrc$^\tagk{1}$, DIP$^\tagk{1}$-mean, DIP$^\tagk{1}$-std+, DIP$^\tagk{1}$-MMD. DIP$^\tagk{1}$-mean has the same target risk and OLSSrc$^\tagk{1}$, because the DIP matching penalty on mean has no effect when the intervention is variance shift noise intervention. DIP$^\tagk{1}$-std+ and DIP$^\tagk{1}$-MMD improves upon DIP$^\tagk{1}$-mean. The right three plots in Figure~\ref{fig:sim_6_2_sv1} show scatterplots of 100 random coefficient $\semBX$ data generations to confirm this observation.
\paragraph{(ix) Multiple source anticausal DA with Y intervention + variance shift noise intervention: }
\begin{figure}[ht]
  \begin{minipage}{0.25\textwidth}
    \centering
    \includegraphics[width=0.99\textwidth]{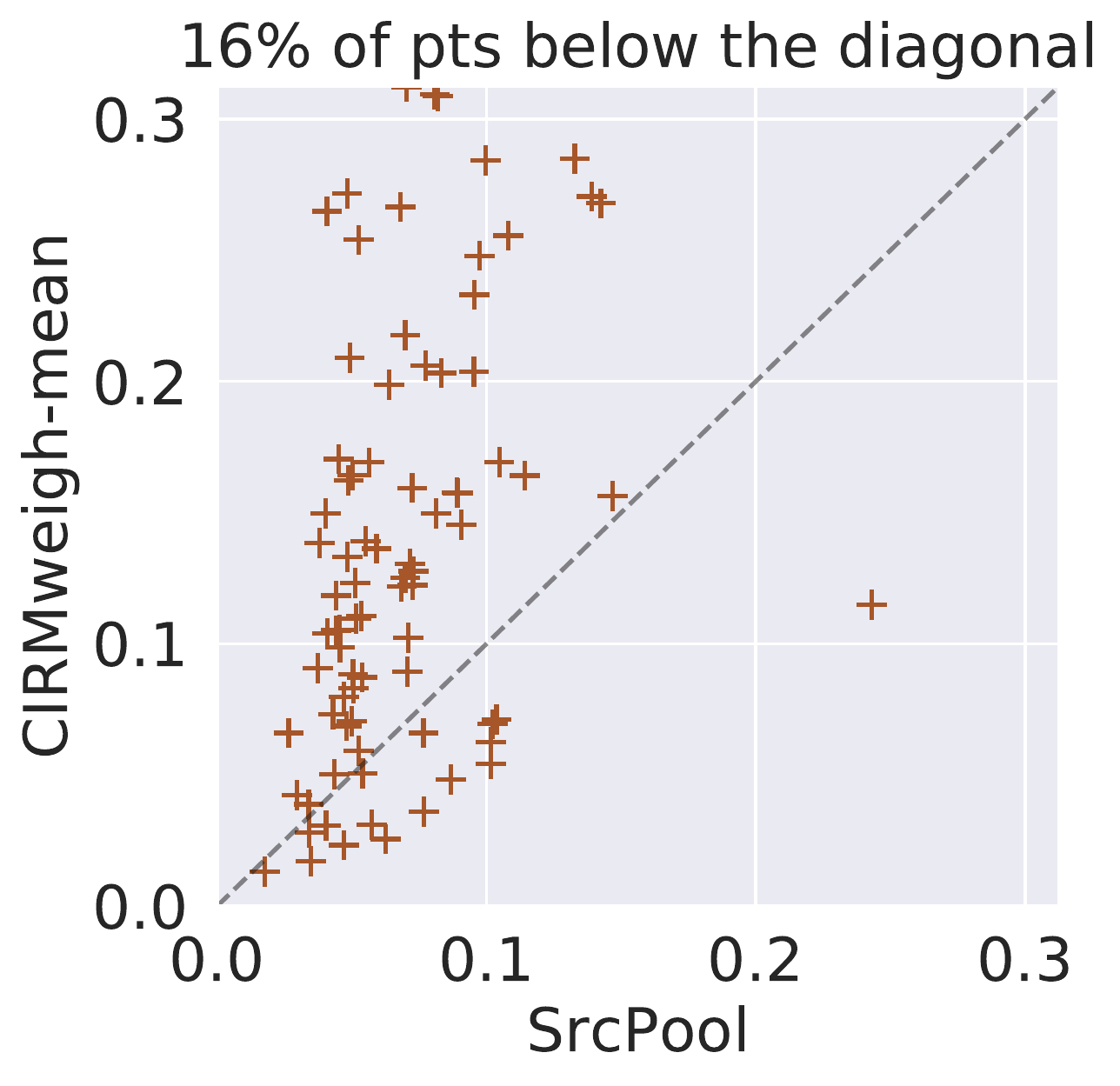}
  \end{minipage}\hfill
  \begin{minipage}{0.25\textwidth}
    \centering
    \includegraphics[width=0.99\textwidth]{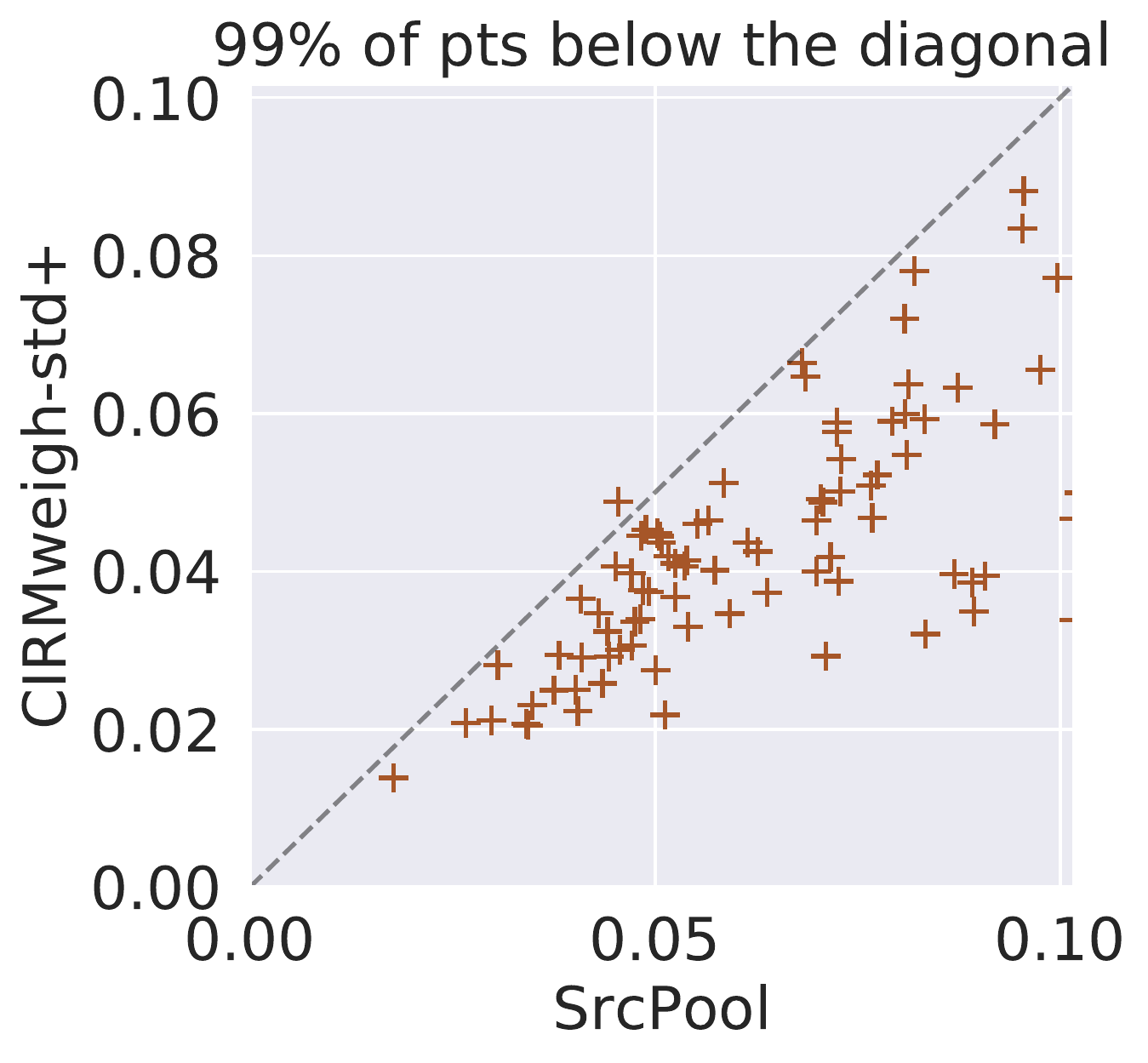}
  \end{minipage}\hfill
  \begin{minipage}{0.25\textwidth}
    \centering
    \includegraphics[width=0.99\textwidth]{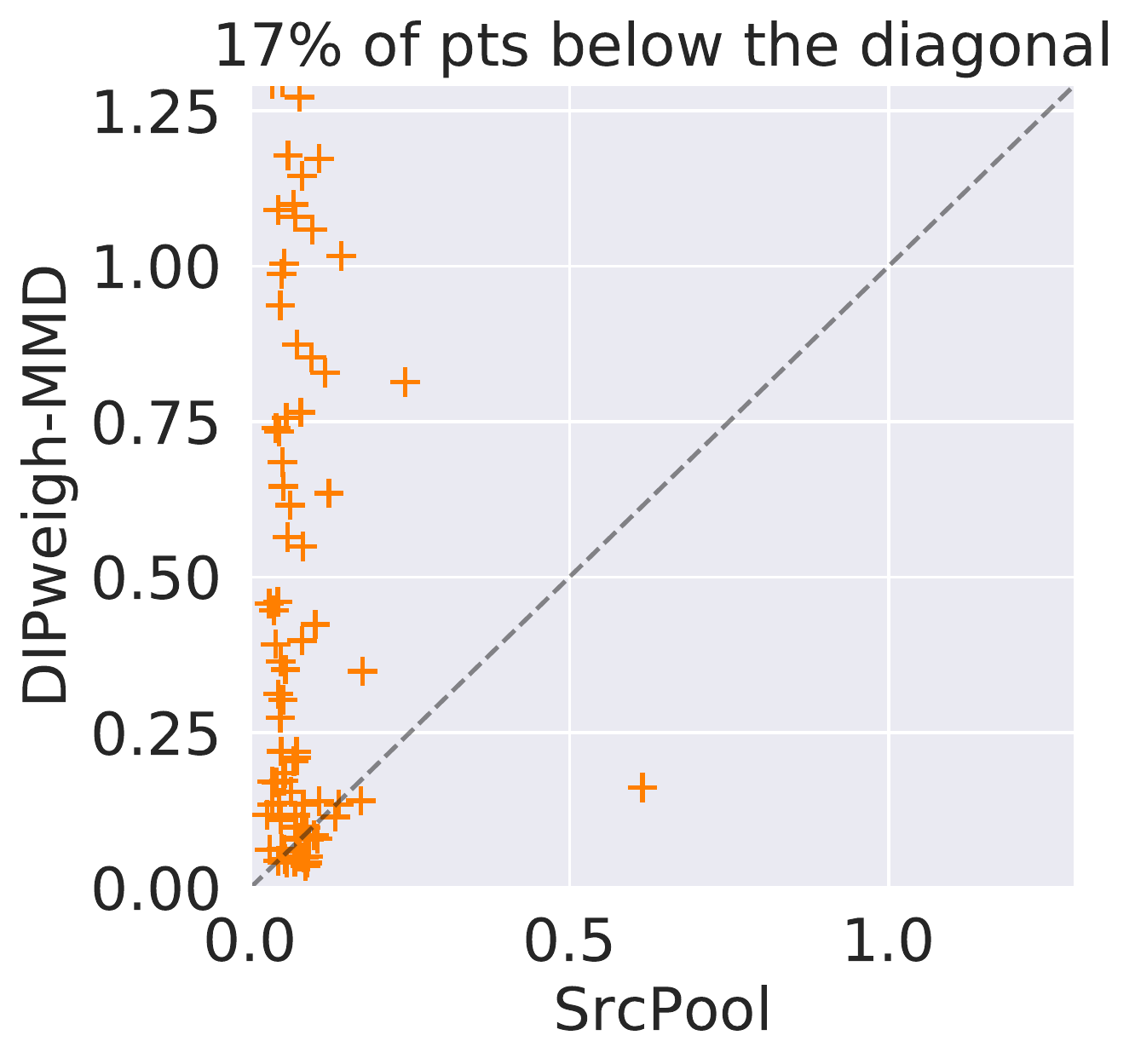}
  \end{minipage}\hfill
  \begin{minipage}{0.25\textwidth}
    \centering
    \includegraphics[width=0.99\textwidth]{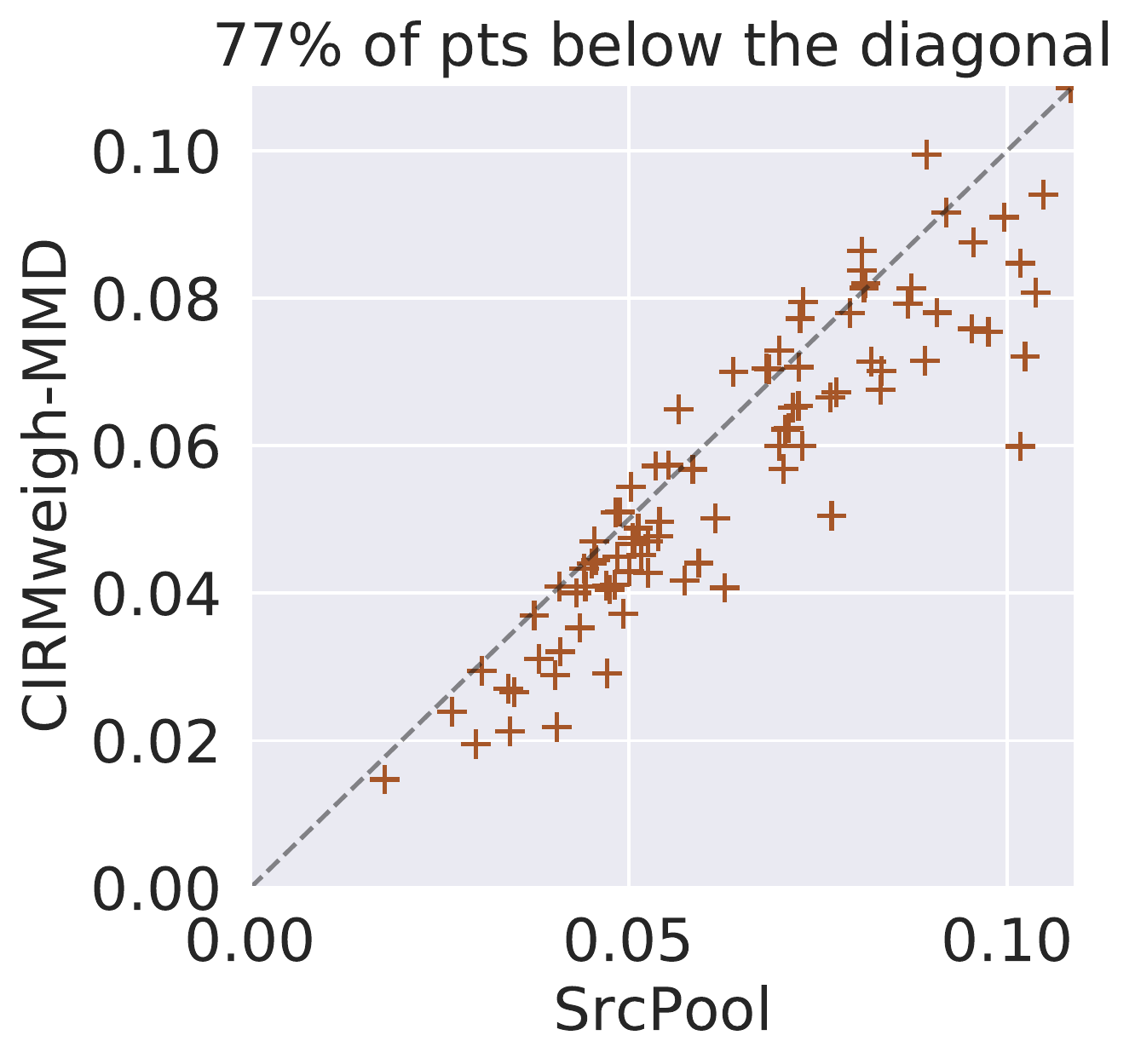}
  \end{minipage}\hfill
  \caption{Target risk comparison in simulation (ix) multiples source causal domain adaptation with variance intervention with Y intervention and with conditionally invariant components (CICs) (the lower the better). CIRMweigh-std+ and CIRMweigh-MMD outperform SrcPool. CIRMweigh-mean does not outperform SrcPool, because it can not handle the variance shift noise intervention. DIPweigh-MMD performs much worse than SrcPool because of the intervention on $Y$.}
  \label{fig:sim_6_2_smv1_y_shift}
\end{figure}
We consider a simulation (d = 20, M = 14) similar to simulation (iii) except that the type of intervention is changed from mean shift noise intervention to variance shift noise intervention. The variance shift noise intervention on $X$ is as specified simulation (viii). The intervention on $Y$ is still mean shift noise intervention as in simulation (iii). The scatterplots in Figure~\ref{fig:sim_6_2_smv1_y_shift} compares CIRMweigh-mean, CIRMweigh-std+, DIPweigh-MMD, CIRMweigh-MMD with SrcPool in 100 runs. CIRMweigh-std+ and CIRMweigh-MMD outperform SrcPool. CIRMweigh-mean does not outperform SrcPool, because it can not handle the variance shift noise intervention. DIPweigh-MMD performs much worse than SrcPool because of the intervention on $Y$.

% subsection linear_sem_simulations (end)
\subsection{MNIST experiments with synthetic image interventions} % (fold)
\label{sub:mnist_experiments_with_synthetic_interventions}

In this section, we generate synthetic image classification datasets by modifying the images from the MNIST dataset~\citep{lecun1998mnist}. Then we compare the performance of various DA methods on this digit classification task. The interventions are added directly on the image pixels. The DA methods are applied to the last-layer features of a pre-trained CNN on the original MNIST dataset. Since our DA methods are only defined for regression problems in the previous sections, for this classification problem, we adapt the loss function to softmax loss and we use one-hot encoding of the label $Y$ in the DIP, CIP and CIRM formulations.

For the MNIST experiments with synthetic image interventions, a priori the linear SCM assumption is no longer valid and the interventions are not necessarily noise interventions. Here we show that the DA methods such as DIP, CIRM still have target performance as predicted by our theorems despite the likely violation of several assumptions.

All the methods in this subsection are implemented via Pytorch~\citep{paszke2019pytorch} and are optimized with Pytorch's stochastic gradient descent. Specifically, stochastic gradient descent (SGD) optimizer is used with step-size (or learning rate) $10^{-4}$, batchsize $500$ and number of epochs $100$.

\subsubsection{MNIST with patch intervention}
\label{subs:MNIST_patch}

\begin{figure}[ht]
  \centering
  \begin{minipage}{0.20\textwidth}
      \centering
      \includegraphics[width=0.99\textwidth]{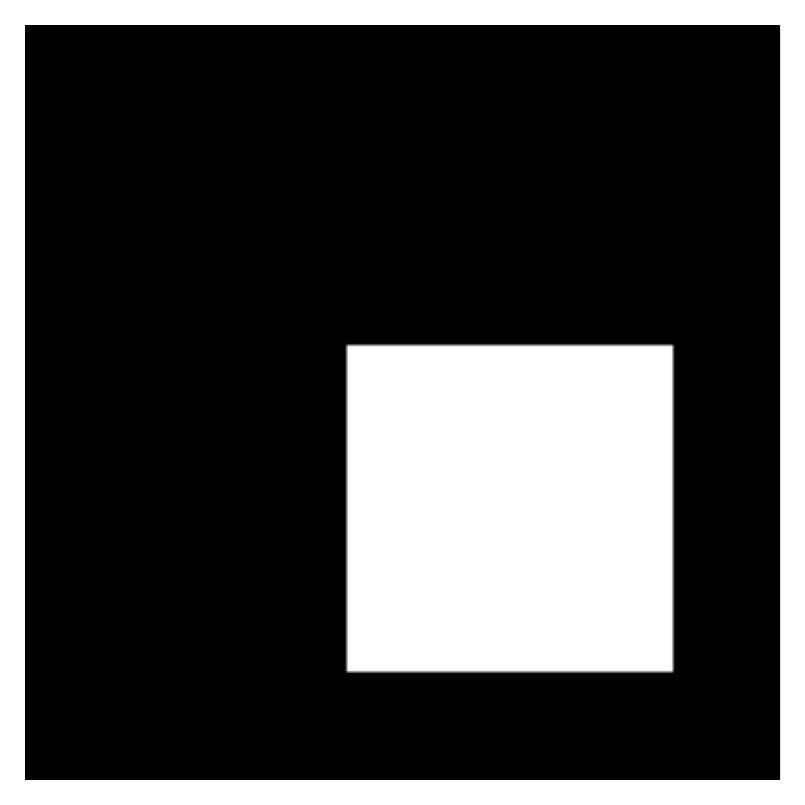}
      \subcaption{mask (a) \\ on $32 \times 32$ image}
  \end{minipage}
  \hfill
  \begin{minipage}{0.20\textwidth}
      \centering
      \includegraphics[width=0.99\textwidth]{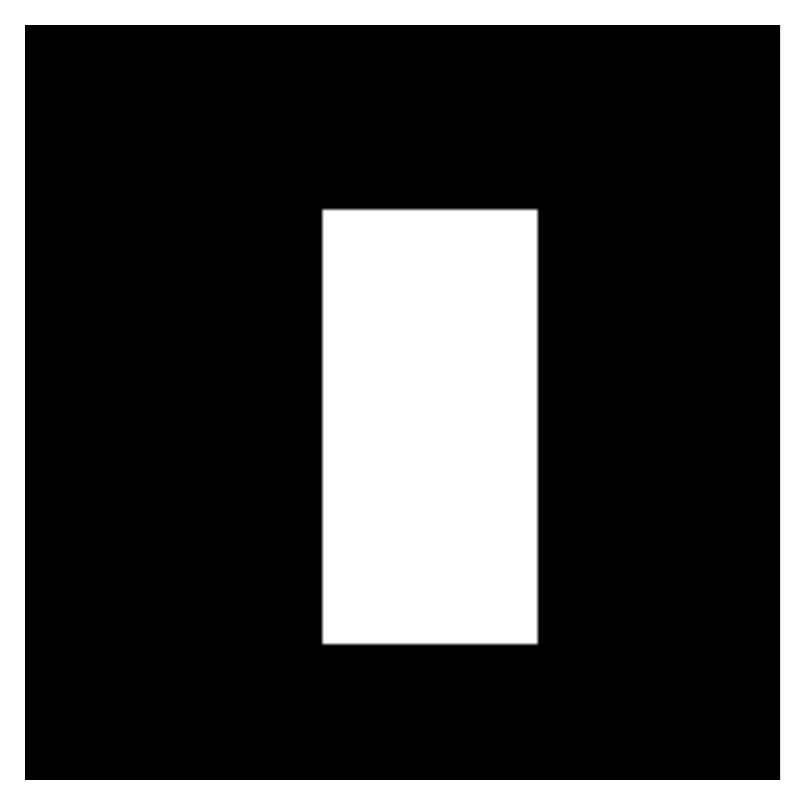}
      \subcaption{mask (b) \\ on $32 \times 32$ image}
  \end{minipage}
  \hfill
  \begin{minipage}{0.20\textwidth}
      \centering
      \includegraphics[width=0.99\textwidth]{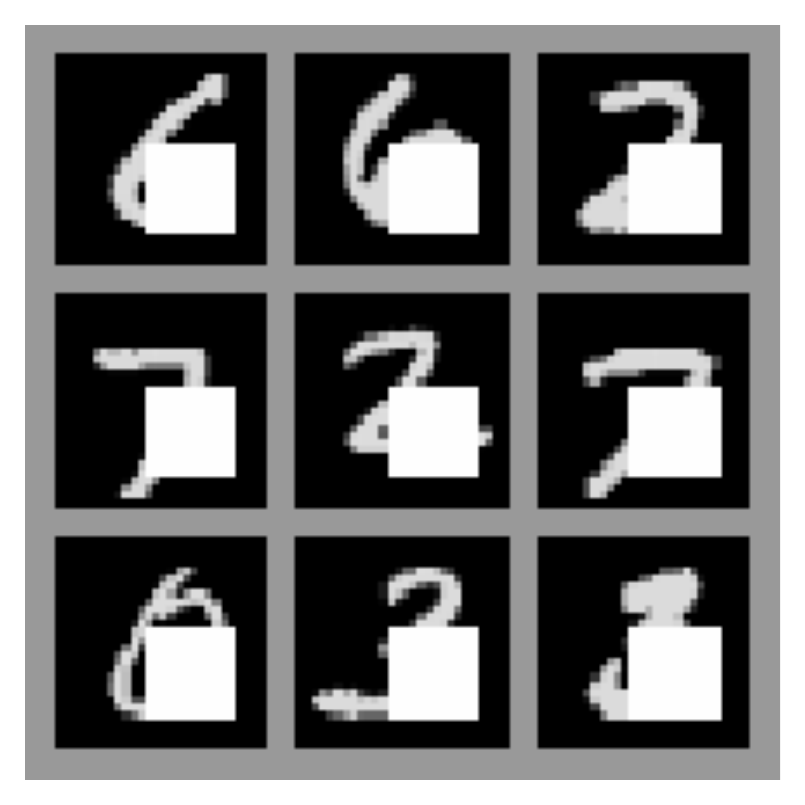}
      \subcaption{data samples \\ masked by mask (a)}
  \end{minipage}
  \hfill
  \begin{minipage}{0.20\textwidth}
      \centering
      \includegraphics[width=0.99\textwidth]{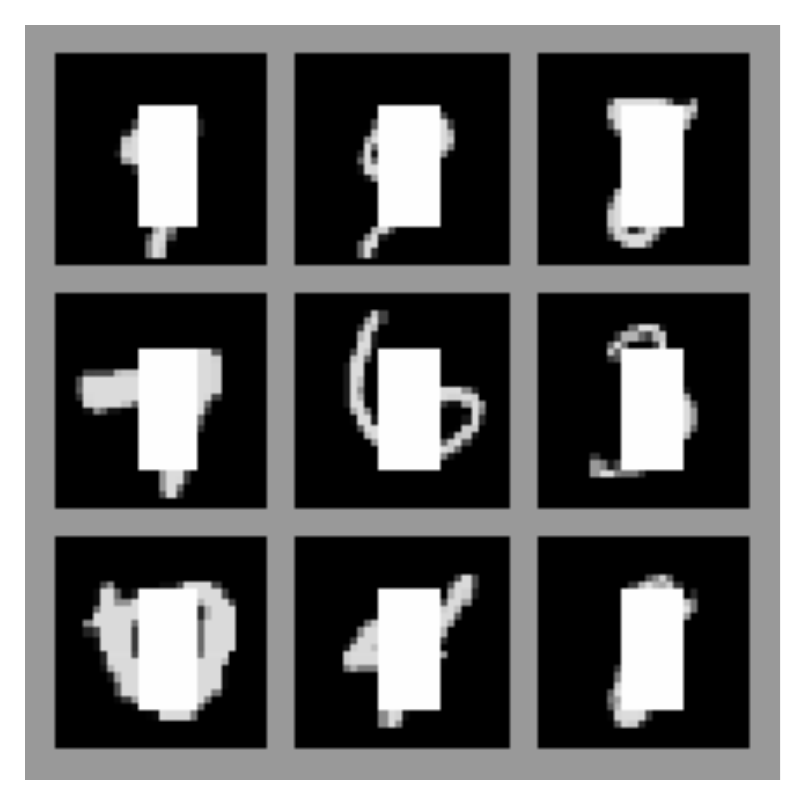}
      \subcaption{data samples \\ masked by mask (b)}
  \end{minipage} \hfill
\caption{The two interventions in the MNIST single source domain adaptation without Y intervention. From left to right, mask (a), mask (b) and the corresponding data samples.}
\label{fig:MNIST_patches_2M_interv}
\end{figure}

\paragraph{MNIST DA with patch intervention without Y intervention:}
We take the original MNIST dataset with 60000 training samples and create two synthetic datasets (source and target) as follows. For the source dataset, each training image is masked by the mask (a) in Figure~\ref{fig:MNIST_patches_2M_interv}. That is, for each training image, the pixels at the white region of the mask (a) are set to white (maximum pixel value). For the target dataset environment, each training image is masked by the mask (b) in Figure~\ref{fig:MNIST_patches_2M_interv}. For each experiment, we take random $20\%$ of samples from the source dataset as the source environment and random $20\%$ of samples from the target dataset as the target environment. The task is to predict the labels of the images in the target environment without observing any target labels. This experiment is repeated $10$ times and we report the boxplot of the $10$ target classification accuracies for each DA method.

We apply the DA methods on the last-layer features of a pre-trained convolutional neural network (CNN). The CNN has two convolutional layers and two fully connected layers similar to the LeNet-5 architecture. It is pre-trained on the original MNIST dataset with test accuracy on the separate 10000 test images from the MNIST dataset being 98.8\%. A priori, it is not clear what type of interventions on the last-layer features happen when we only know the intervention is on the pixels. To fit into our theoretical and conceptual framework, these induced interventions on the last-layer features would need to be approximated by shift noise interventions. The following DA methods are applied:
\begin{itemize}
  \item Original: CNN model pre-trained on the original MNIST dataset without any modification.
  \item Tar: oracle CNN model trained only on the target environment. It is similar to OLSTar because only the weights in the last layer are trained. We changed the name to Tar because the full model is a neural network.
  \item Src$^\tagk{1}$: CNN model trained only on the source environment.
  \item DIP$^\tagk{1}$: CNN model where DIP$^\tagk{1}$-mean-finite is applied using the source label and the last-layer features of source and target environments.
  \item DIP$^\tagk{1}$-MMD: oracle CNN model where where DIP$^\tagk{1}$-MMD-finite is applied using the target label and the last-layer features of source and target environments.
\end{itemize}
Based on the discussion on the regularization parameter choice at the beginning of Section~\ref{sec:numerical_experiments}, we vary the regularization parameter $\lamMatch$ from the set $\braces{10^k}_{k=-5, \cdots, 4}$, and we choose the largest $\lamMatch$ such that the source accuracy has not dropped too much (in this case no more than 1\% of the source accuracy of Src) as the final regularization parameter.

\begin{figure}[ht]
    \begin{minipage}{0.35\textwidth}
    \centering
    \includegraphics[width=0.99\textwidth]{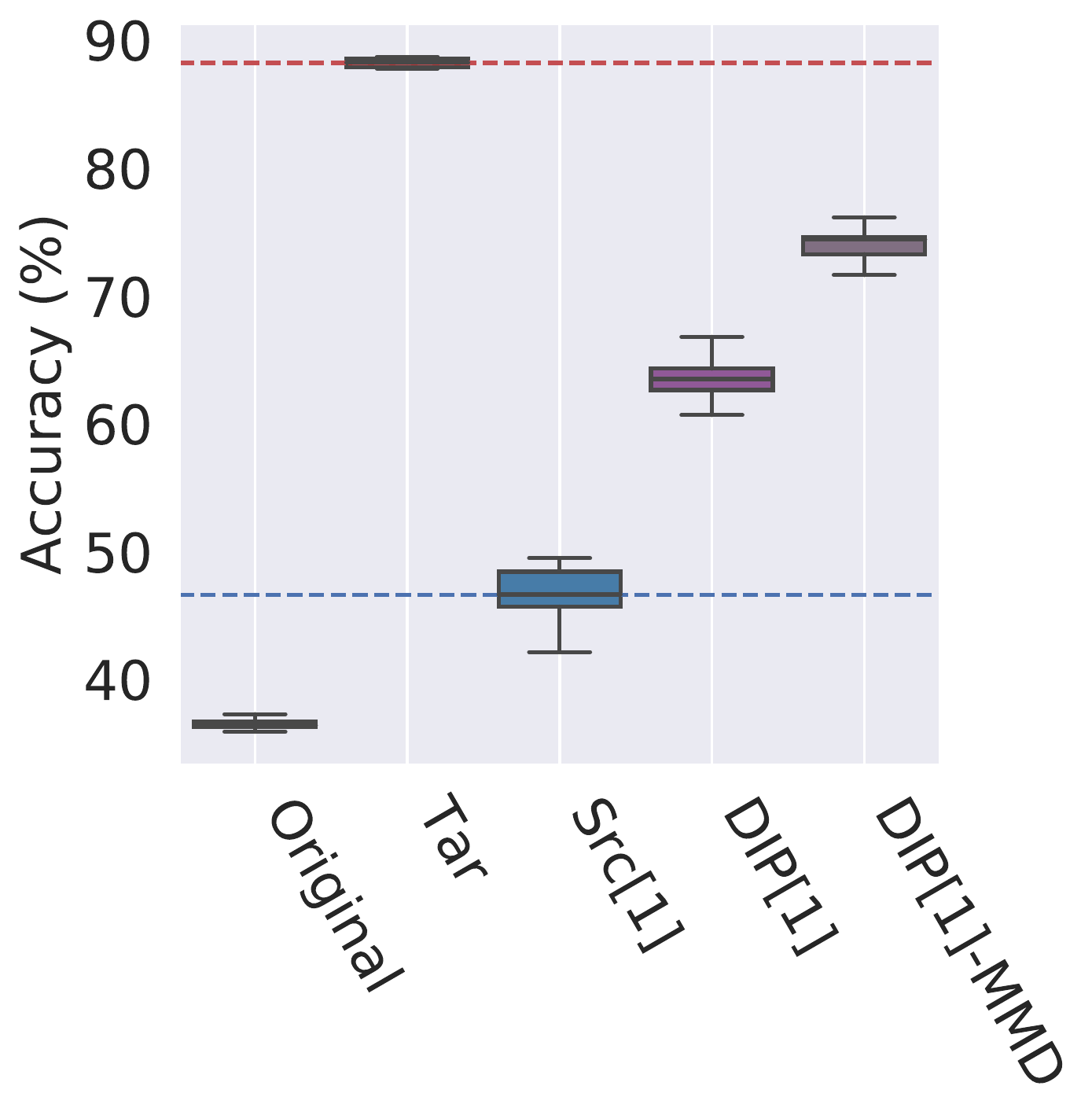}
    % \subcaption{}
    % \label{fig:MNIST_patch_intervention_single}
  \end{minipage}\hfill
  \begin{minipage}{0.65\textwidth}
    \centering
    \includegraphics[width=0.99\textwidth]{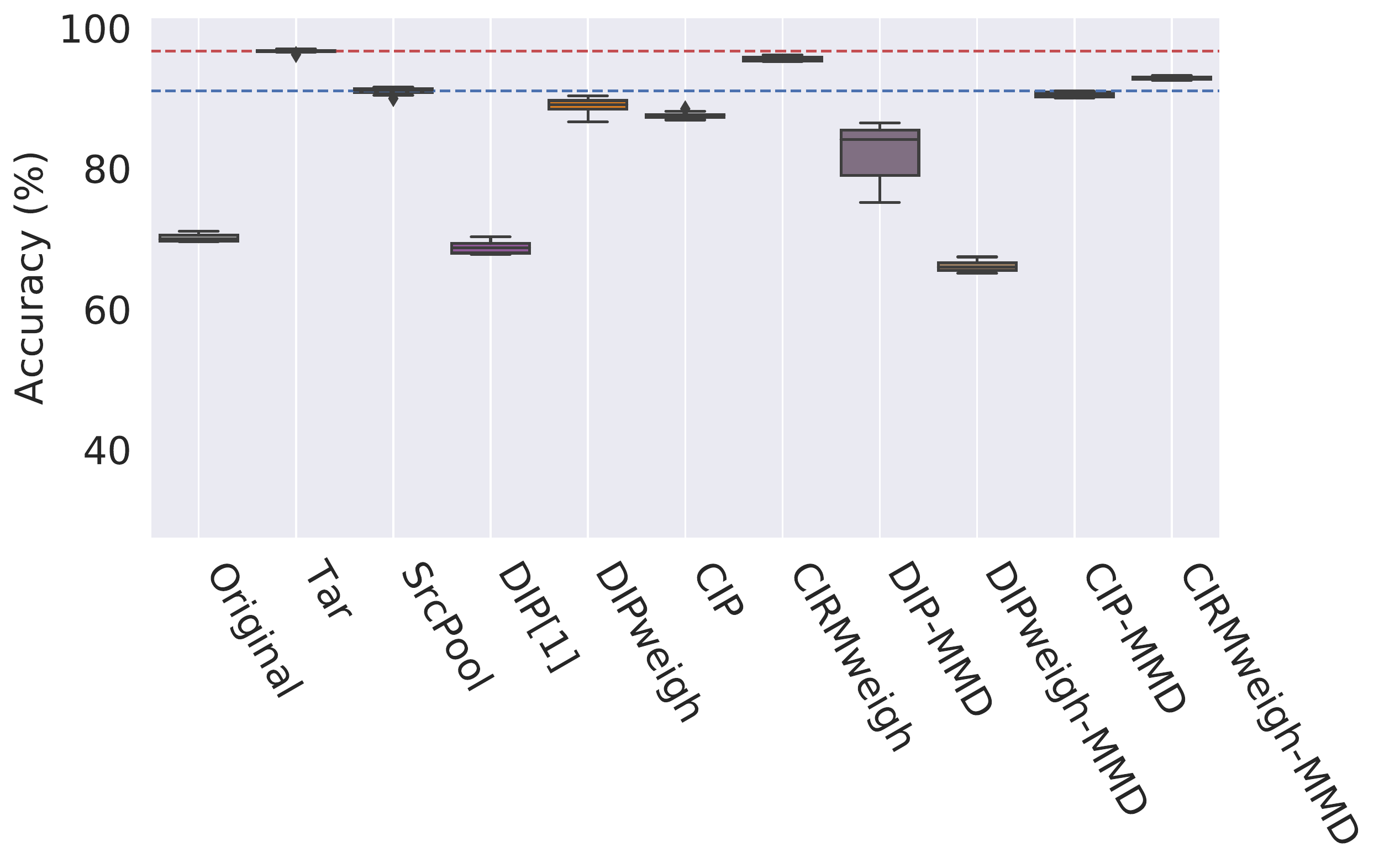}
    % \subcaption{}
    % \label{fig:MNIST_patch_intervention_multiple}
  \end{minipage}\hfill
  \caption{Left: Target accuracy comparison in MNIST experiment with patch intervention and single source without $Y$ intervention. Right: Target accuracy comparison in MNIST experiment with patch intervention and multiple sources with $Y$ intervention.}
  \label{fig:MNIST_patch_intervention_target}
\end{figure}

The left plot in Figure~\ref{fig:MNIST_patch_intervention_target} shows that both DIP$^\tagk{1}$ and DIP$^\tagk{1}$-MMD achieve better target accuracy than Original or Src$^\tagk{1}$.

\begin{figure}[ht]
  \includegraphics[width=0.99\textwidth]{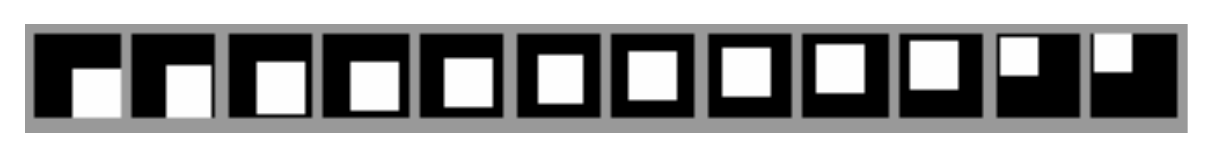}
  \caption{The 12 $X$ interventions in the MNIST multiple source domain adaptation with Y intervention. From left to right, the first 11 are the source $X$ interventions, the last is the target $X$ intervention. The target environment suffers additional $Y$ intervention. }
  \label{fig:MNIST_patches_interv_M12}
\end{figure}

\paragraph{MNIST DA with patch intervention with Y intervention: }
We take the original MNIST dataset with 60000 training samples and create 12 synthetic datasets ($\envs=11$ source environments and one target environment) as follows. For the $m$-th source dataset, each training image is masked by the $m$-th image (from left to right) in Figure~\ref{fig:MNIST_patches_interv_M12}. For the target dataset, each train or test image is masked by the right most image in Figure~\ref{fig:MNIST_patches_interv_M12}. The target dataset suffers additional $Y$ intervention: for digits (3, 4, 5, 6, 8, 9) in the target dataset, $80\%$ of the images in the MNIST dataset are removed from the target dataset. For experiment, we take random $20\%$ of samples from the source datasets as the 11 source environments and $20\%$ of samples from the target dataset as the target environment. This experiment is repeated 10 times and we report the boxplot of 10 target classification accuracies for each DA method.

As the MNIST experiments above, we apply the DA methods on the last-layer features of the same pre-trained convolutional neural network (CNN). In addition to the methods in the MNIST experiments above, we also consider the following methods and their MMD variants:
\begin{itemize}
  \item DIPweigh: CNN model where DIPweigh-mean-finite is applied on the last-layer features.
  \item CIP: CNN model where CIP-mean-finite is applied on the last-layer features.
  \item CIRMweigh: CNN model where CIRM-mean-finite is applied on the last-layer features.
\end{itemize}
The regularization parameter $\lamCIP$ is chosen to be $0.1$ based on source risk. For the regularization parameter $\lamMatch$ in DIPweigh and CIRMweigh, we first output the source environment index with the largest weight as ``the best source''. Then we vary it from the set $\braces{10^k}_{k=-5, \cdots, 4}$, and we choose the largest $\lamMatch$ such that the source accuracy of ``the best source'' is not dropped too much (in this case not more than 1\% of the source accuracy of Src applied to ``the best source'') as the final regularization parameter.

The right plot in Figure~\ref{fig:MNIST_patch_intervention_target} compares the target accuracies of the DA methods in this setting. We observe that CIRMweigh and CIRMweigh-MMD outperforms SrcPool and Original. Due to the intervention on $Y$, the matching penalty of DIPweigh and DIPweigh-MMD is not useful and the two methods perform worse than SrcPool.

\subsubsection{MNIST with rotation intervention}
\label{subs:MNIST_rotation}
\begin{figure}[ht]
    \centering
  \begin{minipage}{0.16\textwidth}
      \centering
      \includegraphics[width=0.99\textwidth]{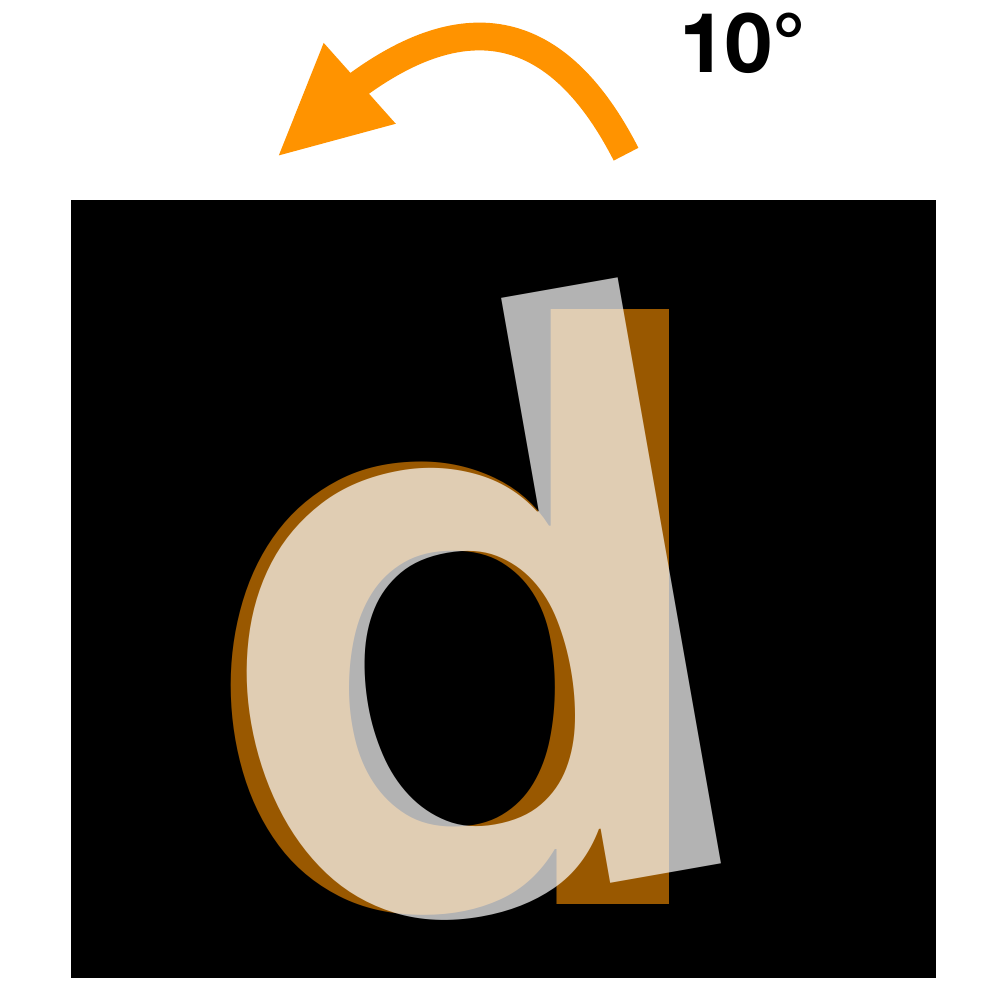}
      \subcaption{Rotation$10^\circ$}
  \end{minipage}
  \hfill
  \begin{minipage}{0.16\textwidth}
      \centering
      \includegraphics[width=0.99\textwidth]{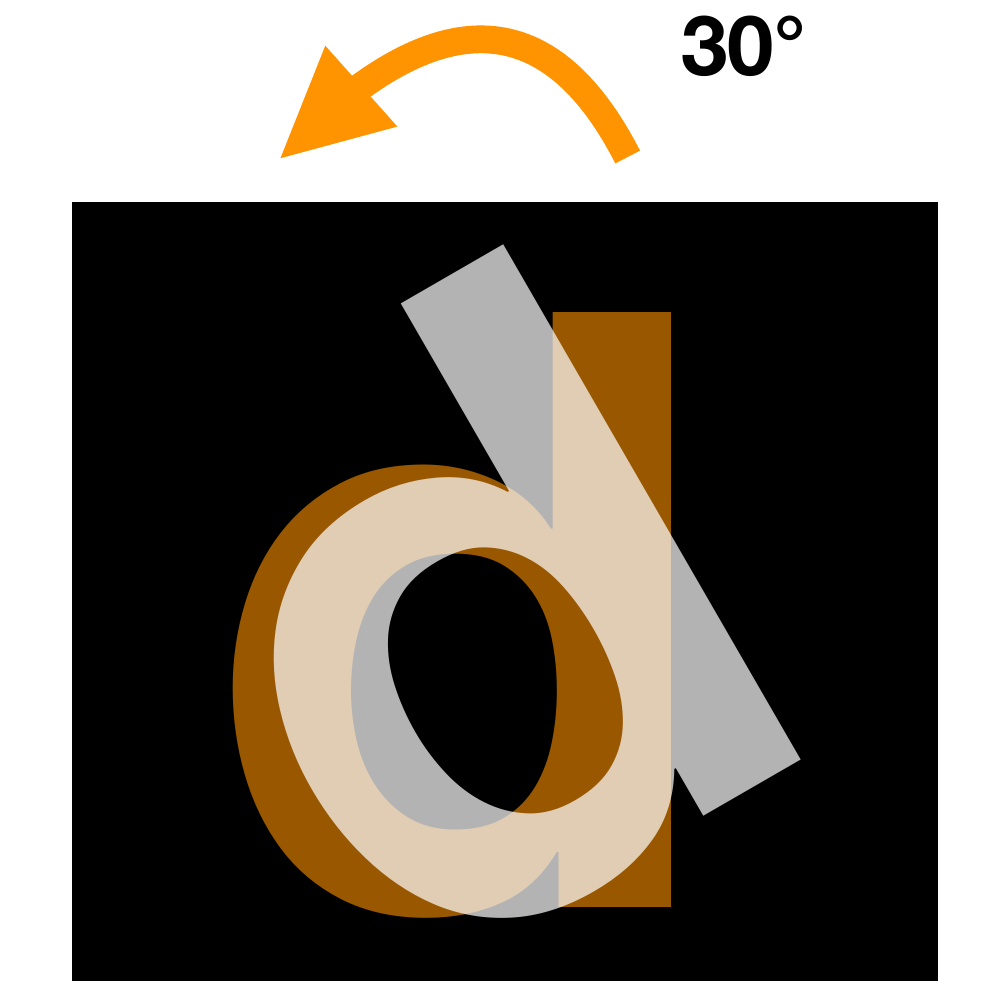}
      \subcaption{Rotation$30^\circ$}
  \end{minipage}
  \hfill
  \begin{minipage}{0.16\textwidth}
      \centering
      \includegraphics[width=0.99\textwidth]{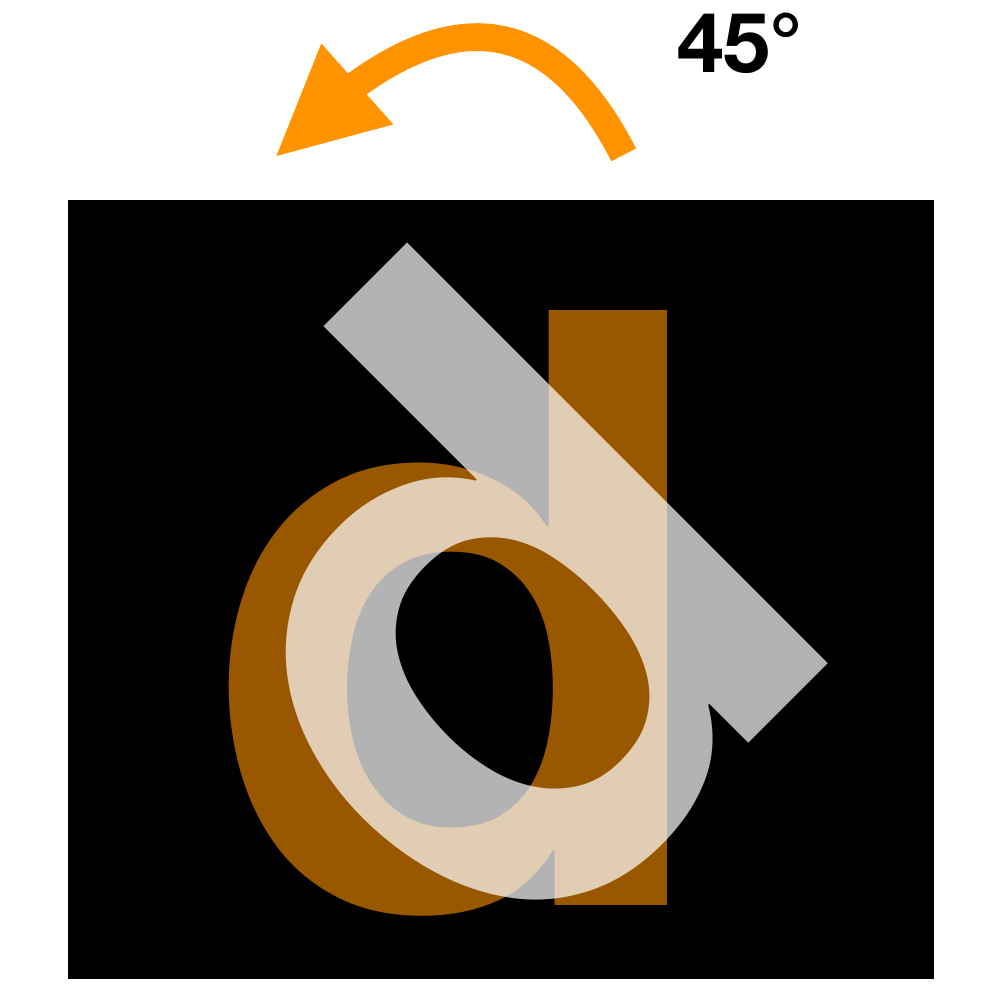}
      \subcaption{Rotation$45^\circ$}
  \end{minipage}
  \hfill
  \begin{minipage}{0.16\textwidth}
      \centering
      \includegraphics[width=0.99\textwidth]{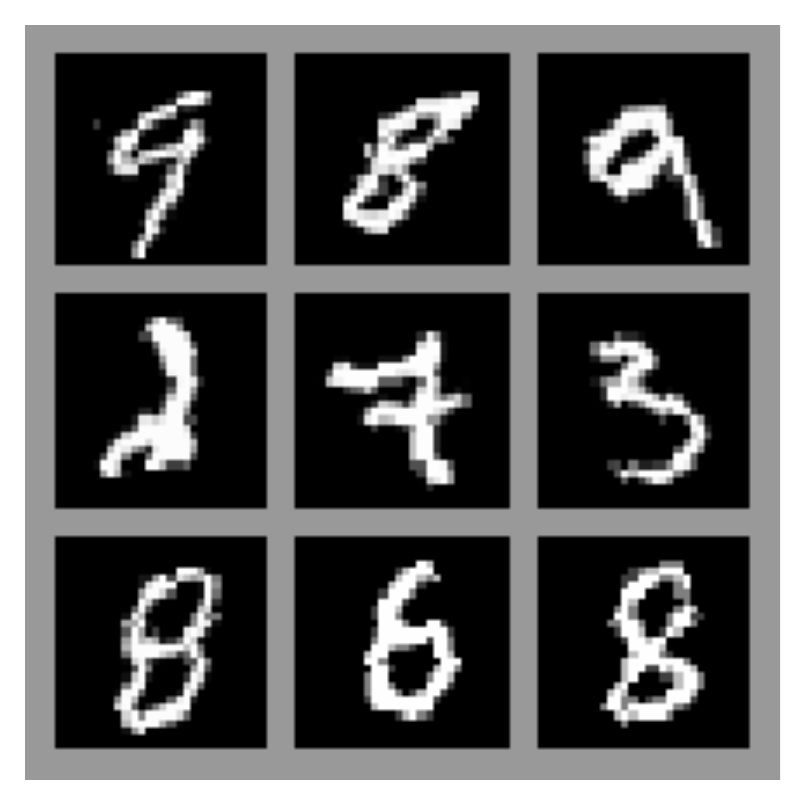}
      \subcaption{samples $10^\circ$}
  \end{minipage}
  \hfill
  \begin{minipage}{0.16\textwidth}
      \centering
      \includegraphics[width=0.99\textwidth]{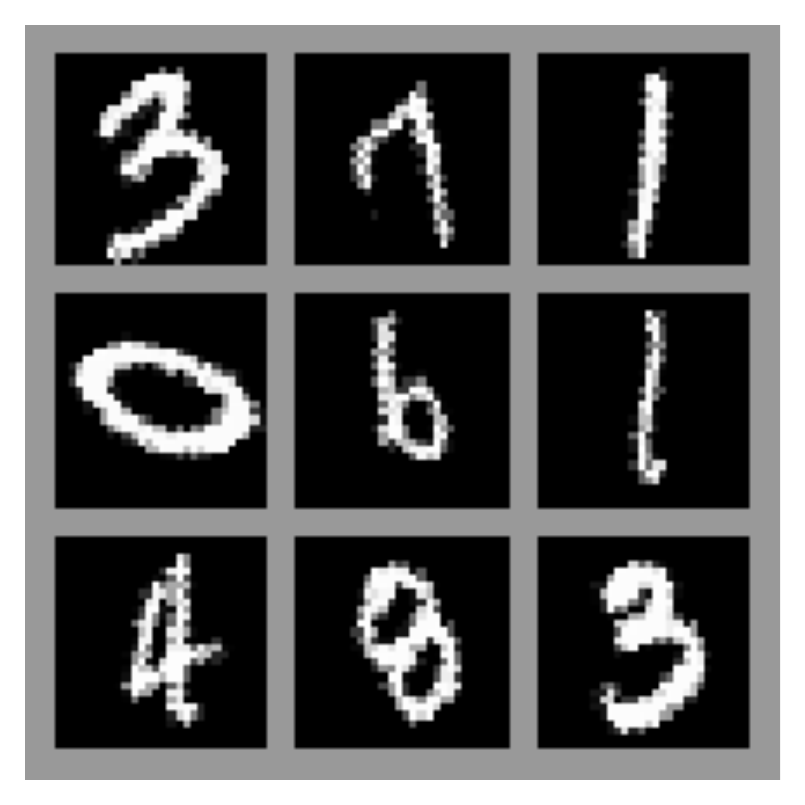}
      \subcaption{samples $30^\circ$}
  \end{minipage}
  \hfill
  \begin{minipage}{0.16\textwidth}
      \centering
      \includegraphics[width=0.99\textwidth]{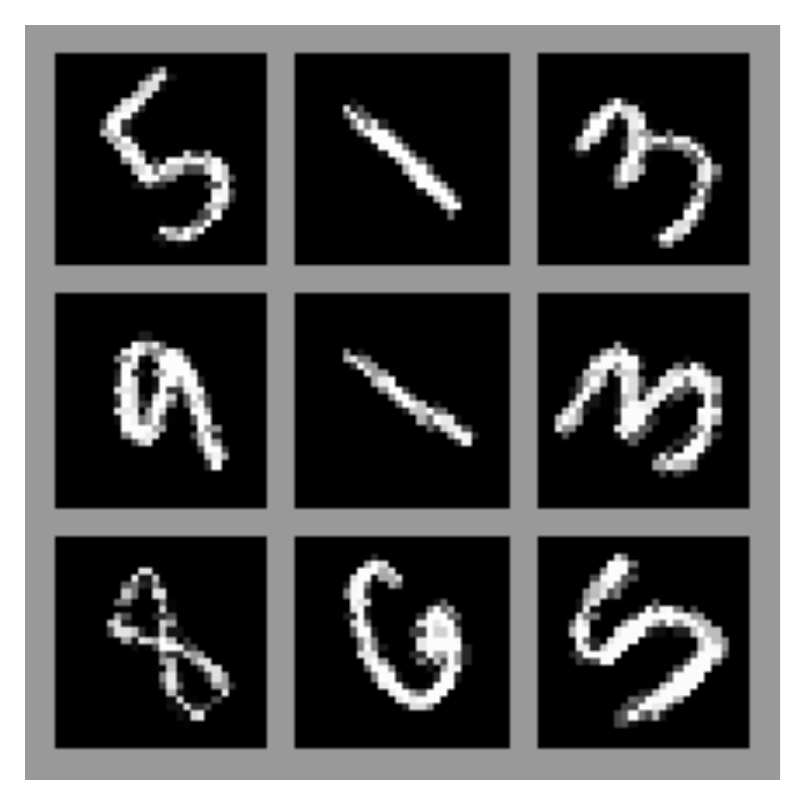}
      \subcaption{samples $45^\circ$}
  \end{minipage} \hfill
  \caption{(a)(b)(c): the three interventions in the MNIST rotation intervention domain adaptation without $Y$ intervention. (d)(e)(f): the corresponding data samples under rotation intervention.}
  \label{fig:MNIST_rotation_2M_interv}
\end{figure}

\paragraph{MNIST DA with rotation intervention without Y intervention:}
We take the original MNIST dataset with 60000 training samples and create three synthetic datasets (2 sources and 1 target) as follows. For each dataset, each training image is rotated by one of the angles $\braces{10, 30, 45}$ anti-clock-wise as shown in Figure~\ref{fig:MNIST_rotation_2M_interv}. The three datasets are named Rotation~$10^\circ$, Rotation~$30^\circ$ and Rotation~$45^\circ$ respectively. For the first experiment, we use Rotation $10^\circ$ as the source environment and Rotation $45^\circ$ as the target environment. For the second experiment, we use Rotation $30^\circ$ as the source environment and Rotation $45^\circ$ as the target environment. For each experiment, we take random $20\%$ of samples from the source dataset as the source environment and random $20\%$ of samples from the target dataset as the target environment. The task is to predict the labels of the images in the target environment without observing any target labels. Except for the change in the type of intervention, the other experimental settings are the same as in MNIST DA with patch intervention without Y intervention in Section~\ref{subs:MNIST_patch}.

\begin{figure}[ht]
  \begin{minipage}{0.24\textwidth}
    \centering
    \includegraphics[width=0.99\textwidth]{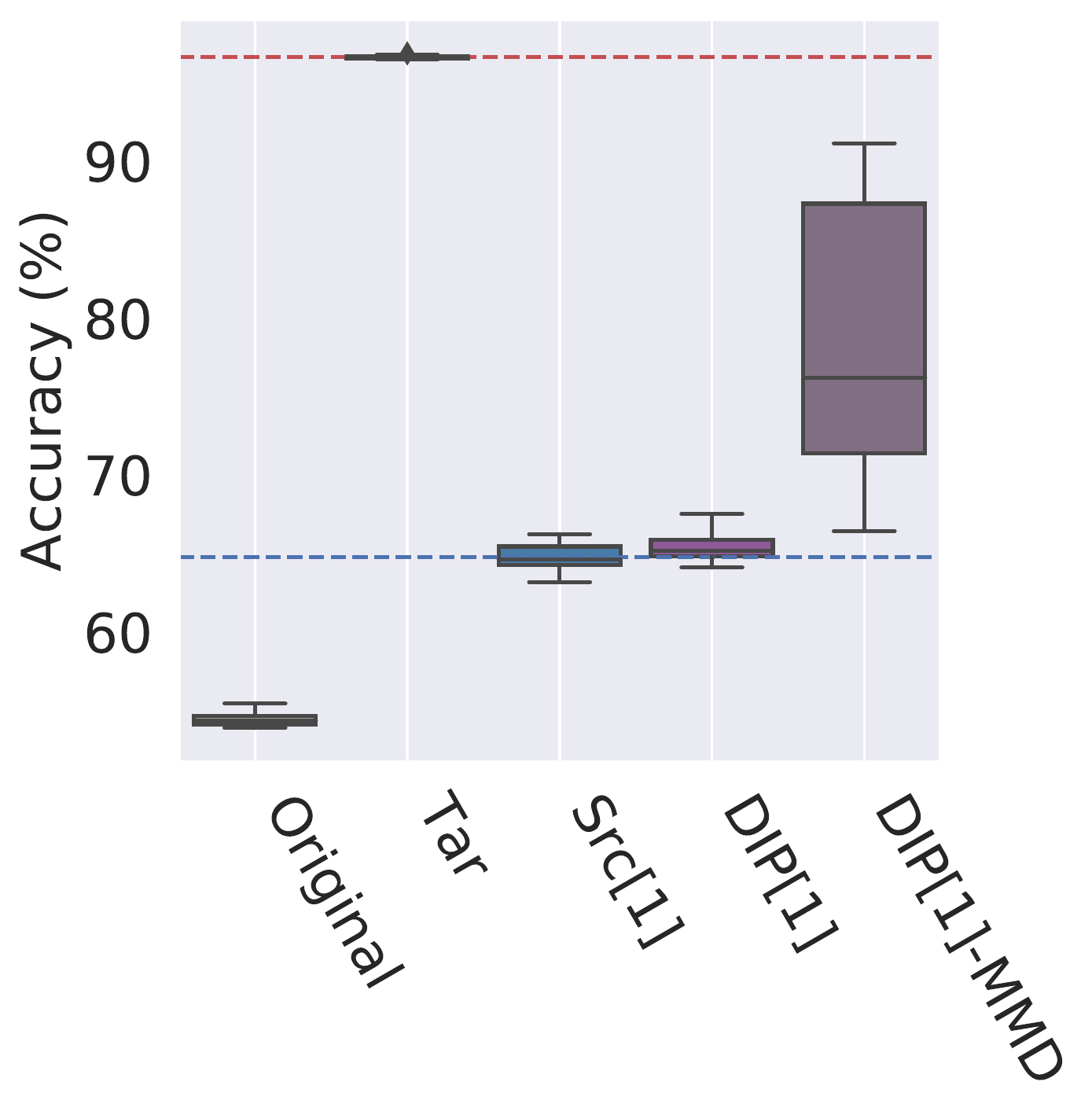}
    \subcaption{src: Rotation $10^\circ$ \\\phantom{aaa} tar: Rotation $45^\circ$}
    \label{fig:MNIST_rotation_intervention_single_10_45}
  \end{minipage}\hfill
  \begin{minipage}{0.24\textwidth}
    \centering
    \includegraphics[width=0.99\textwidth]{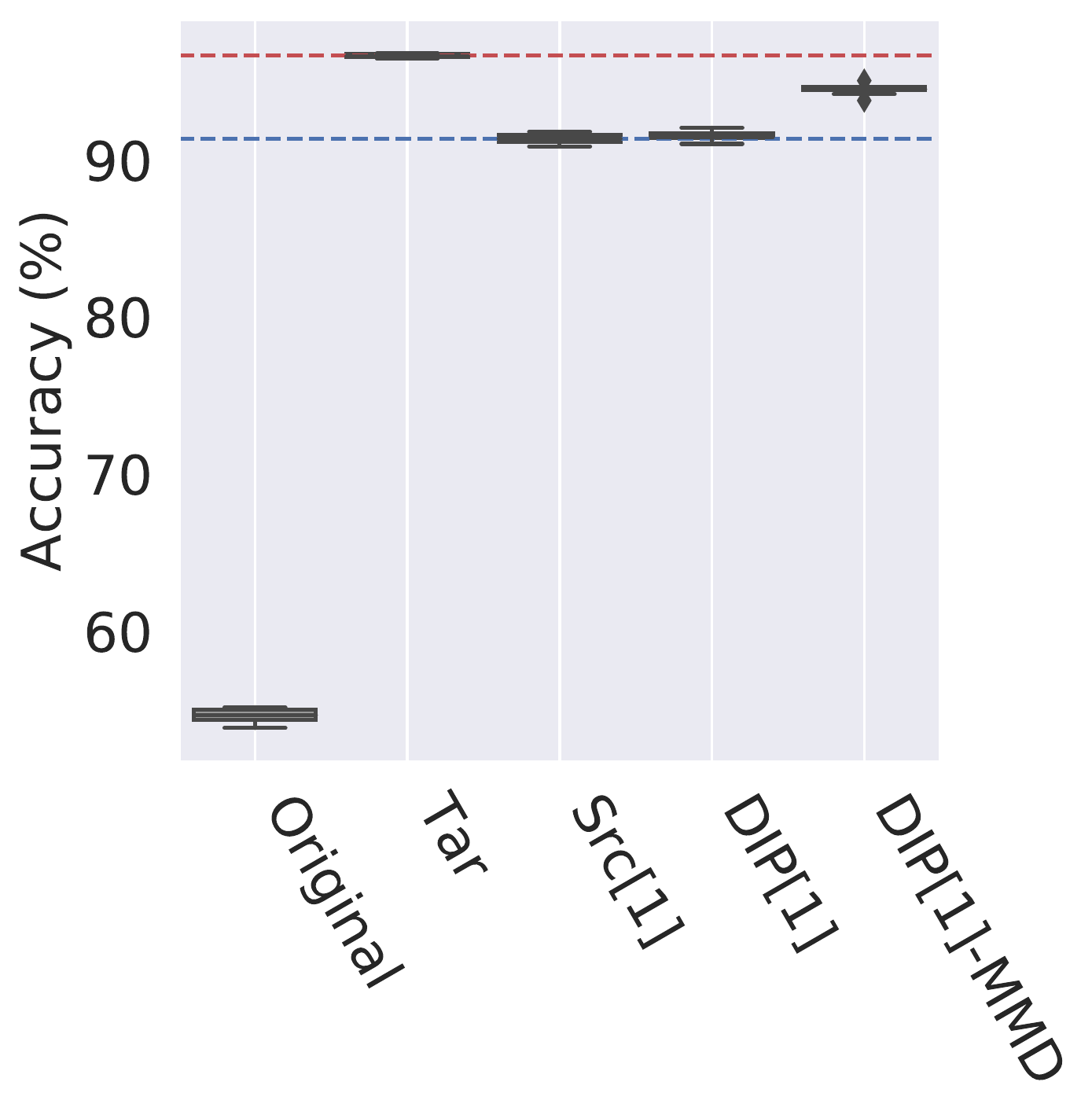}
    \subcaption{src: Rotation $30^\circ$ \\\phantom{aaa} tar: Rotation $45^\circ$}
    \label{fig:MNIST_rotation_intervention_single_30_45}
  \end{minipage}\hfill
  \begin{minipage}{0.50\textwidth}
    \centering
    \includegraphics[width=0.99\textwidth]{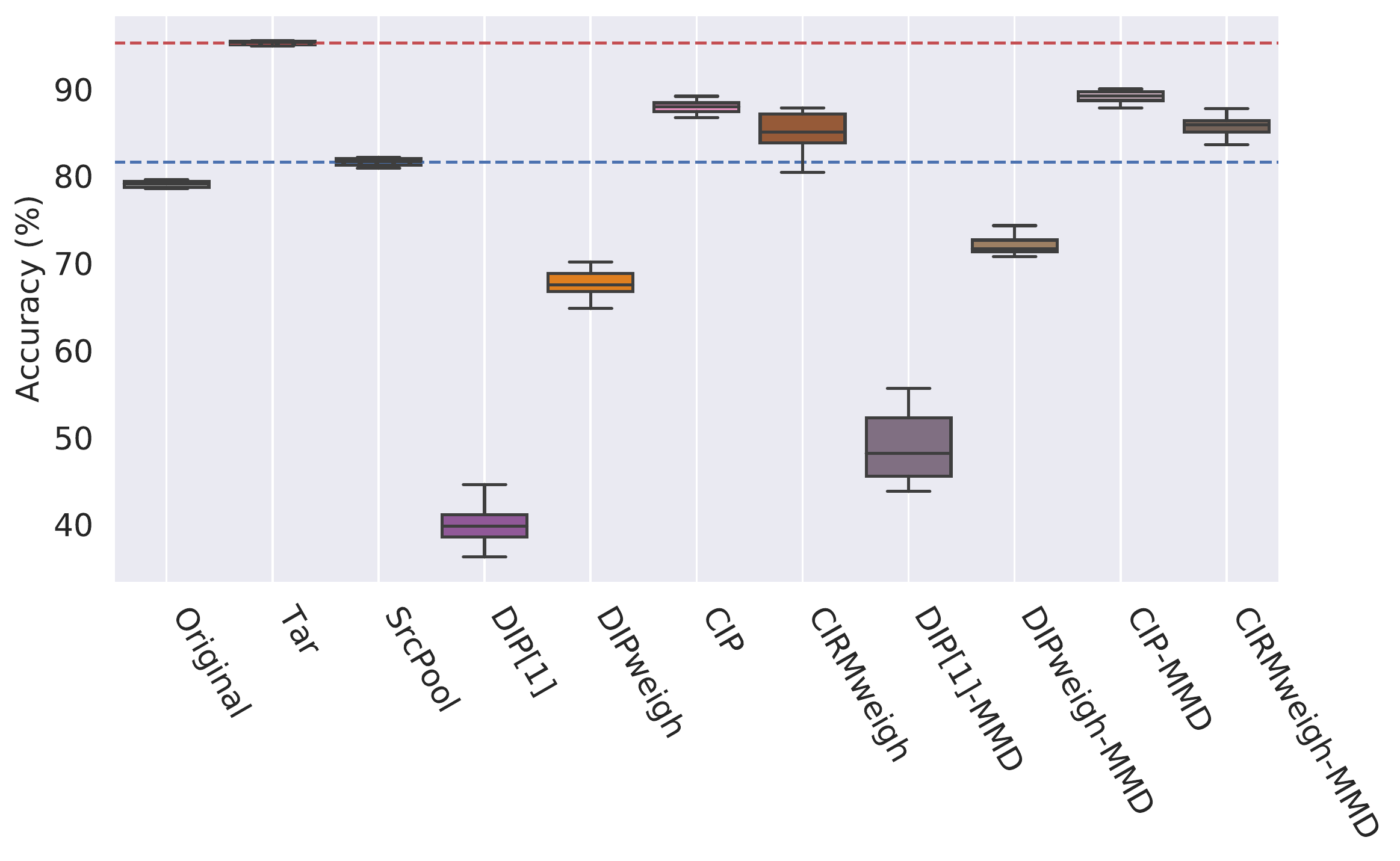}
    \subcaption{MNIST rotation interv. with $Y$ interv.}
    \label{fig:MNIST_rotation_intervention_multiple}
  \end{minipage}\hfill
  \caption{(a) Target accuracy comparison in MNIST experiment with rotation intervention and single source without $Y$ intervention from Rotation 10\% to Rotation 45\%. (b) Target accuracy comparison in MNIST experiment with rotation intervention and single source without $Y$ intervention from Rotation 30\% to Rotation 45\%. (c) Target accuracy comparison MNIST experiment with rotation intervention and multiple source with $Y$ intervention.}
  \label{fig:MNIST_rotation_intervention_target}
\end{figure}

Figure~\ref{fig:MNIST_rotation_intervention_single_10_45} shows the boxplot of the target accuracies of Original, Tar, Src$^\tagk{1}$, DIP$^\tagk{1}$ and DIP$^\tagk{1}$-MMD for the first experiment. DIP$^\tagk{1}$-MMD achieves higher target accuracy than Src$^\tagk{1}$. Figure~\ref{fig:MNIST_rotation_intervention_single_30_45} shows the boxplot of 10 runs of Original, Tar, Src$^\tagk{1}$, DIP$^\tagk{1}$ and DIP$^\tagk{1}$-MMD for the second experiment. DIP$^\tagk{1}$-MMD achieves higher target accuracy than Src$^\tagk{1}$. Comparing Figure~\ref{fig:MNIST_rotation_intervention_single_10_45} and~\ref{fig:MNIST_rotation_intervention_single_30_45}, we also observe that the first experiment is a more difficult classification task than the second experiment as the accuracies achieved by our DA methods are lower in the first experiment.

\paragraph{MNIST DA with rotation intervention with Y intervention:} We take the original MNIST dataset with 60000 training samples and create 5 synthetic datasets ($\envs=4$ source environments and one target environment) as follows. For $m \in \braces{1, \cdots, 4}$, for the $m$-th source dataset, each training image is rotated by $\parenth{m \times 15 - 30}^\circ$ clock-wise. Each image in the target dataset is rotated by $30^\circ$ clock-wise. The target dataset suffers additional $Y$ intervention: for digits (3, 4, 5, 6, 8, 9) in the target dataset, $80\%$ of the images in the MNIST dataset are removed from the target dataset. For experiment, we take random $20\%$ of samples from the source datasets as the 4 source environments and $20\%$ of samples from the target dataset as the target environment. Except for the change in the type of intervention, the other experimental settings are the same as in MNIST DA with patch intervention with Y intervention in Section~\ref{subs:MNIST_patch}.

Figure~\ref{fig:MNIST_rotation_intervention_multiple} shows the boxplot of the target accuracies of Original, Tar, SrcPool, DIP$^\tagk{1}$, DIPweigh, CIP, CIRMweigh, DIP-MMD, CIP-MMD and CIRMweigh-MMD. CIP and CIP-MMD outperforms SrcPool and Original. Due to the intervention on Y, the matching penalty of DIPweigh and DIPweigh-MMD is not useful and the two methods perform worse than SrcPool. CIP, CIRMweigh and the corresponding MMD variants outperform SrcPool.

\subsubsection{MNIST with random translation intervention}
\begin{figure}[ht]
  \begin{minipage}{0.25\textwidth}
    \centering
    \includegraphics[width=0.99\textwidth]{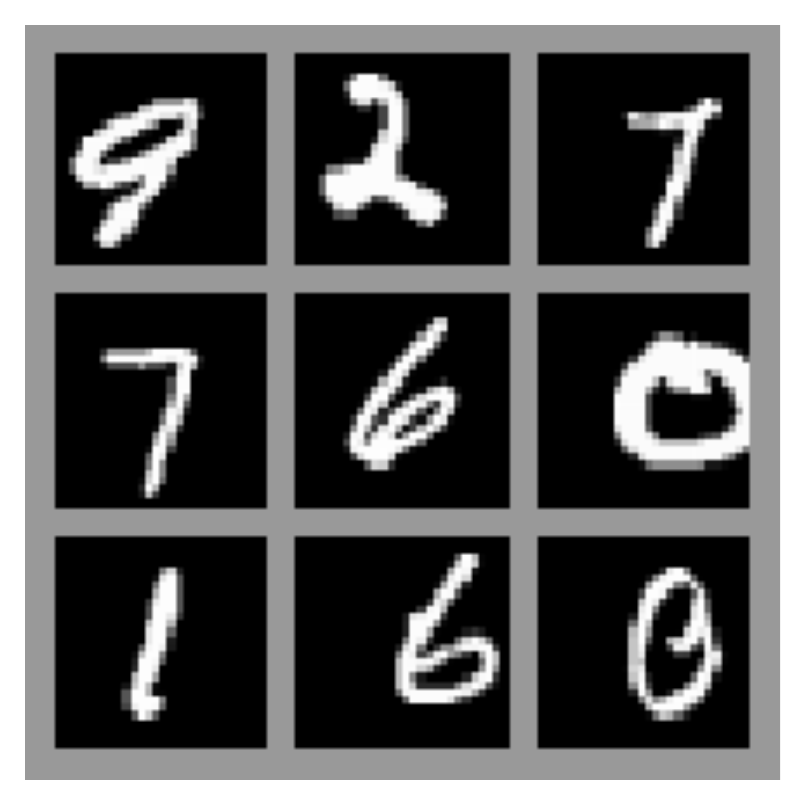}
    \subcaption{source samples after \\ random horizontal \\ translation}
    \label{fig:MNIST_translation_intervention_single_env0}
  \end{minipage}\hfill
  \begin{minipage}{0.25\textwidth}
    \centering
    \includegraphics[width=0.99\textwidth]{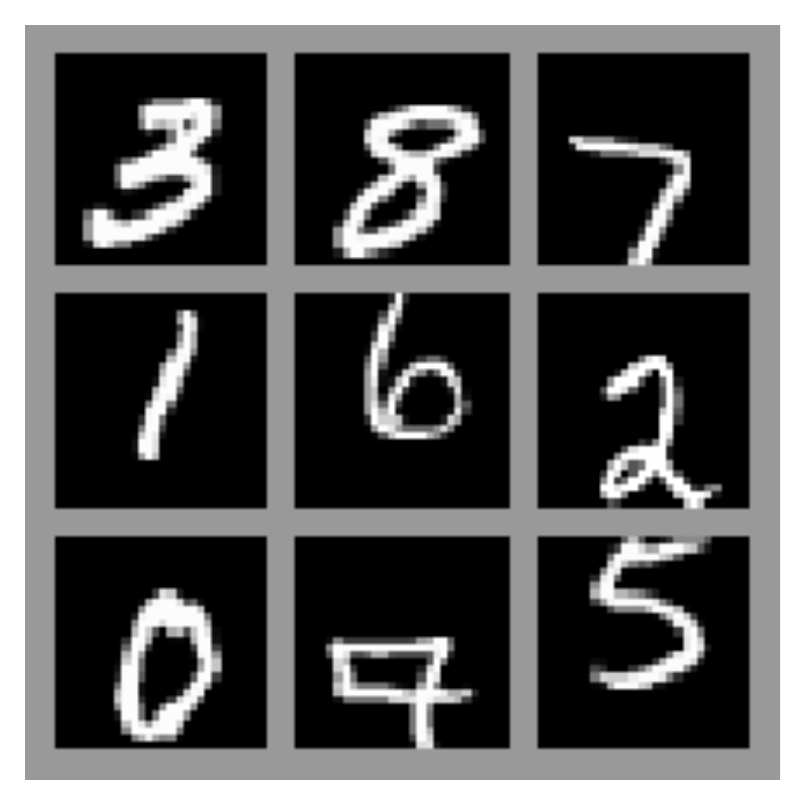}
    \subcaption{target samples after \\ random vertical \\ translation}
    \label{fig:MNIST_translation_intervention_single_env1}
  \end{minipage}\hfill
  \begin{minipage}{0.33\textwidth}
    \centering
    \includegraphics[width=0.99\textwidth]{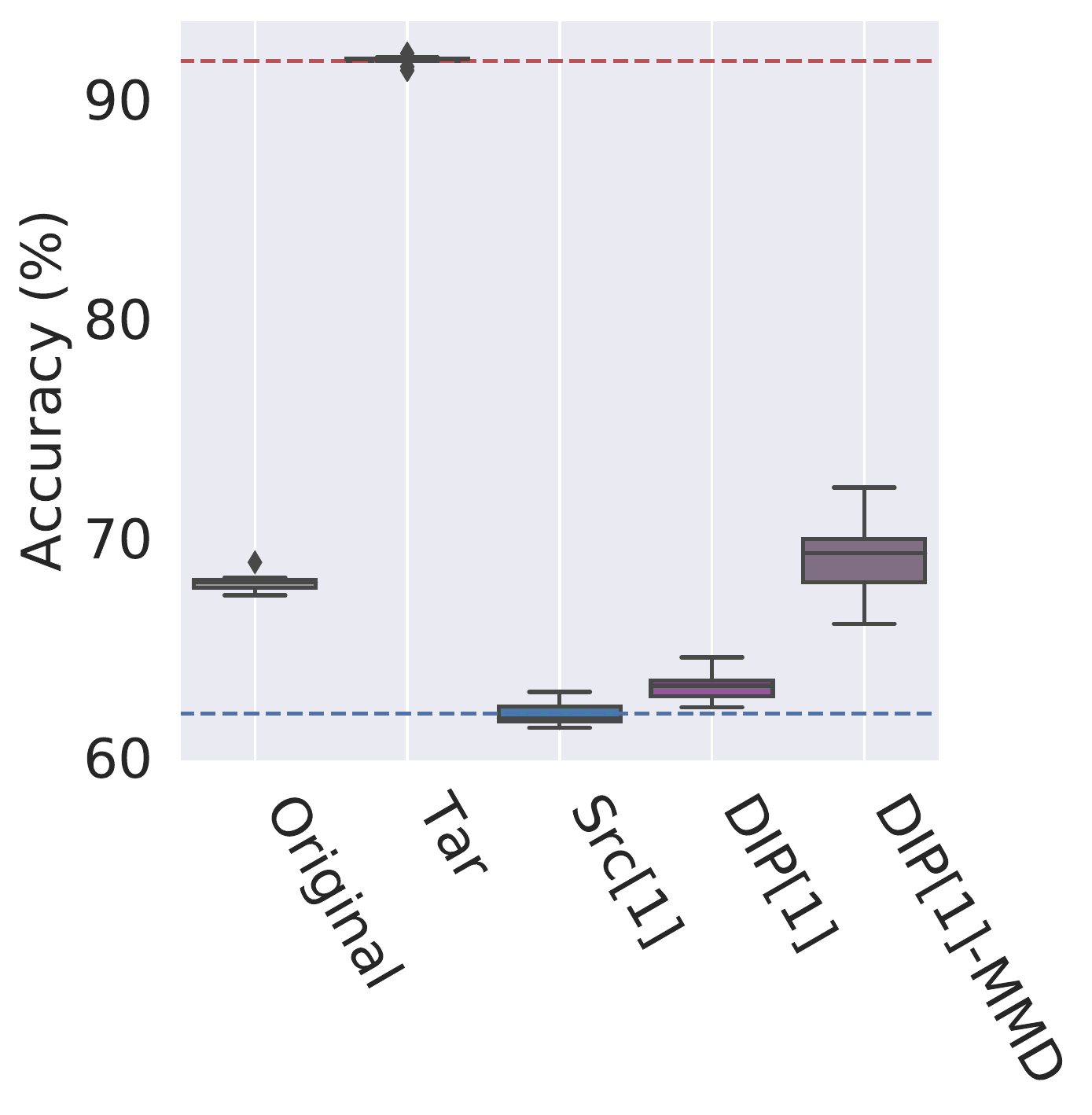}
    \subcaption{MNIST translation interv.}
    \label{fig:MNIST_translation_intervention_single_res}
  \end{minipage}\hfill
  \caption{(a) Source samples after random horizontal translation. (b) Target samples after random vertical translation. (c) Target accuracy comparison in MNIST experiment with random translation intervention and single source without Y intervention. }
  \label{fig:MNIST_translation_intervention_target}
\end{figure}

We take the original MNIST dataset with 60000 training samples and create two synthetic datasets (source and target) as follows. For the source dataset, each training image is translated horizontally with a distance randomly selected from $-0.2 \times \text{image width}$ to $0.2 \times \text{image width}$ as shown in Figure~\ref{fig:MNIST_translation_intervention_single_env0}. For the target dataset environment, each training image is translated vertically with a distance randomly selected from $-0.2 \times \text{image height}$ to $0.2 \times \text{image height}$ as shown in Figure~\ref{fig:MNIST_translation_intervention_single_env1}. For each experiment, we take random $20\%$ of samples from the source dataset as the source environment and random $20\%$ of samples from the target dataset as the target environment. The task is to predict the labels of the images in the target environment without observing any target labels. Except for the change in the type of intervention, the other experimental settings are the same as in MNIST DA with patch intervention without Y intervention in Section~\ref{subs:MNIST_patch}.

Figure~\ref{fig:MNIST_translation_intervention_single_res} shows the boxplot of the target accuracies of Original, Tar, Src$^\tagk{1}$, DIP$^\tagk{1}$ and DIP$^\tagk{1}$-MMD. DIP$^\tagk{1}$-MMD still performs better than Src$^\tagk{1}$, but it barely improves over Original. Since the intervention is random rather than fixed for each environment, intuitively this experiment is more difficult for our DA methods to have high accuracy than the first two experiments with fixed intervention for each environment.

% subsection mnist_experiments_with_synthetic_interventions (end)

\subsection{Experiments on Amazon review dataset with unknown interventions} % (fold)
\label{sub:real_data_experiments_with_unknown_interventions}
In this subsection, we compare DA methods on the regression dataset Amazon Review Data~\citep{ni2019justifying}. Unlike the simulated experiments or the MNIST experiment, neither the data generation process nor the type of interventions is known.

This dataset contains product reviews and metadata from Amazon in the date range May 1996 - Oct 2018. In our experiment, we consider the task of predicting review rating from review text from various product categories (Automotive, Digital Music, Office Products etc.). For each review, the covariates $X$ are the TF-IDF features generated from raw review text; the label $Y$ is the review rating (score from $1-5$). The different product categories are used as source and target environments. We take 15 product categories with the largest sample sizes, use 14 of them as source environments and leave the last one as the target environment.

The samples size $\obs$ is 10000 in each source or target environment. The dimension depends on the TF-IDF feature extractor. Here we use both unigrams and bigrams, and build the vocabulary with terms that have a document frequency not smaller than $0.008$. This results in feature dimension $\dims = 482$. A linear model with $\ell_2$ regularization is used on top of the TF-IDF features to predict ratings.

\begin{figure}[ht]
  \centering
  \begin{minipage}{0.32\textwidth}
      \centering
      \includegraphics[width=0.99\textwidth]{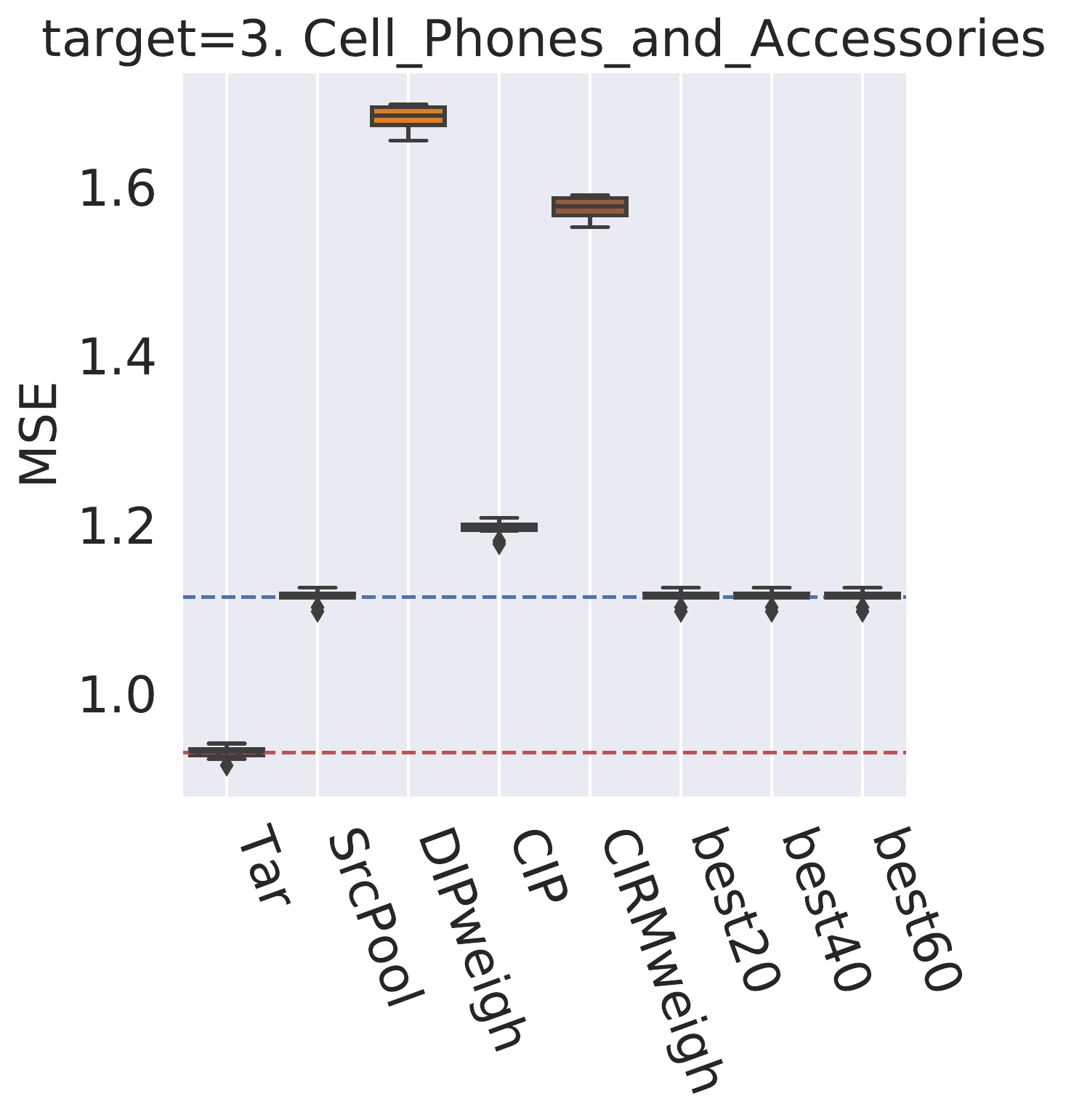}
      % \caption{mask (a)}
  \end{minipage}
  \begin{minipage}{0.32\textwidth}
      \centering
      \includegraphics[width=0.99\textwidth]{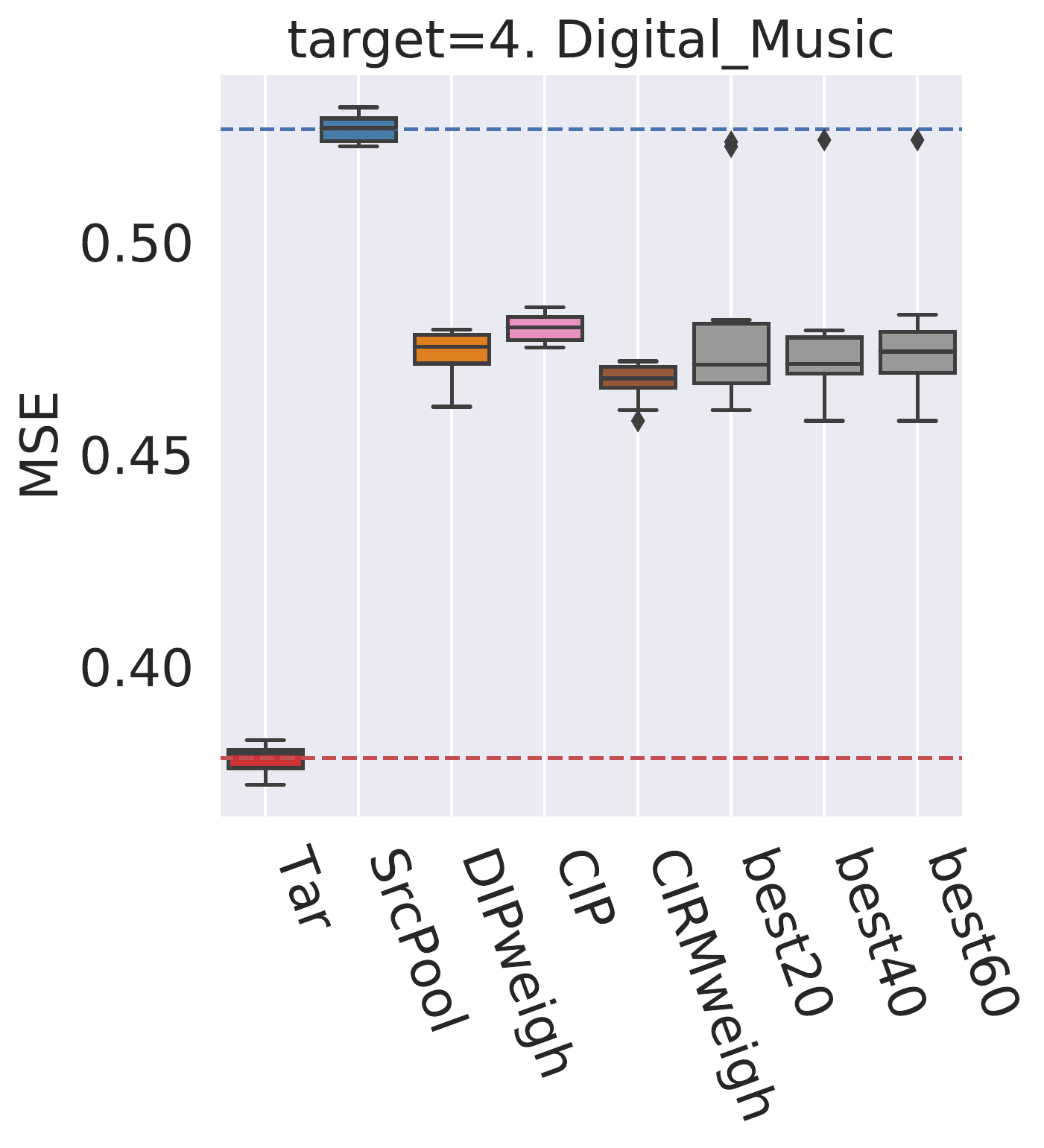}
      % \caption{mask (b)}
  \end{minipage}
  \begin{minipage}{0.32\textwidth}
      \centering
      \includegraphics[width=0.99\textwidth]{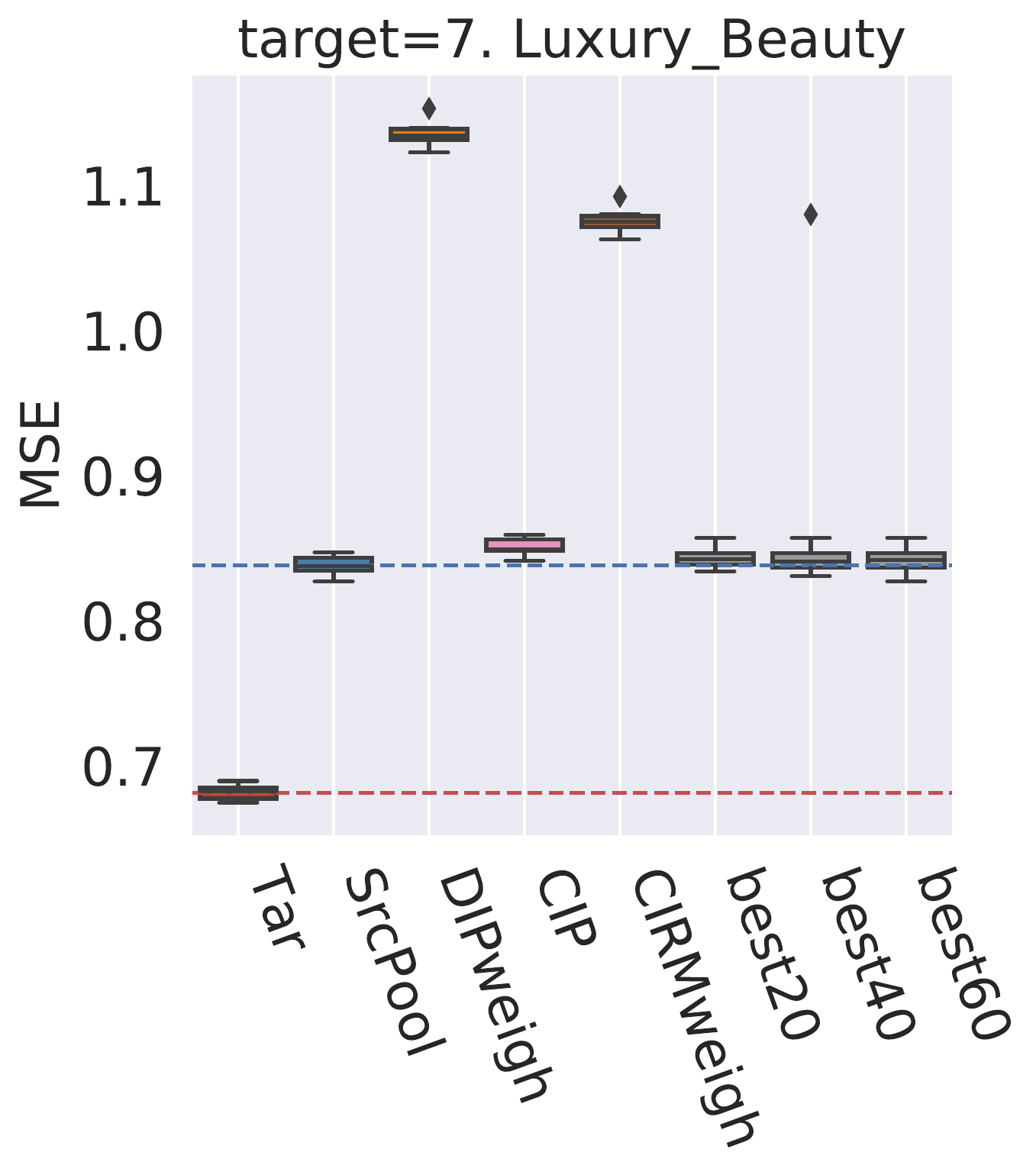}
      % \caption{mask (b)}
  \end{minipage}\hfill
\caption{Target risk in Amazon review data prediction experiment (the lower the better). Without explicit knowledge about the type of interventions, it is no longer clear which domain adaptation method is the best. From left to right, depending on the target environment choice, the best methods are CIP, DIPweigh, CIRMweigh accordingly. }
\label{fig:Amazon_target_risk_comparison}
\end{figure}

Without explicit knowledge about the causal structure or the type of interventions, a priori it is no longer clear which DA method is the best. We apply SrcPool and the advanced DA methods, DIPweigh, CIP, CIRMweigh on the linear model. Figure~\ref{fig:Amazon_target_risk_comparison} shows that the advanced DA methods do not outperform SrcPool except for target environment number 4.

We did an additional experiment with a small portion of target labels revealed. We use the small portion of target labels to choose the best method out of SrcPool, DIPweigh, CIP and CIRMweigh based on the small portion of labeled target data. The last three boxes (labeled as best20, best40 and best60) in each subplot of Figure~\ref{fig:Amazon_target_risk_comparison} show that with 20, 40, 60 target labels revealed, the best method based on the small portion of labeled target data is always not worse than SrcPool and can sometimes outperform SrcPool.

We arrive at a conclusion that in general, without explicit knowledge about causal structure or the type of interventions, it is ambitious to expect advanced DA methods to always outperform SrcPool. However, with a small portion of labeled target data, we show that DA methods still lead to substantial improvements. Our empirical observation agrees with the concurrent empirical study of DA methods by~\cite{wiles2021fine}. In a larger-scale experimental framework with many real and synthetic datasets, they find that DA methods can outperform SrcPool in some settings but there is no single best method across all settings. Hence, knowledge about the distribution shift is necessary for successful DA with guarantees.

% subsubsection amazon_review_data_prediction (end)

% subsection real_data_experiments_with_unknown_interventions (end)

% section numerical_experiments (end)

%%%%%%%%%%%%%%%%%%%%%%%%%%%%%%%%%%%%%%%%%%%%%%%%%%%%%%%%%%%%%%%%%%%%%%%%
%%%%%%%%%%%%%%%%%%%%%%%%%%%%%%%%%%%%%%%%%%%%%%%%%%%%%%%%%%%%%%%%%%%%%%%
\section{Discussion} % (fold)
\label{sec:discussion}
In this paper, we propose a theoretical framework using structural causal models (SCMs) to analyze and obtain insights on prediction performance of several popular domain adaptation methods. First, we show that under the assumption of anticausal prediction, linear SCM and no intervention on $Y$, the popular DA method DIP achieves a low target risk. This theoretical result is compatible with many previous empirical results on the usefulness of DIP for domain adaptation. Second, we derive several conditions where DIP fails to achieve a low target risk: when the prediction problem is causal or mixed-causal-anticausal; when there is label distribution perturbation. To tackle these difficult DA scenarios, we design a new DA method called CIRM, among other modifications. The theoretical analysis is complemented with simulation analysis and real data experiments. The theoretical extension to nonlinear SCMs is a challenging future direction. However our empirical results suggest that the linear SCM assumption provides useful insights even for cases where a linear SCM does not hold.

The real data experiments show that it can be beneficial to use DA methods when the anticausal prediction assumption is satisfied but it can also be dangerous to blindly use DA methods when little is known about the data generation process. Thus, prior knowledge of the underlying causal structure and the types of interventions are often crucial. The prior knowledge of the causal structure can come from domain expert knowledge or causal studies on related datasets with the same variables. How to seamlessly combine causal studies about the causal structure and domain adaptation with appropriate uncertainty quantification is one promising future direction.

In absence of prior information about the causal structure and the types and locations of interventions, one needs to select good DA methods. The assessment of the goodness of a model or algorithm is difficult in general but it can be done when having access to a small fraction of labeled target data. We show empirically that a small fraction of labeled target data substantially helps to select the best DA method. It remains an open question what is the minimal amount of labeled target data points in order to guarantee the best DA method selection.
% section discussion (end)

% Acknowledgements should go at the end, before appendices and references

\acks{Y. Chen and P. B\"uhlmann have received funding from the European ResearchCouncil under the Grant Agreement No 786461 (CausalStats - ERC-2017-ADG).
They both acknowledge scientific interaction and exchange at ``ETH Foundations of Data Science''. They also thank Domagoj Cevid, Christina Heinze-Deml, Wooseok Ha, Jinzhou Li, Nicolai Meinshausen, Armeen Taeb, Fanny Yang and Andrii Zadaianchuk for fruitful discussions and for helpful suggestions on presentation and writing. }


\newpage

\appendix
%!TEX root = ./main_paper.tex

\section{Summary of DA methods in this paper} % (fold)
\label{sec:summary_of_da_methods_in_this_paper}
In this section, we provide a summary of the DA methods presented in this paper. Methods that do not fit in the main paper due to the space limitation are formally introduced here.
\subsection{Population DA methods}
\label{sub:population_DA_methods}
First, we formulate the DIP variants that adapt to other types of intervention as mentioned in Section~\ref{ssub:failure_scenarios_of_dip}.
\begin{itemize}
  \item \textbf{DIP$^\tagk{m}$-std:} the population DIP estimator where the difference between the source and target standard deviations is used as distributional distance. For the $m$-th source environment, it is defined as
  \begin{align}
    \label{eq:estimator_pop_dipstdmatchlin}
    f_{\tagdipstdmatchlin}^\tagk{m}(x) &\defn x \tp \beta_{\tagdipstdmatchlin}^\tagk{m} + \beta_{\tagdipstdmatchlin, 0}^\tagk{m} \notag \\
    \beta_\tagdipstdpmatchlin^\tagk{m}, \beta_{\tagdipstdmatchlin, 0}^\tagk{m} &\defn \argmin_{\beta, \beta_0}\  \Exs_{(X, Y) \sim \distriv{m}}\parenth{Y - X\tp\beta - \beta_0}^2 \notag \\
    &\text{s.t.\ } \Var_{X \sim \distriv{m}_X} \brackets{X \tp\beta} = \Var_{X \sim \distritar_X}\brackets{X \tp\beta}.
  \end{align}
  \item \textbf{DIP$^\tagk{m}$-std$+$:} the population DIP estimator where the differences between the source and target means, standard deviations and 25\% quantiles are used as distributional distance
  \begin{align}
    \label{eq:estimator_pop_dipstdpmatchlin}
    f_{\tagdipstdpmatchlin}^\tagk{m}(x) &\defn x \tp \beta_{\tagdipstdpmatchlin}^\tagk{m} + \beta_{\tagdipstdpmatchlin, 0}^\tagk{m} \notag \\
    \beta_\tagdipstdpmatchlin^\tagk{m}, \beta_{\tagdipstdpmatchlin, 0}^\tagk{m} &\defn \argmin_{\beta, \beta_0}\  \Exs_{(X, Y) \sim \distriv{m}}\parenth{Y - X\tp\beta - \beta_0}^2 \notag \\
    &\text{s.t.\ } \Exs_{X \sim \distriv{m}_X} \brackets{X \tp\beta} = \Exs_{X \sim \distritar_X}\brackets{X \tp\beta} \notag \\
    &\phantom{\text{s.}}\Var_{X \sim \distriv{m}_X} \brackets{X \tp\beta} = \Var_{X \sim \distritar_X}\brackets{X \tp\beta} \notag \\
    &\phantom{\text{s.t.\ \ }}\psi_{25\%} \parenth{{\Xv{m}}\tp \beta} = \psi_{25\%} \parenth{{\Xtar}\tp \beta},
  \end{align}
  where $\psi_{25\%}$ is the 25\% quantile function which takes a random variable and returns its 25\% quantile.
  \item \textbf{DIP$^\tagk{m}$-MMD:} the population DIP estimator where the maximum mean discrepancy (MMD) is used as distributional distance.
  \begin{align}
    \label{eq:estimator_pop_dipMMDmatchlin}
    f_{\tagdipMMDmatchlin}^\tagk{m}(x) &\defn x \tp \beta_{\tagdipMMDmatchlin}^\tagk{m} + \beta_{\tagdipMMDmatchlin, 0}^\tagk{m} \notag \\
    \beta_\tagdipMMDmatchlin^\tagk{m}, \beta_{\tagdipMMDmatchlin, 0}^\tagk{m} &\defn \argmin_{\beta, \beta_0}\  \Exs_{(X, Y) \sim \distriv{m}}\parenth{Y - X\tp\beta - \beta_0}^2 \notag \\
    &\text{s.t.\ } \disMMD\parenth{{\Xv{m}}\tp \beta, {\Xtar}\tp \beta} = 0,
  \end{align}
  where the maximum mean discrepancy (MMD)~\cite{gretton2012kernel} with respect to the reproducing kernel Hilbert space (RHKS) $\RKHS$ between two random variable $Z_1$ and $Z_2$ is
  \begin{align*}
    \disMMD\parenth{Z_1, Z_2} = \sup_{h \in \RKHS} \abss{\Exs \brackets{h({Z_1)}}  -  \Exs \brackets{h({Z_2)}} }.
  \end{align*}
  By default, the RKHS with Gaussian kernel is used throughout this paper.
\end{itemize}
Second we introduce the weighted version of CIRM following DIPweigh~\eqref{eq:estimator_pop_dipweigh}.
\begin{itemize}
  \item \textbf{CIRMweigh-mean:} the population CIRM estimator that weights the source environments based on the source risks. It is defined as follows
  \begin{align}
    \label{eq:estimator_pop_cirmweigh}
    f_{\tagcirmweigh}(x) &\defn \frac{1}{\sum_{m=1}^{\envs} e^{-\eta \cdot s_m}}\sum_{m=1}^{\envs} e^{-\eta \cdot s_m}\parenth{x \tp\betacirmeanmatch{m} + \beta_{\tagcirmeanmatch, 0}^\tagk{m}} \notag \\
    s_m &\defn \risk^\tagk{m}\parenth{f_{\tagcirmeanmatch}^\tagk{m}}.
  \end{align}
  Here $\eta > 0$ is a constant. Choosing $\eta$ to be $\infty$ is equivalent to choosing the source estimator with the lowest source risk.
\end{itemize}
The rational behind the use of source risks to weigh the environments follows from the corollary below.
\begin{corollary}
  \label{cor:multiple_source_anticausal_mean_shift_source_pop_risk}
  Under the data generation Assumption~\ref{ass:assumption_multiple_source_anti_causal}, the $m$-th source population risk~\eqref{eq:source_pop_risk} of CIRM$^\tagk{m}$-mean satisfies
  \begin{align}
    \label{eq:multiple_source_source_pop_risk_cirm}
    \risk^\tagk{m}\parenth{f_{\tagcirmeanmatch}^\tagk{m}} = \tilde{\risk}\parenth{f_{\tagcirmeanmatch}^\tagk{m}} + \frac{\parenth{\interAv{m}_Y - \interAtar_Y}^2}{\parenth{1 + \noisecovY^2 \sembv\tp \noisecovX^{-\frac{1}{2}} \Gcirmv{m} \noisecovX^{-\frac{1}{2}} \sembv }^2}
  \end{align}
\end{corollary}
The proof of Corollary~\ref{cor:multiple_source_anticausal_mean_shift_source_pop_risk} is provided in Appendix~\ref{sub:proof_of_cor_5}. Comparing Corollary~\ref{cor:multiple_source_anticausal_mean_shift_source_pop_risk} with Corollary~\ref{cor:single_source_anticausal_mean_shift_source_pop_risk}, the source risk of CIRM is no longer exactly equal to its target risk. The source risk has an additional term that depends on the intervention on $Y$. However, in the case where $\noisecovX$ has eigenvalues much smaller than $\noisecovY$, this additional term is negligible. In these scenarios, the source risk of CIRM still constitutes a good approximation of the target risk of CIRM. It is still possible to apply CIRM for each $m\in \braces{1, \ldots, \envs}$ and pick the source environment with the lowest source population risk in order to reduce the target population risk. Based on the above ideas, we introduce the following weighted version of CIRM.

Third we introduce the CIP and CIRM extensions to deal with mixed-causal-anticausal DA problems in Section~\ref{sub:mixed_causal_anti_causal_domain_adaptation}.
\begin{itemize}
  \item \textbf{CIP$\diamondsuit$-mean:} the population conditional invariance penalty estimator for the mixed causal anticausal DA setting.
  \begin{align}
    \label{eq:estimator_pop_cipmeanmix}
    f_{\tagcipmeanmix}(x) &\defn x \tp \bmat{\gamma^* - \Gamma^* \betacipmeanmix \\ \betacipmeanmix} + \beta_{\tagcipmeanmix, 0}^\tagk{m} \notag \\
    \betacipmeanmix, \beta_{\tagcipmeanmix, 0} &\defn \argmin_{\beta, \beta_0}\ \frac{1}{\envs} \sum_{k=1}^\envs \Exs_{(X_\tagpar, X_\tagdes, Y) \sim \distriv{k}}\parenth{Y_\tagint - {X_\tagint}\tp\beta - \beta_0}^2 \notag \\
    \text{s.t.\ } \Exs_{(X_\tagpar, X_\tagdes, Y) \sim \distriv{m}} & \brackets{{X_\tagint} \tp\beta \mid Y_\tagint = y} = \Exs_{(X_\tagpar, X_\tagdes, Y) \sim \distriv{1}} \brackets{{X_\tagint} \tp\beta \mid Y_\tagint = y}, \notag \\
    & \forall y \in \real, \forall m \in \braces{1, \ldots, \envs},
  \end{align}
  where $Y_\tagint \defn Y - X_\tagpar \tp \gamma^*$, $X_\tagint \defn X_\tagdes - {\Gamma^*} \tp X_\tagpar$,
  \begin{align}
    \label{eq:def_mix_gamma_star}
    \gamma^*, \gamma_0^* &\defn \argmin_{\gamma, \gamma_0 \in \real^{\dimscau} \times \real} \Exs_{\parenth{X_\tagpar, X_\tagdes, Y} \sim \distri^\text{allsrc}} \parenth{Y - X_\tagpar \tp \gamma - \gamma_0}^2, \notag \\
    \Gamma^*, \Gamma_0^* &\defn \argmin_{\Gamma, \Gamma_0 \in \real^{\dimscau \times (\dims-\dimscau)} \times \real^{\dims-\dimscau}} \Exs_{\parenth{X_\tagpar, X_\tagdes} \sim \distri^\text{allsrc}_X} \vecnorm{X_\tagdes - \Gamma \tp X_\tagpar - \Gamma_0}{2}^2,
  \end{align}
  and $\distri^{\text{allsrc}}$ denotes the uniform mixture of all source distribution.
  \item \textbf{CIRM$\diamondsuit^\tagk{m}$-mean:} the population conditional invariant residual matching estimator using $m$-th source environment for the mixed causal anticausal DA setting.
  \begin{align}
    \label{eq:estimator_pop_cirmeanmatchmix}
    f_{\tagcirmeanmatchmix}^\tagk{m}(x) &\defn x \tp \bmat{\gamma^* - \Gamma^* \betacirmeanmatchmix{m} \\ \betacirmeanmatchmix{m}} + \beta_{\tagcirmeanmatchmix, 0}^\tagk{m} \notag \\
    \betacirmeanmatchmix{m}, \beta_{\tagcirmeanmatchmix, 0}^\tagk{m} &\defn \argmin_{\beta, \beta_0}\  \Exs_{(X_\tagpar, X_\tagdes, Y) \sim \distriv{m}}\parenth{Y_\tagint - X_\tagint\tp\beta - \beta_0}^2 \notag \\
    &\text{s.t.\ } \Exs_{(X_\tagpar, X_\tagdes) \sim \distriv{m}_X} \brackets{\beta \tp \parenth{ X_\tagint - \parenth{X_\tagint \tp\betacipmeanmix} \bcirmeanmatchmix }} \notag \\
    &= \Exs_{(X_\tagpar, X_\tagdes) \sim \distritar_X} \brackets{\beta \tp \parenth{ X_\tagint - \parenth{X_\tagint \tp\betacipmeanmix} \bcirmeanmatchmix }},
\end{align}
  where $Y_\tagint \defn Y - X_\tagpar \tp \gamma^*$, $X_\tagint \defn X_\tagdes - {\Gamma^*} \tp X_\tagpar$ with $\gamma^*$ and $\Gamma^*$ defined in Equation~\eqref{eq:def_mix_gamma_star} and
  \begin{align}
    \label{eq:estimator_pop_cirmeanmatchmix_b_choice}
    \bcirmeanmatchmix \defn \frac{\Exs_{(X_\tagpar, X_\tagdes, Y)\sim \distri^{\text{allsrc}} }\brackets{X_\tagint \cdot (Y_\tagint - \Exs[Y_\tagint])}}{ \Exs_{(X_\tagpar, X_\tagdes, Y)\sim \distri^{\text{allsrc}} } \brackets{\parenth{X_\tagint\tp \betacipmeanmix - \Exs[X_\tagint\tp \betacipmeanmix] } \cdot \parenth{Y_\tagint-\Exs[Y_\tagint]}}},
  \end{align}
  with $\distri^{\text{allsrc}}$ denote the uniform mixture of all source distributions.
\end{itemize}

Finally, we summarize in Table~\ref{tab:summary_pop_da_methods} all the population DA methods introduced in this paper.
\begin{table}[ht]
    \centering
    % \begin{adjustwidth}{-.2in}{-.3in}
    {\renewcommand{\arraystretch}{1.}
    \begin{tabular}{c|cccc}
        \toprule
         \thead{ \bf Original} & \thead{
         \bf Made-ups \\ \bf to assist  \\ \bf explanation} & \thead{
         \bf Weighted \\ \bf variants} & \thead{\bf Matching penalty \\ \bf variants} & \thead{\bf Mixed \\ \bf variants}
        \\ \midrule \\
        DIP$^\tagk{m}$~\eqref{eq:estimator_pop_dipmeanmatchlin}
        & \shortstack{DIPAbs$^\tagk{m}$~\eqref{eq:estimator_pop_dipmeanmatchabs} \\ DIPOracle$^\tagk{m}$~\eqref{eq:estimator_pop_diporaclemeanmatchlin}}
        & DIPweigh~\eqref{eq:estimator_pop_dipweigh}
        & \shortstack{DIP-std~\eqref{eq:estimator_pop_dipstdmatchlin} \\ DIP-std+~\eqref{eq:estimator_pop_dipstdpmatchlin} \\ DIP-MMD~\eqref{eq:estimator_pop_dipMMDmatchlin}}
        & \shortstack{DIP$\diamondsuit^\tagk{m}$~\eqref{eq:estimator_pop_dipmeanmix} \\ DIP$\diamondsuit$weigh}
        \\[5mm]
        \hline \\
        CIP~\eqref{eq:estimator_pop_cipmean}
        & N/A
        & N/A
        & \shortstack{CIP-std \\ CIP-std+ \\ CIP-MMD}
        & CIP$\diamondsuit$~\eqref{eq:estimator_pop_cipmeanmix}
        \\[5mm]
        \hline \\
        CIRM$^\tagk{m}$~\eqref{eq:estimator_pop_cirmeanmatch}
        & N/A
        & CIRMweigh~\eqref{eq:estimator_pop_cirmweigh}
        & \shortstack{CIRM-std \\ CIRM-std+ \\ CIRM-MMD}
        & \shortstack{CIRM$\diamondsuit^\tagk{m}$~\eqref{eq:estimator_pop_cirmeanmatchmix} \\ CIRM$\diamondsuit$weigh}
        \\[2mm]
        \bottomrule
    \end{tabular}
    }
    % \end{adjustwidth}
    \caption{Summary of population DA methods introduced in this paper. The matching penalty variants of CIP and CIRM can be formulated similarly as those for DIP. They are omitted for the sake of space.}
    \label{tab:summary_pop_da_methods}
\end{table}

\subsection{Finite-sample formulation of DA methods}
\label{sub:finite_sample_formulation_of_DA_methods}
We introduce the finite-sample formulations of the population DA methods in the previous subsection so that they can be implemented to reproduce the results in the numerical experiments in Section~\ref{sec:numerical_experiments}. The finite-sample DA formulations are summarized in Table~\ref{tab:summary_finite_DA_methods}.

The finite-sample formulations of DIP-mean, CIP-mean and CIRM-mean are introduced in Section~\ref{sub:finite_sample_formulations_and_hyperparameter_choices}. Here we state the finite-sample formulations of weighted variants, matching penalty and mixed-causal-anticausal variants.

To formulate the finite-sample versions of DIPweigh-mean~\eqref{eq:estimator_pop_dipweigh} and CIRMweigh-mean~\eqref{eq:estimator_pop_cirmweigh}, it is sufficient to replace the corresponding population estimators and the population source risks with finite-sample ones. The mixed-causal-anticausal variants of DIP, CIP and CIRM only requires two additional regressions. They are omitted for the sake of space.

Next, we introduce the finite-sample formulations of the matching penalty variants for DIP.
\begin{itemize}
  \item \textbf{DIP$^\tagk{m}$-std-finite:} the finite-sample formulation of the DIP$^\tagk{m}$-std estimator~\eqref{eq:estimator_finite_dipstdmatchlin}. The difference between the source and target standard deviations is used as distributional distance,
  \begin{align}
    \label{eq:estimator_finite_dipstdmatchlin}
    \hat{f}_{\tagdipstdmatchlin}^\tagk{m}(x) &\defn x \tp \hat{\beta}_{\tagdipstdmatchlin}^\tagk{m} + \hat{\beta}_{\tagdipstdmatchlin, 0}^\tagk{m} \notag \\
    \hat{\beta}_{\tagdipstdmatchlin}^\tagk{m}, \hat{\beta}_{\tagdipstdmatchlin, 0}^\tagk{m} &\defn \argmin_{\beta, \beta_0}\  \frac{1}{\obs_m}\sum_{i=1}^{\obs_m} \parenth{y_i^\tagk{m} - {x_i^\tagk{m}}\tp\beta - \beta_0}^2 + \lamMatch  \parenth{\delta^\tagk{m} - \tilde{\delta}}^2,
  \end{align}
  where $\delta^\tagk{m}$ is the standard deviation of $\parenth{{x_1^\tagk{m}}\tp \beta, {x_2^\tagk{m}}\tp \beta, \ldots, {x_{\obs_m}^\tagk{m}}\tp \beta}$, and $\tilde{\delta}$ is the standard deviation of $\parenth{{\tilde{x}_1}\tp \beta, {\tilde{x}_2}\tp \beta, \ldots, {\tilde{x}_{\obstar}}\tp \beta}$.
  \item \textbf{DIP$^\tagk{m}$-std$+$-finite:} the finite-sample DIP estimator where the differences between the source and target means, standard deviations and 25\% quantiles are used as distributional distance, $\mathcal{V}$ is linear and $\mathcal{U}$ is the singleton of identity mapping. For the $m$-th source environment, it is defined as
  \begin{align}
    \label{eq:estimator_finite_dipstdpmatchlin}
    \hat{f}_{\tagdipstdpmatchlin}^\tagk{m}(x) &\defn x \tp \hat{\beta}_{\tagdipstdpmatchlin}^\tagk{m} + \hat{\beta}_{\tagdipstdpmatchlin, 0}^\tagk{m} \notag \\
    \hat{\beta}_{\tagdipstdpmatchlin}^\tagk{m}, \hat{\beta}_{\tagdipstdpmatchlin, 0}^\tagk{m} &\defn \argmin_{\beta, \beta_0}\  \frac{1}{\obs_m}\sum_{i=1}^{\obs_m} \parenth{y_i^\tagk{m} - {x_i^\tagk{m}}\tp\beta - \beta_0}^2 + \lamMatch  \parenth{\mu^\tagk{m} - \tilde{\mu}}^2  \notag \\
    & \hspace{3cm} + \lamMatch  \parenth{\delta^\tagk{m} - \tilde{\delta}}^2 + \lamMatch  \parenth{\psi_{25\%}^\tagk{m} - \tilde{\psi}_{25\%}}^2,
  \end{align}
  where $\mu^\tagk{m}, \delta^\tagk{m}, \psi_{25\%}^\tagk{m}$ are respectively the mean, standard deviation and 25\% quantile of $\parenth{{x_1^\tagk{m}}\tp \beta, {x_2^\tagk{m}}\tp \beta, \ldots, {x_{\obs_m}^\tagk{m}}\tp \beta}$, and $\tilde{\mu}, \tilde{\delta}, \tilde{\psi}_{25\%}$ are respectively the mean, standard deviation and 25\% quantile of $\parenth{{\tilde{x}_1}\tp \beta, {\tilde{x}_2}\tp \beta, \ldots, {\tilde{x}_{\obstar}}\tp \beta}$.
  \item \textbf{DIP$^\tagk{m}$-MMD-finite:} the finite-sample DIP estimator where mean squared difference is used as distributional distance, $\mathcal{V}$ is linear and $\mathcal{U}$ is the singleton of identity mapping. For the $m$-th source environment, it is defined as
  \begin{align}
    \label{eq:estimator_finite_dipMMDmatchlin}
    \hat{f}_{\tagdipMMDmatchlin}^\tagk{m}(x) &\defn x \tp \hat{\beta}_{\tagdipMMDmatchlin}^\tagk{m} + \hat{\beta}_{\tagdipMMDmatchlin, 0}^\tagk{m} \notag \\
    \hat{\beta}_{\tagdipMMDmatchlin}^\tagk{m}, \hat{\beta}_{\tagdipMMDmatchlin, 0}^\tagk{m} &\defn \argmin_{\beta, \beta_0}\  \frac{1}{\obs_m}\sum_{i=1}^{\obs_m} \parenth{y_i^\tagk{m} - {x_i^\tagk{m}}\tp\beta - \beta_0}^2 + \lamMatch \disMMD\parenth{Z^\tagk{m}, \tilde{Z}},
  \end{align}
  where $Z^\tagk{m}$ is the set of predicted responses in the $m$-th source environment $\braces{{x_1^\tagk{m}}\tp\beta, \ldots, {x_{\obs_m}}\tp\beta}$, similarly $\tilde{Z} = \braces{{\tilde{x}_1}\tp\beta, \ldots, {\tilde{x}_{\obstar}}\tp\beta}$ and the maximum mean discrepancy (MMD)~\cite{gretton2012kernel} with respect to the reproducing kernel Hilbert space (RHKS) $\RKHS$ between these two sets is
  \begin{align*}
    \disMMD\parenth{Z^\tagk{m}, \tilde{Z}} = \sup_{h \in \RKHS} \parenth{\frac{1}{\obs_m} \sum_{i=1}^{\obs_m} h({x_i^\tagk{m}}\tp \beta)  -  \frac{1}{\obstar} \sum_{i=1}^{\obstar} h(\tilde{x_i} \tp \beta) }.
  \end{align*}
  In all experiments, RKHS with Gaussian kernel is used.
\end{itemize}

\begin{table}[ht]
    \centering
    % \begin{adjustwidth}{-.2in}{-.3in}
    {\renewcommand{\arraystretch}{1.}
    \begin{tabular}{c|cccc}
        \toprule
         \thead{ \bf Original \\ (finite)} & \thead{
         \bf Weighted \\ \bf variants} & \thead{\bf Matching penalty \\ \bf variants} & \thead{\bf Mixed \\ \bf variants}
        \\ \midrule \\
        DIP-mean~\eqref{eq:estimator_finite_dipmeanmatchlin}
        & DIPweigh-mean
        & \shortstack{DIP-std~\eqref{eq:estimator_finite_dipstdmatchlin} \\ DIP-std+~\eqref{eq:estimator_finite_dipstdpmatchlin} \\ DIP-MMD~\eqref{eq:estimator_finite_dipMMDmatchlin}}
        & \shortstack{DIP$\diamondsuit^\tagk{m}$-mean \\ DIP$\diamondsuit$weigh-mean}
        \\[5mm]
        \hline \\
        CIP-mean~\eqref{eq:estimator_finite_cipmean}
        & N/A
        & \shortstack{CIP-std \\ CIP-std+ \\ CIP-MMD}
        & CIP$\diamondsuit$-mean
        \\[5mm]
        \hline \\
        CIRM-mean~\eqref{eq:estimator_finite_cirmeanmatch}
        & CIRMweigh-mean
        & \shortstack{CIRM-std \\ CIRM-std+ \\ CIRM-MMD}
        & \shortstack{CIRM$\diamondsuit^\tagk{m}$-mean \\ CIRM$\diamondsuit$weigh-mean}
        \\[2mm]
        \bottomrule
    \end{tabular}
    }
    % \end{adjustwidth}
    \caption{Summary of finite-sample formulations of the DA methods discussed in this paper. The suffixes ``-finite'' of the finite-sample DA methods are omitted in the table for brevity. The matching penalty variants of CIP and CIRM can be formulated similarly as those for DIP. The mixed-causal-anticausal variants of DIP, CIP and CIRM only requires two additional regressions. They are omitted for the sake of space.}
    \label{tab:summary_finite_DA_methods}
\end{table}

% section summary_of_da_methods_in_this_paper (end)

\section{Proofs related to Theorem~\ref{thm:single_source_anticausal_mean_shift}} % (fold)
\label{sec:proofs_related_to_theorem_thm:single_source_anticausal_mean_shift}
In this section, we prove Theorem~\ref{thm:single_source_anticausal_mean_shift} and related corollaries.

\subsection{Proof of Theorem~\ref{thm:single_source_anticausal_mean_shift}}
\label{sub:proof_of_theorem_1}

At a high level, the proof of Theorem~\ref{thm:single_source_anticausal_mean_shift} goes by connecting the target population risk of DIP$^\tagk{1}$ with that of DIPOracle$^\tagk{1}$, and then bounding the target population risk of DIPOracle$^\tagk{1}$ as a regularized version of OLSTar.

Using the linear SCM assumption in Assumption~\ref{ass:assumption_single_source_anti_causal}, each data point in the source environment is generated i.i.d. from the following equation
\begin{align}
  \label{eq:proof_source_x_sem}
  \Xv{1} &= \semBX \Xv{1} + \sembv \Yv{1} + \interAv{1}_X + \noisev{1}_X, \notag \\
  \Yv{1} &= \noisev{1}_Y.
\end{align}
Define $\invIB = \parenth{\Ind_\dims - \semBX}^{-1}$. Solving $\Xv{1}$ from Equation~\eqref{eq:proof_source_x_sem}, we have
\begin{align}
  \label{eq:proof_source_x_sem_sol}
  \Xv{1}  = \invIB \sembv \noisev{1}_Y + \invIB \interAv{1}_X + \invIB \noisev{1}_X.
\end{align}
For $(\beta, \beta_0) \in \real^{\dims} \times \real$, the residual takes the following form
\begin{align*}
  \Yv{1} - \beta\tp \Xv{1} - \beta_0 = \parenth{1 - \beta\tp \invIB \sembv} \noisev{1}_Y - \parenth{\beta\tp \invIB \interAv{1}_X + \beta_0} - \beta\tp \invIB \noisev{1}_X.
\end{align*}
Taking expectation and using the fact that noise has zero mean, we obtain
\begin{align}
  \label{eq:proof_expected_residual_source1}
  \Exs\brackets{\Yv{1} - \beta\tp \Xv{1} - \beta_0}^2 = \parenth{1 - \beta\tp \invIB \sembv}^2 \noisecovY^2 + \parenth{\beta\tp \invIB \interAv{1}_X + \beta_0}^2 + \beta \tp \invIB \noisecovX \invIB\tp \beta.
\end{align}
Similarly, we obtain the target expected residual
\begin{align}
  \label{eq:proof_expected_residual_tar}
  \Exs\brackets{\Ytar - \beta\tp \Xtar - \beta_0}^2 = \parenth{1 - \beta\tp \invIB \sembv}^2 \noisecovY^2 + \parenth{\beta\tp \invIB \interAtar_X + \beta_0}^2 + \beta \tp \invIB \noisecovX \invIB\tp \beta.
\end{align}
\paragraph{Risk of OLSTar:} Using the expression for the target residual, the OLSTar estimator in Equation~\eqref{eq:estimator_pop_olstarget} becomes the solution of the following quadratic program
\begin{align}
  \label{eq:proof_quadratic_program_causal}
  \min_{\beta, \beta_0}  \parenth{1 - \beta\tp \invIB \sembv}^2 \noisecovY^2 + \parenth{\beta\tp \invIB \interAtar_X + \beta_0}^2 + \beta \tp \invIB \noisecovX \invIB\tp \beta.
\end{align}
Solving the quadratic program and with matrix inversion lemma, we obtain
\begin{align*}
  \beta_\tagolstarget &= \noisecovY^2 \parenth{\Ind_\dims - \semBX}\tp \parenth{\noisecovX + \noisecovY^2 \sembv \sembv\tp }^{-1} \sembv \\
  &= \parenth{\Ind_\dims - \semBX}\tp \frac{\noisecovY^2 \noisecovX^{-1}\sembv}{1 + \noisecovY^2 \sembv\tp \noisecovX^{-1} \sembv } \\
  \beta_{\tagolstarget, 0} &= -\beta_\tagolstarget\tp \invIB \interAtar_X.
\end{align*}
The target population loss of OLSTar is
\begin{align}
  \label{eq:proof_olstarget_risk}
  \tilde{\risk}\parenth{f_{\tagolstarget}} &= \noisecovY^2 - \noisecovY^4 \sembv \tp \parenth{\noisecovX + \noisecovY^2 \sembv \sembv \tp}^{-1} \sembv \notag \\
  & = \frac{\noisecovY^2}{1 + \noisecovY^2 \sembv\tp \noisecovX^{-1} \sembv },
\end{align}
where the last equality follows from matrix inversion lemma.
\paragraph{Risk of OLSSrc$^\tagk{1}$:} The minimization problem of OLSSrc can be similarly solved by changing the target variables to source ones in Equation~\eqref{eq:proof_quadratic_program_causal}. We obtain
\begin{align*}
  \beta_\tagolssource^\tagk{1} &= \noisecovY^2 \parenth{\Ind_\dims - \semBX}\tp \parenth{\noisecovX + \noisecovY^2 \sembv \sembv\tp }^{-1} \sembv \\
  &= \parenth{\Ind_\dims - \semBX}\tp \frac{\noisecovY^2 \noisecovX^{-1}\sembv}{1 + \noisecovY^2 \sembv\tp \noisecovX^{-1} \sembv } \\
  \beta_{\tagolssource, 0}^\tagk{1} &= -{\beta_\tagolssource^\tagk{1} }\tp \invIB \interAv{1}_X.
\end{align*}
Because of the difference in the intercept term, the target population risk of OLSSrc has one additional term
\begin{align}
  \label{eq:proof_olssrc_risk}
  \tilde{\risk}\parenth{f_{\tagolssource}^\tagk{1}} = \frac{\noisecovY^2}{1 + \noisecovY^2 \sembv\tp \noisecovX^{-1} \sembv } + \frac{\parenth{\noisecovY^2 \sembv \tp \noisecovX^{-1} \parenth{\interAv{1}_X - \interAtar_X}}^2}{\parenth{1 + \noisecovY^2 \sembv \tp \noisecovX^{-1} \sembv}^2}.
\end{align}

\paragraph{Risk of DIP$^\tagk{1}$:} Using Equation~\eqref{eq:proof_source_x_sem_sol}, for any $\beta \in \real^\dims$, we have
\begin{align*}
  \beta\tp \Xv{1} = \beta\tp \invIB \sembv + \beta\tp \invIB \interAv{1}_X + \beta \tp \invIB \noisev{1}_X.
\end{align*}
Together with the similar equation on target, we obtain
\begin{align}
  \label{eq:proof_expectation_beta_x}
  \Exs\brackets{\beta\tp \Xv{1}}  = \Exs\brackets{\beta\tp \Xtar} + \beta\tp \invIB \parenth{\interAv{1}_X - \interAtar_X}.
\end{align}
Combining Equation~\eqref{eq:proof_expected_residual_source1} and~\eqref{eq:proof_expectation_beta_x}, we observe that the DIP$^\tagk{1}$ problem~\eqref{eq:estimator_pop_dipmeanmatchlin} becomes a constrained quadratic program
\begin{align}
  \label{eq:proof_quadratic_program_dipmeanmatchlin}
  &\min_{\beta, \beta_0}  \parenth{1 - \beta\tp \invIB \sembv}^2 \noisecovY^2 + \parenth{\beta\tp \invIB \interAv{1}_X + \beta_0}^2 + \beta \tp \invIB \noisecovX \invIB\tp \beta \notag \\
  &\text{s.t. } \beta \tp \invIB \parenth{\interAv{1}_X - \interAtar} = 0.
\end{align}
Note that because of the constraint, this quadratic program is the same for the source data and for the target data. As a consequence, we have
\begin{align*}
  \tilde{\risk}\parenth{f_{\tagdipmeanmatchlin}^\tagk{1}} = \tilde{\risk}\parenth{f_{\tagdiporaclemeanmatchlin}^\tagk{1}}.
\end{align*}
Now we solve the constrained quadratic program in Equation~\eqref{eq:proof_quadratic_program_dipmeanmatchlin}. First, we observe that the minimization on $\beta_0$ can be easily solved. Second, since $\invIB$ is invertible, we can re-parametrize the original problem and obtain the following problem
\begin{align}
  \label{eq:proof_quadratic_program_dipmeanmatchlin_reparametrized}
  &\min_{\gamma}  \parenth{1 - \gamma\tp \sembv}^2 \noisecovY^2 + \gamma\tp \noisecovX \gamma \notag \\
  &\text{s.t. } \gamma\tp \parenth{\interAv{1}_X - \interAtar} = 0.
\end{align}
Define $u = \frac{\interAv{1} - \interAtar}{\vecnorm{\interAv{1} - \interAtar}{2}}$. Using Gram-Schmidt orthogonalization, we can complete the vector $u$ to form an orthonormal basis $\parenth{u, q_1, \ldots, q_{\dims-1}}$. Let $\Qdip \in \real^{\dims \times (\dims-1)}$ be the matrix formed with $i$-th column being $q_i$. Then the mapping
\begin{align*}
  \real^{\dims-1} &\rightarrow \real^\dims \\
  \zeta &\mapsto \Qdip \zeta
\end{align*}
constitute a bijection between $\real^{\dims-1}$ and the set $\braces{\gamma \in \real^\dims \mid \gamma \tp u = 0}$. This bijection allows us to transform the constrained quadratic program~\eqref{eq:proof_quadratic_program_dipmeanmatchlin_reparametrized} to the following unconstrained one
\begin{align*}
  \min_{\zeta \in \real^{\dims^{-1}}} \parenth{1 - \zeta\tp \Qdip\tp \sembv}^2 \noisecovY^2 + \zeta \tp \Qdip \tp \noisecovX \Qdip \zeta.
\end{align*}
Solving the quadratic program by setting gradient to zero, we obtain the minimizer
\begin{align*}
  \zeta^* = \noisecovY^2 \parenth{\Qdip \tp \parenth{\noisecovY^2 \sembv \sembv\tp + \noisecovX}\Qdip }^{-1} \Qdip\tp \sembv
\end{align*}
and
\begin{align*}
  \betadipmeanmatchlin{1} &= \noisecovY^2 \parenth{\Ind_\dims - \semBX}\tp \Qdip \parenth{\Qdip \tp (\noisecovX + \noisecovY^2 \sembv \sembv\tp) \Qdip }^{-1} \Qdip \tp \sembv  \\
  &= \parenth{\Ind_\dims - \semBX}\tp \frac{\noisecovY^2 \Qdip \parenth{\Qdip \tp \noisecovX \Qdip}^{-1} \Qdip\tp \sembv }{1 + \noisecovY^2 \sembv \tp \Qdip \parenth{\Qdip \tp \noisecovX \Qdip}^{-1} \Qdip \tp \sembv} \\
  \beta_{\tagdipmeanmatchlin, 0}^\tagk{1} &= -{\betadipmeanmatchlin{1}} \tp \invIB \interAv{1}_X,\\
  \betadiporaclemeanmatchlin{1} &= \betadipmeanmatchlin{1}  \\
  \beta_{\tagdiporaclemeanmatchlin, 0}^\tagk{1} &= -{\betadiporaclemeanmatchlin{1}} \tp \invIB \interAtar_X.
\end{align*}
Consequently, the target population can be obtained by replacing $\sembv$ with $\Qdip \tp \sembv$ and $\noisecovX$ with $\Qdip \tp \noisecovX $ in Equation~\eqref{eq:proof_olstarget_risk}
\begin{align*}
  \tilde{\risk}\parenth{f_{\tagdipmeanmatchlin\tagk{1}}} = \tilde{\risk}\parenth{f_{\tagdiporaclemeanmatchlin\tagk{1}}} = \frac{\noisecovY^2}{1 + \noisecovY^2 \sembv\tp \noisecovX^{-\frac{1}{2}} \Gdip \noisecovX^{-\frac{1}{2}} \sembv }.
\end{align*}
where $\Gdip = \noisecovX^{1/2} \Qdip \parenth{\Qdip \tp \noisecovX \Qdip}^{-1} \Qdip \tp \noisecovX^{1/2}$ is a projection matrix with rank $\dims-1$.

\subsection{Proof of Corollary~\ref{cor:single_source_anticausal_mean_shift}}
\label{sub:proof_of_cor_1}
Equation~\eqref{eq:single_source_pop_risk_olsoracle_gaussian},~\eqref{eq:single_source_pop_risk_olssrc_gaussian} and~\eqref{eq:single_source_pop_risk_dipmeanmatchlin_gaussian_eq} follow directly by plugging in $\noisecovX = \frac{\noisecovY^2}{\rho}\Ind_\dims$ in the corresponding equations in Theorem~\ref{thm:single_source_anticausal_mean_shift}. For the high probability bound, we use several tail inequalities. Let $a= \interAv{1}_X - \interAtar_X$. Since $a$ is generated randomly from $\Normal(0, \tau^2 \Ind_\dims)$, then for a fixed vector $v \in \real^\dims$, $a\tp v$ follows $\Normal(0, \tau^2 v\tp v)$. The standard Gaussian tail bound on $a\tp v$ gives, for $t > 0$,
\begin{align}
  \label{eq:proof_cor_std_gaussian_tail}
  \Prob\parenth{ \abss{\frac{a\tp v}{\tau \parenth{v\tp v}^{1/2}}} \geq t} \leq 2 \exp\parenth{-t^2/2}.
\end{align}

Since $\vecnorm{a}{2}^2$ follows Chi-square distribution with $\dims$-degree of freedom, the standard chi-square tail bound (see e.g.~\cite{laurent2000adaptive}) gives, for $t > 0$,
\begin{align}
  \label{eq:proof_cor_std_chi_sq_tail}
  \Prob\parenth{\frac{\vecnorm{a}{2}^2}{\tau^2 \dims} \leq 1 - \frac{t}{\sqrt{\dims}}} \leq \exp\parenth{-t^2/8}.
\end{align}
Combining Equation~\eqref{eq:proof_cor_std_gaussian_tail} and~\eqref{eq:proof_cor_std_chi_sq_tail}, with probability at least $1 - \exp(-t^2/8) - 2 \exp(-t^2/2)$, we have
\begin{align*}
  \frac{\parenth{a\tp v}^2}{\vecnorm{a}{2}^2} \leq \frac{t^2}{\dims - \sqrt{\dims} t} \parenth{v\tp v}.
\end{align*}
For $t$ constant satisfying $0 < t \leq \frac{\sqrt{\dims}}{2}$, we have
\begin{align*}
  \frac{t^2}{\dims - \sqrt{\dims} t} \leq \frac{2t^2}{\dims}.
\end{align*}
Plugging the above high probability bound into Equation~\eqref{eq:single_source_pop_risk_olssrc_gaussian} and~\eqref{eq:single_source_pop_risk_dipmeanmatchlin_gaussian_eq} with $v = \sembv$, with probability at least $1 - \exp(-t^2/8) - 2 \exp(-t^2/2)$, we have
\begin{align*}
  \tilde{\risk}\parenth{f_{\tagolssource}^\tagk{1}} &\leq \frac{\noisecovY^2}{1 + \rho \vecnorm{\sembv}{2}^2} + \frac{\tau^2 t^2 \vecnorm{\sembv}{2}^2}{\parenth{1 + \rho \vecnorm{\sembv}{2}^2}^2}, \text{ and } \\
  \tilde{\risk}\parenth{f_{\tagdipmeanmatchlin}^\tagk{1}} &\leq \frac{\noisecovY^2}{1 + \rho \parenth{1 - \frac{2t^2}{\dims}}\vecnorm{\sembv}{2}^2}.
\end{align*}
We conclude and obtain the form needed in the corollary by a change of variable from $2t^2$ to $t$.

\subsection{Proof of Corollary~\ref{cor:single_source_anticausal_mean_shift_source_pop_risk}} % (fold)
\label{sub:proof_of_cor_2}
Corollary~\ref{cor:single_source_anticausal_mean_shift_source_pop_risk} shows that the source population risk of DIP$^\tagk{1}$ is the same as target population risk of DIP$^\tagk{1}$. For this, it suffices to observe that the source expected residual~\eqref{eq:proof_expected_residual_source1} and the target expected residual~\eqref{eq:proof_expected_residual_tar} only differ by the term $\beta\tp \invIB \interAv{1}_X$ and the term $\beta\tp \invIB \interAtar_X$. Since the DIP constraint enforces $\beta\tp \invIB \parenth{\interAv{1}_X - \interAtar_X} = 0$ as shown in Equation~\eqref{eq:proof_quadratic_program_dipmeanmatchlin}, we obtain that
\begin{align}
  \risk^\tagk{1}\parenth{f_{\tagdipmeanmatchlin}^\tagk{1}} = \tilde{\risk}\parenth{f_{\tagdipmeanmatchlin}^\tagk{1}}.
\end{align}

% subsection proof_of_corollary_cor:single_source_anticausal_mean_shift_source_pop_risk (end)

\subsection{Proof of Corollary~\ref{cor:single_source_anticausal_mean_shift_intervention_on_y}}
\label{sub:proof_of_cor_3}

Using the linear SEM assumption in Assumption~\ref{ass:assumption_single_source_anti_causal} and the additional intervention on $Y$, each data point in the source environment is generated i.i.d. from the following equation
\begin{align}
  \label{eq:proof_source_x_sem_prop1}
  \Xv{1} &= \semBX \Xv{1} + \sembv \Yv{1} + \interAv{1}_X + \noisev{1}_X, \notag \\
  \Yv{1} &= \interAv{1}_Y + \noisev{1}_Y.
\end{align}
For $(\beta, \beta_0) \in \real^{\dims} \times \real$, the source residual takes the following form
\begin{align*}
  \Yv{1} - \beta\tp \Xv{1} - \beta_0 = \parenth{1 - \beta\tp \invIB \sembv} \noisev{1}_Y + \parenth{\parenth{1 - \beta\tp \invIB \sembv} \interAv{1}_Y - \beta\tp \invIB \interAv{1}_X - \beta_0} - \beta\tp \invIB \noisev{1}_X.
\end{align*}

The DIP$^\tagk{1}$ problem~\eqref{eq:estimator_pop_dipmeanmatchlin} becomes a constrained quadratic program
\begin{align}
  \label{eq:proof_quadratic_program_dipmeanmatchlin_prop1}
  &\min_{\beta, \beta_0}  \parenth{1 - \beta\tp \invIB \sembv}^2 \noisecovY^2 + \parenth{\parenth{1 - \beta\tp \invIB \sembv} \interAv{1}_Y - \beta\tp \invIB \interAv{1}_X - \beta_0}^2 + \beta \tp \invIB \noisecovX \invIB\tp \beta \notag \\
  &\text{s.t. } \beta \tp \invIB \parenth{\interAv{1}_X + \interAv{1}_Y \sembv} = \beta \tp \invIB \parenth{\interAtar_X + \interAtar_Y \sembv}.
\end{align}
Note that because of the intervention on $Y$, unlike in Theorem~\ref{thm:single_source_anticausal_mean_shift}, this quadratic program is no longer the same for DIP$^\tagk{1}$ and DIPOracle$^\tagk{1}$. However, the constrained quadratic program~\eqref{eq:proof_quadratic_program_dipmeanmatchlin_prop1} can be solved similarly as we did in Appendix~\ref{sub:proof_of_theorem_1} around Equation~\eqref{eq:proof_quadratic_program_dipmeanmatchlin} by introducing
\begin{align*}
  u = \frac{\interAv{1}_X + \interAv{1}_Y \sembv - \interAtar_X - \interAtar_Y \sembv}{ \vecnorm{\interAv{1}_X + \interAv{1}_Y \sembv - \interAtar_X - \interAtar_Y \sembv}{2}}
\end{align*}
and $\QdipY \in \real^{\dims \times \dims-1}$ is the matrix with columns formed by the vectors that complete the vector $u$ to an orthonormal basis of $\real^\dims$. Following the rest of the proof in Appendix~\ref{sub:proof_of_theorem_1}, we obtain
\begin{align}
  \label{eq:proof_betadip_Yshift}
   \betadipmeanmatchlin{1}
  &= \parenth{\Ind_\dims - \semBX}\tp \frac{\noisecovY^2 \QdipY \parenth{\QdipY \tp \noisecovX \QdipY}^{-1} \QdipY\tp \sembv }{1 + \sembv \tp \QdipY \parenth{\QdipY \tp \noisecovX \QdipY}^{-1} \QdipY \tp \sembv} \notag \\
  \beta_{\tagdipmeanmatchlin, 0}^\tagk{1} &= \parenth{1 - {\betadipmeanmatchlin{1}} \tp \invIB \sembv } \interAv{1}_Y -{\betadipmeanmatchlin{1}} \tp \invIB \interAv{1}_X.
\end{align}
Note that the intercept $\beta_{\tagdipmeanmatchlin, 0}^\tagk{1}$ has an extra term due to the intervention on $\interAv{1}_Y$. For $(\beta, \beta_0) \in \real^{\dims} \times \real$, the target residual takes the following form
\begin{align*}
  \Ytar - \beta\tp \Xtar - \beta_0 = \parenth{1 - \beta\tp \invIB \sembv} \noisetar_Y + \parenth{\parenth{1 - \beta\tp \invIB \sembv} \interAtar_Y - \beta\tp \invIB \interAtar_X - \beta_0} - \beta\tp \invIB \noisetar_X.
\end{align*}
Plugging $(\betadipmeanmatchlin{1}, \beta_{\tagdipmeanmatchlin, 0}^\tagk{1})$ of Equation~\eqref{eq:proof_betadip_Yshift} into the above residual, we obtain that the population target risk which has an extra term that depends on $\interAv{1}_Y - \interAtar_Y$
\begin{align*}
  \tilde{\risk}\parenth{f_{\tagdipmeanmatchlin\tagk{1}}} = \frac{\noisecovY^2}{1 + \noisecovY^2 \sembv\tp \noisecovX^{-\frac{1}{2}} \GdipY \noisecovX^{-\frac{1}{2}} \sembv } + \parenth{\interAv{1}_Y - \interAtar_Y}^2,
\end{align*}
where $\GdipY = \noisecovX^{1/2} \QdipY \parenth{\QdipY  \tp \noisecovX \QdipY}^{-1} \QdipY\tp \noisecovX^{1/2}$ is a projection matrix with rank $\dims-1$.

% section proofs_related_to_theorem_thm:single_source_anticausal_mean_shift (end)

\section{Proofs related to Theorem~\ref{thm:mutiple_source_anticausal_mean_shift}}
\label{sec:proofs_related_to_theorem_thm:mutiple_source_anticausal_mean_shift}
In this section, we prove Theorem~\ref{thm:mutiple_source_anticausal_mean_shift} and related corollaries.
\subsection{Proof of Theorem~\ref{thm:mutiple_source_anticausal_mean_shift}}
\label{sub:proof_of_theorem_2}
Using the linear SEM assumption in Assumption~\ref{ass:assumption_multiple_source_anti_causal}, each data point in the $m$-th source environment is generated i.i.d. from the following equation
\begin{align}
  \label{eq:proof2_source_x_sem}
  \Xv{m} &= \semBX \Xv{m} + \sembv \Yv{m} + \interAv{m}_X + \noisev{m}_X, \notag \\
  \Yv{m} &= \interAv{m}_Y + \noisev{m}_Y.
\end{align}
Define $H = \parenth{\Ind_\dims - \semBX}^{-1}$. For $(\beta, \beta_0) \in \real^{\dims} \times \real$, the residual takes the following form
\begin{align}
  \label{eq:proof2_residual_sem}
  \Yv{m} - \beta\tp \Xv{m} - \beta_0 = \parenth{1 - \beta\tp \invIB \sembv} \parenth{\interAv{m}_Y + \noisev{m}_Y} - \parenth{\beta\tp \invIB \interAv{m}_X + \beta_0} - \beta\tp \invIB \noisev{m}_X.
\end{align}
The target residual has a similar form
\begin{align}
  \label{eq:proof2_target_residual_sem}
  \Ytar - \beta\tp \Xtar - \beta_0 = \parenth{1 - \beta\tp \invIB \sembv} \parenth{\interAtar_Y + \noisetar_Y} - \parenth{\beta\tp \invIB \interAtar_X + \beta_0} - \beta\tp \invIB \noisetar_X.
\end{align}
\paragraph{Risk of OLSTar: } Using the expression for the target residual, the OLSTar estimator in Equation~\eqref{eq:estimator_pop_olstarget} becomes the solution of the following quadratic program
\begin{align*}
  \min_{\beta, \beta_0} &\parenth{1 - \beta\tp \invIB \sembv}^2 \noisecovY^2 + \frac{1}{\envs} \sum_{m=1}^\envs \parenth{ \parenth{1 - \beta\tp \invIB \sembv} \interAtar_Y - \beta\tp\invIB \interAtar_X - \beta_0}^2 + \beta\tp\invIB \noisecovX \invIB\tp \beta.
\end{align*}
Solving the quadratic program by setting the gradient to zero, we obtain
\begin{align*}
    \beta_\tagolstarget &= \noisecovY^2 \parenth{\Ind_\dims - \semBX}\tp \parenth{\noisecovX + \noisecovY^2 \sembv \sembv\tp }^{-1} \sembv  \\
  \beta_{\tagolstarget, 0} &= \parenth{1 - \beta_\tagolstarget\tp \invIB \sembv} \interAtar_Y - \beta_\tagolstarget\tp \invIB \interAtar_X.
\end{align*}
Despite the difference in the intercept term, the corresponding target risk is the same as in Theorem~\ref{thm:single_source_anticausal_mean_shift}
\begin{align*}
  \tilde{\risk}\parenth{f_{\tagolstarget}} = \frac{\noisecovY^2}{1 + \noisecovY^2 \sembv\tp \noisecovX^{-1} \sembv }.
\end{align*}

\paragraph{Risk of CIP: } Using the SEM~\eqref{eq:proof2_source_x_sem}, the constraint of CIP in Equation~\eqref{eq:estimator_pop_cipmean} becomes
\begin{align*}
  \beta \tp \invIB \interAv{m}_X = \beta \tp \invIB \interAv{1}_X,\quad \forall m \in \braces{2, \ldots, \envs}.
\end{align*}
Together with the residual expression in Equation~\eqref{eq:proof2_residual_sem}, the CIP objective~\eqref{eq:estimator_pop_cipmean} is equivalent to
\begin{align}
  \label{eq:proof2_CIP_simplified_form}
  \min_{\beta, \beta_0} &\parenth{1 - \beta\tp \invIB \sembv}^2 \noisecovY^2 + \frac{1}{\envs} \sum_{m=1}^\envs \parenth{ \parenth{1 - \beta\tp \invIB \sembv} \interAv{m}_Y - \beta\tp\invIB \interAv{m}_X - \beta_0}^2 + \beta\tp\invIB \noisecovX \invIB\tp \beta \notag \\
  \text{s.t. } & \beta\tp \invIB \parenth{\interAv{m}_X - \interAv{1}_X} = 0, \quad \forall m \in \braces{2, \ldots, \envs}.
\end{align}
First, the one-dimensional quadratic program on $\beta_0$ can be solved easily by setting derivative to zero and we obtain
\begin{align*}
  \beta_0^* = \frac{1}{\envs} \sum_{m=1}^{\envs} \parenth{1 - {\beta^*}\tp \invIB \sembv } \interAv{m}_Y - {\beta^*}\tp \invIB \interAv{m}_X.
\end{align*}
Plugging the expression of $\beta_0$ back to Equation~\eqref{eq:proof2_CIP_simplified_form}, the minimization on $\beta$ becomes
\begin{align*}
  \min_{\beta} &\parenth{1 - \beta\tp \invIB \sembv}^2 \parenth{\noisecovY^2 + \Delta_Y} + \beta\tp\invIB \noisecovX \invIB\tp \beta \notag \\
  \text{s.t. } & \beta\tp \invIB \parenth{\interAv{m}_X - \interAv{1}_X} = 0, \ \forall m \in \braces{2, \ldots, \envs},
\end{align*}
where
\begin{align}
  \label{eq:proof2_delta_y_def}
  \Delta_Y = \frac{1}{\envs} \sum_{m=1}^{\envs} \parenth{\interAv{m}_Y - \bar{\interA}_Y}^2 \text{ and } \bar{\interA}_Y = \frac{1}{\envs} \sum_{k=1}^\envs \interAv{k}_Y.
\end{align}
Second, since $\invIB$ is invertible, we can re-parametrize the optimization on $\beta$ as follows
\begin{align}
  \label{eq:proof2_quadratic_program_cip_reparametrized_gamma}
  \min_{\gamma \in \real^\dims} & \parenth{1 - \gamma\tp \sembv}^2 \parenth{\noisecovY^2 + \Delta_Y)} + \gamma\tp \noisecovX \gamma \notag \\
  \text{s.t. } & \gamma \tp \parenth{\interAv{m}_X - \interAv{1}_X} = 0, \ \forall m \in \braces{2, \ldots, \envs}.
\end{align}
Let $P \in \real^{\dims \times p}$ be the matrix formed with an orthonormal basis of the $p$-dimensional subspace $\text{span}\parenth{\interAv{2}-\interAv{1}, \ldots, \interAv{m}-\interAv{1}}$.
Let $\Qcip \in \real^{\dims \times (\dims - p)}$ be the matrix with columns formed by completing the columns of $P$ to a basis of $\real^\dims$ via Gram-Schmidt orthogonalization. The following mapping
\begin{align*}
  \real^{\dims - p} &\rightarrow \real^\dims \\
  \zeta &\mapsto \Qcip \zeta
\end{align*}
constitutes a bijection between $\real^{\dims - p}$ and the set $\braces{\gamma \in \real^{\dims} \mid P \tp \gamma = 0}$. With the change of variable, the constrained optimization in Equation~\eqref{eq:proof2_quadratic_program_cip_reparametrized_gamma} is equivalent to the unconstrained one
\begin{align*}
  \min_{\zeta \in \real^{\dims-p}} \parenth{1 - \zeta \tp \Qcip \tp \sembv}^2 \parenth{\noisecovY^2 + \Delta_Y} + \zeta \tp \Qcip \tp \noisecovX \Qcip \zeta.
\end{align*}
Solving the unconstrained quadratic program by setting gradient to zero, we obtain the minimizer
\begin{align*}
  \zeta^* &= \parenth{\noisecovY^2 + \Delta_Y} \parenth{\Qcip \tp \parenth{\parenth{\noisecovY^2 + \Delta_Y} \sembv \sembv\tp + \noisecovX}\Qcip }^{-1} \Qcip\tp \sembv\\
  &= \parenth{\noisecovY^2 + \Delta_Y} \frac{\Qcip \parenth{{\Qcip}\tp \noisecovX \Qcip }^{-1} \Qcip \tp \sembv}{1 + \parenth{\noisecovY^2 + \Delta_Y} \sembv \tp \Qcip \parenth{{\Qcip}\tp \noisecovX \Qcip }^{-1} \Qcip \tp \sembv},
\end{align*}
where the last equality uses the matrix inversion lemma.
Finally, transforming the variables back, the CIP estimator is
\begin{align}
  \label{eq:proof2_cipmean_beta_sol}
  \betacipmean &= \parenth{\noisecovY^2 + \Delta_Y} \frac{\parenth{\Ind_\dims- \semBX}\tp \Qcip \parenth{{\Qcip}\tp \noisecovX \Qcip }^{-1} \Qcip \tp \sembv}{1 + \parenth{\noisecovY^2 + \Delta_Y} \sembv \tp \Qcip \parenth{{\Qcip}\tp \noisecovX \Qcip }^{-1} \Qcip \tp \sembv} \\
  \beta_{\tagcipmean, 0} &= \parenth{1 - {\betacipmean} \tp \invIB \sembv} \bar{\interA}_Y - {\betacipmean} \tp \invIB \interAv{m}_X \notag.
\end{align}
According to the form of the target residual~\eqref{eq:proof2_target_residual_sem}, the target population risk for $(\beta, \beta_0)$ takes the following form
\begin{align*}
  \parenth{1 - \beta\tp \invIB \sembv}^2 \noisecovY^2 + \parenth{ \parenth{1 - \beta\tp \invIB \sembv} \interAtar_Y - \beta\tp\invIB \interAtar_X - \beta_0}^2 + \beta\tp\invIB \noisecovX \invIB\tp \beta.
\end{align*}
Plugging in the CIP estimator into the equation above, we obtain the target population risk of CIP
\begin{align*}
  \tilde{\risk}\parenth{f_{\tagcipmean}} = \frac{\noisecovY^2 + \Delta_Y }{1 + \parenth{\noisecovY^2 + \Delta_Y} \sembv\tp \noisecovX^{-\frac{1}{2}} \Gcip \noisecovX^{-\frac{1}{2}} \sembv } + \frac{ \parenth{\interAtar_Y - \bar{\interA}_Y}^2- \Delta_Y }{\parenth{1 + \parenth{\noisecovY^2 + \Delta_Y} \sembv\tp \noisecovX^{-\frac{1}{2}} \Gcip \noisecovX^{-\frac{1}{2}}\sembv }^2},
\end{align*}
where $\Gcip = \noisecovX^{1/2} \Qcip \parenth{\Qcip \tp \noisecovX \Qcip}^{-1} \Qcip \tp \noisecovX^{1/2}$ is a projection matrix with rank $\dims-p$.

\paragraph{Risk of CIRM: } First we derive $\bcirmeanmatch$ in Equation~\eqref{eq:estimator_pop_cirmeanmatch_b_choice}. From the CIP expression in Equation~\eqref{eq:proof2_cipmean_beta_sol} and the SEM~\eqref{eq:proof2_source_x_sem}, we obtain the following expression for ${\Xv{m}}\tp \betacipmean$
\begin{align*}
  \frac{\parenth{\noisecovY^2 + \Delta_Y} \parenth{  \noisev{m}_Y \sembv\tp \noisecovX^{-1/2} \Gcip \noisecovX^{-1/2} \sembv + {\noisev{m}_X} \tp \noisecovX^{-1/2} \Gcip \noisecovX^{-1/2} \sembv + {\interAv{m}_X}\tp \noisecovX^{-1/2} \Gcip \noisecovX^{-1/2} \sembv} }{ 1+ \parenth{\noisecovY^2 + \Delta_Y}  \sembv\tp \noisecovX^{-1/2} \Gcip \noisecovX^{-1/2} \sembv}.
\end{align*}
The deviation from its expectation is
\begin{align*}
  &{\Xv{m}}\tp \betacipmean - \Exs\brackets{{\Xv{m}}\tp \betacipmean } \\
  &= \frac{\parenth{\noisecovY^2 + \Delta_Y} \sembv\tp \noisecovX^{-1/2} \Gcip \noisecovX^{-1/2} \sembv}{ 1+ \parenth{\noisecovY^2 + \Delta_Y}  \sembv\tp \noisecovX^{-1/2} \Gcip \noisecovX^{-1/2} \sembv} \cdot \noisev{m}_Y + \frac{\parenth{\noisecovY^2 + \Delta_Y} {\noisev{m}_X}\tp \noisecovX^{-1/2} \Gcip \noisecovX^{-1/2} \sembv}{ 1+ \parenth{\noisecovY^2 + \Delta_Y}  \sembv\tp \noisecovX^{-1/2} \Gcip \noisecovX^{-1/2} \sembv}.
\end{align*}
Plugging the above equation into Equation~\eqref{eq:estimator_pop_cirmeanmatch_b_choice}, together with
\begin{align}
  \label{proof2_sem_X_expr_cirm}
  \Xv{m} = \invIB \sembv \Yv{m} + \invIB \interAv{m}_X + \invIB \noisev{m}_X
\end{align}
and $\Yv{m} - \Exs\brackets{\Yv{m}} = \noisev{m}_Y$, we obtain
\begin{align}
  \label{proof2_cirm_b_sol}
  \bcirmeanmatch = \parenth{\frac{1 + \parenth{\noisecovY^2 + \Delta_Y} \sembv\tp \noisecovX^{-1/2} \Gcip \noisecovX^{-1/2} \sembv }{\parenth{\noisecovY^2 + \Delta_Y} \sembv\tp \noisecovX^{-1/2} \Gcip \noisecovX^{-1/2} \sembv}} \invIB \sembv.
\end{align}
Note that the vector $\bcirmeanmatch$ is co-linear with the $Y$ component in Equation~\eqref{proof2_sem_X_expr_cirm}. We remark that the idea of CIRM is to use ${\Xtar}\tp \betacipmean$ as a proxy for the unobserved $\Ytar$ to remove the $\Ytar$ part in the covariates so that the DIP matching based ideas still can be applied.

With the $\bcirmeanmatch$ expression~\eqref{proof2_cirm_b_sol} and the $\betacipmean$ expression~\eqref{eq:proof2_cipmean_beta_sol}, the LHS of the CIRM constraint~\eqref{eq:estimator_pop_cirmeanmatch} becomes
\begin{align}
  \label{eq:proof2_cirm_corrected_residual}
  &\Xv{m} - \parenth{{\Xv{m}}\tp \betacipmean} \bcirmeanmatch \notag \\
  &= \parenth{\Yv{m} - \parenth{{\Xv{m}}\tp \betacipmean } \cdot \frac{1 + \parenth{\noisecovY^2 + \Delta_Y} \sembv\tp \noisecovX^{-1/2} \Gcip \noisecovX^{-1/2} \sembv}{\parenth{\noisecovY^2 + \Delta_Y)}  \sembv\tp \noisecovX^{-1/2} \Gcip \noisecovX^{-1/2} \sembv}} \invIB \sembv  + \invIB \interAv{m}_X + \invIB \noisev{m}_X \notag \\
   & = \parenth{\betacipmean \tp \invIB \interAv{m}_X + \betacipmean \tp \invIB \noisev{m}_X} \invIB \sembv  +  \invIB \interAv{m}_X + \invIB \noisev{m}_X \notag \\
   & \stackrel{(i)}{=} \parenth{\betacipmean \tp \invIB \interAv{1}_X + \betacipmean \tp \invIB \noisev{m}_X} \invIB \sembv  +  \invIB \interAv{m}_X + \invIB \noisev{m}_X,
\end{align}
where the last inequality (i) follows from the fact that the constraint in CIP~\eqref{eq:estimator_pop_cipmean} forces
\begin{align*}
  {\betacipmean}\tp \invIB \interAv{m}_X = {\betacipmean}\tp \invIB \interAv{1}_X, \quad \forall m \in \braces{1, \ldots, \envs}.
\end{align*}
An expression similar to Equation~\eqref{eq:proof2_cirm_corrected_residual} can be obtained for the target environments by taking into account that $\interAtar_X \in \text{span}\parenth{\interAv{1}_X, \ldots, \interAv{\envs}_X }$,
\begin{align}
  \label{eq:proof2_cirm_corrected_residual_tar}
  \Xtar - \parenth{{\Xtar}\tp \betacipmean} \bcirmeanmatch = \invIB \sembv \parenth{ \betacipmean \tp \invIB \interAv{1}_X + \betacipmean \tp \invIB \noisetar_X} + \invIB \interAtar_X + \invIB \noisetar_X.
\end{align}
Multiply Equation~\eqref{eq:proof2_cirm_corrected_residual} and~\eqref{eq:proof2_cirm_corrected_residual_tar} by $\beta$ and take expectation, we obtain a simplified form of the CIRM constraint~\eqref{eq:estimator_pop_cirmeanmatch} as follows
\begin{align*}
  \beta \tp \invIB \interAv{m}_X =  \beta \tp \invIB \interAtar_X.
\end{align*}
Note that the CIRM constraint is effectively matching the covariate interventions as DIP constraint did when $Y$ is not intervened on. As a consequence, given $\betacipmean$ and $\bcirmeanmatch$, the CIRM$^\tagk{m}$ estimator~\eqref{eq:estimator_pop_cirmeanmatch} is very similar to DIP$^\tagk{m}$~\eqref{eq:proof_quadratic_program_dipmeanmatchlin} except that the intervention on $Y$ still appears in the residual. The CIRM$^\tagk{m}$ estimator~\eqref{eq:estimator_pop_cirmeanmatch} can be written as follows
\begin{align}
  \label{eq:proof2_quadratic_program_cirmeanmatch_final}
  &\min_{\beta, \beta_0}  \parenth{1 - \beta\tp \invIB \sembv}^2 \noisecovY^2 + \parenth{\parenth{1 - \beta\tp \invIB \sembv} \interAv{m}_Y - \beta\tp \invIB \interAv{m}_X - \beta_0}^2 + \beta \tp \invIB \noisecovX \invIB\tp \beta \notag \\
  &\text{s.t. } \beta \tp \invIB \parenth{\interAv{m}_X - \interAtar_X} = 0.
\end{align}
After solving for $\beta_0$, the quadratic program for $\beta$ is exactly the same as in the DIP proof in Appendix~\ref{sub:proof_of_theorem_1} around Equation~\eqref{eq:proof_quadratic_program_dipmeanmatchlin}. Thus we obtain
\begin{align}
  \label{eq:proof_betacirm}
  \betacirmeanmatch{m} &= \parenth{\Ind_\dims - \semBX}\tp \frac{\noisecovY^2 \Qcirmv{m} \parenth{\Qcirmv{m} \tp \noisecovX \Qcirmv{m}}^{-1} \Qcirmv{m}\tp \sembv }{1 + \noisecovY^2 \sembv \tp \Qcirmv{m} \parenth{\Qcirmv{m} \tp \noisecovX \Qcirmv{m} }^{-1} \Qcirmv{m} \tp \sembv} \notag \\
  \beta_{\tagcirmeanmatch, 0}^\tagk{m} &= \parenth{1 - {\betacirmeanmatch{m}} \tp \invIB \sembv} \interAv{m}_Y -{\betacirmeanmatch{m}} \tp \invIB \interAv{m}_X,
\end{align}
the corresponding CIRM target population risk is
\begin{align*}
  \tilde{\risk}\parenth{f_{\tagcirmeanmatch\tagk{m}}} = \frac{\noisecovY^2}{1 + \noisecovY^2 \sembv\tp \noisecovX^{-\frac{1}{2}} \Gcirmv{m} \noisecovX^{-\frac{1}{2}} \sembv } + \frac{\parenth{\interAv{m}_Y - \interAtar_Y}^2}{\parenth{1 + \noisecovY^2 \sembv\tp \noisecovX^{-\frac{1}{2}} \Gcirmv{m} \noisecovX^{-\frac{1}{2}} \sembv }^2},
\end{align*}
where $\Qcirmv{m}$ and $\Gcirmv{m}$ are in the same way as $\Qcirmv{1}$ and $\Gcirmv{1}$ in Theorem~\ref{thm:single_source_anticausal_mean_shift}.

\mbox{$\Gcirmv{m} = \noisecovX^{1/2} \Qcirmv{m} \parenth{\Qcirmv{m}  \tp \noisecovX \Qcirm{m} }^{-1} \Qcirmv{m} \tp \noisecovX^{1/2}$} is a projection matrix. $\Qcirmv{m} \in \real^{\dims \times \dims-1}$ is the matrix with columns formed by the vectors that complete the vector $u$ to an orthonormal basis where
\begin{align*}
  u^\tagk{m} = \begin{cases}
    \frac{\interAv{m} - \interAtar}{ \vecnorm{\interAv{m} - \interAtar}{2}}, & \text{if } \interAv{m} \neq \interAtar \\
    0, & \text{otherwise.}
  \end{cases}
\end{align*}

\subsection{Proof of Corollary~\ref{cor:multiple_source_anticausal_mean_shift}} % (fold)
\label{sub:proof_of_corollary_cor:multiple_source_anticausal_mean_shift}
Equation~\eqref{eq:multiple_source_pop_risk_olsoracle_gaussian},~\eqref{eq:multiple_source_pop_risk_cip_gaussian} and~\eqref{eq:multiple_source_pop_risk_cirm_gaussian} follow directly by plugging in $\noisecovX = \frac{\noisecovY^2}{\rho}\Ind_\dims$ and intervention on $Y$ equals to zero in the corresponding formula in Theorem~\ref{thm:mutiple_source_anticausal_mean_shift}. Recall that $\Pcip \in \real^{\dims \times p}$ is the matrix with columns formed with the orthonormal basis of $\text{span}\parenth{\interAv{2}_X - \interAv{1}_X, \ldots, \interAv{\envs}_X - \interAv{1}_X}$. Let $A \in \real^{\dims \times (\envs-1)}$ with $(m-1)$-th column $\interAv{m}_X - \interAv{1}_X$. According the assumption that $\interAv{2}_X - \interAv{1}_X, \ldots \interAv{\envs}_X - \interAv{1}_X$ are generated independently from the standard Gaussian distribution,
$\frac{1}{\dims} A\tp A$ follows a multivariate Wishart distribution and its eigenvalue tail bound is well known (see e.g. Chapter 6 in~\cite{wainwright2019high}). We have
\begin{align*}
  \Prob\parenth{\lambda_\text{min}\parenth{A\tp A} \leq \dims\parenth{\parenth{1-\delta} - \sqrt{\frac{\envs-1}{\dims}}}^2} \leq \exp\parenth{-\dims\delta^2/2}.
\end{align*}
\begin{align*}
  \Prob\parenth{\lambda_\text{max}\parenth{A\tp A} \geq \dims\parenth{\parenth{1+\delta} + \sqrt{\frac{\envs-1}{\dims}}}^2} \leq \exp\parenth{-\dims\delta^2/2}.
\end{align*}
It implies with probability at least $1 - 2 \exp(-\dims\delta^2/2)$, for $\delta < \frac{1}{2} - \sqrt{\frac{\envs-1}{\dims}}$, we have
\begin{align*}
  \lambda_\text{min}\parenth{A\tp A} > \frac{\dims}{2} \text{ and } \lambda_\text{max}\parenth{A\tp A} < \frac{3\dims}{2}.
\end{align*}
This also implies that with the same probability $A\tp A$ is full rank and $p = \envs - 1$.

For a fixed vector $v \in \real^\dims$, we have
\begin{align*}
  \vecnorm{\Pcip \tp v}{2}^2 = v\tp A \parenth{A\tp A}^{-1} A \tp v.
\end{align*}
Note that
\begin{align*}
  A \tp v = \sum_{m=2}^\envs \parenth{\interAv{m}_X - \interAv{1}_X}\tp v
\end{align*}
is a sum of $(\envs-1)$ i.i.d. Gaussian random variables with mean $0$ and variance $v\tp v$. Using the standard chi-square tail bound, we have
\begin{align*}
  \Prob\parenth{\vecnorm{A\tp v}{2}^2 \geq (\envs-1) v\tp v \parenth{1 + \frac{t}{\sqrt{\envs-1}}}} \leq \exp\parenth{-t^2/8}.
\end{align*}
\begin{align*}
  \Prob\parenth{\vecnorm{A\tp v}{2}^2 \leq (\envs-1) v\tp v \parenth{1 - \frac{t}{\sqrt{\envs-1}}}} \leq \exp\parenth{-t^2/8}.
\end{align*}
Combining the high probability bounds above, we have, with probability at least $1 - 2\exp\parenth{-\dims\delta^2/2} - 2\exp\parenth{-t^2/8}$,
\begin{align*}
  \frac{(\envs-1)}{3\dims} v\tp v \leq \vecnorm{\Pcip \tp v}{2}^2 \leq \frac{3 (\envs-1)}{\dims} v\tp v,
\end{align*}
given that $\delta < \frac{1}{2} - \sqrt{\frac{\envs-1}{\dims}}$ and $t \leq \frac{\sqrt{\envs-1}}{2}$. Note that it is possible to ensure $1 - 2\exp\parenth{-\dims\delta^2/2} - 2\exp\parenth{-t^2/8}$ to be close to $1$ when $\dims$ and $\envs$ are both large and $\envs \ll \dims$. Taking $t = \frac{\sqrt{\envs-1}}{2}$ and $\dims = 6\envs$, this probability is larger than $1 - 2 \exp(-\dims/32) - 2\exp(-\envs/32)$.

The high probability bound for $\parenth{u\tp \sembv}^2$ can be obtained similarly as in the proof of Corollary~\ref{cor:single_source_anticausal_mean_shift}. According to the proof of Corollary~\ref{cor:single_source_anticausal_mean_shift} in Appendix~\ref{sub:proof_of_cor_1}, for $0 < t \leq \dims/2$, with probability at least $1 - \exp(-t/16) - 2 \exp(-t/4)$, we have
\begin{align*}
  \parenth{u\tp \sembv}^2 \leq \frac{t}{\dims} \vecnorm{\sembv}{2}^2.
\end{align*}
Putting the two high probability bound together, we conclude Corollary~\ref{cor:multiple_source_anticausal_mean_shift}.

% subsection proof_of_corollary_cor:multiple_source_anticausal_mean_shift (end)
\subsection{Proof of Corollary~\ref{cor:multiple_source_anticausal_mean_shift_source_pop_risk}} % (fold)
\label{sub:proof_of_cor_5}
This corollary follows from the proof of Theorem~\ref{thm:mutiple_source_anticausal_mean_shift} in Appendix~\ref{sub:proof_of_theorem_2}.
According to Equation~\eqref{eq:proof2_quadratic_program_cirmeanmatch_final}, the $m$-th source population risk can be written as
\begin{align*}
  \parenth{1 - \beta\tp \invIB \sembv}^2 \noisecovY^2 + \parenth{\parenth{1 - \beta\tp \invIB \sembv} \interAv{m}_Y - \beta\tp \invIB \interAv{m}_X - \beta_0}^2 + \beta \tp \invIB \noisecovX \invIB\tp \beta.
\end{align*}
Similarly, the target population risk can be written as
\begin{align*}
  \parenth{1 - \beta\tp \invIB \sembv}^2 \noisecovY^2 + \parenth{\parenth{1 - \beta\tp \invIB \sembv} \interAtar_Y - \beta\tp \invIB \interAtar_X - \beta_0}^2 + \beta \tp \invIB \noisecovX \invIB\tp \beta.
\end{align*}
The CIRM constraint ensures that $\beta \tp \invIB \parenth{\interAv{m}_X - \interAtar_X} = 0$ according to Equation~\eqref{eq:proof2_quadratic_program_cirmeanmatch_final}. Hence the only difference between the source population risk and target population risk lies in the $\interAv{m}_Y$ and $\interAtar_Y$ dependent terms. Plugging the CIRM solution~\eqref{eq:proof_betacirm} into the source and target population risks, we obtain
\begin{align*}
  \risk^\tagk{m}\parenth{f_{\tagcirmeanmatch}^\tagk{m}} &= \tilde{\risk}\parenth{f_{\tagcirmeanmatch}^\tagk{m}} + \parenth{1 - \betacirmeanmatch{m} \tp \invIB \sembv}^2 \parenth{\interAv{m}_Y - \interAtar_Y}^2 \\
  &=\tilde{\risk}\parenth{f_{\tagcirmeanmatch}^\tagk{m}} + \frac{\parenth{\interAv{m}_Y - \interAtar_Y}^2}{\parenth{1 + \noisecovY^2 \sembv\tp \noisecovX^{-\frac{1}{2}} \Gcirmv{m} \noisecovX^{-\frac{1}{2}} \sembv }^2}.
\end{align*}

% subsection proof_of_corollary_cor:multiple_source_anticausal_mean_shift_source_pop_risk (end)

\subsection{Proof of Corollary~\ref{cor:single_source_mixed_mean_shift}} % (fold)
\label{sub:proof_of_cor_6}
\paragraph{Risk of DIP$\diamondsuit$:}
Since $X_\tagpar^\tagk{m}$ is uncorrelated with $\noisev{m}_Y$, regressing $Y^\tagk{m}$ on $X_\tagpar^\tagk{m}$ gives back the coefficients in the data generation SCM. Solving Equation~\eqref{eq:solve_gamma_dipmix}, we obtain
\begin{align*}
  \gamma^\tagk{m} = \sembh_\tagpar.
\end{align*}
Similarly, since $X_\tagpar^\tagk{m}$ is uncorrelated with $X_\tagdes^\tagk{m}$, regressing of $X_\tagdes^\tagk{m}$ on $X_\tagpar^\tagk{m}$ gives back the coefficients in the data generation SCM. Solving Equation~\eqref{eq:solve_Gamma_dipmix}, we obtain
\begin{align*}
  {\Gamma^\tagk{m}}\tp = \parenth{\Ind_{\dims-\dimscau} - \semBX_\tagdes}^{-1}\sembv_\tagdes \sembh_\tagpar\tp + \parenth{\Ind_{\dims-\dimscau} - \semBX_\tagdes}^{-1}\semB_\text{d-p}.
\end{align*}
Use the definition of intermediate random variables in Equation~\eqref{eq:def_intermediate_X_Y}, we observe that these variables satisfy the anticausal data generation Assumption~\ref{ass:assumption_single_source_anti_causal}
\begin{align*}
  \bmat{\Xv{m}_{\tagint} \\ \Yv{m}_{\tagint}} &= \bmat{\semBX_\tagdes & \sembv_\tagdes \\ 0 & 0} \bmat{\Xv{m}_{\tagint} \\ \Yv{m}_{\tagint}} + \bmat{\interAv{m}_{X, \tagdes} \\ \interAv{m}_Y} + \bmat{\noisev{m}_{X, \tagdes} \\ \noisev{m}_Y} \\
  \bmat{\Xtar_{\tagint} \\ \Ytar_{\tagint}} &= \bmat{\semBX_\tagdes & \sembv_\tagdes \\ 0 & 0} \bmat{\Xtar_{\tagint} \\ \Ytar_{\tagint}} + \bmat{\interAtar_{X, \tagdes} \\ \interAtar_Y} + \bmat{\noisetar_{X, \tagdes} \\ \noisetar_Y}.
\end{align*}
Together with the assumption of no intervention on $Y$, we can apply Theorem~\ref{thm:single_source_anticausal_mean_shift} on the intermediate random variables to obtain the target risk of DIP$\diamondsuit$.
\paragraph{Risk of OLSTar:} The data generation Assumption~\ref{ass:assumption_multiple_source_mixed} gives the following equations
\begin{align*}
  \Xtar_\tagpar &= H_\tagpar\interAtar_{X, \tagpar} + H_\tagpar\noisetar_{X, \tagpar} \\
  \Xtar_\tagdes &= H_\tagdes \sembv_\tagdes Y + H_\tagdes \semBX_{\text{d-p}} \Xtar_\tagpar + H_\tagdes\interAtar_{X, \tagdes} + H_\tagdes\noisetar_{X, \tagdes} \\
  \Ytar &= \sembh_\tagpar \tp \Xtar_\tagpar + \noisetar_Y,
\end{align*}
where $H_\tagpar = \parenth{\Ind_\dimscau-\semBX_\tagpar}^{-1}$, $H_\tagdes = \parenth{\Ind_{\dims-\dimscau}-\semBX_\tagdes}^{-1}$.
The target population risk for $\parenth{\beta_\tagpar, \beta_\tagdes, \beta_0}$ becomes
\begin{align*}
  &\Exs\parenth{\Ytar - \beta_\tagpar\tp \Xtar_\tagpar - \beta_\tagdes\tp \Xtar_\tagdes - \beta_0}^2 \\
  &= \parenth{1 - \beta_\tagdes\tp H_\tagdes \sembh_\tagdes }^2 \noisecovY^2 + \Exs\parenth{\sembh_\tagpar\tp\Xtar_\tagpar - \beta_\tagpar\tp \Xtar_\tagpar - \beta_\tagdes\tp H_\tagdes \sembv_\tagdes \sembh_\tagpar\tp \Xtar_\tagpar - \beta_\tagdes\tp H_\tagdes \semBX_{\text{d-p}} \Xtar_\tagpar}^2 \\
  &+ \parenth{\beta_\tagdes\tp H_\tagdes \interAtar_{X,\tagdes} + \beta_0}^2 + \beta_\tagdes\tp H_\tagdes \noisecovX_\tagdes H_\tagdes\tp \beta_\tagdes.
\end{align*}
The minimization on $\beta_\tagpar$ and $\beta_0$ can be solved easily and it remains a quadratic program on $\beta_\tagdes$. Since this quadratic program is similar to that in the proof of Theorem~\ref{thm:single_source_anticausal_mean_shift}, we conclude by the referring to the part where we solve the quadratic program in Appendix~\ref{sub:proof_of_theorem_1}.

% subsection proof_of_corollary_cor:multiple_source_mixed_mean_shift (end)

\section{Additional CIRM variants: RII, RIIRMI, CIRMI} % (fold)
\label{sec:cirm_variants}

The target population risk of CIP and CIRM estimator discussed above still have dependence on $\interAv{1}_Y - \interAtar_Y$. As we have explained after Theorem~\ref{thm:mutiple_source_anticausal_mean_shift}, this is because Assumption~\ref{ass:assumption_multiple_source_anti_causal} does not rule out the unidentifiable scenario. When we add additional assumptions such as $\sembv \notin \text{span}\parenth{\interAv{2}_X - \interAv{1}_X, \ldots, \interAv{\envs}_X - \interAv{1}_X}$, we show that there are new methods which take advantage of this assumption to get rid of $\interAv{1}_Y - \interAtar_Y$ dependence in the risk. We introduce three new estimators.

\begin{itemize}
  \item \textbf{RII-mean:} the population residual invariance and independent estimator where mean is matched across environments
  \begin{align}
    \label{eq:estimator_pop_riimean}
    f_{\tagriimean}(x) &\defn x \tp\betariimean + \beta_{\tagriimean, 0} \notag \\
    \betariimean, \beta_{\tagriimean, 0} &\defn \argmin_{\beta, \beta_0}\  \frac{1}{\envs} \sum_{k=1}^\envs \Exs_{(X, Y) \sim \distriv{k}}\parenth{Y - X\tp\beta - \beta_0}^2 \notag \\
    &\text{s.t.\ } \Exs_{(X, Y) \sim \distriv{m}} \brackets{Y - X \tp\beta} = \Exs_{(X, Y) \sim \distriv{1}} \brackets{Y - X \tp\beta} \notag  \\
    &\text{and\ } \Exs_{(X, Y) \sim \distriv{m}} \brackets{(Y - \Exs_{Y \sim \distriv{m}_Y}[Y])\cdot (Y - X \tp\beta)} = 0, \forall m \in \braces{2, \ldots, \envs}.
  \end{align}
  The idea of matching the residual has appeared in the invariant causal prediction paper~\cite{peters2016causal} and in the anchor regression paper~\cite{rothenhausler2018anchor}. The residual independence penalty is also the core idea in the invariant causal prediction~\cite{peters2016causal} and Greenfeld and Shalit~\cite{greenfeld2019robust}.
  \item \textbf{RIIRMI$^\tagk{m}$-mean:} the population residual invariant independent residual matching estimator with additional residual independence using $m$-th source environment
  \begin{align}
    \label{eq:estimator_pop_riirmi}
    f_{\tagriirmi}^\tagk{m}(x) &\defn x \tp\betariirmi{m} + \beta_{\tagriirmi, 0}^\tagk{m} \notag \\
    \betariirmi{m}, \beta_{\tagriirmi, 0}^\tagk{m} &\defn \argmin_{\beta, \beta_0}\  \Exs_{(X, Y) \sim \distriv{m}}\parenth{Y - X\tp\beta - \beta_0}^2 \notag \\
    \text{s.t.\ } \Exs_{X \sim \distriv{m}_X} & \brackets{\beta \tp \parenth{ X - \parenth{X \tp\betariimean} \briirmi }} = \Exs_{X \sim \distritar_X} \brackets{\beta \tp \parenth{ X - \parenth{X \tp\betariimean} \briirmi }} \notag \\
    &\text{and\ } \Exs_{(X, Y) \sim \distriv{m}} \brackets{(Y - \Exs_{Y \sim \distriv{m}_Y}[Y])\cdot (Y - X \tp\beta)} = 0,
\end{align}
  where, with $\distri^{\text{source}}$ denote the uniform mixture of all source distributions,
  \begin{align}
    \label{eq:estimator_pop_riirmi_b_choice}
    \briirmi \defn \frac{\Exs_{(X, Y)\sim \distri^{\text{source}} }\brackets{X \cdot (Y - \Exs[Y])}}{ \Exs_{Y \sim \distri^{\text{source}}_Y}\brackets{\parenth{Y-\Exs[Y]}^2}}.
  \end{align}
  \item \textbf{CIRMI$^\tagk{m}$-mean:} the population conditional invariant residual matching estimator with additional residual independence using $m$-th source environment
  \begin{align}
    \label{eq:estimator_pop_cirmeanmatch_ri}
    f_{\tagcirmi}^\tagk{m}(x) &\defn x \tp\betacirmi{m} + \beta_{\tagcirmi, 0}^\tagk{m} \notag \\
    \betacirmi{m}, \beta_{\tagcirmi, 0}^\tagk{m} &\defn \argmin_{\beta, \beta_0}\  \Exs_{(X, Y) \sim \distriv{m}}\parenth{Y - X\tp\beta - \beta_0}^2 \notag \\
    \text{s.t.\ } \Exs_{X \sim \distriv{m}_X} & \brackets{\beta \tp \parenth{ X - \parenth{X \tp\betacipmean} \bcirmi }} = \Exs_{X \sim \distritar_X} \brackets{\beta \tp \parenth{ X - \parenth{X \tp\betacipmean} \bcirmi }} \notag \\
    &\text{and\ } \Exs_{(X, Y) \sim \distriv{m}} \brackets{(Y - \Exs_{Y \sim \distriv{m}_Y}[Y])\cdot (Y - X \tp\beta)} = 0,
\end{align}
  where, with $\distri^{\text{source}}$ denote the uniform mixture of all source distributions,
  \begin{align}
    \label{eq:estimator_pop_cirmeanmatch_ri_b_choice}
    \bcirmi \defn \frac{\Exs_{(X, Y)\sim \distri^{\text{source}} }\brackets{X \cdot (Y - \Exs[Y])}}{ \Exs_{(X, Y)\sim \distri^{\text{source}} } \brackets{\parenth{X\tp \betacipmean - \Exs[X\tp \betacipmean] } \cdot \parenth{Y-\Exs[Y]}}}.
  \end{align}
\end{itemize}
Note that a common feature of the three estimators above is that they all have the residual independent constraint of the form
\begin{align*}
  \Exs_{(X, Y) \sim \distriv{m}} \brackets{(Y - \Exs_{Y \sim \distriv{m}_Y}[Y])\cdot (Y - X \tp\beta)} = 0.
\end{align*}
This constraint takes advantage of the anticausal prediction setting and restricts the estimator to ignorant of the intervention on $Y$.

\begin{theorem}
  \label{thm:mutiple_source_anticausal_mean_shift_bnotin}
  Under data generation Assumption~\ref{ass:assumption_multiple_source_anti_causal} and the additional assumption
  \begin{align}
    \label{ass:mutiple_source_anticausal_mean_shift_bnotin_assumption}
    \sembv \notin \text{span}\parenth{\interAv{2}_X - \interAv{1}_X, \ldots, \interAv{\envs}_X - \interAv{1}_X},
  \end{align}
  the population target risk of \textbf{RII-mean}, \textbf{RIIRMI$^\tagk{1}$-mean} and \textbf{CIRMI$^\tagk{1}$-mean} satisfies
  \begin{align}
    \label{eq:multiple_source_pop_risk_bnotin_riimean}
    \tilde{\risk}\parenth{f_{\tagriimean}} =  \frac{\sembv \tp \parenth{\Ind_\dims - \Prii \Prii \tp} \noisecovX^{1/2} \parenth{\Ind_\dims - \Grii} \noisecovX^{1/2}\parenth{\Ind_\dims - \Prii \Prii \tp} \sembv}{\vecnorm{\parenth{\Ind_\dims - \Prii \Prii \tp} \sembv}{2}^4},
  \end{align}
  where $\Prii \in \real^{\dims \times p}$ equals to $\Pcip$ which is the matrix formed by an orthonormal basis of the $p$-dimensional subspace \mbox{$\text{span}\parenth{\interAv{2}_X-\interAv{1}_X, \ldots, \interAv{\envs}_X - \interAv{1}_X}$}, $\Grii = \noisecovX^{1/2} \Qrii \parenth{\Qrii\tp \noisecovX \Qrii}^{-1} \Qrii\tp \noisecovX^{1/2}$ is a projection matrix of rank $\dims- p -1$, $\Qrii \in \real^{\dims \times (\dims - p - 1)}$ is the matrix with columns formed by completing the columns of $\Prii$ and $\frac{\parenth{\Ind_\dims - \Prii \Prii \tp} \sembv }{\vecnorm{\parenth{\Ind_\dims - \Prii \Prii \tp} \sembv}{2}}$ to an orthonormal basis of $\real^\dims$ via Gram-Schmidt orthogonalization,
  \begin{align}
    \label{eq:multiple_source_pop_risk_bnotin_riirmi}
    \tilde{\risk}(f_\tagriirmi^\tagk{m}) = \tilde{\risk}(f_\tagcirmi^\tagk{m}) = \frac{\sembv\tp \parenth{\Ind_\dims - u u\tp}  \noisecovX^{1/2} \parenth{\Ind_\dims - \Griirmi} \noisecovX^{1/2}\parenth{\Ind_\dims - u u\tp} \sembv}{\vecnorm{\parenth{\Ind_\dims - u u\tp} \sembv}{2}^4},
  \end{align}
  where $u = \frac{\interAv{m}_X - \interAtar_X}{\vecnorm{\interAv{m}_X - \interAtar_X}{2}}$, $\Griirmi = \noisecovX^{1/2} \Qriirmi \parenth{\Qriirmi \tp \noisecovX \Qriirmi}^{-1} \Qriirmi\tp \noisecovX^{1/2}$ is a projection matrix of rank $\dims- 2$, $\Qriirmi \in \real^{\dims \times (\dims - 2)}$ is the matrix with columns formed by completing $u$ and $v$ to an orthonormal basis of $\real^\dims$ via Gram-Schmidt orthogonalization.
\end{theorem}
We present a corollary that puts additional assumptions on how the interventions are positioned to make the results in Theorem~\ref{thm:mutiple_source_anticausal_mean_shift_bnotin} easier to understand.
\begin{corollary}
  \label{cor:multiple_source_anticausal_mean_shift_bnotin}
  In addition to Assumption~\ref{ass:assumption_multiple_source_anti_causal} and the assumption in Equation~\eqref{ass:mutiple_source_anticausal_mean_shift_bnotin_assumption}, assume $\noisecovX = \frac{\noisecovY^2}{\rho} \Ind_\dims$ with $\rho > 0$, then
  \begin{align}
    \label{eq:multiple_source_pop_risk_bnotin_rii_gaussian}
    \tilde{\risk}\parenth{f_{\tagriimean}} = \frac{\noisecovY^2}{\rho \vecnorm{\sembv}{2}^2 - \rho \parenth{\Prii\tp \sembv}^2}
  \end{align}
  \begin{align}
    \label{eq:multiple_source_pop_risk_bnotin_cirmi_gaussian}
    \tilde{\risk}(f_\tagriirmi^\tagk{1}) = \tilde{\risk}(f_\tagcirmi^\tagk{1}) = \frac{\noisecovY^2}{\rho \vecnorm{\sembv}{2}^2 - \rho \parenth{u\tp \sembv}^2},
  \end{align}
   Additionally, if $\interAv{2}_X - \interAv{1}_X, \ldots, \interAv{\envs}_X - \interAv{1}_X, \xi$ are generated independently from the standard Gaussian distribution and $\interAtar_X - \interAv{1} = \Pcip \xi$, then with high probability, we have
   \begin{align*}
    \text{dim}\parenth{\text{span}\parenth{\interAv{2}_X-\interAv{1}_X, \ldots, \interAv{\envs}_X - \interAv{1}_X}} = \envs - 1.
  \end{align*}
   \begin{align}
    \label{eq:multiple_source_pop_risk_bnotin_rii_gaussian_ineq}
    \frac{\noisecovY^2}{\rho \parenth{1 - \frac{(\envs-1)}{3\dims}} \vecnorm{\sembv}{2}^2} \leq \tilde{\risk}\parenth{f_{\tagriimean}} \leq \frac{\noisecovY^2}{\rho \parenth{1 - \frac{3(\envs-1)}{\dims}} \vecnorm{\sembv}{2}^2}
  \end{align}
  \begin{align}
    \label{eq:multiple_source_pop_risk_bnotin_cirmi_gaussian_ineq}
    \tilde{\risk}(f_\tagriirmi^\tagk{m}) = \tilde{\risk}(f_\tagcirmi^\tagk{m}) \leq \frac{\noisecovY^2}{\rho \parenth{1 - \frac{c}{\dims}} \vecnorm{\sembv}{2}^2}
  \end{align}
  where $c$ is a constant.
\end{corollary}
The target population risks of RII, RIIRMI, CIRMI are compared with CIRM under the assumptions of Corollary~\ref{cor:multiple_source_anticausal_mean_shift_bnotin}. Under the additional assumption~\eqref{ass:mutiple_source_anticausal_mean_shift_bnotin_assumption}, the new DA methods RIIRMI and CIRMI effectively remove the $\interAtar_Y - \interAv{m}_Y$ dependency when compared to CIRM. However, RIIRMI and CIRMI have slightly worse target population risk when there is no intervention on the label $\interAv{1}_Y = \interAv{2}_Y=\ldots=\interAtar_Y=0$.

\begin{table}[t]
    \centering
    % \begin{adjustwidth}{-.2in}{-.3in}
    {\renewcommand{\arraystretch}{.8}
    \begin{tabular}{cllc}
        \toprule
         & \thead{ \bf Estimator}  & \thead{
         \bf Target population risk upper bound\\interventions under general position}
        \\ \midrule
        \multirow{6}{*}{\makecell{Intervention on $Y$ \\ (Corollary~\ref{cor:multiple_source_anticausal_mean_shift_bnotin}, $\envs$ sources)}}
        &\thead{DIP$^\tagk{m}$}
        &$\displaystyle\frac{\noisecovY^2}{1 + \rho(1 - \frac{\const}{\dims})\vecnorm{\sembv}{2}^2} + \parenth{\interAtar_Y - \interAv{m}_Y}^2$
        \\[5mm]
        &\thead{CIP}
        & $\displaystyle\frac{\noisecovY^2}{1 + \rho(1 - \frac{\const(\envs-1)}{\dims})\vecnorm{\sembv}{2}^2} + \frac{ \parenth{\interAtar_Y - \bar{a}_Y}- \Delta_Y}{\brackets{1 + \rho(1 - \frac{c(\envs-1)}{\dims})\vecnorm{\sembv}{2}^2}^2}$
        \\[5mm]
        &\thead{CIRM$^\tagk{m}$}
        & $\displaystyle\frac{\noisecovY^2}{1 + \rho(1 - \frac{\const}{\dims})\vecnorm{\sembv}{2}^2} + \frac{\parenth{\interAtar_Y - \interAv{m}_Y}^2}{\brackets{1 + \rho(1 - \frac{c}{\dims})\vecnorm{\sembv}{2}^2}^2}$
        \\[5mm]
        &\thead{RII}
        & $\displaystyle\frac{\noisecovY^2}{\rho(1 - \frac{\const(\envs-1)}{\dims})\vecnorm{\sembv}{2}^2}$
        \\[5mm]
        &\thead{RIIRMI$^\tagk{m}$}
        & $\displaystyle\frac{\noisecovY^2}{\rho(1 - \frac{\const}{\dims})\vecnorm{\sembv}{2}^2}$
        \\[5mm]
        &\thead{CIRMI$^\tagk{m}$}
        & $\displaystyle\frac{\noisecovY^2}{\rho(1 - \frac{\const}{\dims})\vecnorm{\sembv}{2}^2}$
        \\[2mm]
        \bottomrule
    \end{tabular}
    }
    % \end{adjustwidth}
    \caption{Summary of target population risk for different estimators for anticausal domain adaptation under the assumptions of Corollary~\ref{cor:multiple_source_anticausal_mean_shift_bnotin}. Here $\const$ is a constant. For simplicity, we only compare the target population risk upper bound under high probability when the interventions are generated i.i.d. Gaussian.}
    \label{tab:summary_target_expanded}
\end{table}

\subsection{Proof of Theorem~\ref{thm:mutiple_source_anticausal_mean_shift_bnotin}}
\label{sub:proof_of_theorem_3}
\paragraph{Risk of RII: } Using the SEM~\eqref{eq:proof2_source_x_sem} and the residual expression~\eqref{eq:proof2_residual_sem}, the RII objective~\eqref{eq:estimator_pop_riimean} becomes
\begin{align*}
  \min_{\beta, \beta_0} &\parenth{1 - \beta\tp \invIB \sembv}^2 \noisecovY^2 + \frac{1}{\envs} \sum_{m=1}^\envs \parenth{ \parenth{1 - \beta\tp \invIB \sembv} \interAv{m}_Y - \beta\tp\invIB \interAv{m}_X - \beta_0}^2 + \beta\tp\invIB \noisecovX \invIB\tp \beta \notag \\
  \text{s.t. } & \parenth{1 - \beta\tp \invIB \sembv} \parenth{\interAv{m}_Y - \interAv{1}_Y} + \beta\tp \invIB \parenth{\interAv{m}_X - \interAv{1}_X} = 0, \ \forall m \in \braces{2, \ldots, \envs} \\
  \text{and } & \parenth{1 - \beta\tp \invIB \sembv}^2 \noisecovY^2 = 0.
\end{align*}
The second constraint above ensures that
\begin{align*}
  1 - \beta\tp \invIB \sembv = 0.
\end{align*}
This observation allows us to simplify the RII minimization problem above and obtain
\begin{align}
  \label{eq:proof3_RII_simplified_form}
  \min_{\beta, \beta_0} &\frac{1}{\envs} \sum_{m=1}^\envs \parenth{ 0 - \beta\tp\invIB \interAv{m}_X - \beta_0}^2 + \beta\tp\invIB \noisecovX \invIB\tp \beta \notag \\
  \text{s.t. } & \beta\tp \invIB \parenth{\interAv{m}_X - \interAv{1}_X} = 0, \ \forall m \in \braces{2, \ldots, \envs} \notag \\
  \text{and } & \beta\tp \invIB \sembv = 1.
\end{align}
The objective~\eqref{eq:proof3_RII_simplified_form} is a quadratic program with linear constraints. First, the one-dimensional quadratic program on $\beta_0$ can be solved easily by setting the derivative to zero.
\begin{align*}
  \beta_0^* &= \frac{1}{\envs} \sum_{m=1}^\envs {\beta^*} \tp \invIB \interAv{m}_X \\
  & = {\beta^*} \tp \invIB \interAv{1}_X,
\end{align*}
where the last equality follows from the first constraint in Equation~\eqref{eq:proof3_RII_simplified_form}. Second, since $\invIB$ is invertible, we can re-parametrize the optimization on $\beta$ as follows
\begin{align}
  \label{eq:proof3_RII_simplified_form_reparametrized}
  \min_{\gamma \in \real^\dims} &\gamma \tp \noisecovX \gamma \\
  \text{s.t. } & \gamma\tp \parenth{\interAv{m}_X - \interAv{1}_X} = 0, \ \forall m \in \braces{2, \ldots, \envs} \notag \\
  \text{and } & \gamma \sembv = 1.
\end{align}
Let $\Prii$ be the matrix with columns formed by an orthonormal basis of the $p$-dimensional subspace \mbox{$\text{span}\parenth{\interAv{2}-\interAv{1}, \ldots, \interAv{\envs} - \interAv{1}}$}. Define
\begin{align*}
  \vrii \defn \frac{\parenth{\Ind_\dims - \Prii \Prii \tp} \sembv }{\vecnorm{\parenth{\Ind_\dims - \Prii \Prii \tp} \sembv}{2}}.
\end{align*}
$\vrii$ is well defined because $\vecnorm{\parenth{\Ind_\dims - \Prii \Prii \tp} \sembv}{2} \neq 0$ by assumption~\ref{ass:mutiple_source_anticausal_mean_shift_bnotin_assumption}. By construction, we have $\Prii \tp \vrii = 0$ and also
\begin{align}
  \label{eq:proof3_vrii_dot_product}
  \parenth{\frac{\vrii}{\vecnorm{\parenth{\Ind_\dims - \Prii \Prii \tp} \sembv}{2}}}\tp \sembv = 1
\end{align}.

Let $\Qrii \in \real^{\dims \times (\dims - p - 1)}$ be the matrix with columns formed by completing the columns of $\Prii$ and $\vrii$ to an orthonormal basis of $\real^\dims$ via Gram-Schmidt orthogonalization. Because of Equation~\eqref{eq:proof3_vrii_dot_product}, the following map
\begin{align*}
  \real^{\dims - p - 1} &\rightarrow \real^\dims \\
  \zeta &\mapsto \frac{\vrii}{\vecnorm{\parenth{\Ind_\dims - \Prii \Prii \tp} \sembv}{2}} + \Qrii \zeta
\end{align*}
constitutes a bijection between $\real^{\dims - p - 1}$ and the set $\braces{\gamma \in \real^\dims \mid P\tp \gamma = 0, \gamma \tp \sembv = 1}$. With the change of variable, the constrained quadratic program in Equation~\eqref{eq:proof3_RII_simplified_form_reparametrized} is equivalent to the following unconstrained one
\begin{align}
  \label{eq:proof3_RII_simplified_form_reparametrized_unconstrained}
  \min_{\zeta \in \real^{\dims - p - 1}} \parenth{\frac{\vrii}{\vecnorm{\parenth{\Ind_\dims - \Prii \Prii \tp} \sembv}{2}} + \Qrii \zeta}\tp \noisecovX \parenth{\frac{\vrii}{\vecnorm{\parenth{\Ind_\dims - \Prii \Prii \tp} \sembv}{2}} + \Qrii \zeta}
\end{align}
Solving the unconstrained quadratic program by setting gradient to zero, we obtain the minimizer
\begin{align*}
  \zeta^* = - \frac{\parenth{{\Qrii}\tp \noisecovX \Qrii}^{-1} {\Qrii}\tp \noisecovX \vrii}{\vecnorm{\parenth{\Ind_\dims - \Prii \Prii \tp} \sembv}{2}}.
\end{align*}
Transforming the variables back, the RII estimator is
\begin{align}
  \label{eq:proof3_riimean_beta_sol}
  \betariimean &= \parenth{\Ind_\dims - \semBX}\tp \frac{\noisecovX^{-1/2}\parenth{\Ind_\dims - \Grii} \noisecovX^{1/2} \vrii}{\vecnorm{\parenth{\Ind_\dims - \Prii \Prii \tp} \sembv}{2}} \\
  \beta_{\tagriimean, 0} &= - \betariimean \tp \invIB \interAv{1}_X \notag,
\end{align}
where $\Grii = \noisecovX^{1/2} \Qrii \parenth{\Qrii\tp \noisecovX \Qrii}^{-1} \Qrii\tp \noisecovX^{1/2}$ is a projection matrix of rank $\dims- p -1$.
According to the form of the target residual~\eqref{eq:proof2_target_residual_sem}, the target population risk for $(\beta, \beta_0)$ takes the following form
\begin{align*}
  \parenth{1 - \beta\tp \invIB \sembv}^2 \noisecovY^2 + \parenth{ \parenth{1 - \beta\tp \invIB \sembv} \interAtar_Y - \beta\tp\invIB \interAtar_X - \beta_0}^2 + \beta\tp\invIB \noisecovX \invIB\tp \beta.
\end{align*}
Plugging in the RII estimator into the equation above, because $\betariimean\tp \invIB \sembv = 1$ and $\interAtar_X - \interAv{1}_X \in \text{span}(\interAv{2}_X - \interAv{1}_X, \ldots, \interAv{m}_X - \interAv{1}_X)$, we obtain the target population risk of RII
\begin{align*}
  \tilde{\risk}(f_\tagriimean) = \frac{\vrii\tp \noisecovX^{1/2} \parenth{\Ind_\dims - \Grii} \noisecovX^{1/2}\vrii}{\vecnorm{\parenth{\Ind_\dims - \Prii \Prii \tp} \sembv}{2}^2}.
\end{align*}
\paragraph{Risk of RIIRMI: } We start by deriving $\briirmi$ defined in Equation~\eqref{eq:estimator_pop_riirmi_b_choice}. It calculates the correlation between $X$ and $Y$. Using the SEM~\eqref{eq:proof2_source_x_sem}, we obtain
\begin{align}
  \label{eq:proof3_riirmi_b_sol}
  \briirmi = \invIB \sembv.
\end{align}
Note that just like in CIRM, the vector $\briirmi$ is also co-linear with the $Y$ component in the covariate expression in Equation~\eqref{proof2_sem_X_expr_cirm}. So the idea of RIIRMI is very similar to that of CIRM: it uses $\Xtar\tp \betariimean$ as a proxy for the unobserved $\Ytar$ to remove the $\Ytar$ part in the covariates so that the DIP matching based ideas can still be applied.

With the $\briirmi$ expression~\eqref{eq:proof3_riirmi_b_sol} and the $\betariimean$ expression~\eqref{eq:proof3_riimean_beta_sol}, the LHS of the first RIIRMI constraint~\eqref{eq:estimator_pop_riirmi} becomes
\begin{align}
  \label{eq:proof3_riirm_corrected_residual}
  &\Xv{m} - \parenth{{\Xv{m}}\tp \betariimean} \briirmi \notag \\
  &= \parenth{\Yv{m} - {\Xv{m}}\tp \betariimean } \invIB \sembv + \invIB \interAv{m}_X + \invIB \noisev{m}_X \notag \\
  &= \parenth{\parenth{1 - \betariimean \tp \invIB \sembv }\Yv{m} - \betariimean\tp \invIB \interAv{m} - \betariimean\tp \invIB \noisev{m} } \invIB \sembv + \invIB \interAv{m}_X + \invIB \noisev{m}_X \notag \\
  &\stackrel{(i)}{=} \parenth{0 - \betariimean\tp \invIB \interAv{1} - \betariimean\tp \invIB \noisev{m} } \invIB \sembv + \invIB \interAv{m}_X + \invIB \noisev{m}_X,
\end{align}
where the last inequality follows from the two constraints in the RII estimator~\eqref{eq:estimator_pop_riimean}.
Similar expression can be obtained for the target environments by taking into account that $\interAtar_X \in \text{span}\parenth{\interAv{1}_X, \ldots, \interAv{\envs}_X }$,
\begin{align}
  \label{eq:proof3_riirm_corrected_residual_tar}
  \Xtar - \parenth{{\Xtar}\tp \betariimean} \briirmi = \parenth{ \betariimean \tp \invIB \interAv{1} + \betariimean \tp \invIB \noisetar} \invIB \sembv  + \invIB \interAtar_X + \invIB \noisetar_X.
\end{align}
Multiply Equation~\eqref{eq:proof3_riirm_corrected_residual} and~\eqref{eq:proof3_riirm_corrected_residual_tar} by $\beta$ and take expectation, we obtain a simplified form of the first RIIRMI constraint~\eqref{eq:estimator_pop_cirmeanmatch}
\begin{align*}
  \beta \tp \invIB \interAv{m}_X =  \beta \tp \invIB \interAtar_X.
\end{align*}
Note that the first RIIRMI constraint is the same as the one in DIP or CIRM. With the above observation, RIIRMI$^\tagk{1}$ objective~\eqref{eq:estimator_pop_riirmi} becomes
\begin{align}
  \label{eq:proof3_quadratic_program_riirmi_final}
  &\min_{\beta, \beta_0}  \parenth{0 - \beta\tp \invIB \interAv{m}_X - \beta_0}^2 + \beta \tp \invIB \noisecovX \invIB\tp \beta \notag \\
  &\text{s.t. } \beta \tp \invIB \parenth{\interAv{m}_X - \interAtar} = 0 \notag \\
  &\text{and }  \beta\tp \invIB \sembv = 1.
\end{align}
This objective is similar to the one~\eqref{eq:proof2_quadratic_program_cirmeanmatch_final} in the proof of CIRM, with the only difference that there is one additional constraint $\beta\tp \invIB \sembv = 1$. This objective is also similar to the one~\eqref{eq:proof3_RII_simplified_form} in the proof of RII, with the difference that there is only one constraint of the type $\beta \tp \invIB (\interAv{m}_X -\interAtar_X) = 0$. Following the proof of RII to solve the quadratic program with linear constraints, we define
\begin{align*}
  v \defn \frac{\parenth{\Ind_\dims - u u \tp }\sembv}{\vecnorm{\parenth{\Ind_\dims - u u \tp }\sembv}{2}},
\end{align*}
where $u = \frac{\interAv{m}_X - \interAtar_X}{\vecnorm{\interAv{m}_X - \interAtar_X}{2}}$
and define $\Qriirmi \in \real^{\dims \times (\dims - 2)}$ be the matrix with columns formed by completing $u$ and $v$ to an orthonormal basis of $\real^\dims$ via Gram-Schmidt orthogonalization, then the RIIRMI estimator is
\begin{align}
  \label{eq:proof3_riirmi_beta_sol}
  \betariirmi{m} &= \parenth{\Ind_\dims - \semBX}\tp \frac{\noisecovX^{-1/2}\parenth{\Ind_\dims - \Griirmi} \noisecovX^{1/2} \parenth{\Ind_\dims - u u \tp} \sembv }{\vecnorm{\parenth{\Ind_\dims - u u \tp} \sembv}{2}^2 } \\
  \beta_{\tagriirmi, 0}^\tagk{m} &= - {\betariirmi{m}} \tp \invIB \interAv{m}_X\notag .
\end{align}
where $\Griirmi = \noisecovX^{1/2} \Qriirmi \parenth{\Qriirmi \tp \noisecovX \Qriirmi}^{-1} \Qriirmi\tp \noisecovX^{1/2}$ is a projection matrix of rank $\dims- 2$.
According to the form of the target residual~\eqref{eq:proof2_target_residual_sem}, the target population risk for $(\beta, \beta_0)$ takes the following form
\begin{align*}
  \parenth{1 - \beta\tp \invIB \sembv}^2 \noisecovY^2 + \parenth{ \parenth{1 - \beta\tp \invIB \sembv} \interAtar_Y - \beta\tp\invIB \interAtar_X - \beta_0}^2 + \beta\tp\invIB \noisecovX \invIB\tp \beta.
\end{align*}
Plugging in the RIIRMI[1] estimator into the equation above, because $\betariimean\tp \invIB \sembv = 1$ and $\beta\tp \invIB \interAv{m}_X = \beta\tp \invIB \interAtar_X$, we obtain the target population risk of RIIRMI[1]
\begin{align*}
  \tilde{\risk}(f_\tagriirmi^\tagk{m}) = \frac{\sembv\tp \parenth{\Ind_\dims - u u\tp}  \noisecovX^{1/2} \parenth{\Ind_\dims - \Griirmi} \noisecovX^{1/2}\parenth{\Ind_\dims - u u\tp} \sembv}{\vecnorm{\parenth{\Ind_\dims - u u\tp} \sembv}{2}^4}.
\end{align*}
\paragraph{Risk of CIRMI: } The proof for CIRMI follows easily by combining parts of proofs in CIRM and RIIRMI. The CIRMI$^\tagk{1}$ estimator has one additional constraint compared to the CIRM$^\tagk{1}$ estimator in Equation~\eqref{eq:proof2_quadratic_program_cirmeanmatch_final}
\begin{align}
  \label{eq:proof3_quadratic_program_cirmi_final}
  &\min_{\beta, \beta_0}  \parenth{1 - \beta\tp \invIB \sembv}^2 \noisecovY^2 + \parenth{\parenth{1 - \beta\tp \invIB \sembv} \interAv{1}_Y - \beta\tp \invIB \interAv{1}_X - \beta_0}^2 + \beta \tp \invIB \noisecovX \invIB\tp \beta \notag \\
  &\text{s.t. } \beta \tp \invIB \parenth{\interAv{1}_X - \interAtar} = 0 \notag \\
  &\text{and } 1 - \beta\tp \invIB \sembv = 0.
\end{align}
In fact, the above optimization problem is exactly the same as RIIRMI in Equation~\eqref{eq:proof3_quadratic_program_riirmi_final}. Consequently, the CIRMI solution is the same as that of RIIRMI.
\begin{align}
  \label{eq:proof3_cirmi_beta_sol}
  \betacirmi{1} &= \betariirmi{1} \\
  \beta_{\tagcirmi, 0}^\tagk{1} &= \beta_{\tagriirmi, 0}^\tagk{1} \notag ,
\end{align}
\begin{align*}
  \tilde{\risk}(f_\tagcirmi^\tagk{1}) = \tilde{\risk}(f_\tagriirmi^\tagk{1}).
\end{align*}

\subsection{Proof of Corollary~\ref{cor:multiple_source_anticausal_mean_shift_bnotin}} % (fold)
\label{sub:proof_of_corollary_cor:multiple_source_anticausal_mean_shift_bnotin}
The proof of Corollary~\ref{cor:multiple_source_anticausal_mean_shift_bnotin} follows similarly as that of Corollary~\ref{cor:multiple_source_anticausal_mean_shift} in Appendix~\ref{sub:proof_of_corollary_cor:multiple_source_anticausal_mean_shift}.

% subsection proof_of_corollary_cor:multiple_source_anticausal_mean_shift_bnotin (end)

\vskip 0.2in
\bibliography{causal}

\end{document}